\documentclass[10pt,twocolumn,letterpaper]{article}

\usepackage{iccv}
\usepackage{times}
\usepackage{epsfig}
\usepackage{graphicx}
\usepackage{amsmath}
\usepackage{amssymb}
\usepackage{amsfonts}
\usepackage{threeparttable}
\usepackage{multirow}
\usepackage{tabularx}
\usepackage{booktabs} 
\usepackage[english]{babel}
\usepackage{caption} 
\usepackage{arydshln} 
\usepackage{ifthen}
\usepackage{algorithmic}
\usepackage{algorithm}
\newboolean{arxiv}
\setboolean{arxiv}{true}

\usepackage[pagebackref=true,breaklinks=true,letterpaper=true,colorlinks,bookmarks=false]{hyperref}

\iccvfinalcopy 

\def\iccvPaperID{59} 

\ificcvfinal\fi

\usepackage{capt-of,etoolbox}
\makeatletter
\apptocmd\@maketitle{{\myfigure{}\par}}{}{}
\makeatother

\begin{document}

\title{PointFlow: 3D Point Cloud Generation with Continuous Normalizing Flows}

\makeatletter
\newcommand{\printfnsymbol}[1]{%
  \textsuperscript{\@fnsymbol{#1}}%
}
\makeatother
\author{
Guandao Yang$^{1,2}$\thanks{Equal contribution.}\ , 
Xun Huang$^{1,2}$\printfnsymbol{1}, 
Zekun Hao$^{1,2}$, 
Ming-Yu Liu$^3$, 
Serge Belongie$^{1,2}$, 
Bharath Hariharan$^1$\\
$^1$Cornell University 
\hspace{15pt}
$^2$Cornell Tech
\hspace{15pt}
$^3$NVIDIA
}
\newcolumntype{Y}{>{\centering\arraybackslash}X}
\definecolor{mygray}{gray}{0.6}
\newcommand{\todo}[1]{\textcolor{red}{todo:#1}}
\newcommand{\gy}[1]{\textcolor{blue}{GY:#1}}
\newcommand{\revxun}[1]{\textcolor{magenta}{(XH: #1)}}
\newcommand{\xun}[1]{\textcolor{magenta}{#1}}
\newcommand{\mlcomment}[1]{\textcolor{cyan}{\bf ML: #1}}
\newcommand{\ml}[1]{\textcolor{cyan}{#1}} 
\newcommand{\bh}[1]{\textcolor{green}{BH: #1}} 
\newcommand\myfigure{%
	\centering
	\newcommand{\sizea}{0.138\linewidth}
	\includegraphics[width=\sizea]{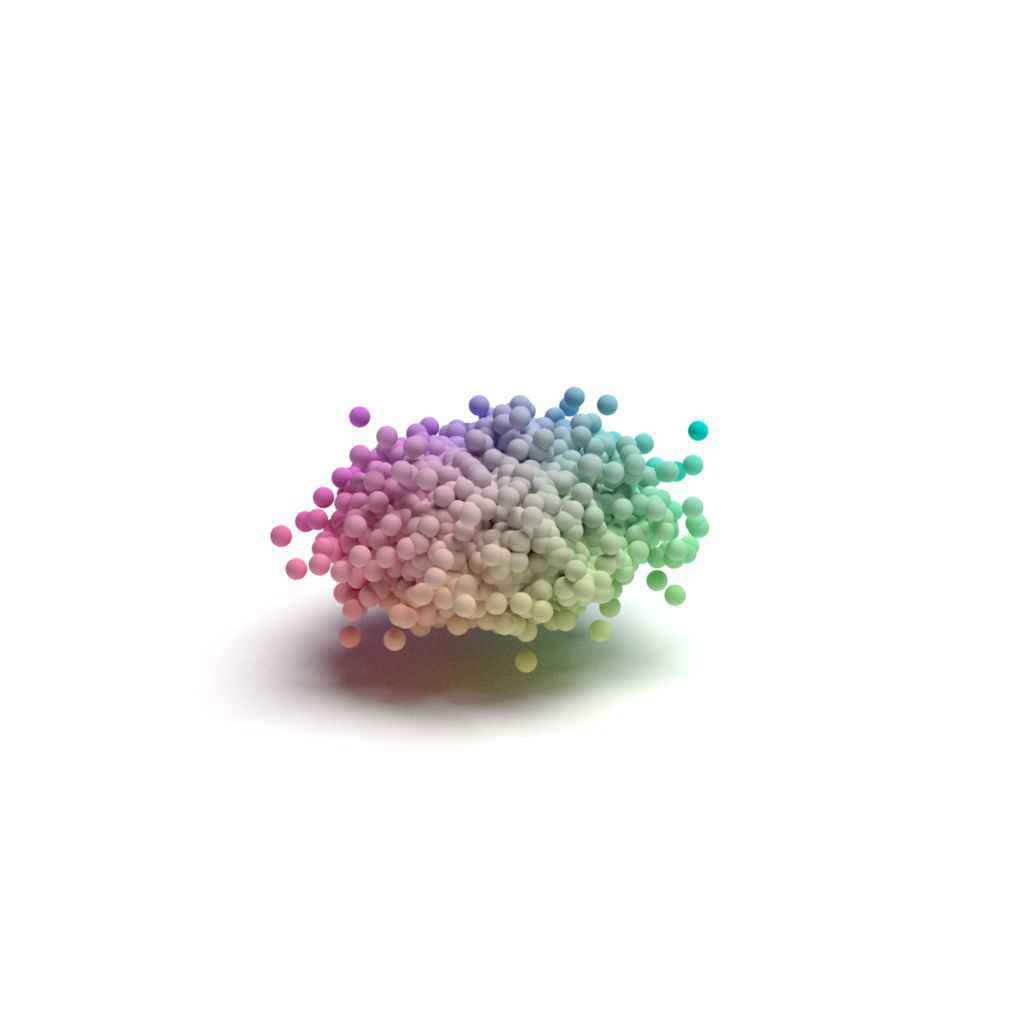}
	\includegraphics[width=\sizea]{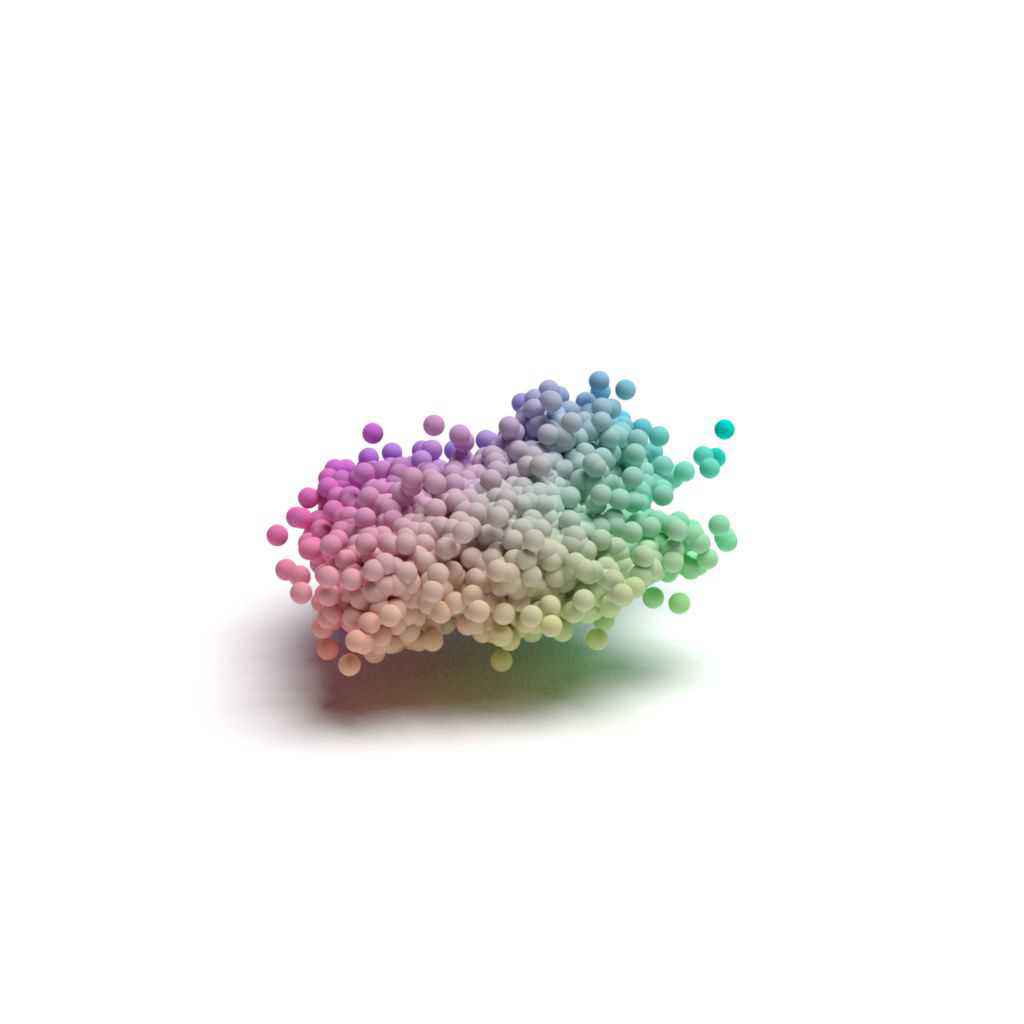}
	\includegraphics[width=\sizea]{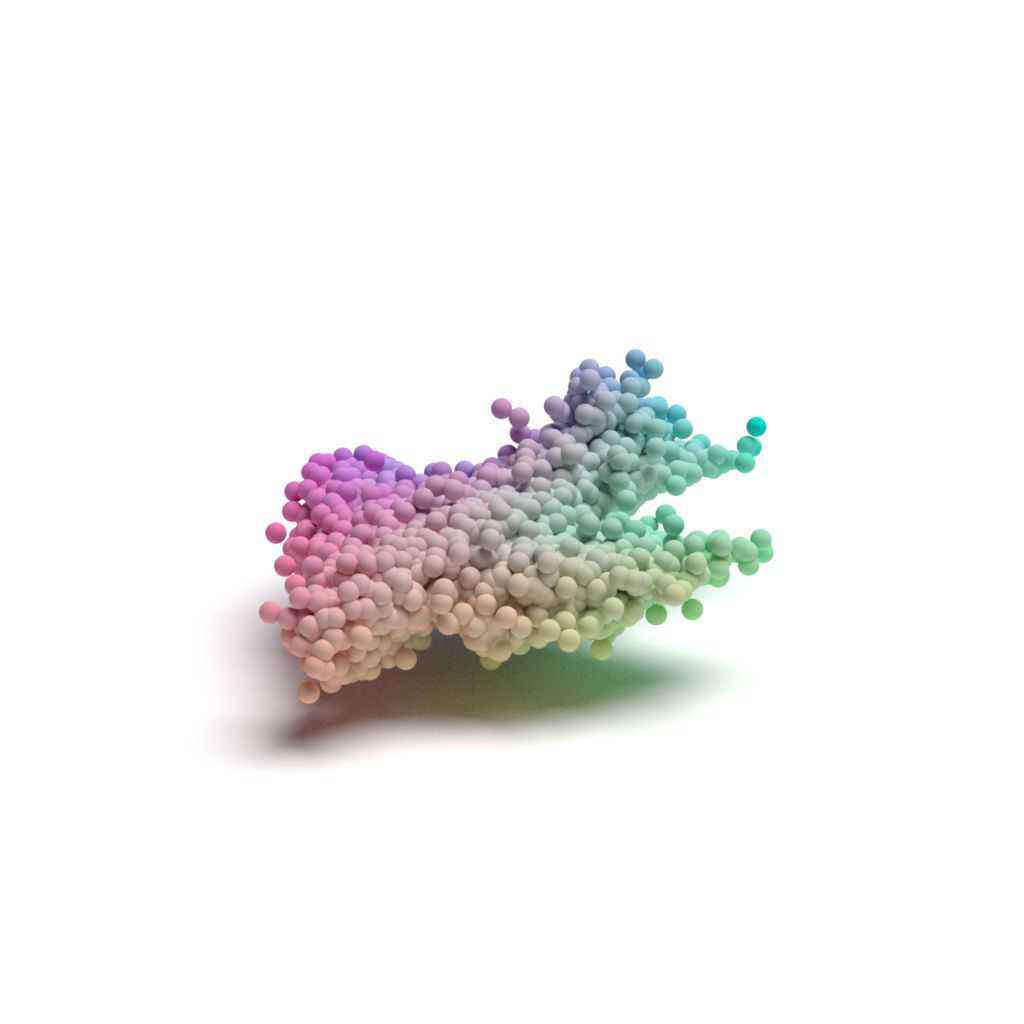}
	\includegraphics[width=\sizea]{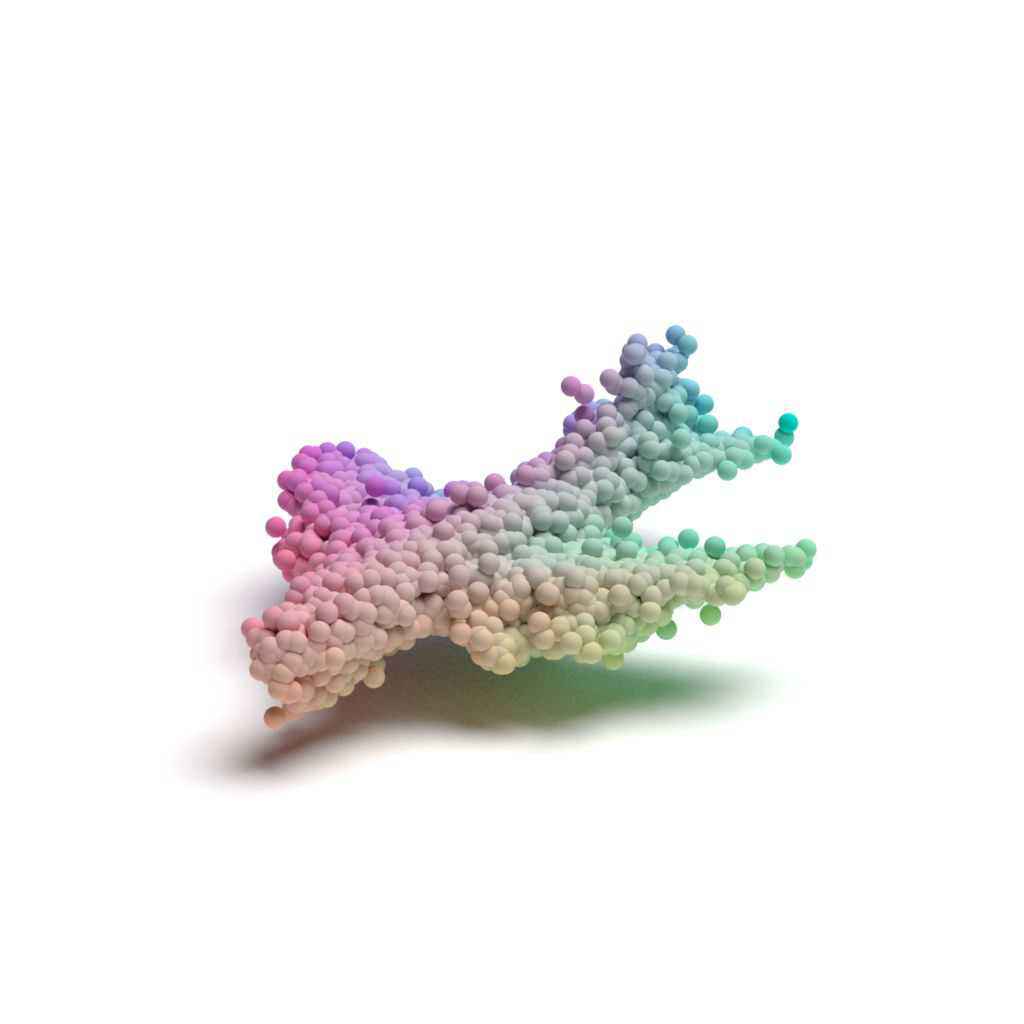}
	\includegraphics[width=\sizea]{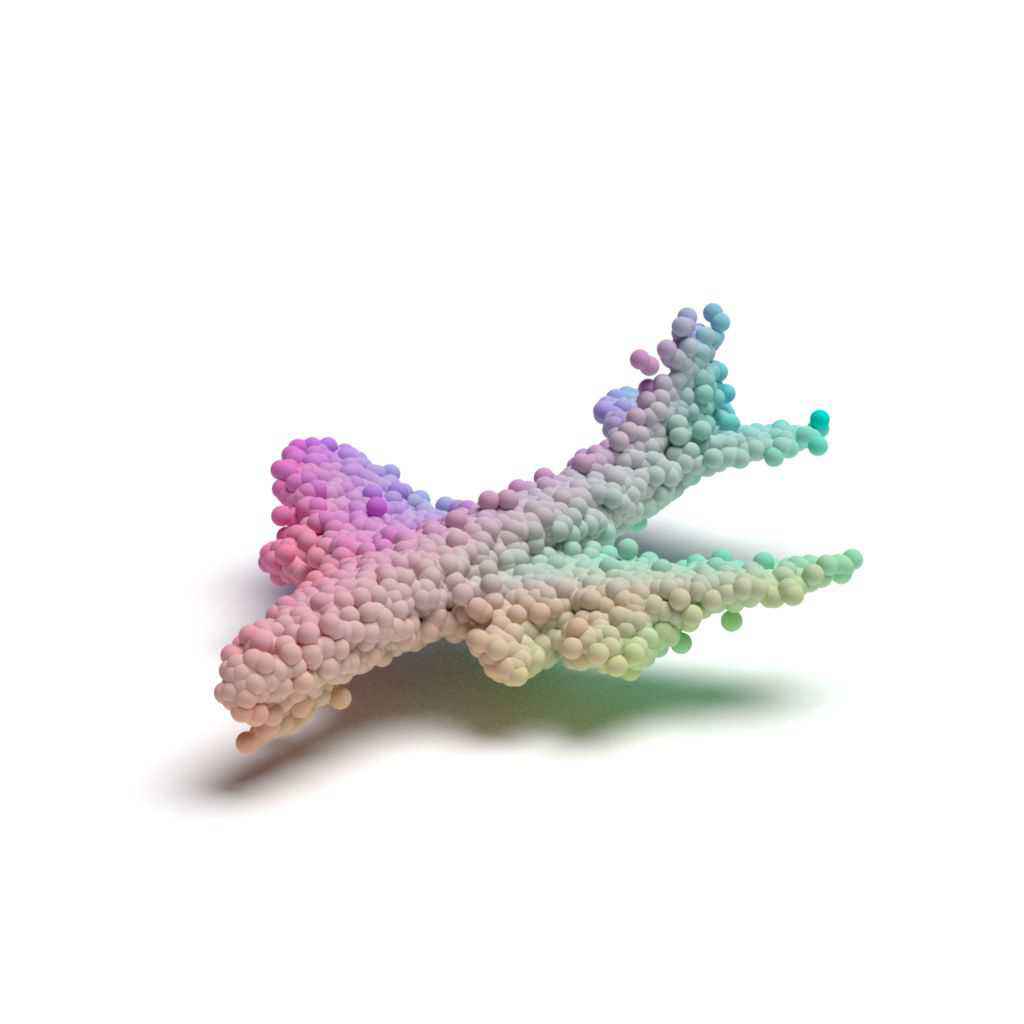}
	\includegraphics[width=\sizea]{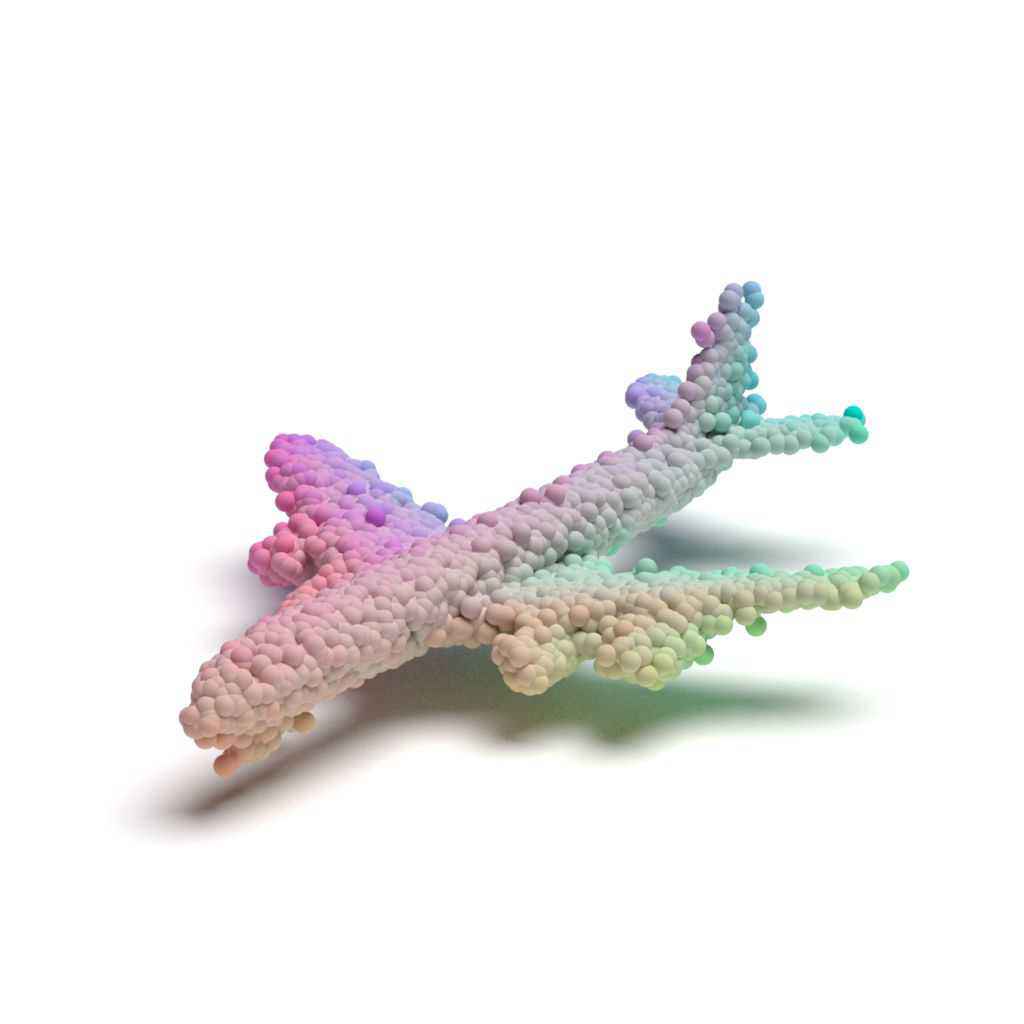}
	\includegraphics[width=\sizea]{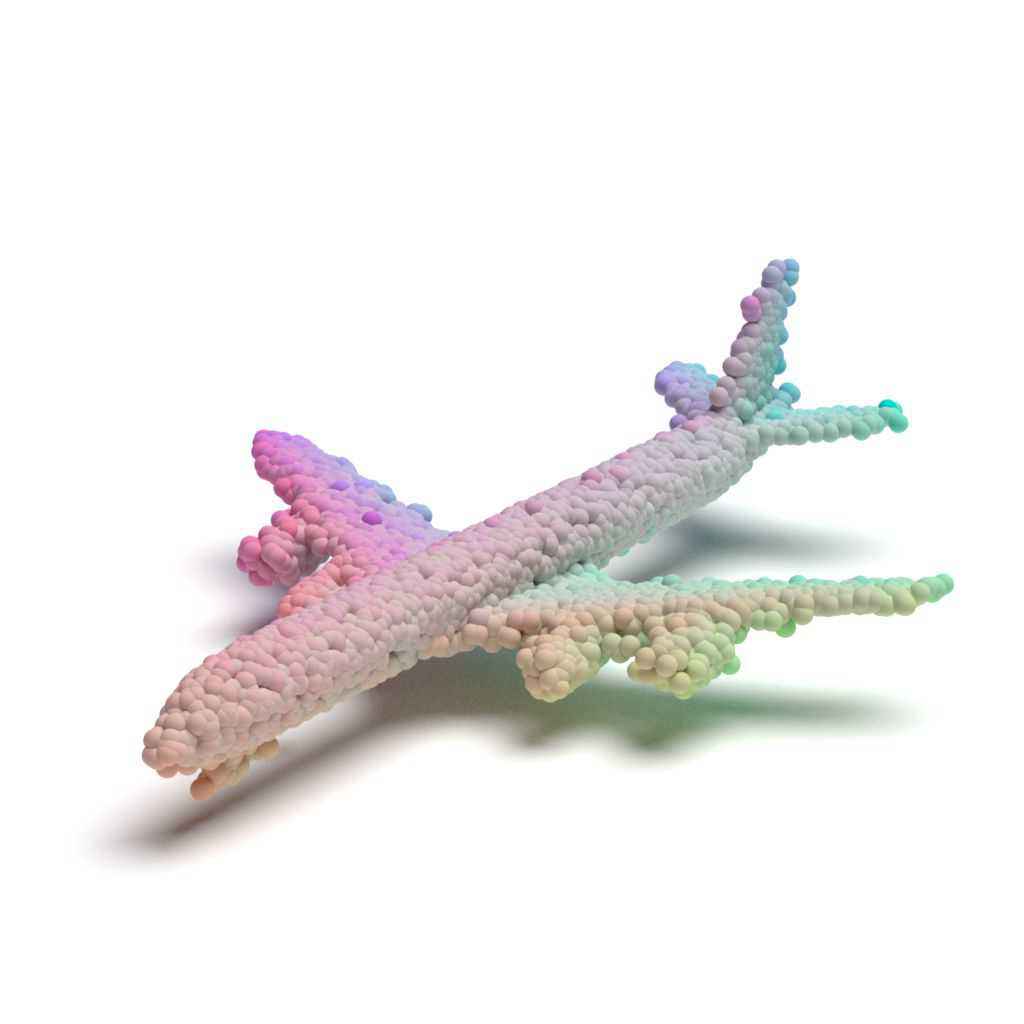}\\
	\includegraphics[width=\sizea]{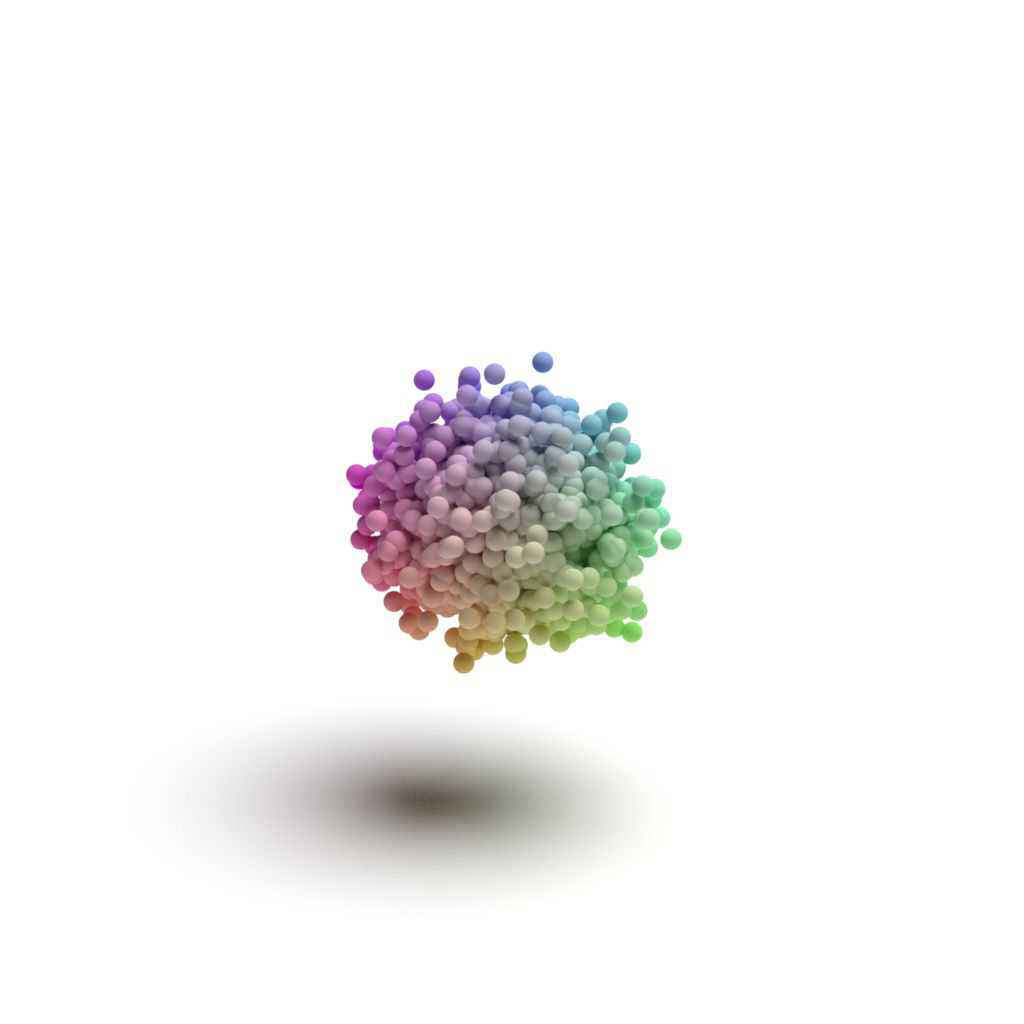}
	\includegraphics[width=\sizea]{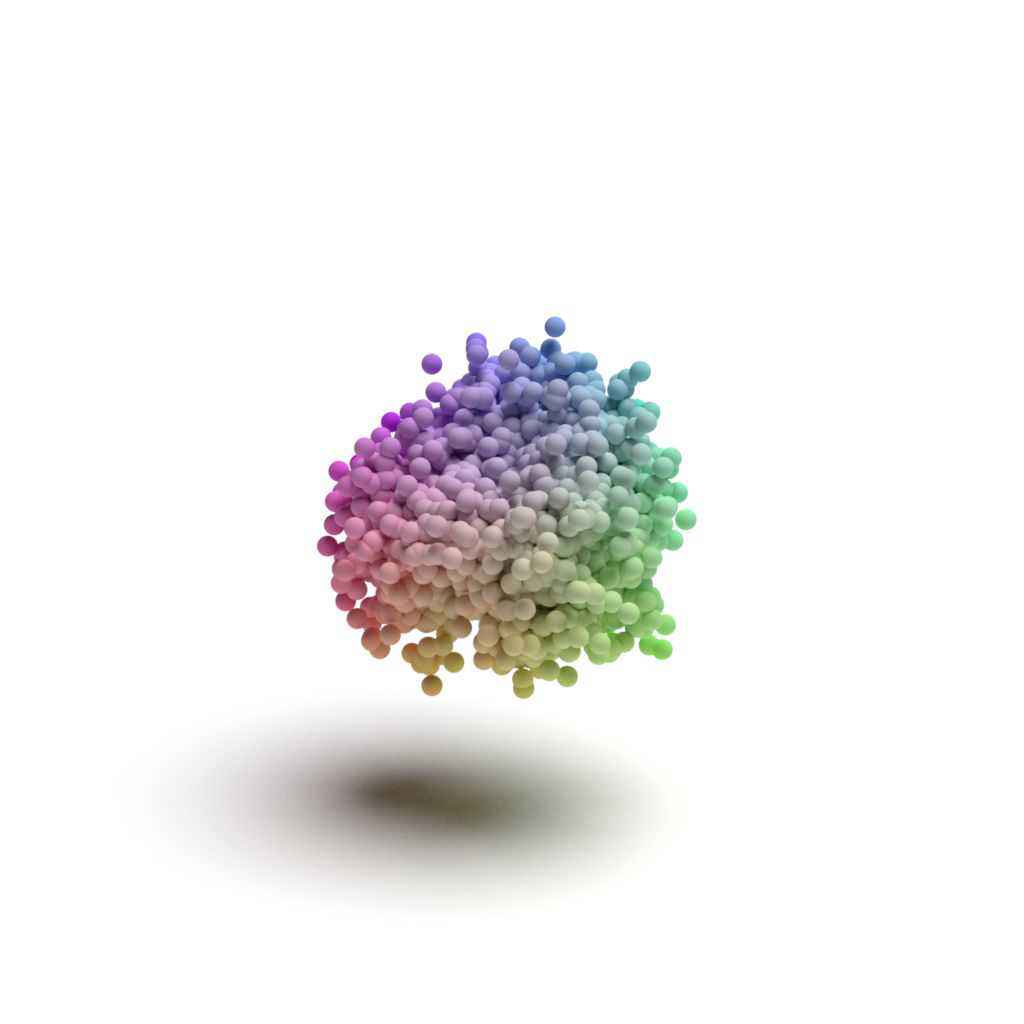}
	\includegraphics[width=\sizea]{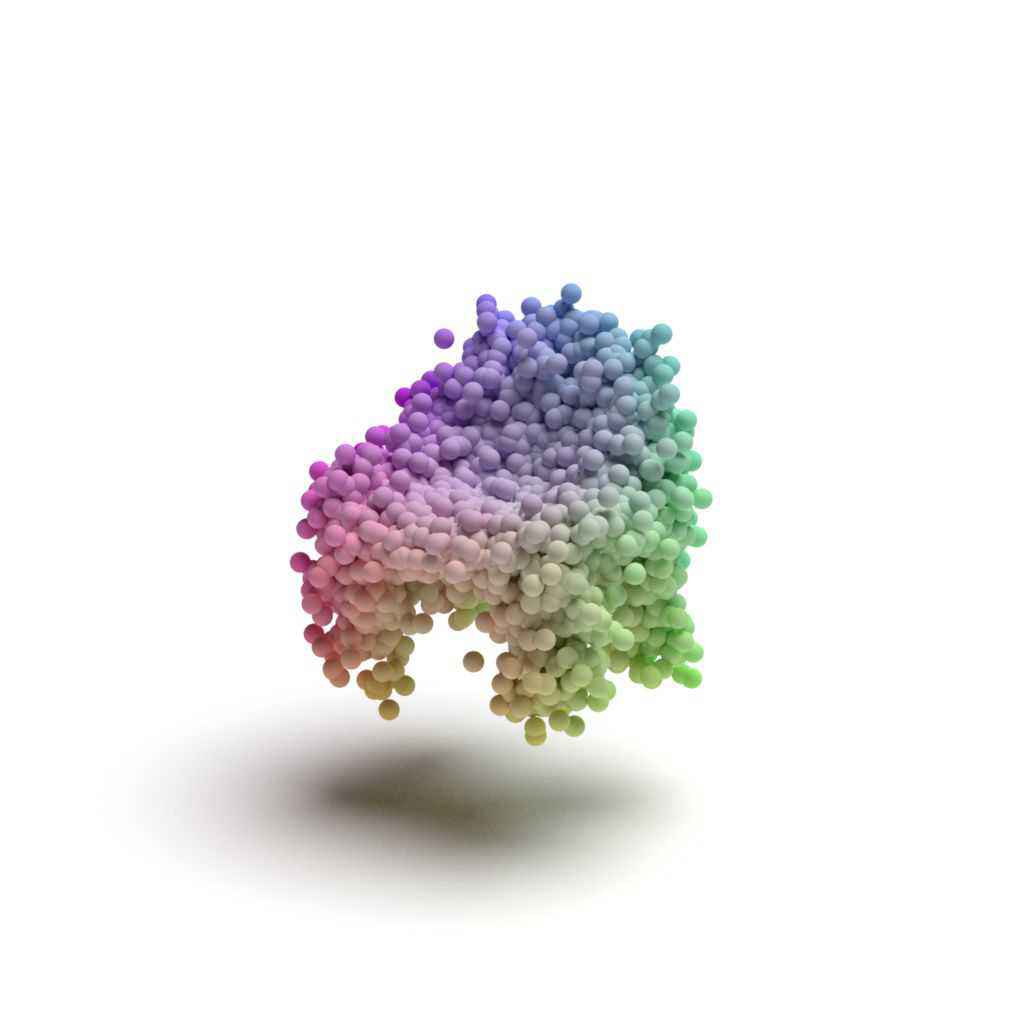}
	\includegraphics[width=\sizea]{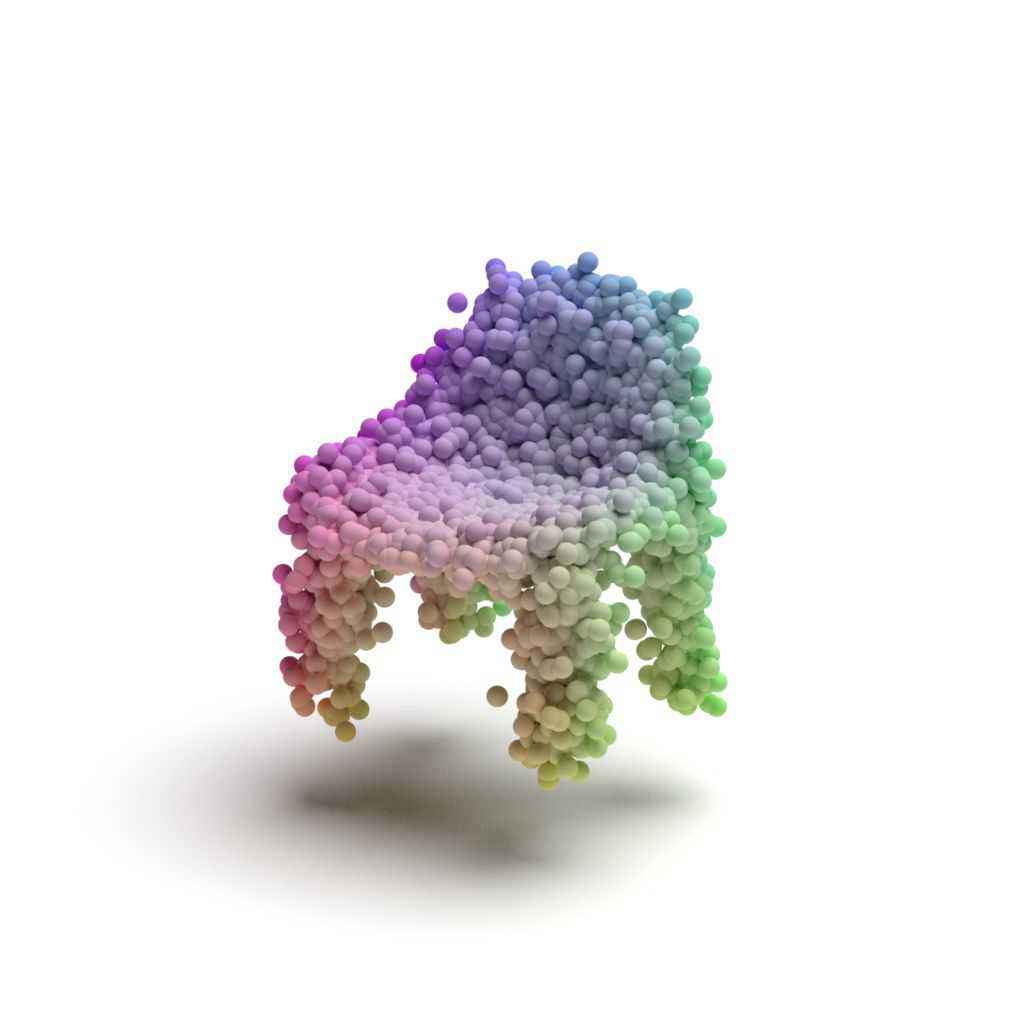}
	\includegraphics[width=\sizea]{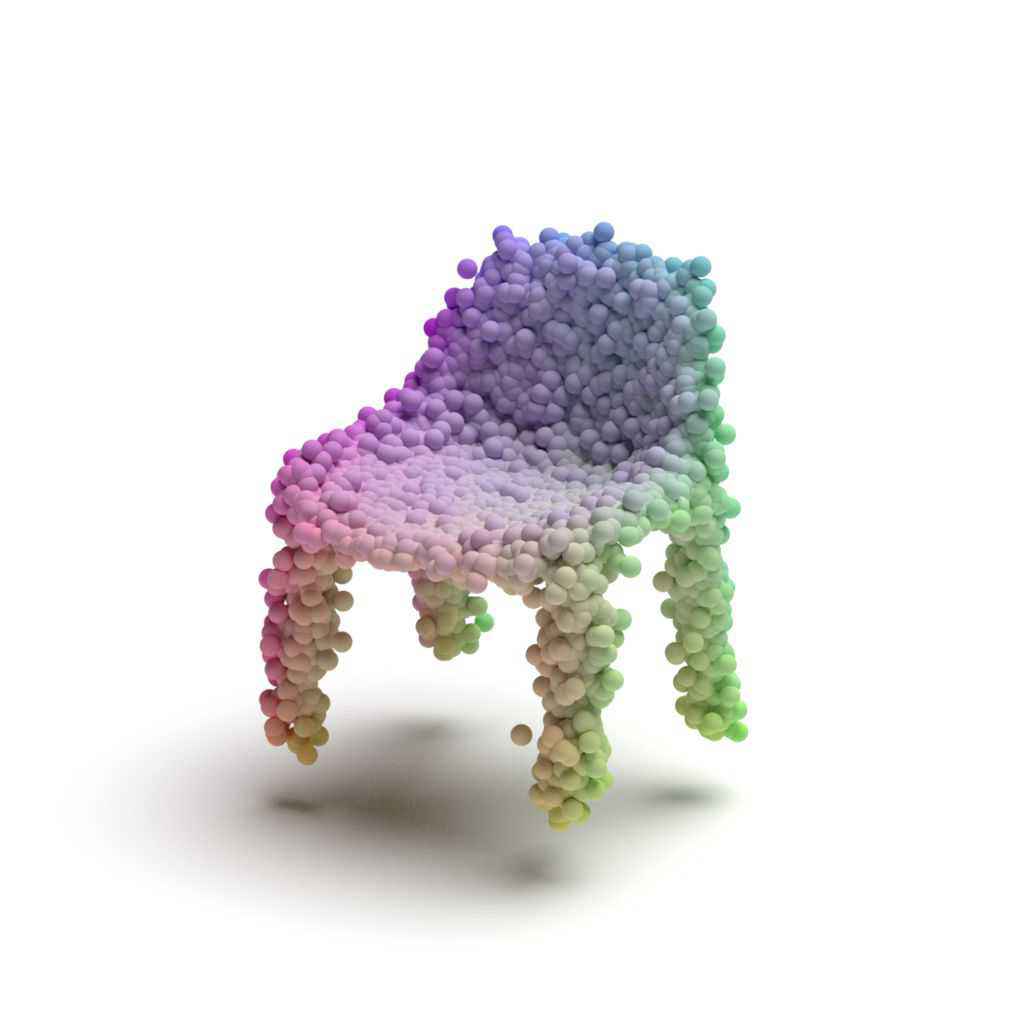}
	\includegraphics[width=\sizea]{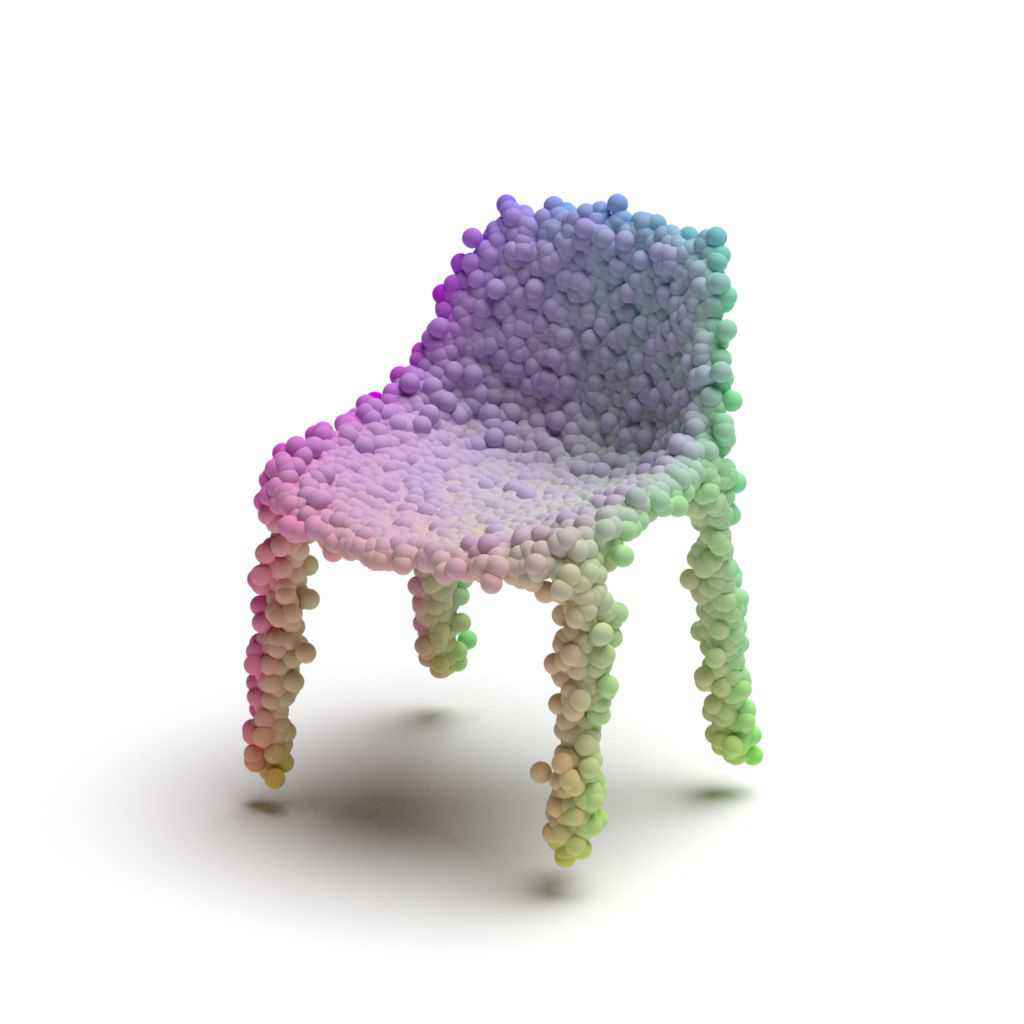}
	\includegraphics[width=\sizea]{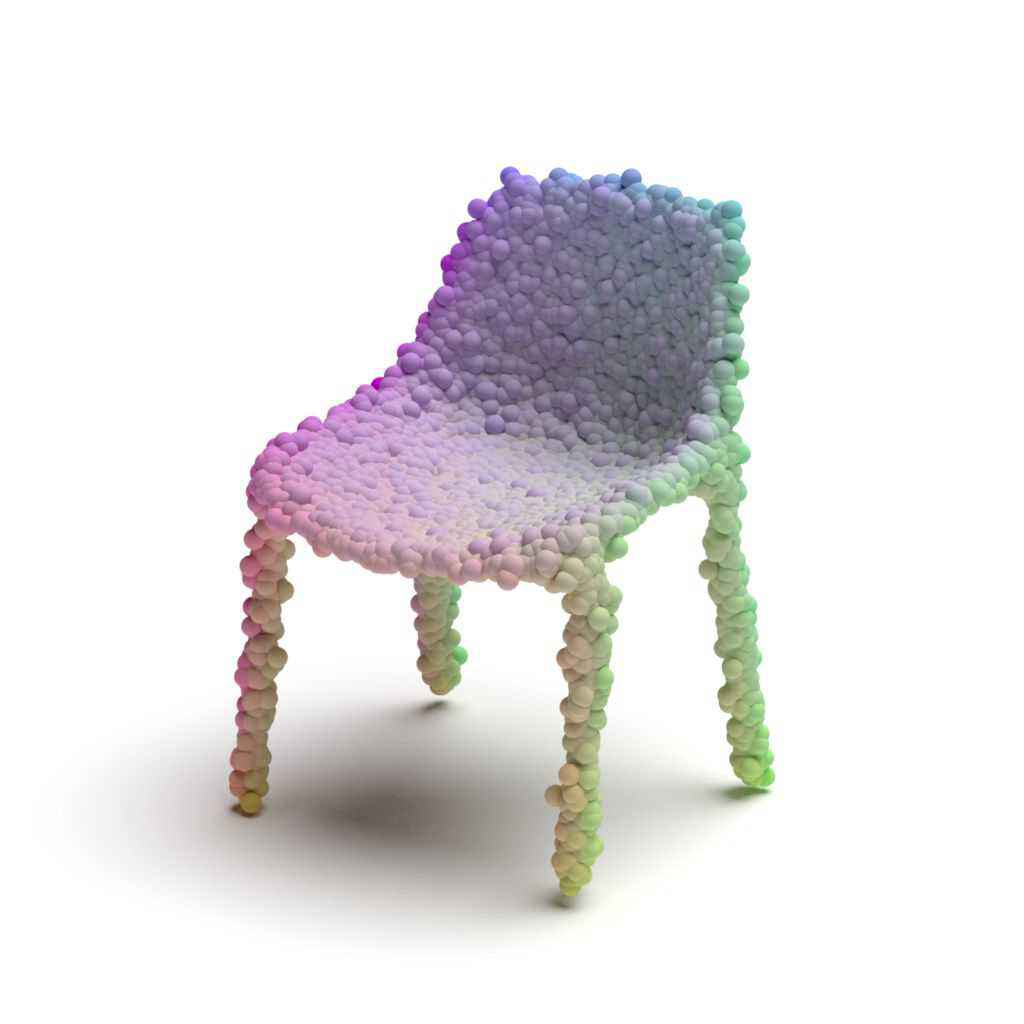}\\
	\includegraphics[width=\sizea]{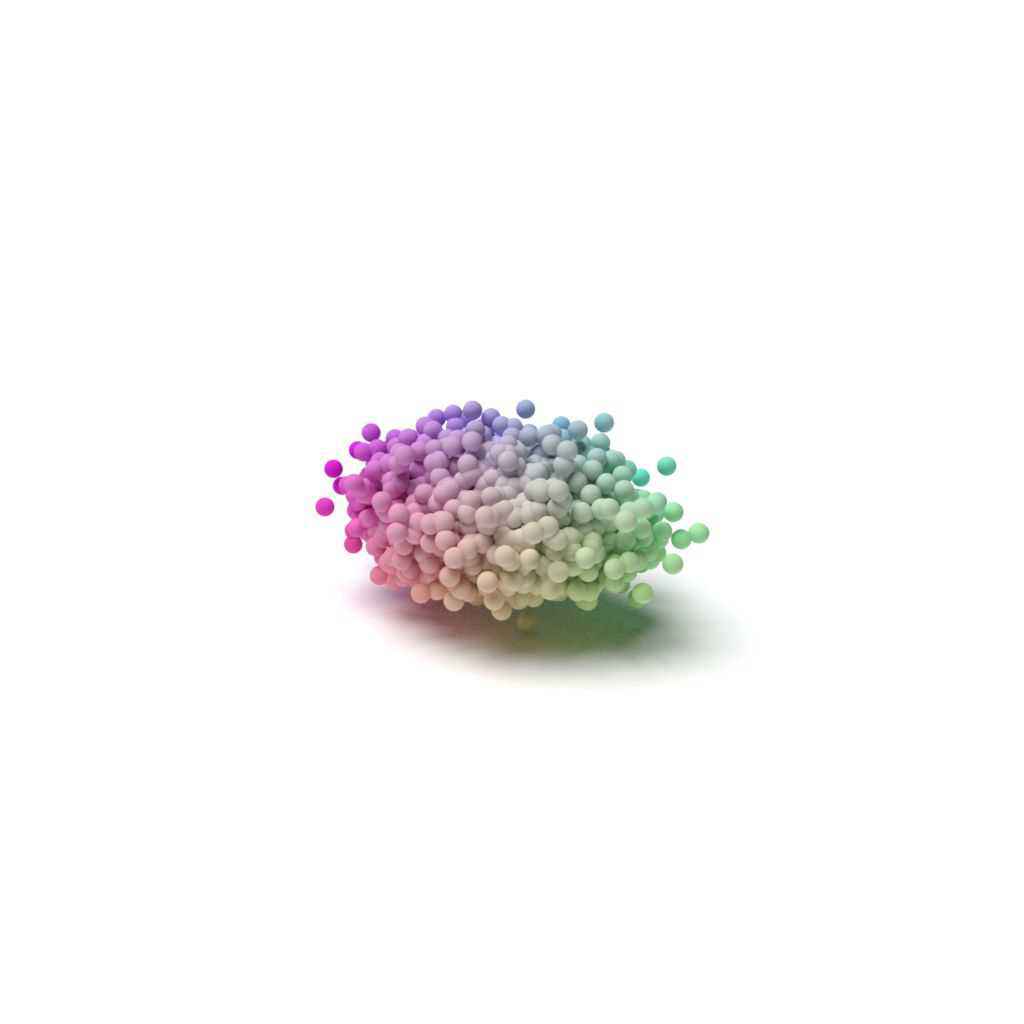}
	\includegraphics[width=\sizea]{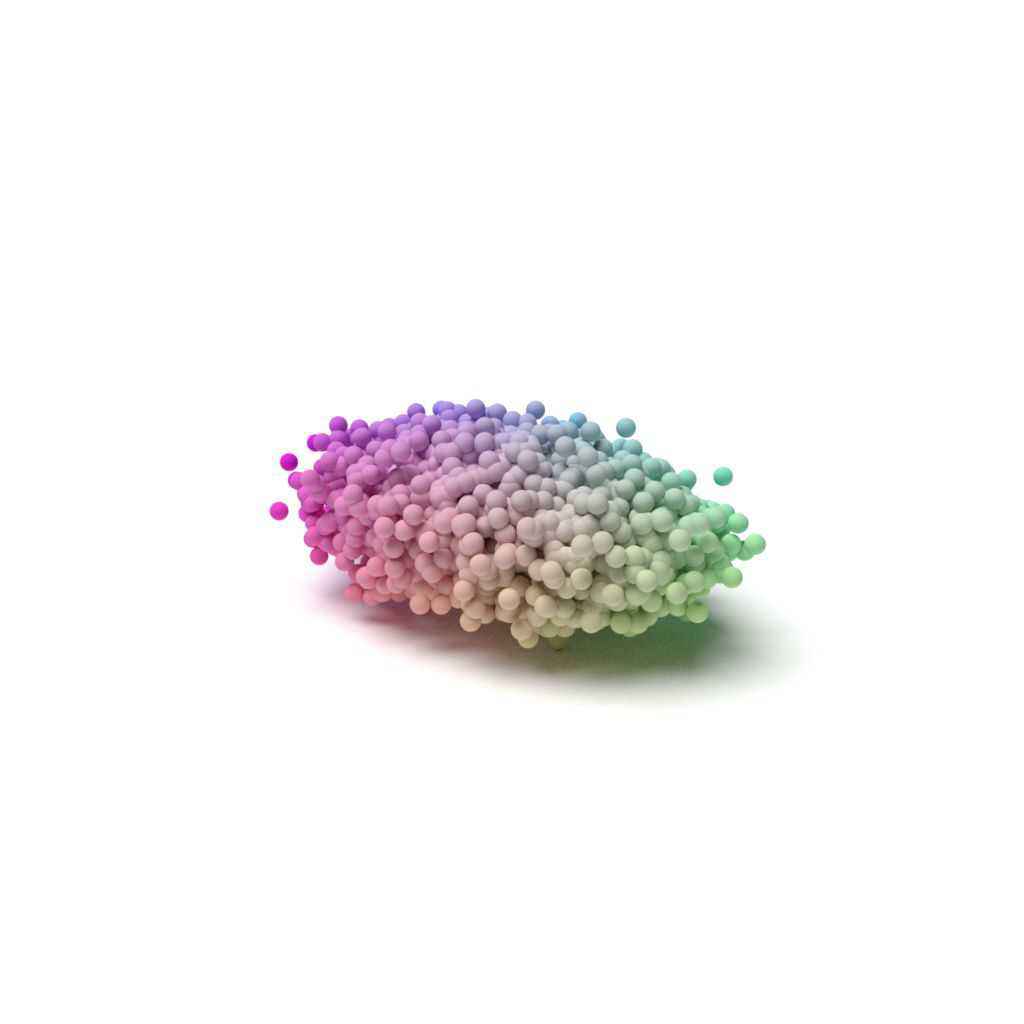}
	\includegraphics[width=\sizea]{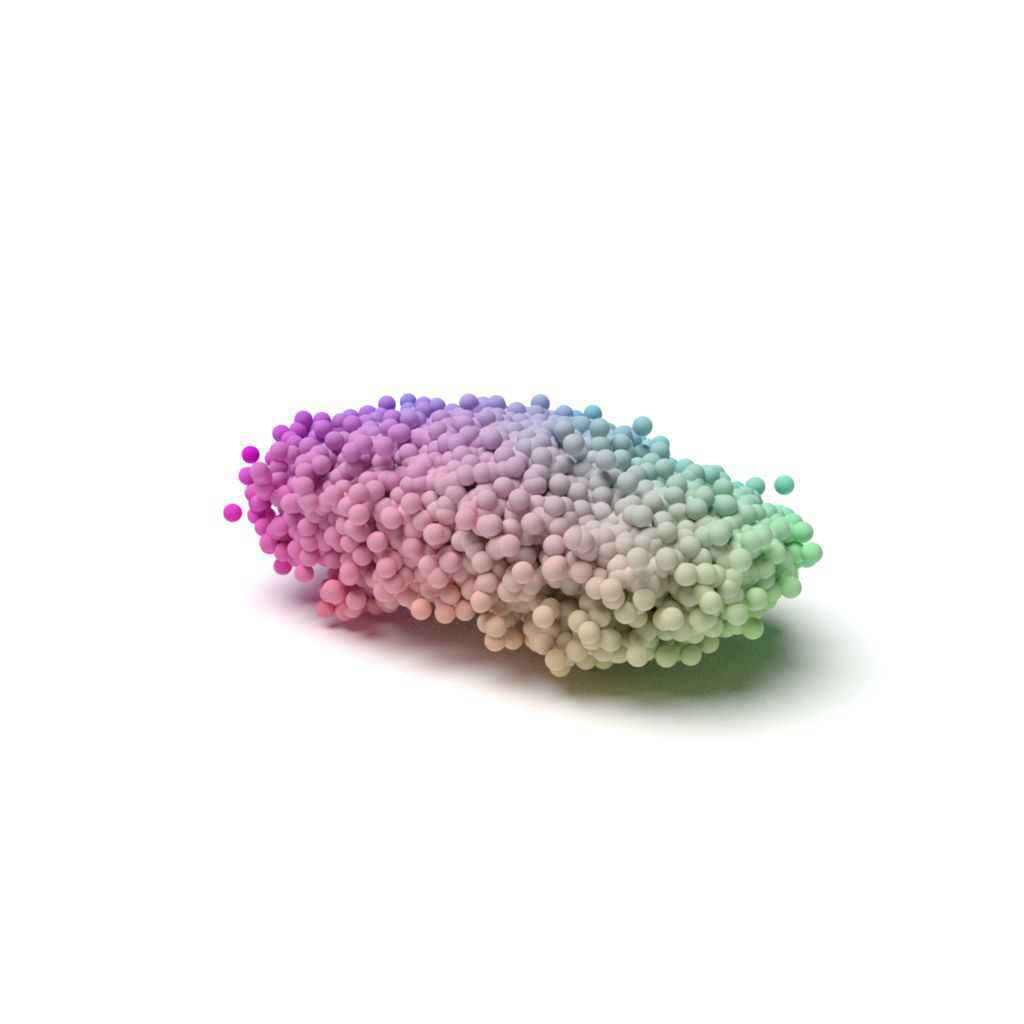}
	\includegraphics[width=\sizea]{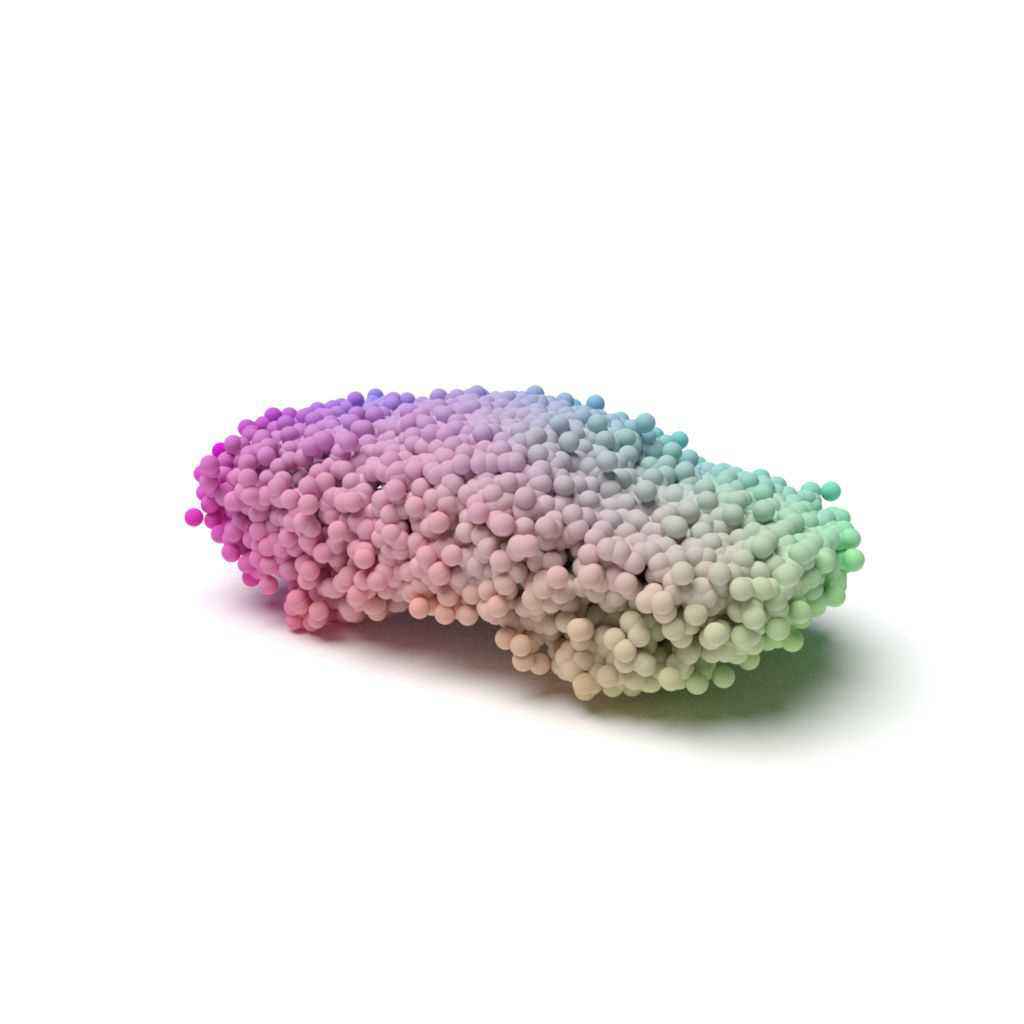}
	\includegraphics[width=\sizea]{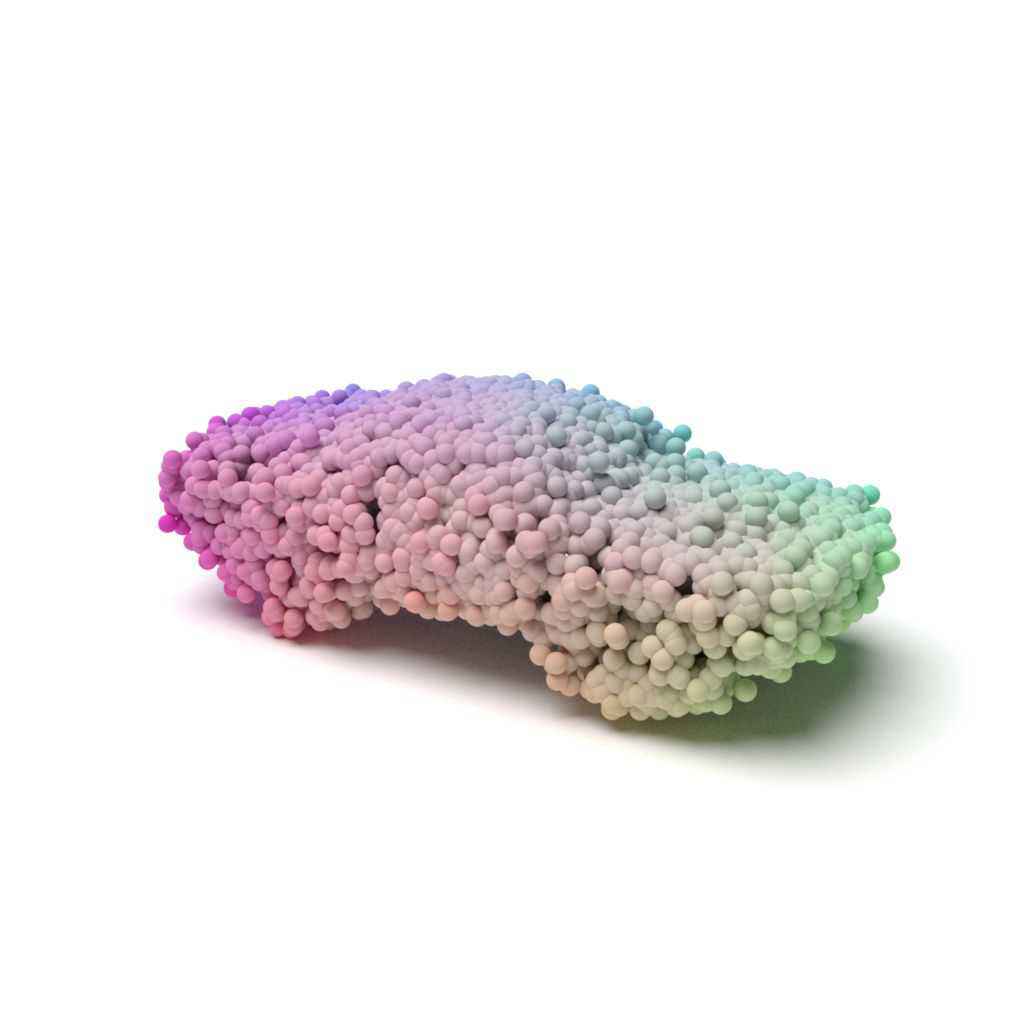}
	\includegraphics[width=\sizea]{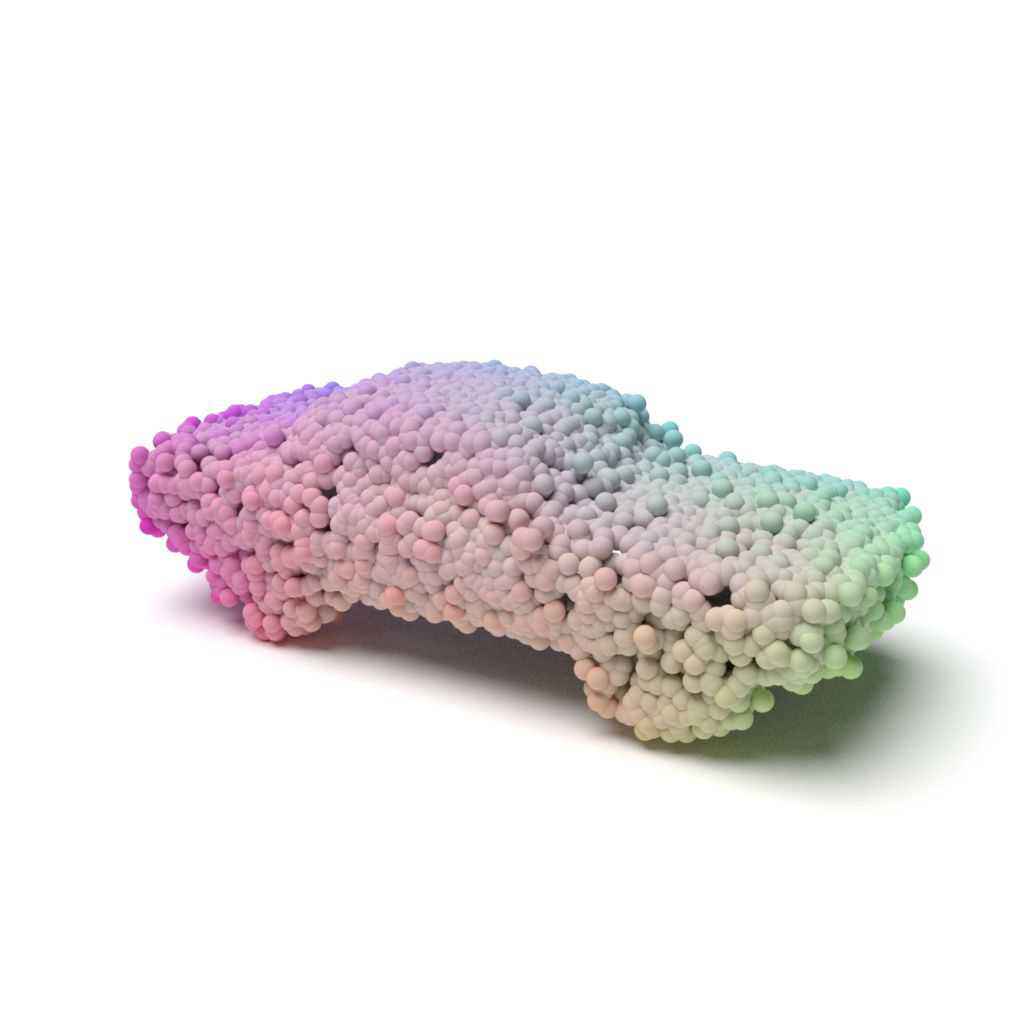}
	\includegraphics[width=\sizea]{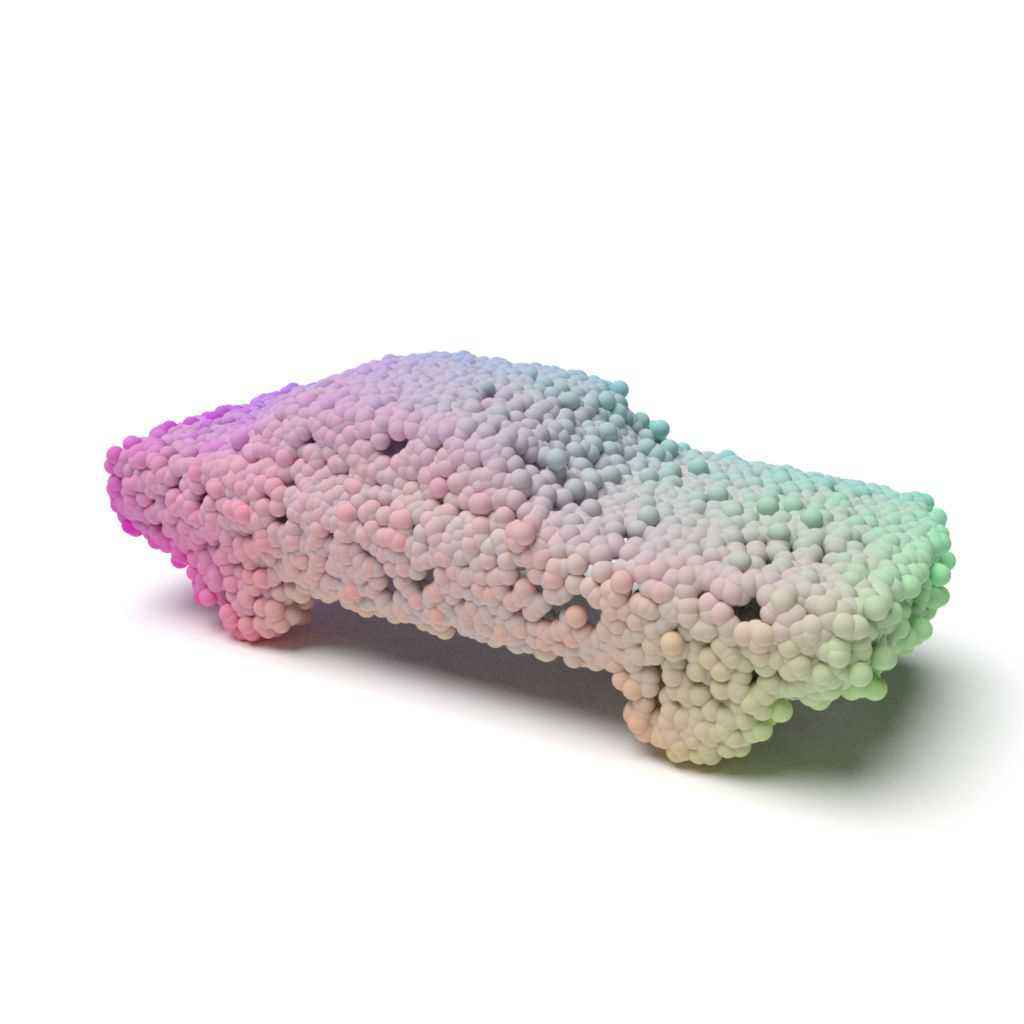}\\
	\captionof{figure}{Our model transforms points sampled from a simple prior to realistic point clouds through continuous normalizing flows. The videos of the transformations can be viewed on our project website: \url{https://www.guandaoyang.com/PointFlow/}.}
	\label{fig:teaser}
	\vspace{0.2in}}
\maketitle

\begin{abstract}
	As 3D point clouds become the representation of choice for multiple vision and graphics applications, the ability to synthesize or reconstruct high-resolution, high-fidelity point clouds becomes crucial. Despite the recent success of deep learning models in discriminative tasks of point clouds, generating point clouds remains challenging.
	This paper proposes a principled probabilistic framework to generate 3D point clouds by modeling them as a distribution of distributions. Specifically, we learn a two-level hierarchy of distributions where the first level is the distribution of shapes and the second level is the distribution of points given a shape. This formulation allows us to both sample shapes and sample an arbitrary number of points from a shape. Our generative model, named PointFlow, learns each level of the distribution with a continuous normalizing flow. The invertibility of normalizing flows enables the computation of the likelihood during training and allows us to train our model in the variational inference framework. Empirically, we demonstrate that PointFlow achieves state-of-the-art performance in point cloud generation. We additionally show that our model can faithfully reconstruct point clouds and learn useful representations in an unsupervised manner.
	The code is available at \url{https://github.com/stevenygd/PointFlow}.
\end{abstract}

\section{Introduction}\label{sec:intro}
Point clouds are becoming popular as a 3D representation because they can capture a much higher resolution than voxel grids and are a stepping stone to more sophisticated representations such as meshes. 
Learning a generative model of point clouds could benefit a wide range of point cloud synthesis tasks such as reconstruction and super-resolution, by providing a better \textit{prior} of point clouds.
However, a major roadblock in generating point clouds is the complexity of the space of point clouds.
A cloud of points corresponding to a chair is best thought of as samples from a distribution that corresponds to the surface of the chair, and the chair itself is best thought of as a sample from a distribution of chair shapes.
As a result, in order to generate a chair according to this formulation, we need to characterize a \textit{distribution of distributions}, which is under-explored by existing generative models.

In this paper, we propose PointFlow, a principled generative model for 3D point clouds that learns a distribution of distributions: the former being the distribution of shapes and the latter being the distribution of points given a shape.
Our key insight is that instead of directly parametrizing the distribution of points in a shape, we model this distribution as an \emph{invertible parameterized transformation} of 3D points from a prior distribution (e.g., a 3D Gaussian). 
Intuitively, under this model, generating points for a given shape involves sampling points from a generic Gaussian prior, and then \emph{moving} them according to this parameterized transformation to their new location in the target shape, as illustrated in Figure~\ref{fig:teaser}. 
In this formulation, a given shape is then simply the variable that parametrizes such transformation, and a category is simply a distribution of this variable.
Interestingly, we find that representing this distribution too as a transformation of a prior distribution leads to a more expressive model of shapes.
In particular, we use the recently proposed continuous normalizing flow framework to model both kinds of transformations~\cite{NormalizingFlow, neuralODE, ffjord}. 

This parameterization confers several advantages.
The invertibility of these transformations allows us to not just sample but also estimate probability densities.
The ability to estimate probability densities in turn allows us to train these models in a principled manner using the variational inference framework~\cite{VAE}, where we maximize a variational lower bound on the log-likelihood of a training set of point clouds.
This probabilistic framework for training further lets us avoid the complexities of training GANs or hand-crafting good distance metrics for measuring the difference between two sets of points. Experiments show that PointFlow outperforms previous state-of-the-art generative models of point clouds, and achieves compelling results in point cloud reconstruction and unsupervised feature learning.

\section{Related work}
\noindent\textbf{Deep learning for point clouds.}
Deep learning has been introduced to improve performance in various point cloud discriminative tasks including classification~\cite{pointnet,pointnet++,foldingnet,deepsets}, segmentation~\cite{pointnet,shoef2019pointwise}, and critical points sampling~\cite{learning2sample}.
Recently, substantial progress has been made in point cloud synthesis tasks such as auto-encoding~\cite{achlioptas_L3DP, foldingnet, atlasnet}, single-view 3D reconstruction~\cite{pointsetgen,GALGA,DeformNetFD,EfficientDP,MultiresolutionTN}, stereo reconstruction~\cite{usenko2015reconstructing}, and point cloud completion~\cite{PU-net, EC-Net}.
Many point cloud synthesis works convert a point distribution to a $N\times 3$ matrix by sampling $N$ ($N$ is pre-defined) points from the distribution so that existing generative models are readily applicable.
For example, Gadelha \etal~\cite{MultiresolutionTN} apply variational auto-encoders~(VAEs)~\cite{VAE} and Zamorski \etal~\cite{3D-AAE} apply adversarial auto-encoders~(AAEs)~\cite{AAE} to point cloud generation.
Achlioptas \etal~\cite{achlioptas_L3DP} explore generative adversarial networks~(GANs)~\cite{GAN,WGAN,iWGAN} for point clouds in both raw data space and latent space of a pre-trained auto-encoder. In the above methods, the auto-encoders are trained with heuristic loss functions that measure the distance between two point sets, such as Chamfer distance~(CD) or earth mover's distance~(EMD).
Sun~\etal~\cite{pointgrow} apply auto-regressive models~\cite{pixelCNN} with a discrete point distribution to generate one point at a time, also using a fixed number of points per shape. 

However, treating a point cloud as a fixed-dimensional matrix has several drawbacks. 
First, the model is restricted to generate a fixed number of points. 
Getting more points for a particular shape requires separate up-sampling models such as \cite{PU-net, EC-Net, patch-based-upsampling}.
Second, it ignores the permutation invariance property of point sets, which might lead to suboptimal parameter efficiency.
Heuristic set distances are also far from ideal objectives from a generative modeling perspective since they make the original probabilistic interpretation of VAE/AAE no longer applicable when used as the reconstruction objective. In addition, exact EMD is slow to compute while approximations could lead to biased or noisy gradients. CD has been shown to incorrectly favor point clouds that are overly concentrated in the mode of the marginal point distribution~\cite{achlioptas_L3DP}.

Some recent works introduce sophisticated decoders consisting of a cascade~\cite{foldingnet} or a mixture~\cite{atlasnet} of smaller decoders to map one (or a mixture of) 2-D uniform distribution(s) to the target point distribution, overcoming the shortcomings of using a fixed number of points. However, they still rely on heuristic set distances that lack a probabilistic guarantee. Also, their methods only learn the distribution of points for each shape, but not the distribution of shapes. Li~\etal~\cite{PC-GAN} propose a ``sandwiching'' reconstruction objective that combines a variant of WGAN~\cite{WGAN} loss with EMD. They also train another GAN in the latent space to learn shape distribution, similar to Achlioptas \etal~\cite{achlioptas_L3DP}. In contrast, our method is simply trained end-to-end by maximizing a variational lower bound on the log-likelihood, does not require multi-stage training, and does not have any instability issues common for GAN-based methods.

\vspace{1mm}
\noindent\textbf{Generative models.}
There are several popular frameworks of deep generative models, including generative adversarial networks~\cite{GAN,WGAN,StyleGAN}, variational auto-encoders~\cite{VAE,stochastic_bp}, auto-regressive models~\cite{PixelRNN,pixelCNN}, and flow-based models~\cite{Nice,NormalizingFlow,RealNVP,GLOW}.
In particular, flow-based models and auto-regressive models can both perform exact likelihood evaluation, while flow-based models are much more efficient to sample from. Flow-based models have been successfully applied to a variety of generation tasks such as image generation~\cite{GLOW,RealNVP,Nice}, video generation~\cite{VideoFlow}, and
voice synthesis~\cite{WaveGlow}.
Also, there has been recent work that combines flows with other generative models, such as GANs~\cite{FlowGAN,danihelka2017comparison}, auto-regressive models~\cite{NAF,MAF,IAF}, and VAEs~\cite{IAF,SylvesterNF,VLAE,NormalizingFlow,SylvesterNF,neuralODE,ffjord}.

Most existing deep generative models aim at learning the distribution of fixed-dimensional variables. Learning the \textit{distribution of distributions}, where the data consists of a \textit{set of sets}, is still underexplored. Edwards and Storkey~\cite{nerual_stats} propose a hierarchical VAE named Neural Statistician that consumes a set of sets. They are mostly interested in the few-shot case where each set only has a few samples. Also, they are focused on classifying sets or generating new samples from a given set. While our method is also applicable to these tasks, our focus is on learning the distribution of sets and generating new sets~(point clouds in our case). In addition, our model employs a tighter lower bound on the log-likelihood, thanks to the use of normalizing flow in modeling both the reconstruction likelihood and the prior.

\section{Overview}
\label{sec:overview}
Consider a set of shapes $\mathcal{X}=\{X_i\}_{i=1}^{N}$ from a particular class of object, where each shape is represented as a set of 3D points $X_i = \{x_j^i\}_{j=1}^{M_i}$.
As discussed in Section~\ref{sec:intro}, each point $x_j^i \in \mathbb{R}^3$ is best thought of as being sampled from a point distribution $Q^i(x)$, usually a uniform distribution over the surface of an object $X_i$.
Each shape $X_i$ is itself a sample from a distribution over shapes $Q(X)$ that captures what shapes in this category look like.

Our goal is to learn the distribution of shapes, each shape itself being a distribution of points. In other words, our generative model should be able to both sample shapes and sample an arbitrary number of points from a shape. 

We propose to use continuous normalizing flows to model the distribution of points given a shape. A continuous normalizing flow can be thought of as a vector field in the $3$-D Euclidean space, which induces a distribution of points through transforming a generic prior distribution~(\eg, a standard Gaussian). To sample points from the induced distribution, we simply sample points from the prior and move them according to the vector field. Moreover, the continuous normalizing flow is invertible, which means we can move data points back to the prior distribution to compute the exact likelihood. This model is highly intuitive and interpretable, allowing a close inspection of the generative process as shown in Figure~\ref{fig:teaser}.

We parametrize each continuous normalizing flow with a latent variable that represents the shape.
As a result, modeling the distribution of shapes can be reduced to modeling the distribution of the latent variable. Interestingly, we find continuous normalizing flow also effective in modeling the latent distribution.
Our full generative model thus consists of two levels of continuous normalizing flows, one modeling the shape distribution by modeling the distribution of the latent variable, and the other modeling the point distribution given a shape. 

In order to optimize the generative model, we construct a variational lower bound on the log-likelihood by introducing an inference network that infers a latent variable distribution from a point cloud. Here, we benefit from the fact that the invertibility of the continuous normalizing flow enables likelihood computation. This allows us to train our model end-to-end in a stable manner, unlike previous work based on GANs that requires two-stage training~\cite{achlioptas_L3DP,PC-GAN}. As a side benefit, we find the inference network learns a useful representation of point clouds in an unsupervised manner. 

In Section~\ref{sec:background} we introduce some background on continuous normalizing flows and variational auto-encoders. We then describe our model and training in detail in Section~\ref{sec:objectives}.

\section{Background}
\label{sec:background}
\subsection{Continuous normalizing flow}\label{sec:method-CNF}

A normalizing flow~\cite{NormalizingFlow} is a series of invertible mappings that transform an initial known distribution to a more complicated one.
Formally, let $f_1, \dots, f_n$ denote a series of invertible transformations we want to apply to a latent variable $y$ with a distribution $P(y)$. $x = f_n \circ f_{n-1} \circ \dots \circ f_1(y)$ is the output variable. Then the probability density of the output variable is given by the change of variables formula:
\begin{align}
\log{P(x)} &= \log{P(y)} - \sum_{k=1}^n \log{\left|\det{\frac{\partial f_k}{\partial y_{k-1}}}\right|}\,,
\label{eq:flow-logprob}
\end{align}
where $y$ can be computed from $x$ using the inverse flow: $y = f_1^{-1} \circ \dots \circ f_{n}^{-1}(x)$. In practice, $f_1, \dots, f_n$ are usually instantiated as neural networks with an architecture that makes the determinant of the Jacobian $\left|\det{\frac{\partial f_k}{\partial y_{k-1}}}\right|$ easy to compute. 
The normalizing flow has been generalized from a discrete sequence to a continuous transformation~\cite{ffjord,neuralODE} by defining the transformation $f$ using a continuous-time dynamic $\frac{\partial y(t)}{\partial t} = f(y(t), t)$, where $f$ is a neural network that has an unrestricted architecture. The continuous normalizing flow~(CNF) model for $P(x)$ with a prior distribution $P(y)$ at the start time can be written as:
\begin{align}
x &= y(t_0) + \int_{t_0}^{t_1} f(y(t),t) d t, \quad y(t_0) \sim P(y) \nonumber \\
\log{P(x)} &= \log{P(y(t_0))} - \int_{t_0}^{t_1} \operatorname{Tr}\left(\frac{\partial f}{\partial y
	(t)}\right)dt \label{eq:cnf-prob}
\end{align}
and $y(t_0)$ can be computed using the inverse flow $y(t_0) = x + \int_{t_1}^{t_0} f(y(t), t) dt$. A black-box ordinary differential equation~(ODE) solver can been applied to estimate the outputs and the input gradients of a continuous normalizing flow~\cite{ffjord,neuralODE}.

\subsection{Variational auto-encoder}
Suppose we have a random variable $X$ that we are building generative models for.
The variational auto-encoder (VAE) is a framework that allows one to learn $P(X)$ from a dataset of observations of $X$~\cite{VAE,stochastic_bp}.
The VAE models the data distribution via a latent variable $z$ with a prior distribution $P_\psi(z)$, and a decoder $P_\theta(X|z)$ which captures the (hopefully simpler) distribution of $X$ given $z$.
During training, it additionally learns an inference model (or encoder) $Q_\phi(z|X)$. The encoder and decoder are jointly trained to maximize a lower bound on the log-likelihood of the observations{\small 
	\begin{align}
	\log{P_\theta(X)} &\geq \log{P_\theta(X)} - D_{KL}(Q_\phi(z|X) || P_\theta(z|X)) \nonumber \\
	&= \mathbb{E}_{Q_\phi(z|x)}\left[\log{P_\theta(X|z)}\right] - D_{KL}\left(Q_\phi(z|X)||P_\psi(z)\right) \nonumber \\
	&\triangleq \mathcal{L}(X;\phi,\psi,\theta)\,, \label{eq:elbo}
	\end{align}}which is also called the evidence lower bound (ELBO). 
One can interpret ELBO as the sum of the negative reconstruction error (the first term) and a latent space regularizer (the second term). In practice, $Q_\phi(z|X)$ is usually modeled as a diagonal Gaussian $\mathcal{N}(z|\mu_\phi(X), \sigma_\phi(X))$ whose mean and standard deviation are predicted by a neural network with parameters $\phi$.
To efficiently optimize the ELBO, sampling from $Q_\phi(z|X)$ is done by reparametrizing $z$ as $z=\mu_\phi(X)+\sigma_\phi(X)\cdot\epsilon$, where $\epsilon\sim\mathcal{N}(0, \mathit{I})$. 

\section{Model}\label{sec:objectives}
We now have the paraphernalia needed to define our generative model of point clouds.
Using the terminology of the VAE, we need three modules: the encoder $Q_\phi(z|X)$ that encodes a point cloud into a shape representation $z$, a prior $P_\psi(z)$ over shape representations, and a decoder $P_\theta(X|z)$ that models the distribution of points given the shape representation.
We use a simple permutation-invariant encoder to predict $Q_\phi(z|X)$, following the architecture in Achlioptas~\etal~\cite{achlioptas_L3DP}.
We use continuous normalizing flows for both the prior $P_\psi(z)$ and the generator $P_\theta(X|z)$, which are described below.

\subsection{Flow-based point generation from shape representations}\label{sec:method-likelihood}
We first decompose the reconstruction log-likelihood of a point set into the sum of log-likelihood of each point
\begin{equation}
	\log{P_\theta(X|z)} = \sum_{x\in X} \log{P_\theta(x|z)}\,.
	\label{eq:decompose}
\end{equation}
We propose to model $P_\theta(x|z)$ using a conditional extension of CNF.
Specifically, a point $x$ in the point set $X$ is the result of transforming some point $y(t_0)$ in the prior distribution $P(y) = \mathcal{N}(0, \mathit{I})$ using a CNF conditioned on $z$:
{\small \begin{equation}
x = G_\theta(y(t_0); z) \triangleq y(t_0) + \int_{t_0}^{t_1} g_\theta(y(t), t, z)dt, y(t_0)\sim P(y)\,,\nonumber
\end{equation}}where $g_\theta$ defines the continuous-time dynamics of the flow $G_\theta$ conditioned on $z$.
Note that the inverse of $G_\theta$ is given by $G_\theta^{-1}(x; z) = x + \int_{t_1}^{t_0} g_\theta(y(t), t, z)dt$ with $y(t_1) = x$.
The reconstruction likelihood of follows equation~\eqref{eq:cnf-prob}:
\begin{equation}
\log{P_\theta(x|z)} = \log{P(G_\theta^{-1}(x;z))} - \int_{t_0}^{t_1} \operatorname{Tr}\left(\frac{\partial g_\theta}{\partial y(t)}\right) dt\,.
\label{eq:recon}
\end{equation}
Note that $\log{P(G_\theta^{-1}(x;z))}$ can be computed in closed form with the Gaussian prior.

\begin{figure*}[t]
	\centering
	\includegraphics[width=\linewidth]{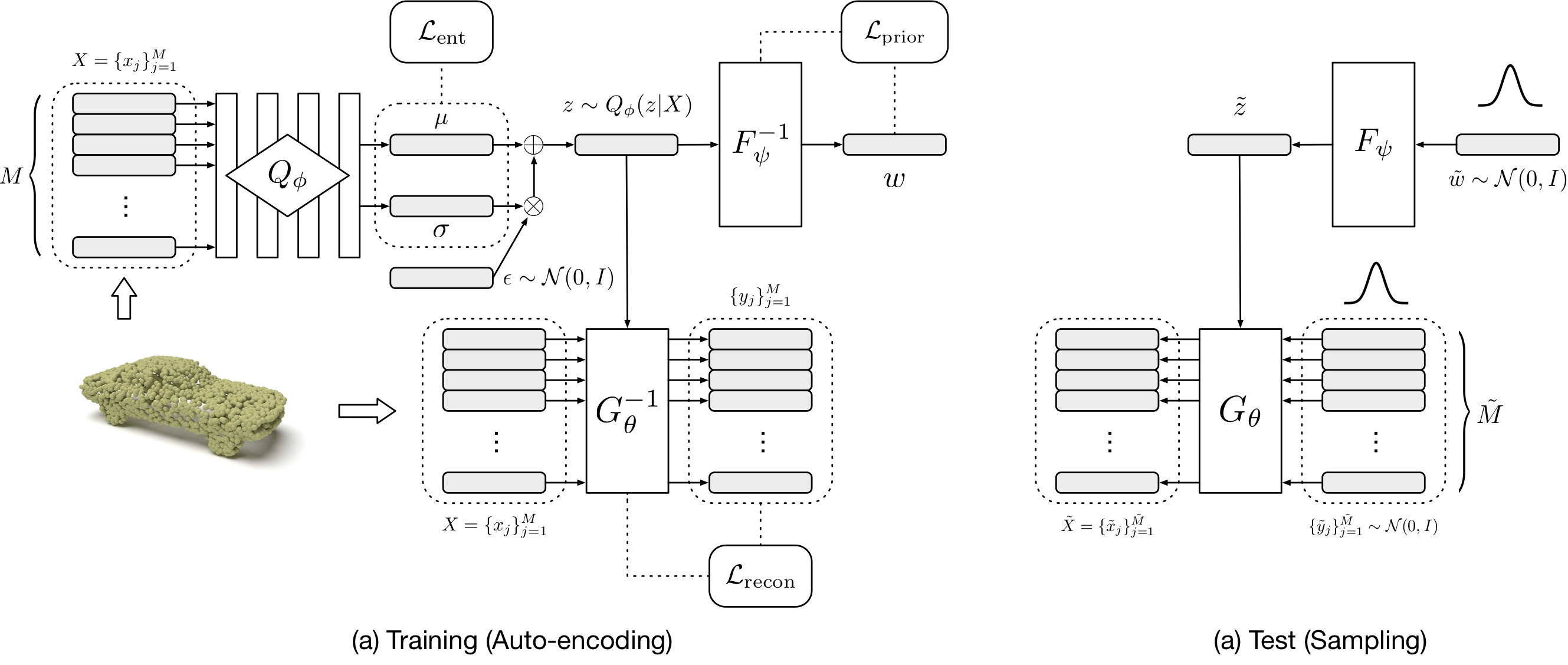}
	\caption{
		Model architecture. (a) At training time, the encoder $Q_\phi$ infers a posterior over shape representations given an input point cloud $X$, and samples a shape representation $z$ from it. We then compute the probability of $z$ in the prior distribution~($\mathcal{L}_{\text{prior}}$) through a inverse CNF $F_{\psi}^{-1}$, and compute the reconstruction likelihood of $X$~($\mathcal{L}_{\text{recon}}$) through another inverse CNF $G_{\theta}^{-1}$ conditioned on $z$. The model is trained end-to-end to maximize the evidence lower bound~(ELBO), which is the sum of $\mathcal{L}_{\text{prior}}$, $\mathcal{L}_{\text{recon}}$, and $\mathcal{L}_{\text{ent}}$ (the entropy of the posterior $Q_\phi(z|X)$). (b) At test time, we sample a shape representation $\tilde{z}$ by sampling $\tilde{w}$ from a Gaussian prior and transforming it with $F_\psi$. To sample points from the shape represented by $\tilde{z}$, we first sample points from the $3$-D Gaussian prior and then move them according to the CNF parameterized by $\tilde{z}$.
	}
	\label{fig:network}
\end{figure*}

\subsection{Flow-based prior over shapes}\label{sec:method-prior}

Although it is possible to use a simple Gaussian prior over shape representations, it has been shown that a restricted prior tends to limit the performance of VAEs~\cite{VLAE}. 
To alleviate this problem, we use another CNF to parametrize a learnable prior.
Formally, we rewrite the KL divergence term in Equation~\ref{eq:elbo} as
{\small 
\begin{align}
D_{KL}(Q_\phi(z|x)||P_\psi(z)) &= -\mathbb{E}_{Q_\phi(z|x)}\left[\log{P_\psi(z)}\right] - H[Q_\phi(z|X)]\,,
\label{eq:kl}
\end{align}}where $H$ is the entropy and $P_\psi(z)$ is the prior distribution with learnable parameters $\psi$, obtained by transforming a simple Gaussian $P(w) = \mathcal{N}(0, \mathit{I})$ with a CNF:
{\small\begin{align*}
z = F_\psi(w(t_0)) \triangleq w(t_0) + \int_{t_0}^{t_1} f_\psi(w(t), t)dt, w(t_0)\sim P(w)\,,
\end{align*}}where $f_\psi$ defines the continuous-time dynamics of the flow $F_\psi$. Similarly as described above, the inverse of $F_\psi$ is given by $F_\psi^{-1}(z) = z + \int_{t_1}^{t_0} f_\psi(w(t), t)dt$ with $w(t_1) = z$. The log probability of the prior distribution can be computed by:
\begin{equation}
\log P_\psi(z) = \log{P\left(F_\psi^{-1}(z)\right)} - \int_{t_0}^{t_1} \operatorname{Tr}\left(\frac{\partial f_\psi}{\partial w(t)}\right)dt\,.
\label{eq:prior}
\end{equation}
\subsection{Final training objective}
Plugging Equation~\ref{eq:decompose}, \ref{eq:recon}, \ref{eq:kl}, \ref{eq:prior} into Equation \ref{eq:elbo}, the ELBO of a point set $X$ can be finally written as{\small 
\begin{align}
\mathcal{L}(X;\phi, \psi, \theta) & = \mathbb{E}_{Q_\phi(z|x)}\left[\log{P_\psi(z)} + \log{P_\theta(X|z)}\right] + H[Q_\phi(z|X)] \nonumber\\
&= \mathbb{E}_{Q_\phi(z|X)}[\log{P\left(F_\psi^{-1}(z)\right)} - \int_{t_0}^{t_1} \operatorname{Tr}\left(\frac{\partial f_\psi}{\partial w(t)}\right)dt\nonumber \\
& +  \sum_{x\in X}(
\log{P(G_\theta^{-1}(x;z))} - \int_{t_0}^{t_1} \operatorname{Tr}\left(\frac{\partial g_\theta}{\partial y(t)}\right) dt)] \nonumber \\
& +  H[Q_\phi(z|X)]\,.
\end{align}}Our model is trained end-to-end by maximizing the ELBO of all point sets in the dataset
\begin{align}
\phi^*, \psi^*, \theta^* =  \arg\max_{\phi, \psi, \theta}\sum_{X\in \mathcal{X}}\mathcal{L}(X;\phi, \psi, \theta).
\end{align}
We can interpret this objective as the sum of three parts:
\begin{enumerate}
	\item \textbf{Prior:} $\mathcal{L}_{\text{prior}}(X;\psi, \phi) \triangleq \mathbb{E}_{Q_\phi(z|x)}[\log{P_\psi(z)}]$ encourages the encoded shape representation to have a high probability under the prior, which is modeled by a CNF as described in Section~\ref{sec:method-prior}.
	We use the reparameterization trick~\cite{VAE} to enable a differentiable Monte Carlo estimate of the expectation:
	\begin{align*}
	\mathbb{E}_{Q_\phi(z|x)}[\log{P_\psi(z)}] \approx \frac{1}{L}\sum_{l=1}^{L} \log{P_\psi(\mu + \epsilon_l \odot \sigma)}\,,
	\end{align*}
	where $\mu$ and $\sigma$ are mean and standard deviation of the isotropic Gaussian posterior $Q_\phi(z|x)$ and $L$ is simply set to $1$. $\epsilon_i$ is sampled from the standard Gaussian distribution $\mathcal{N}(0, \mathit{I})$. 
	
	\item \textbf{Reconstruction likelihood:} $\mathcal{L}_{\text{recon}}(X; \theta, \phi) \triangleq \mathbb{E}_{Q_\phi(z|x)}[\log{P_\theta(X|z)}]$ is the reconstruction log-likelihood of the input point set, computed as described in Section~\ref{sec:method-likelihood}. The expectation is also estimated using Monte Carlo sampling.
	
	\item \textbf{Posterior Entropy:} $\mathcal{L}_{\text{ent}}(X;\phi) \triangleq H[Q_\phi(z|X)]$ is the entropy of the approximated posterior:
	\begin{align*}{\textstyle
		H[Q_\phi(z|X)] =  \frac{d}{2}(1+\ln{(2\pi)})+\sum_{i=1}^{d}\ln{\sigma_i}\,.
	}\end{align*}
\end{enumerate}

All the training details (\eg, hyper-parameters, model architectures) are included in \ifthenelse{\boolean{arxiv}}{Section~\ref{sec:hyper-param} of the appendix}{the supplementary materials}.

\subsection{Sampling}

To sample a shape representation, we first draw $\tilde{w} \sim \mathcal{N}(0, \mathit{I})$ then pass it through $F_\psi$ to get $\tilde{z} = F_\psi(\tilde{w})$.
To generate a point given a shape representation $\tilde{z}$, we first sample a point $\tilde{y} \in \mathbb{R}^{3}$ from $\mathcal{N}(0, \mathit{I})$, then pass $\tilde{y}$ through $G_\theta$ conditioned on $\tilde{z}$ to produce a point on the shape : $\tilde{x} = G_\theta(\tilde{w}; z)$.
To sample a point cloud with size $\tilde{M}$, we simply repeat it for $\tilde{M}$ times.
Combining these two steps allows us to sample a point cloud with $\tilde{M}$ points from our model:
{\small 
\begin{align*}
\tilde{X} = \{G_\theta(\tilde{y}_j;F_\psi(\tilde{w}))\}_{1 \leq j\leq \tilde{M}}, \tilde{w} \sim \mathcal{N}(0, \mathit{I}), \forall j, \tilde{y}_j \sim \mathcal{N}(0, \mathit{I})\,.
\end{align*}}\section{Experiments}

In this section, we first introduce existing metrics for evaluating point cloud generation, discuss their limitations, and introduce a new metric that overcomes these limitations. We then compare the proposed method with previous state-of-the-art generative models of point clouds, using both previous metrics and the proposed one. We additionally evaluate the reconstruction and representation learning ability of the auto-encoder part of our model.

\subsection{Evaluation metrics}

Following prior work, we use Chamfer distance (CD) and earth mover's distance (EMD) to measure the similarity between point clouds. 
Formally, they are defined as follows:
\begin{align*}
\text{CD}(X,Y) &= \sum_{x\in X} \min_{y\in Y} \|x-y\|_2^2 + \sum_{y\in Y}\min_{x \in X} \|x-y\|_2^2, \\
\text{EMD}(X,Y) &= \min_{\phi: X\to Y} \sum_{x\in X} \|x-\phi(x)\|_2,
\end{align*}
where $X$ and $Y$ are two point clouds with the same number of points and $\phi$ is a bijection between them. Note that most previous methods use either CD or EMD in their training objectives, which tend to be favored if evaluated under the same metric. Our method, however, does not use CD or EMD during training.

\begin{table*}[t]
	\centering
	\caption{Generation results. $\uparrow$: the higher the better, $\downarrow$: the lower the better. The best scores are highlighted in bold. Scores of the real shapes that are worse than some of the generated shapes are marked in gray. MMD-CD scores are multiplied by $10^3$; MMD-EMD scores are multiplied by $10^2$; JSDs are multiplied by $10^2$.}
	\label{table:generation}
	\begin{small}
		\begin{tabularx}{\textwidth}{ll*{11}{Y}}
			\toprule
			&  & \multicolumn{2}{c}{\# Parameters (M)} & \multirow{2}{*}{JSD~($\downarrow$)} & \multicolumn{2}{c}{MMD ($\downarrow$)} & \multicolumn{2}{c}{COV (\%, $\uparrow$)} & \multicolumn{2}{c}{1-NNA (\%, $\downarrow$)} \\ 
			\cmidrule(lr){3-4} \cmidrule(lr){6-7} \cmidrule(lr){8-9} \cmidrule(l){10-11} 
			Category                  & Model                  & Full              & Gen              &                                 & CD   & EMD  & CD                  & EMD                & CD                     & EMD                   \\ \midrule
			\multirow{6}{*}{Airplane} & r-GAN                  & 7.22              & 6.91             & 7.44                            & 0.261             & 5.47              & 42.72               & 18.02              & 93.58                  & 99.51                 \\
			& l-GAN~(CD)              & 1.97              & 1.71             & 4.62                            & 0.239             & 4.27              & 43.21               & 21.23              & 86.30                  & 97.28                 \\
			& l-GAN~(EMD)             & 1.97              & 1.71             & \textbf{3.61}                   & 0.269             & 3.29              & \textbf{47.90}      & \textbf{50.62}     & 87.65                  & 85.68                 \\
			& PC-GAN                 & 9.14              & 1.52             & 4.63                            & 0.287             & 3.57              & 36.46               & 40.94              & 94.35                  & 92.32                 \\
			& PointFlow~(ours)                   & \textbf{1.61}              & \textbf{1.06}             & 4.92                            & \textbf{0.217}    & \textbf{3.24}     & 46.91               & 48.40              & \textbf{75.68}         & \textbf{75.06}        \\ 
			\cmidrule(l){2-11} 
			& Training set              & -                 & -                & \textcolor{mygray}{6.61}                            & \textcolor{mygray}{0.226}             & 3.08              & \textcolor{mygray}{42.72}               & \textcolor{mygray}{49.14}              & 70.62                  & 67.53                 \\ 
			\midrule
			\multirow{6}{*}{Chair}    & r-GAN                  & 7.22              & 6.91             & 11.5                            & 2.57              & 12.8              & 33.99               & 9.97               & 71.75                  & 99.47                 \\
			& l-GAN~(CD)              & 1.97              & 1.71             & 4.59                           & 2.46              & 8.91              & 41.39               & 25.68              & 64.43                  & 85.27                 \\
			& l-GAN~(EMD)             & 1.97              & 1.71             & 2.27                           & 2.61              & \textbf{7.85}     & 40.79               & 41.69              & 64.73                  & 65.56                 \\
			& PC-GAN                 & 9.14              & 1.52             & 3.90                            & 2.75              & 8.20              & 36.50               & 38.98              & 76.03                  & 78.37                 \\
			& PointFlow~(ours)                   & \textbf{1.61}     & \textbf{1.06}    & \textbf{1.74}                   & \textbf{2.42}     & 7.87              & \textbf{46.83}      & \textbf{46.98}     & \textbf{60.88}         & \textbf{59.89}        \\ 
			\cmidrule(l){2-11} 
			& Training set           & -                 & -                & 1.50                            & 1.92              & 7.38              & 57.25               & 55.44              & 59.67                  & 58.46                 \\ 
			\midrule
			\multirow{6}{*}{Car}     & r-GAN             & 7.22             & 6.91                            & 12.8              & 1.27              & 8.74                & 15.06              & 9.38                   & 97.87                  & 99.86                 \\
			& l-GAN~(CD)              & 1.97              & 1.71             & 4.43                           & 1.55              & 6.25              & 38.64               & 18.47              & 63.07                  & 88.07                 \\
			& l-GAN~(EMD)             & 1.97              & 1.71             & 2.21                            & 1.48              & 5.43              & 39.20               & 39.77              & 69.74                  & 68.32                 \\
			& PC-GAN                 & 9.14              & 1.52             & 5.85                            & 1.12              & 5.83              & 23.56               & 30.29              & 92.19                  & 90.87                 \\
			& PointFlow~(ours)                   & \textbf{1.61}              & \textbf{1.06}             & \textbf{0.87}                   & \textbf{0.91}     & \textbf{5.22}     & \textbf{44.03}      & \textbf{46.59}     & \textbf{60.65}         & \textbf{62.36}        \\ 
			\cmidrule(l){2-11} 
			& Training set              & -                 & -                & 0.86                            & \textcolor{mygray}{1.03}              & \textcolor{mygray}{5.33}              & 48.30               & 51.42              & 57.39                  & 53.27                 \\ 
			\bottomrule
		\end{tabularx}
	\end{small}
\end{table*}

Let $S_g$ be the set of generated point clouds and $S_r$ be the set of reference point clouds with $|S_r| = |S_g|$. To evaluate generative models, we first consider the three metrics introduced by Achlioptas \etal~\cite{achlioptas_L3DP}:
\begin{itemize}
	\item\textbf{Jensen-Shannon Divergence (JSD)} are computed between the marginal point distributions:
	\begin{align*}
	\text{JSD}(P_g,P_r) = \frac{1}{2}D_{KL}(P_r||M) + \frac{1}{2}D_{KL}(P_g||M)\,,
	\end{align*}
	where $M=\frac{1}{2}(P_r + P_g)$. $P_r$ and $P_g$ are marginal distributions of points in the reference and generated sets, approximated by discretizing the space into $28^3$ voxels and assigning each point to one of them.	However, it only considers the marginal point distributions but not the distribution of individual shapes. A model that always outputs the ``average shape'' can obtain a perfect JSD score without learning any meaningful shape distributions.
	\item\textbf{Coverage (COV)} measures the fraction of point clouds in the reference set that are matched to at least one point cloud in the generated set. For each point cloud in the generated set, its nearest neighbor in the reference set is marked as a match:
	\begin{align*}
	\text{COV}(S_g, S_r) = \frac{|\{\arg\min_{Y \in S_r} D(X,Y) | X \in S_g \}|}{|S_r|},
	\end{align*}
	where $D(\cdot, \cdot)$ can be either CD or EMD. While coverage is able to detect mode collapse, it does not evaluate the quality of generated point clouds. In fact, it is possible to achieve a perfect coverage score even if the distances between generated and reference point clouds are arbitrarily large.
	\item\textbf{Minimum matching distance (MMD)} is proposed to complement coverage as a metric that measures quality. For each point cloud in the reference set, the distance to its nearest neighbor in the generated set is computed and averaged:
	\begin{equation}
	\text{MMD}(S_g, S_r) = \frac{1}{|S_r|}\sum_{Y\in S_r} \min_{X\in S_g} D(X,Y)\,,\nonumber
	\end{equation}
	where $D(\cdot, \cdot)$ can be either CD or EMD. However, MMD is actually very insensitive to low-quality point clouds in $S_g$, since they are unlikely to be matched to real point clouds in $S_r$. In the extreme case, one can imagine that $S_g$ consists of mostly very low-quality point clouds with one additional point cloud in each mode of $S_r$, yet has a reasonably good MMD score.
\end{itemize}

As discussed above, all existing metrics have their limitations. As will be shown later, we also empirically find all these metrics sometimes give generated point clouds even better scores than real point clouds, further casting doubt on whether they can ensure a fair model comparison. We therefore introduce another metric that we believe is better suited for evaluating generative models of point clouds:

\begin{itemize}
	\item \textbf{1-nearest neighbor accuracy (1-NNA)} is proposed by Lopez-Paz and Oquab~\cite{tst} for two-sample tests, assessing whether two distributions are identical. 
	It has also been explored as a metric for evaluating GANs~\cite{1nnacc}.
	Let $S_{-X} = S_r \cup S_g - \{X\}$ and $N_X$ be the nearest neighbor of $X$ in $S_{-X}$. $1$-NNA is the leave-one-out accuracy of the 1-NN classifier:
	\begin{align*}
	\text{1-NNA}(&S_g, S_r) \\
	=&\frac{\sum_{X\in S_g} \mathbb{I}[N_X \in S_g] +  \sum_{Y\in S_r} \mathbb{I}[N_Y \in S_r]}{|S_g|+|S_r|} ,
	\end{align*}
	where $\mathbb{I}[\cdot]$ is the indicator function.
	For each sample, the 1-NN classifier classifies it as coming from $S_r$ or $S_g$ according to the label of its nearest sample.
	If $S_g$ and $S_r$ are sampled from the same distribution, the accuracy of such a classifier should converge to $50\%$ given a sufficient number of samples. The closer the accuracy is to $50\%$, the more similar $S_g$ and $S_r$ are, and therefore the better the model is at learning the target distribution. In our setting, the nearest neighbor can be computed using either CD or EMD. Unlike JSD, 1-NNA considers the similarity between shape distributions rather than between marginal point distributions. Unlike COV and MMD, 1-NNA directly measures distributional similarity and takes both diversity and quality into account.
\end{itemize}

\subsection{Generation}
We compare our method with three existing generative models for point clouds: raw-GAN~\cite{achlioptas_L3DP}, latent-GAN~\cite{achlioptas_L3DP}, and PC-GAN~\cite{PC-GAN}, using their official implementations that are either publicly available or obtained by contacting the authors. 
We train each model using point clouds from one of the three categories in the ShapeNet~\cite{shapenet} dataset: \emph{airplane}, \emph{chair}, and \emph{car}.
The point clouds are obtained by sampling points uniformly from the mesh surface.
All points in each category are normalized to have zero-mean per axis and unit-variance globally.
Following prior convention~\cite{achlioptas_L3DP}, we use $2048$ points for each shape during both training and testing, although our model is able to sample an arbitrary number of points.
We additionally report the performance of point clouds sampled from the training set, which is considered as an upper bound since they are from the target distribution.

\begin{figure}[t]
	\centering	
	\newcommand{\sizea}{0.235\linewidth}
	\setlength{\arrayrulewidth}{.5pt}%
	\setlength{\tabcolsep}{1pt}
	\renewcommand{\arraystretch}{0}
	\begin{tabular}{cccc}
		\includegraphics[width=\sizea]{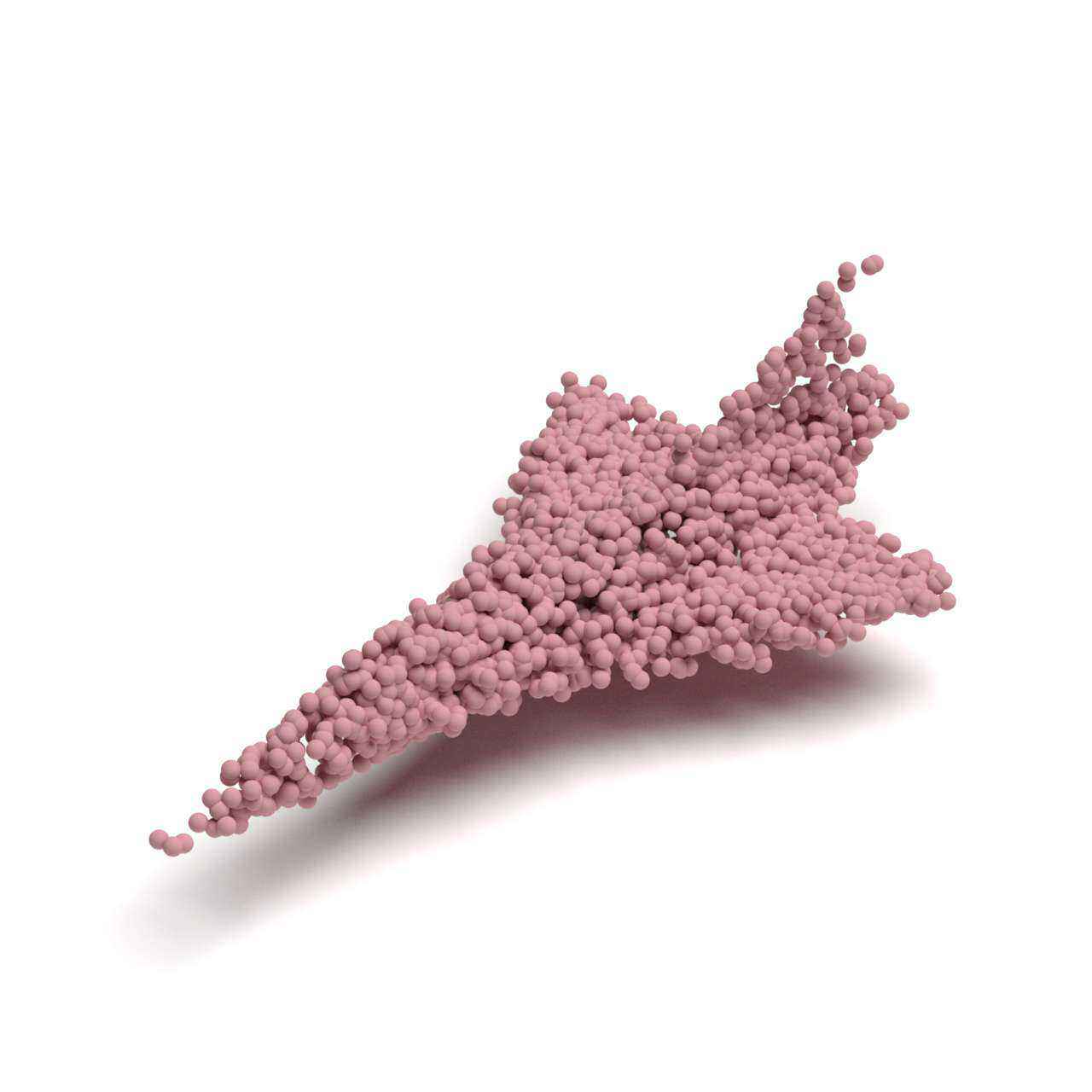} &
		\includegraphics[width=\sizea]{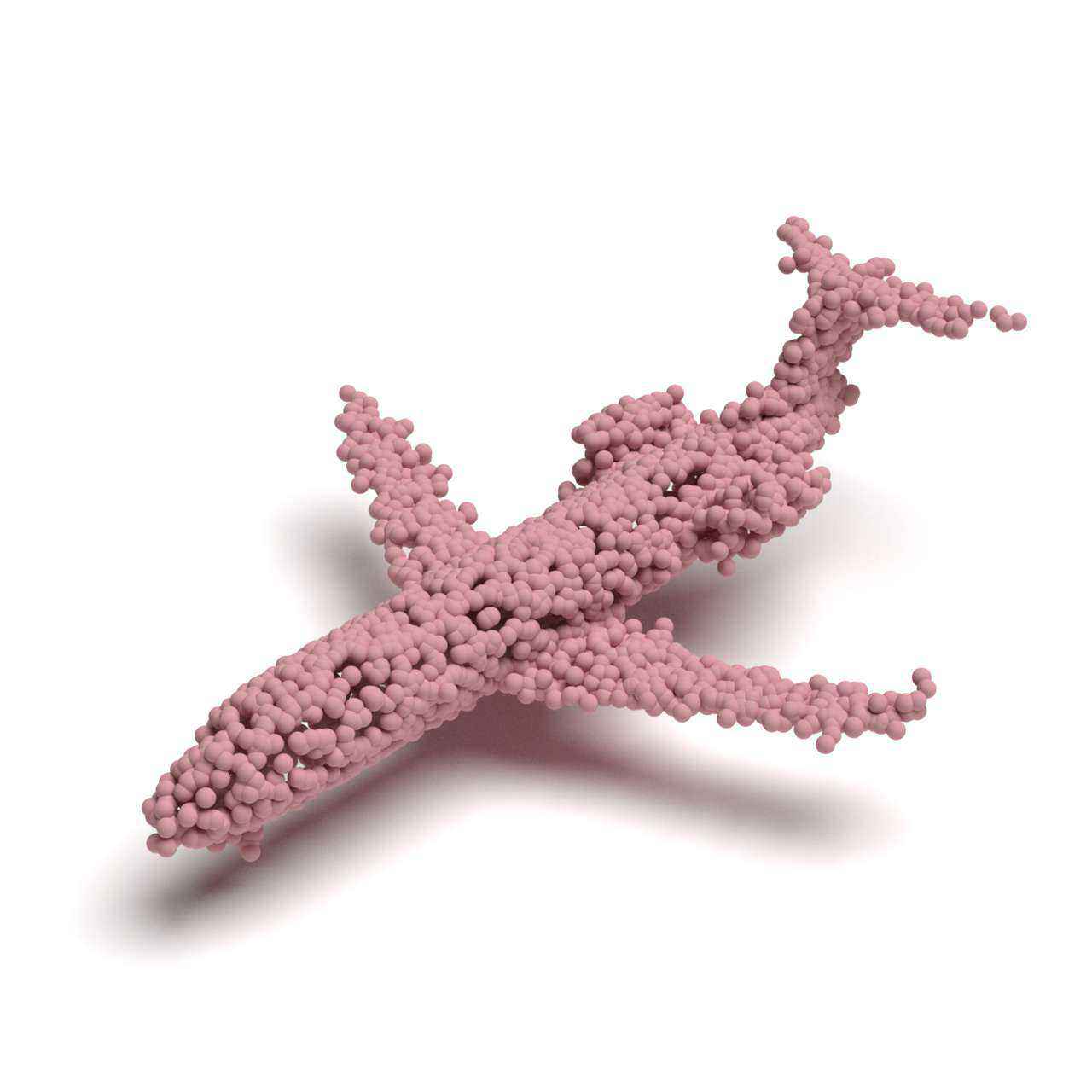} &
		\includegraphics[width=\sizea]{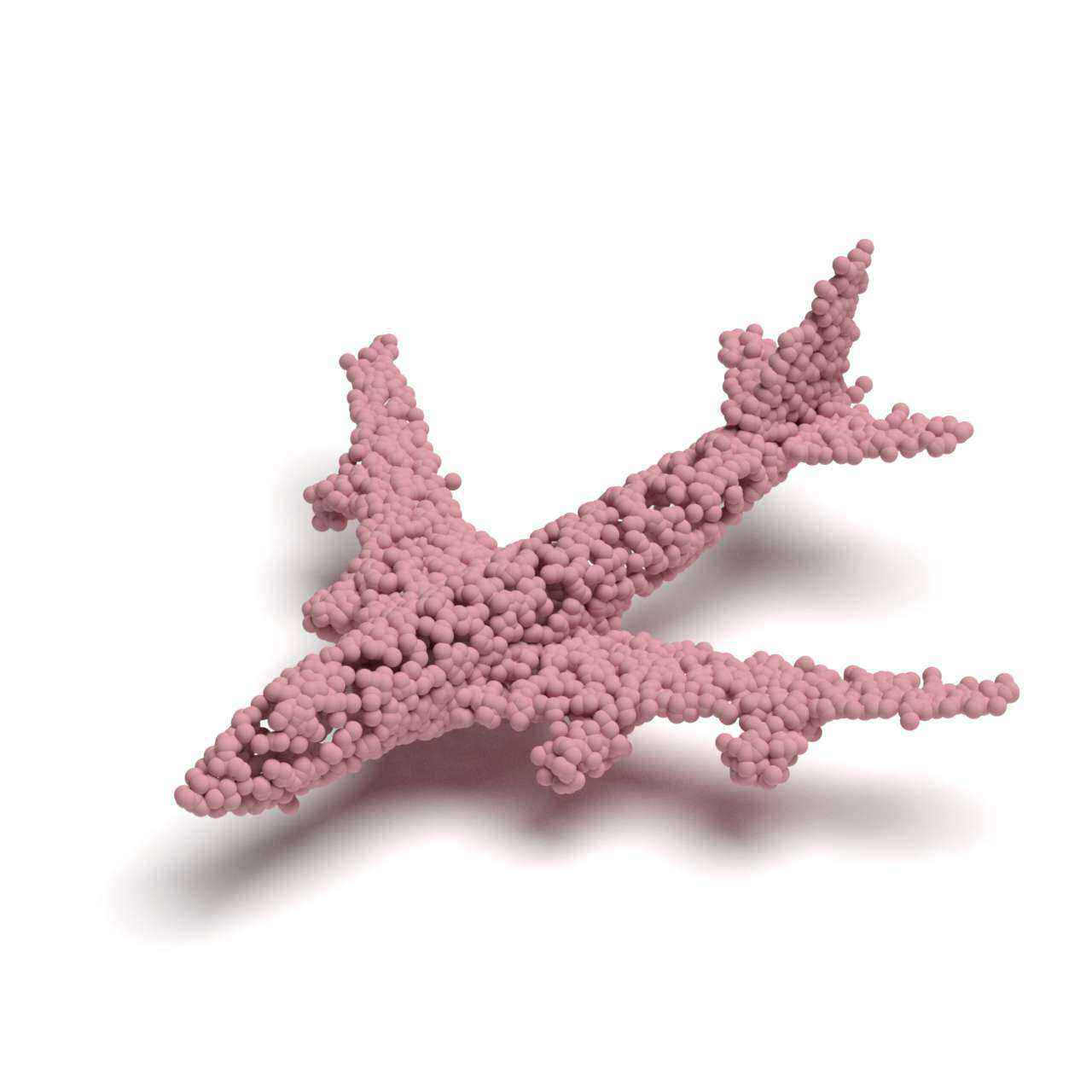} &
		\includegraphics[width=\sizea]{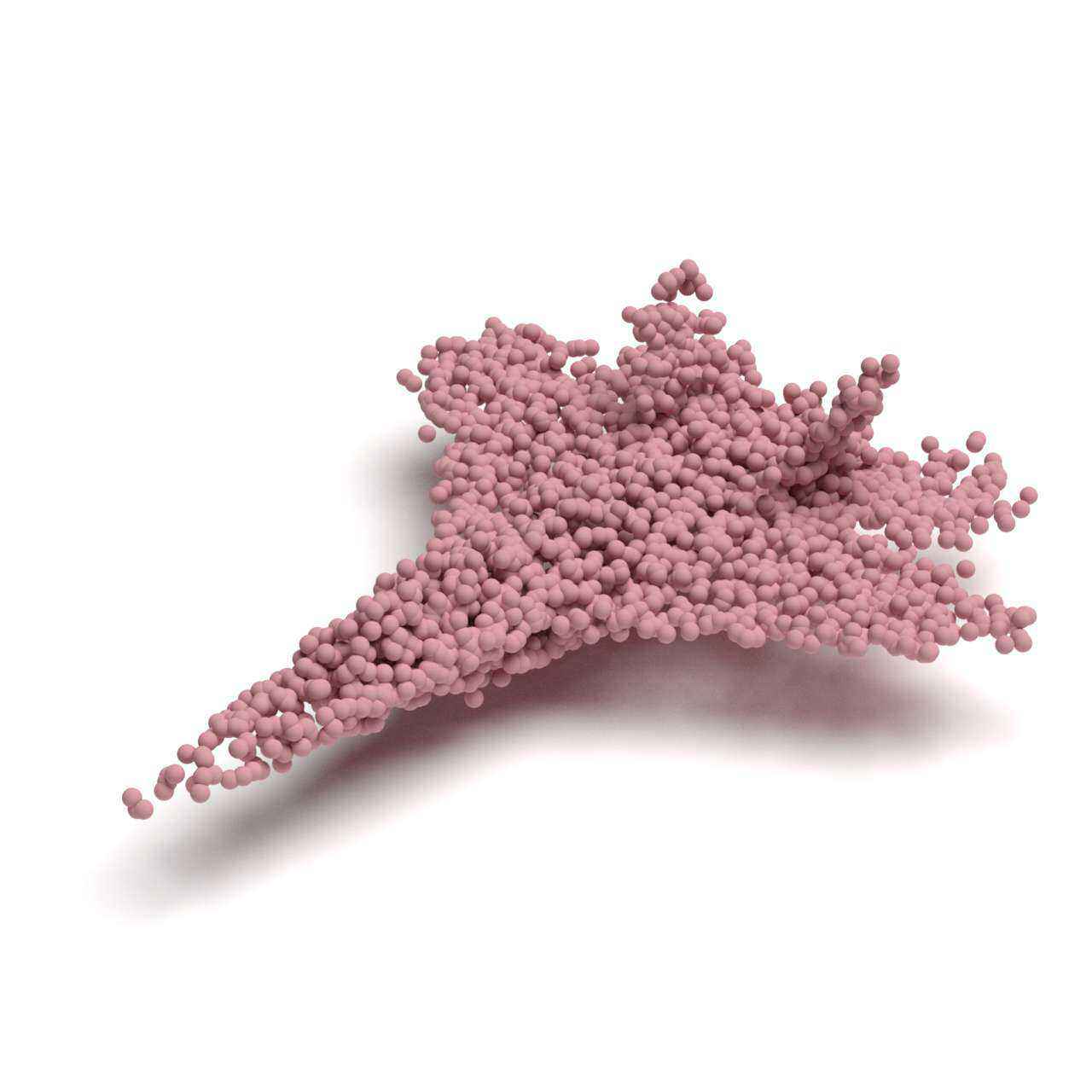} \\
		\includegraphics[width=\sizea]{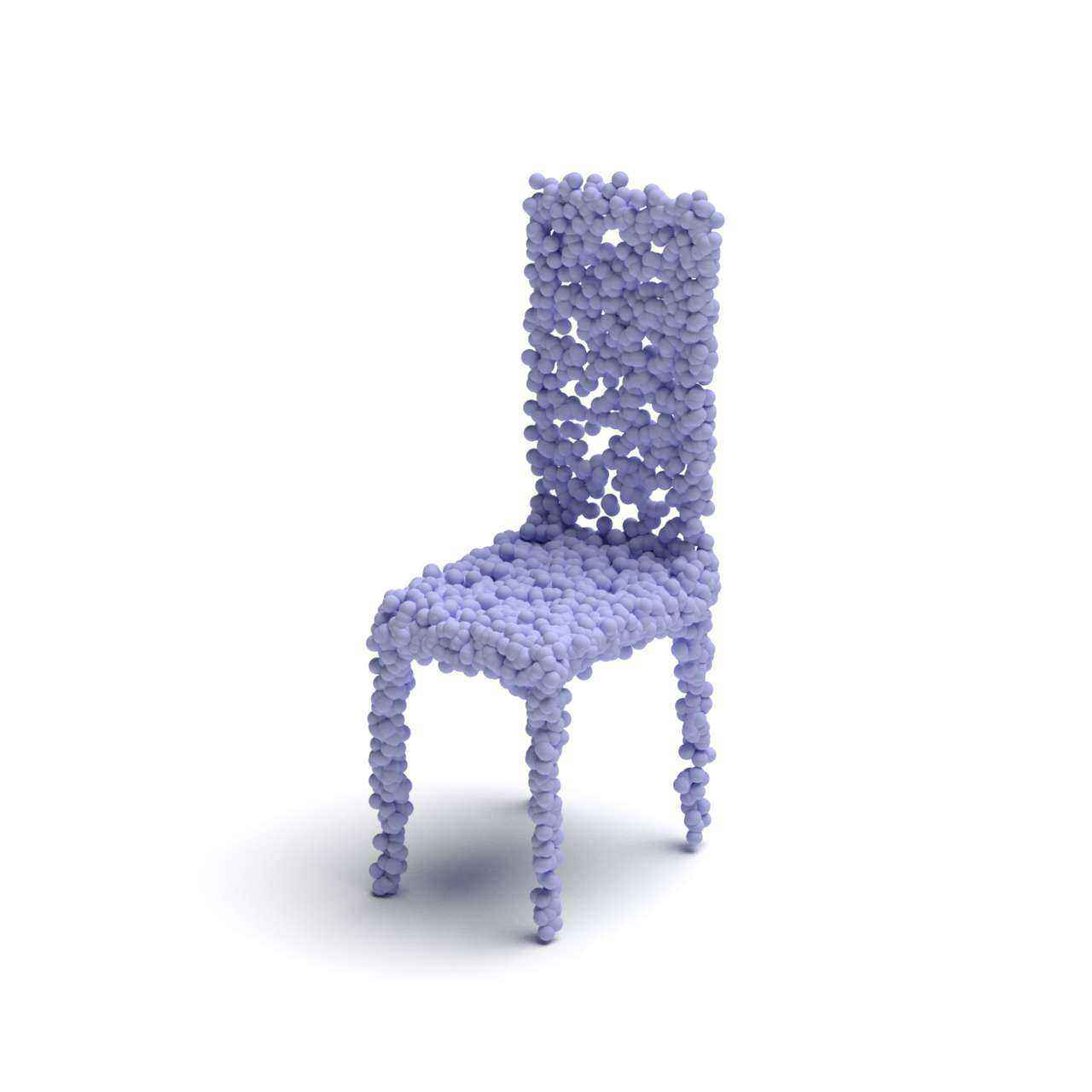} &
		\includegraphics[width=\sizea]{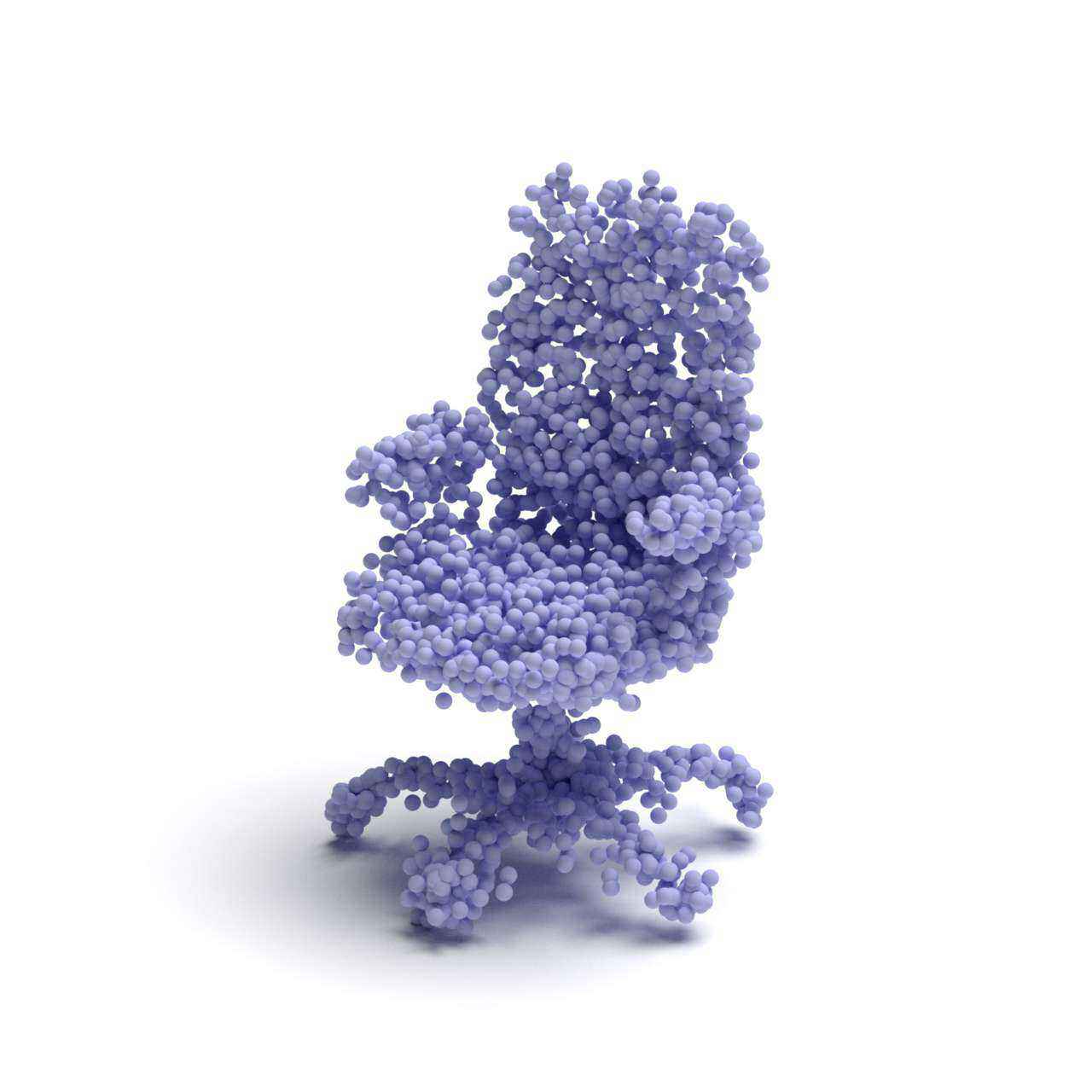} &
		\includegraphics[width=\sizea]{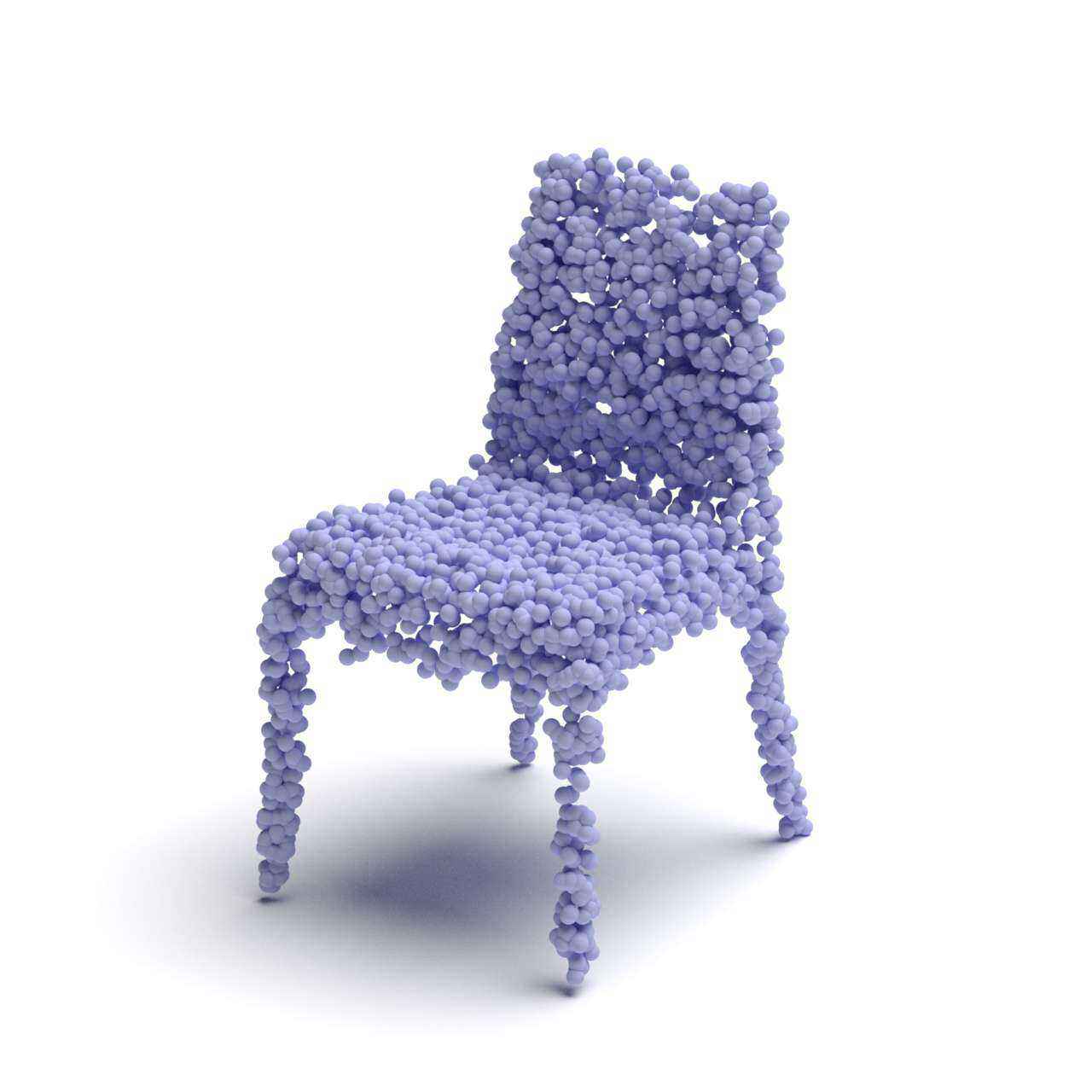} &		
		\includegraphics[width=\sizea]{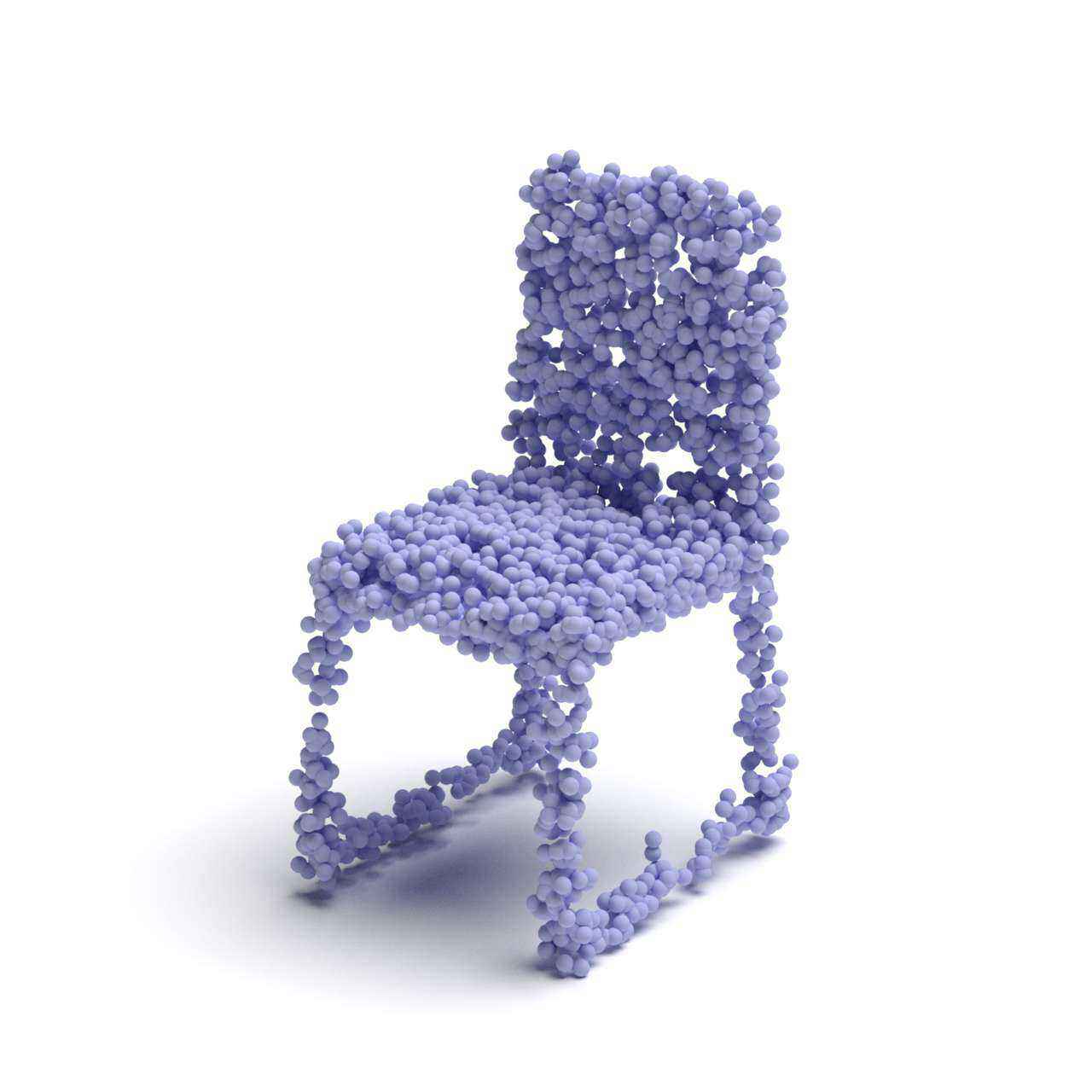}\\
		\includegraphics[width=\sizea]{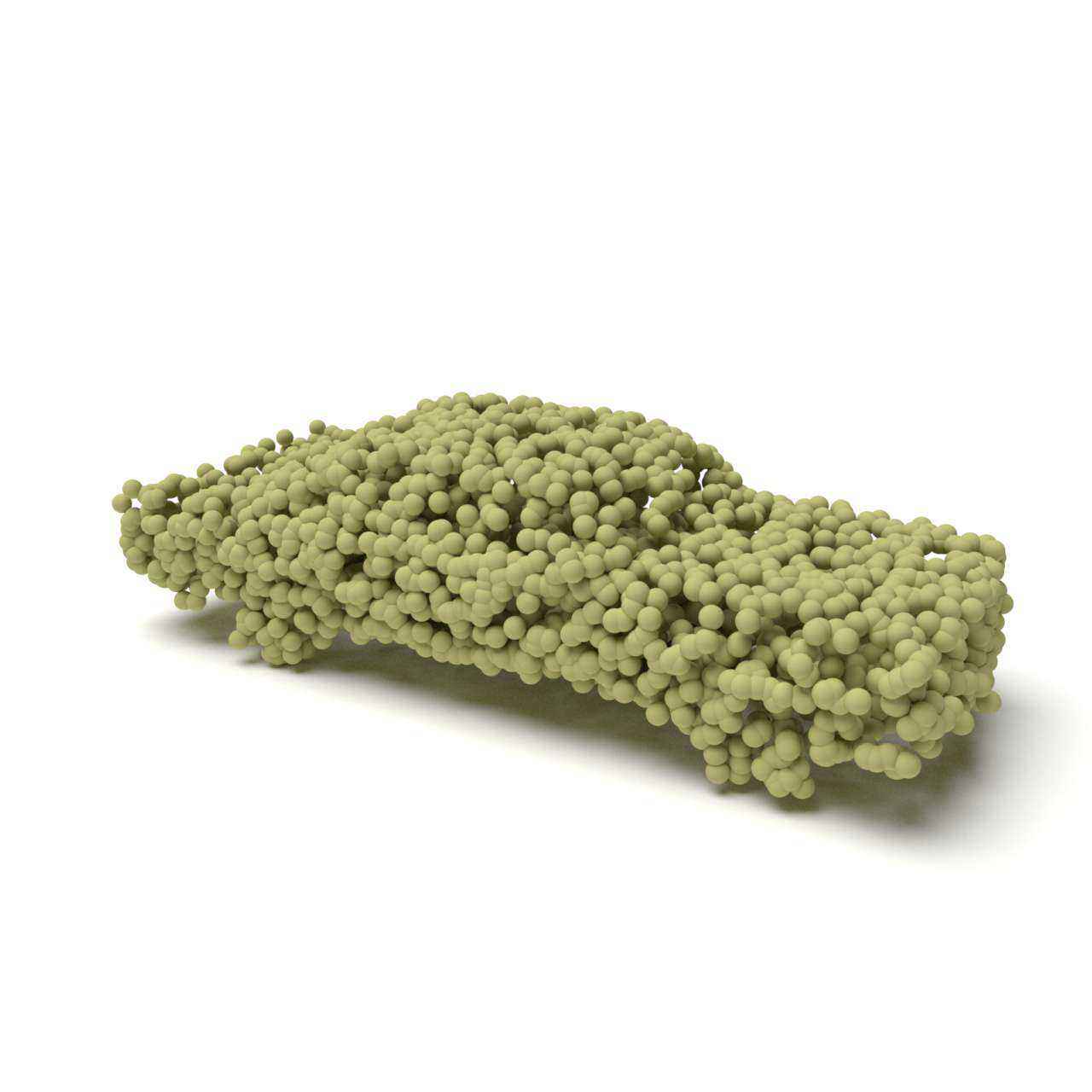} &
		\includegraphics[width=\sizea]{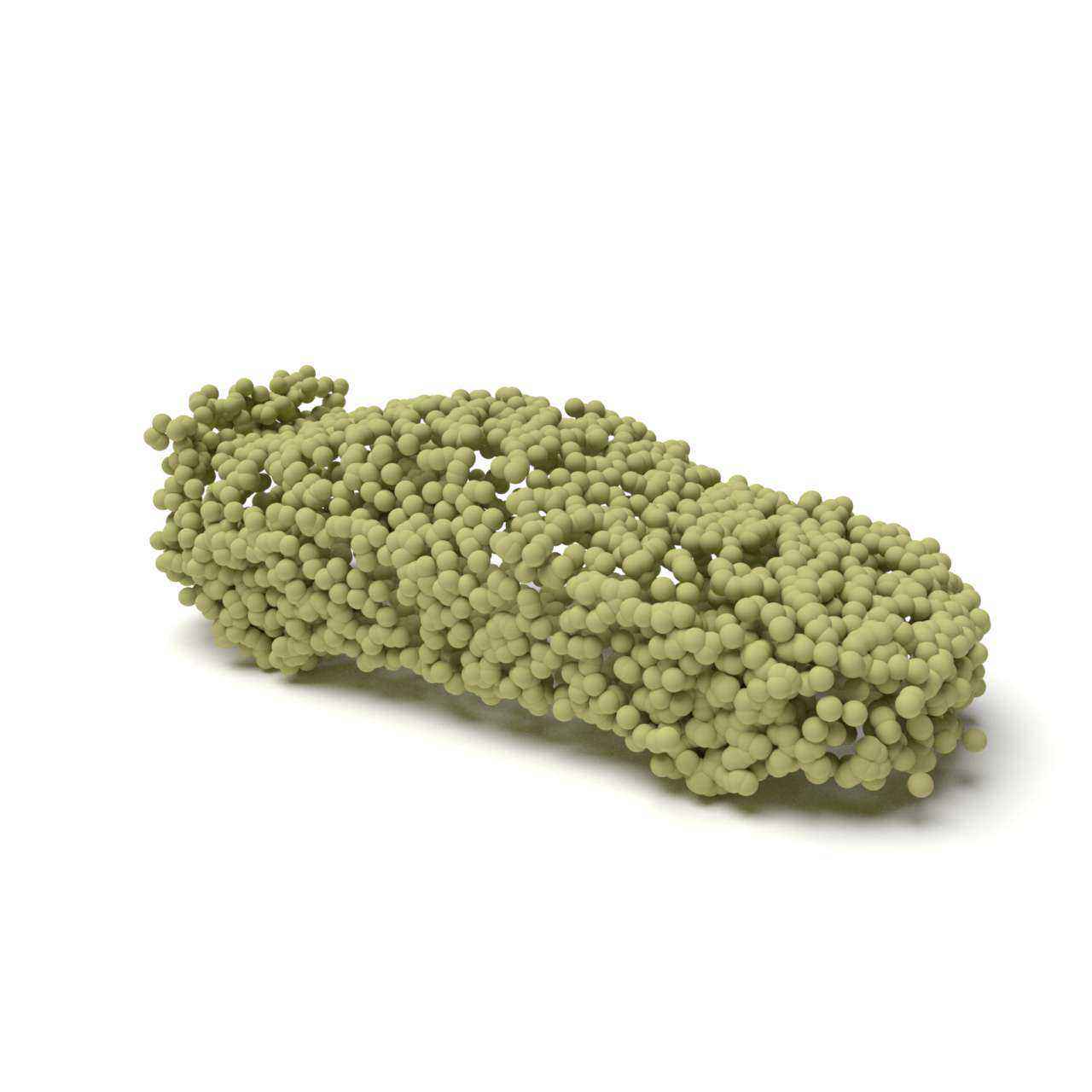} &
		\includegraphics[width=\sizea]{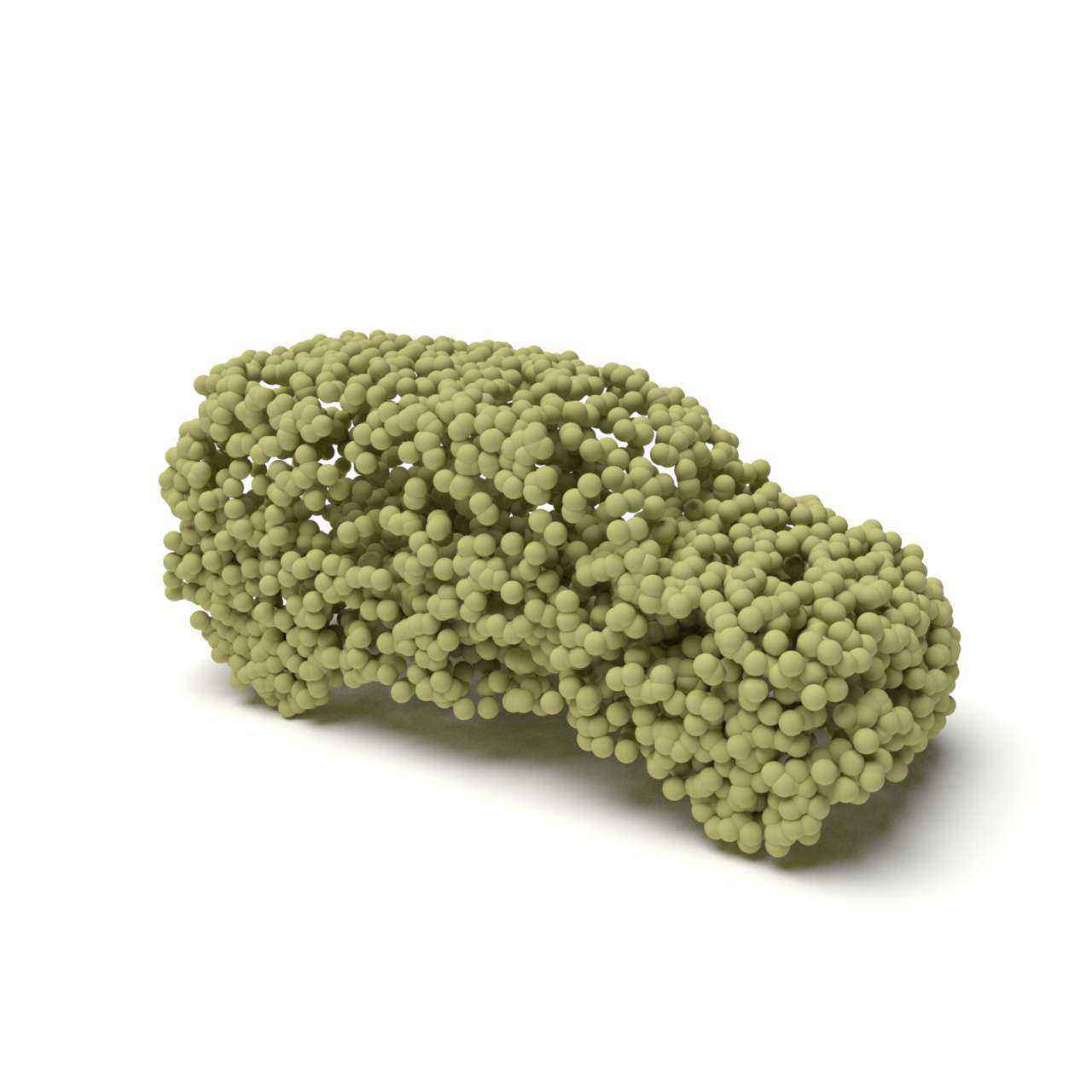} &
		\includegraphics[width=\sizea]{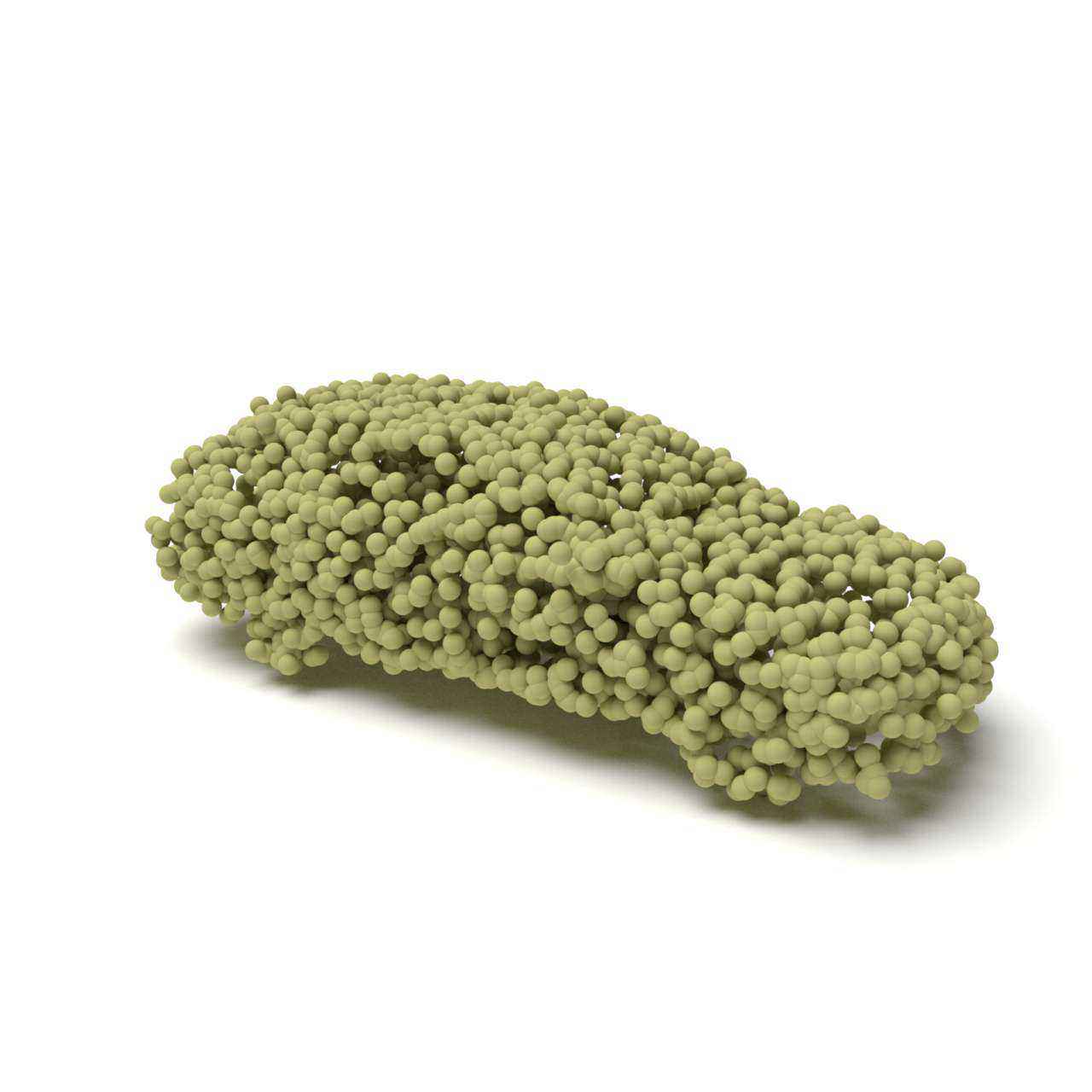}
	\end{tabular}	
	\caption{Examples of point clouds generated by our model. From top to bottom: airplane, chair, and car.}
	\label{fig:generation}
\end{figure}

\begin{figure}[t]
	\centering	
	\newcommand{\sizea}{0.235\linewidth}
	\setlength{\arrayrulewidth}{.5pt}%
	\setlength{\tabcolsep}{1pt}
	\renewcommand{\arraystretch}{0}
	\begin{tabular}{cc;{2pt/2pt}cc}
		\includegraphics[width=\sizea]{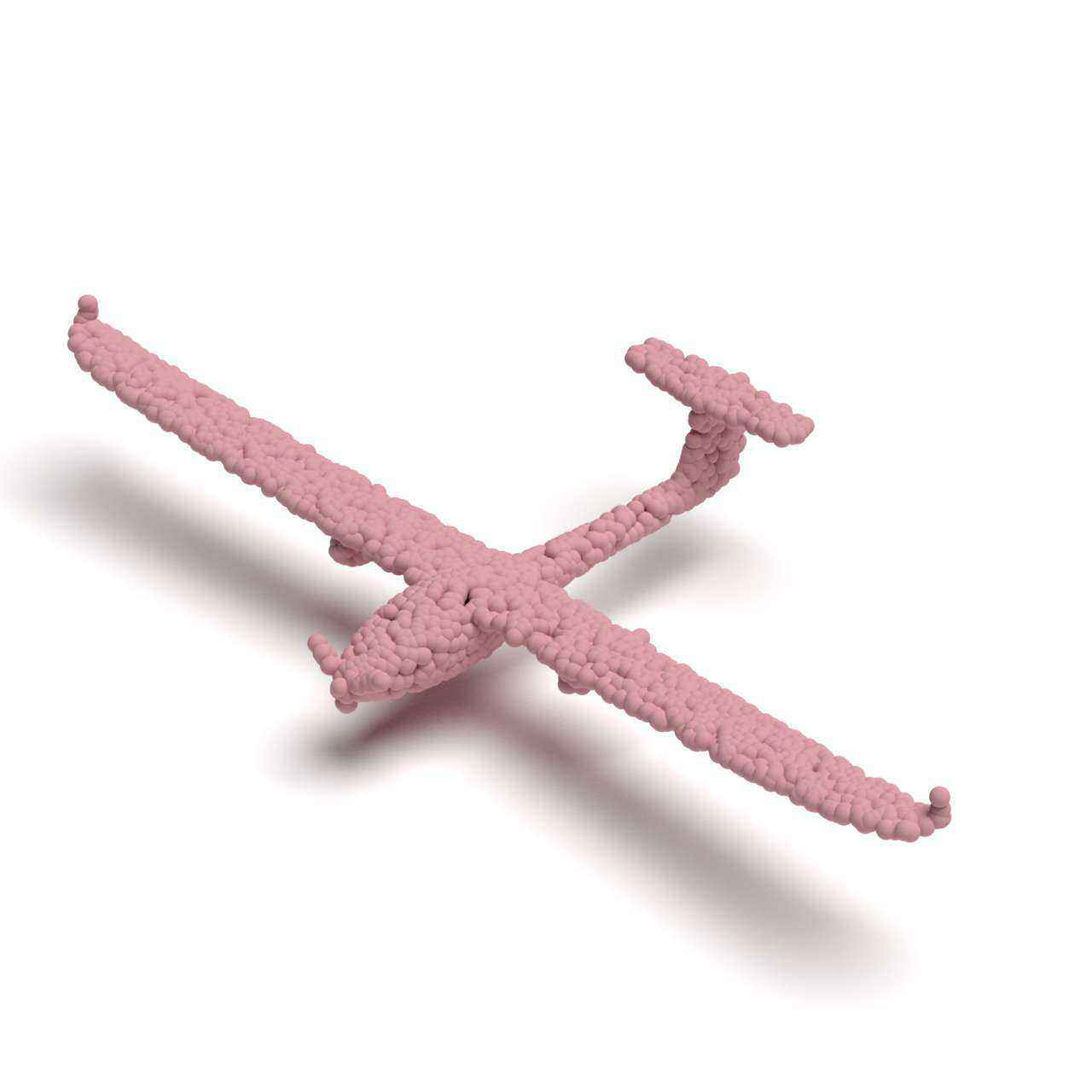} &
		\includegraphics[width=\sizea]{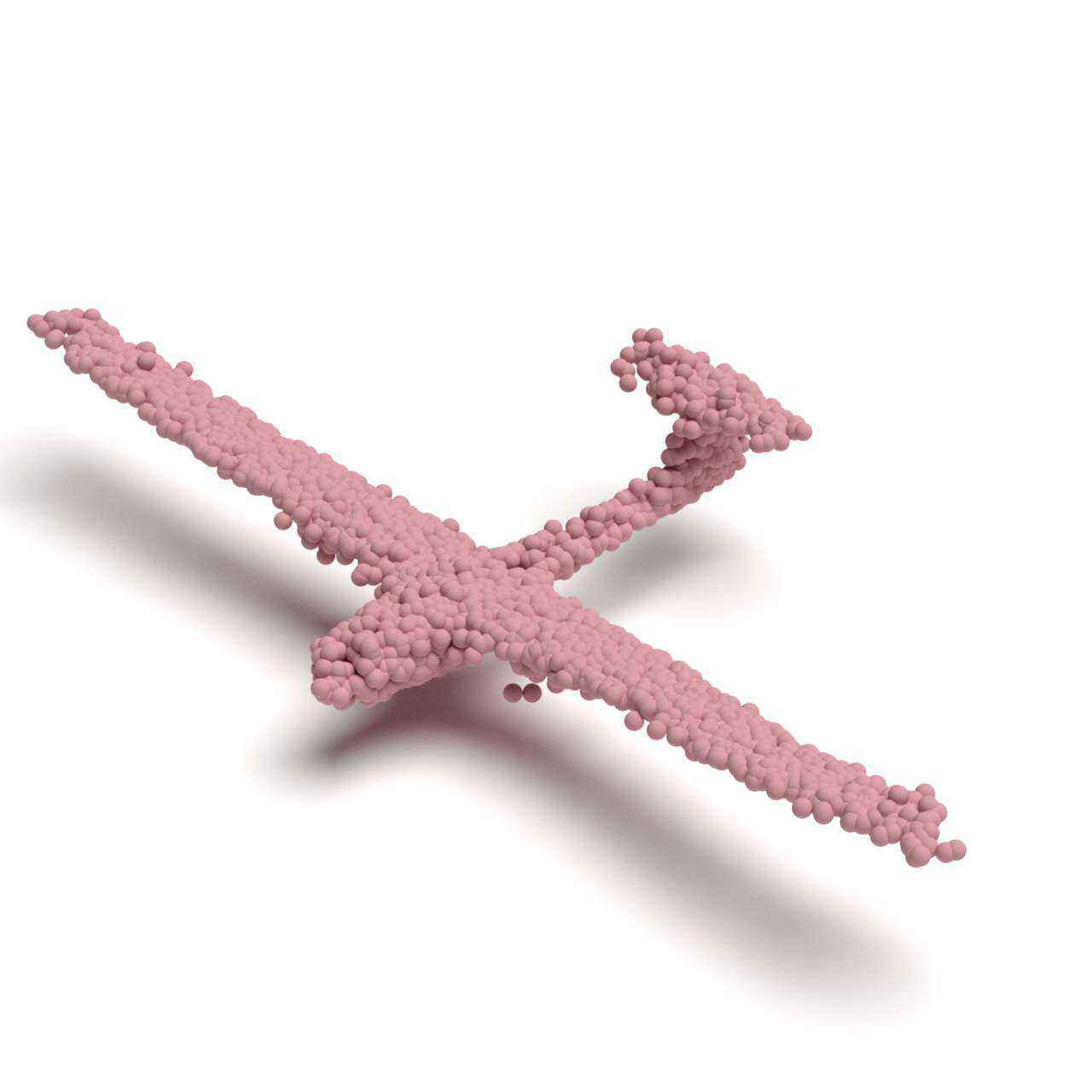} &
		\includegraphics[width=\sizea]{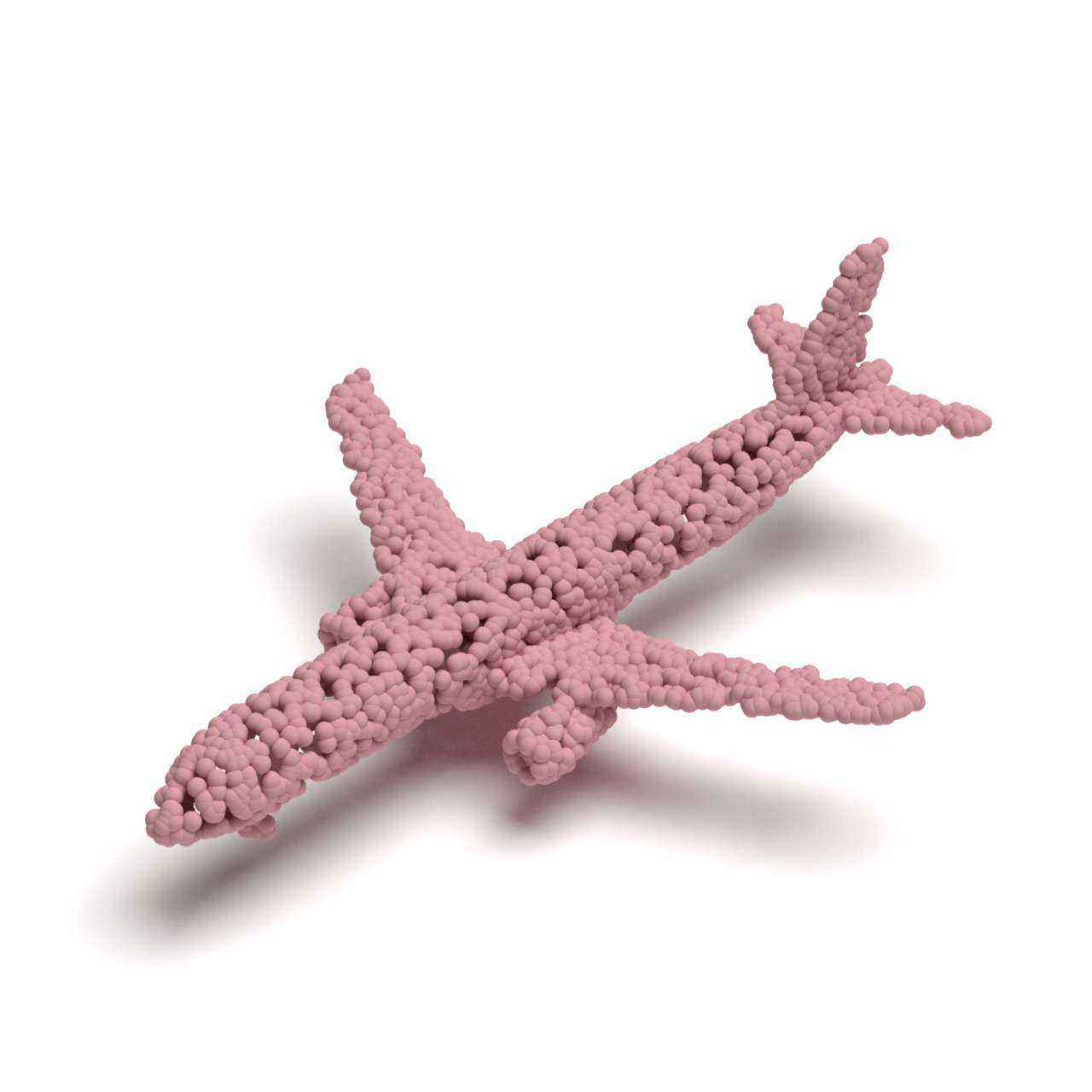} &
		\includegraphics[width=\sizea]{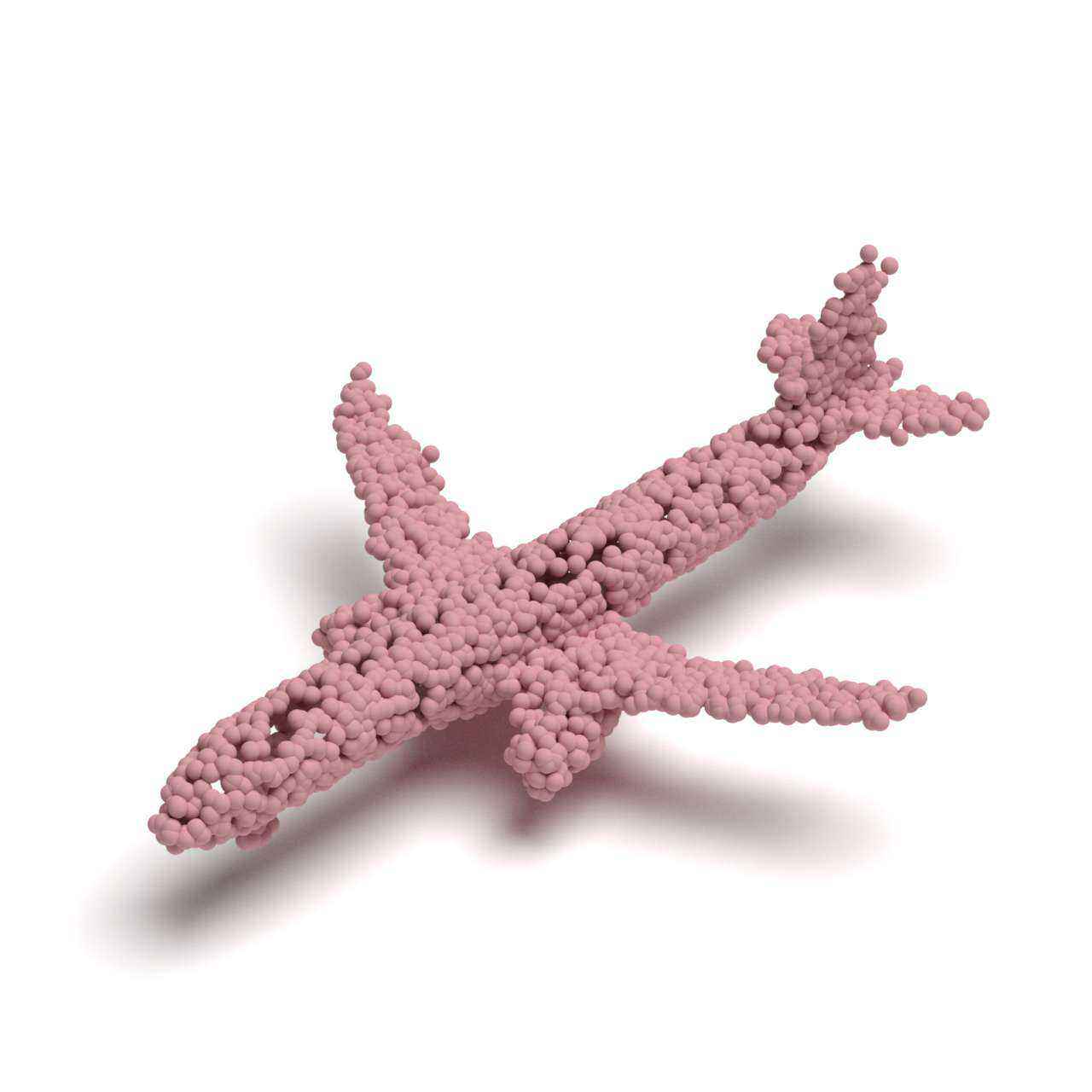} \\
		\includegraphics[width=\sizea]{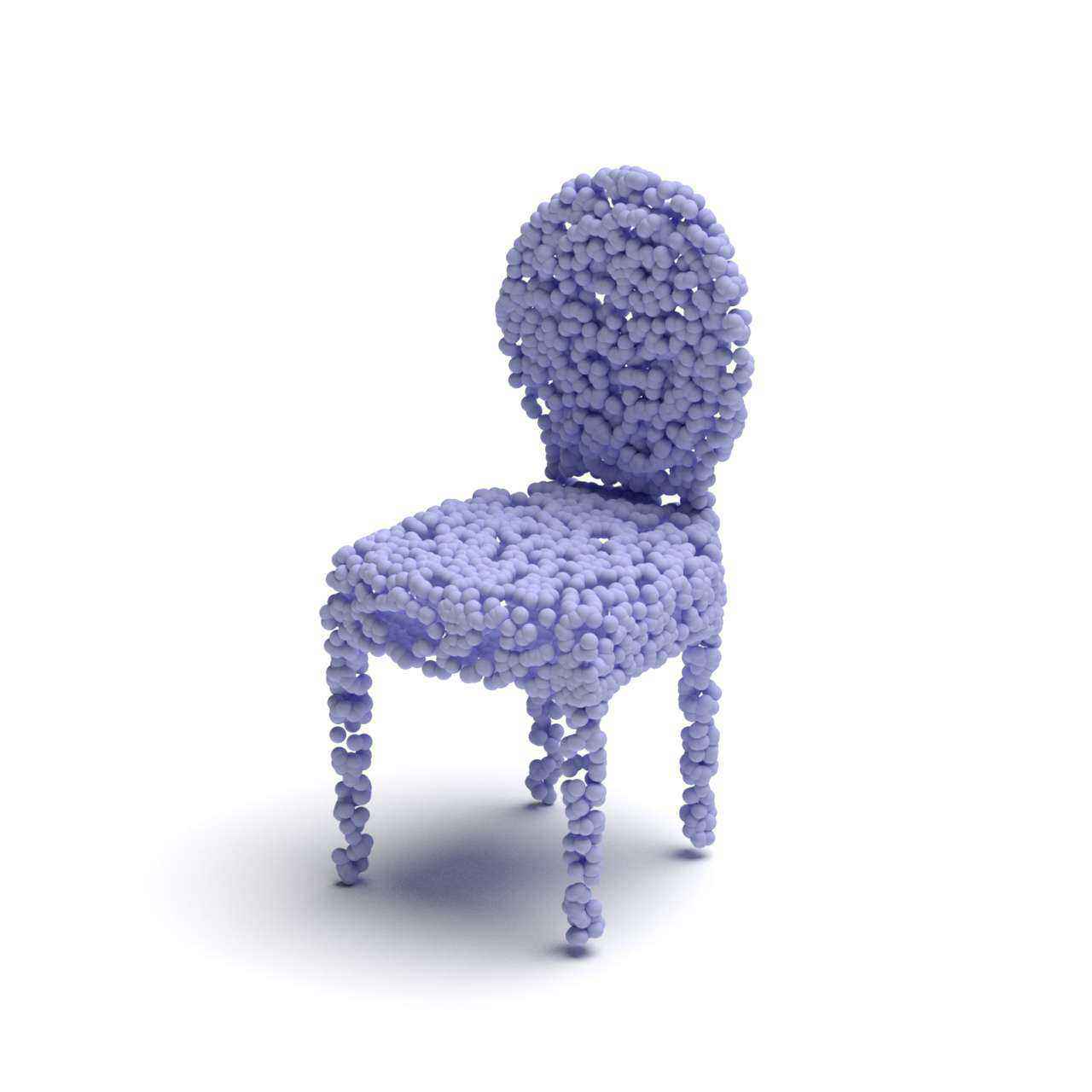} &
		\includegraphics[width=\sizea]{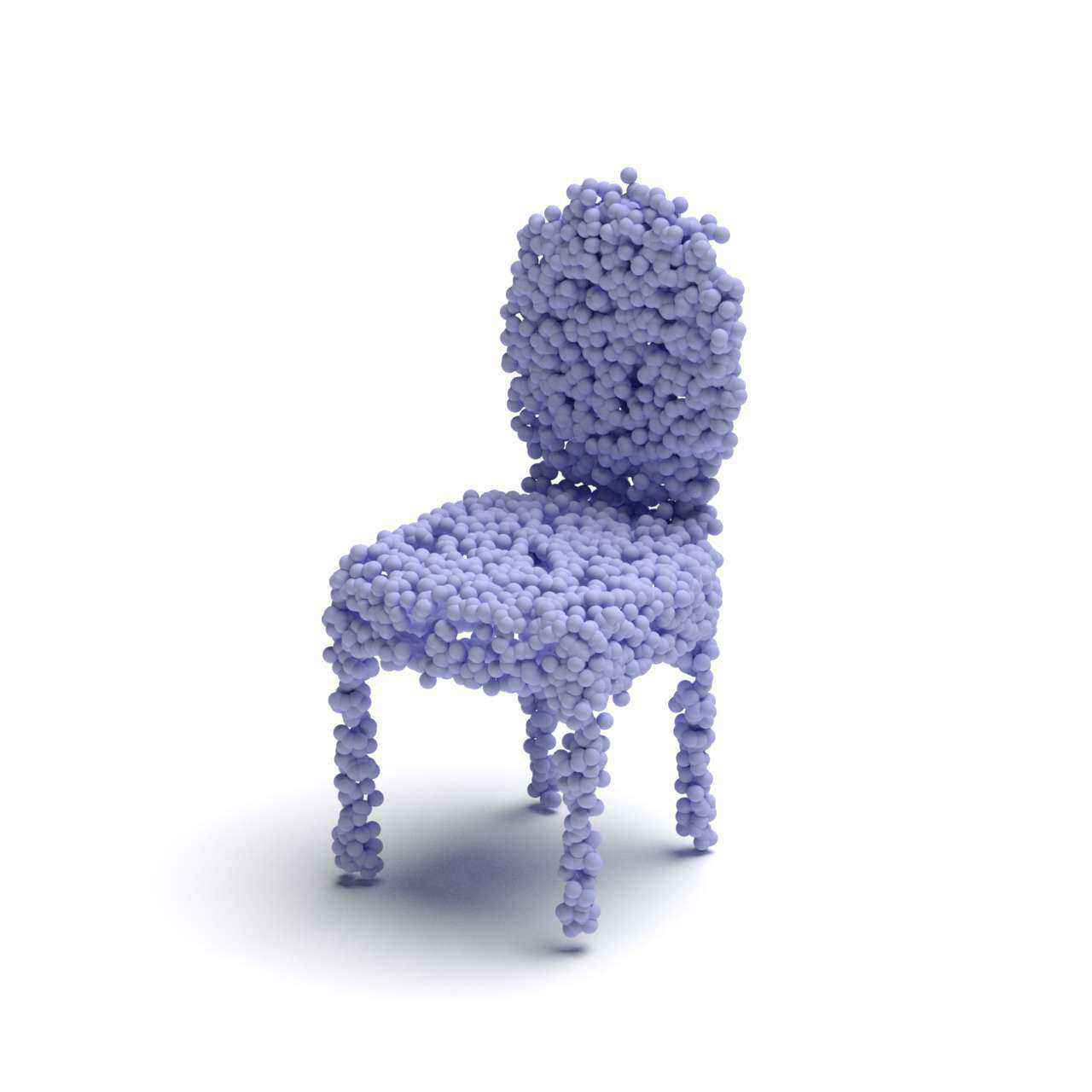} &
		\includegraphics[width=\sizea]{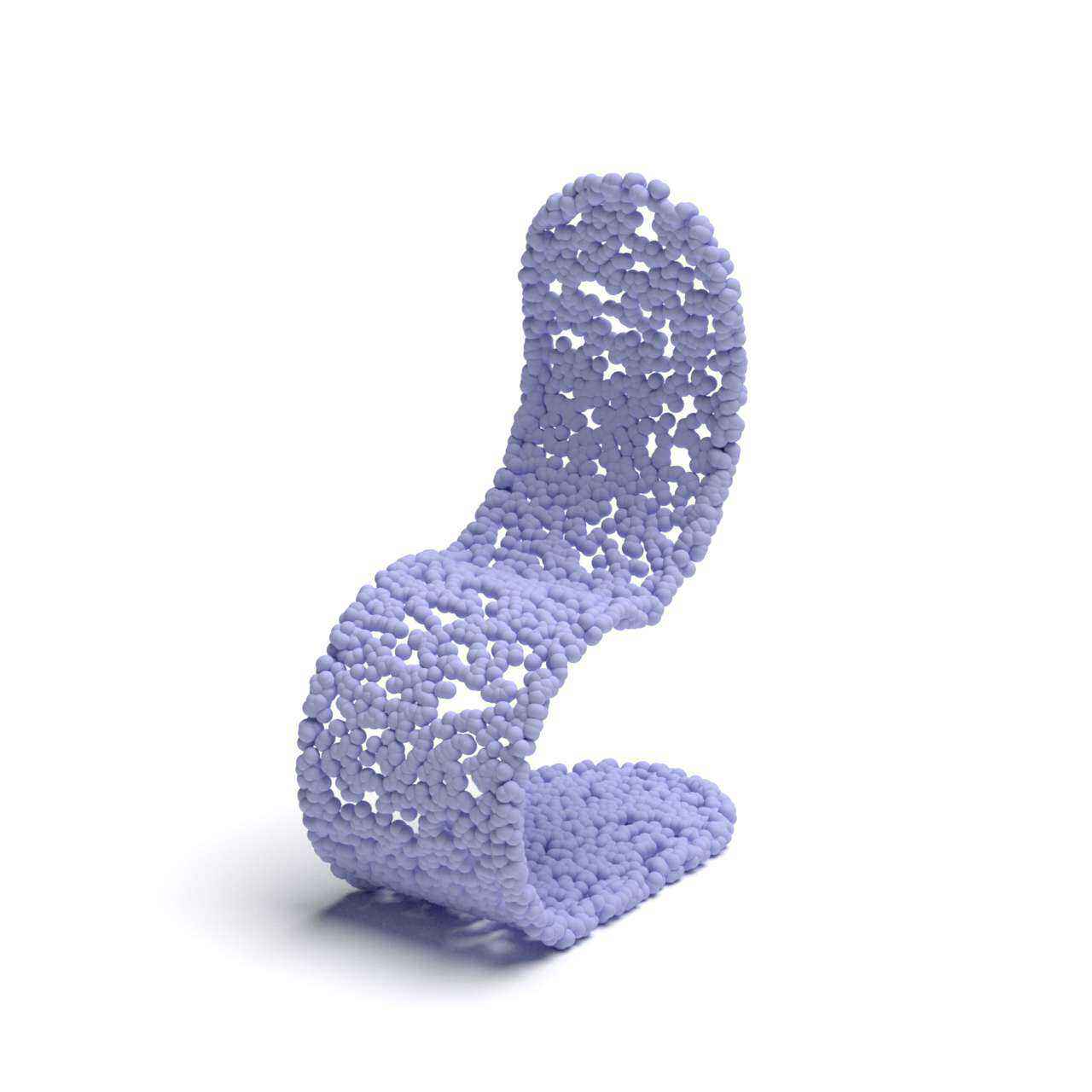} &
		\includegraphics[width=\sizea]{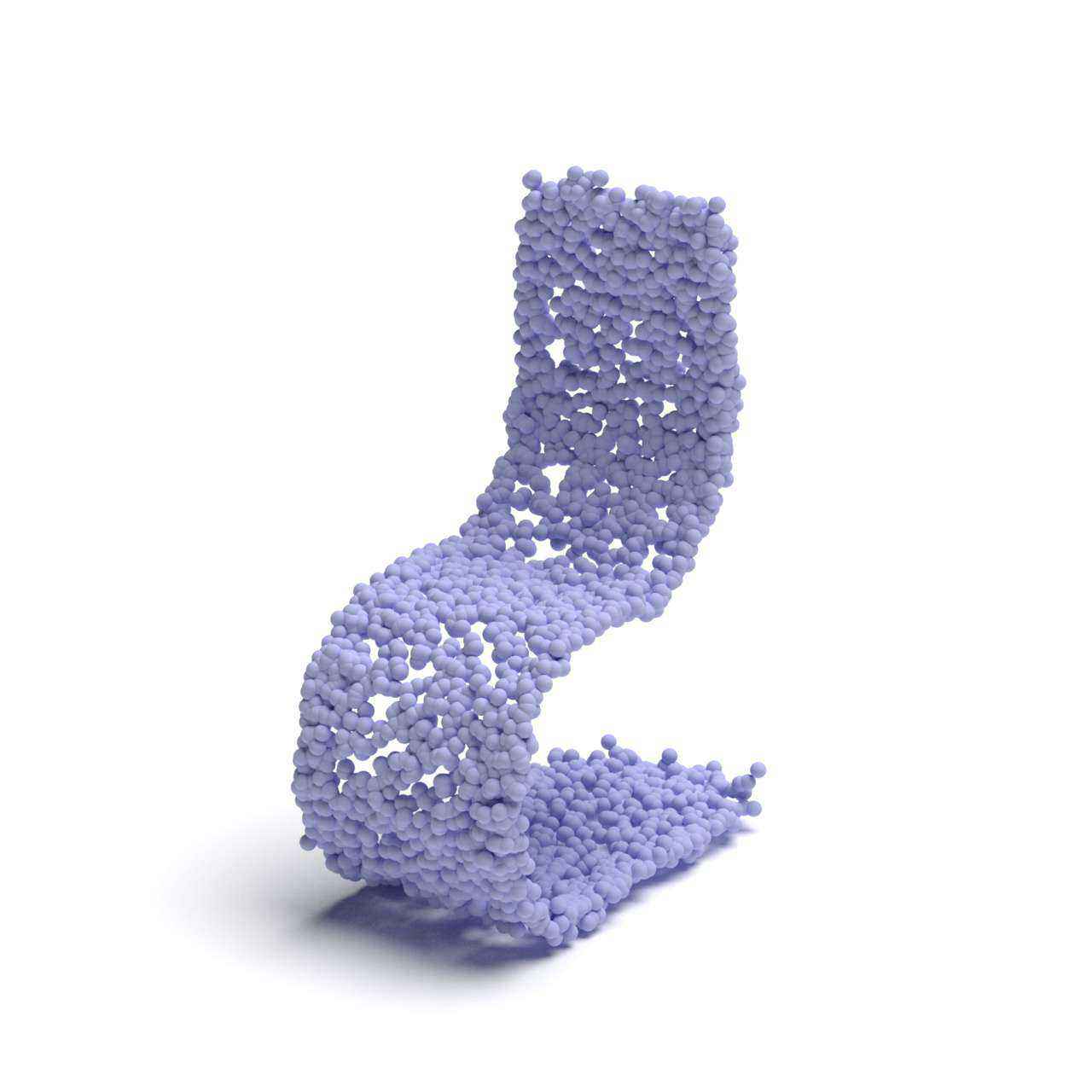}\\
		\includegraphics[width=\sizea]{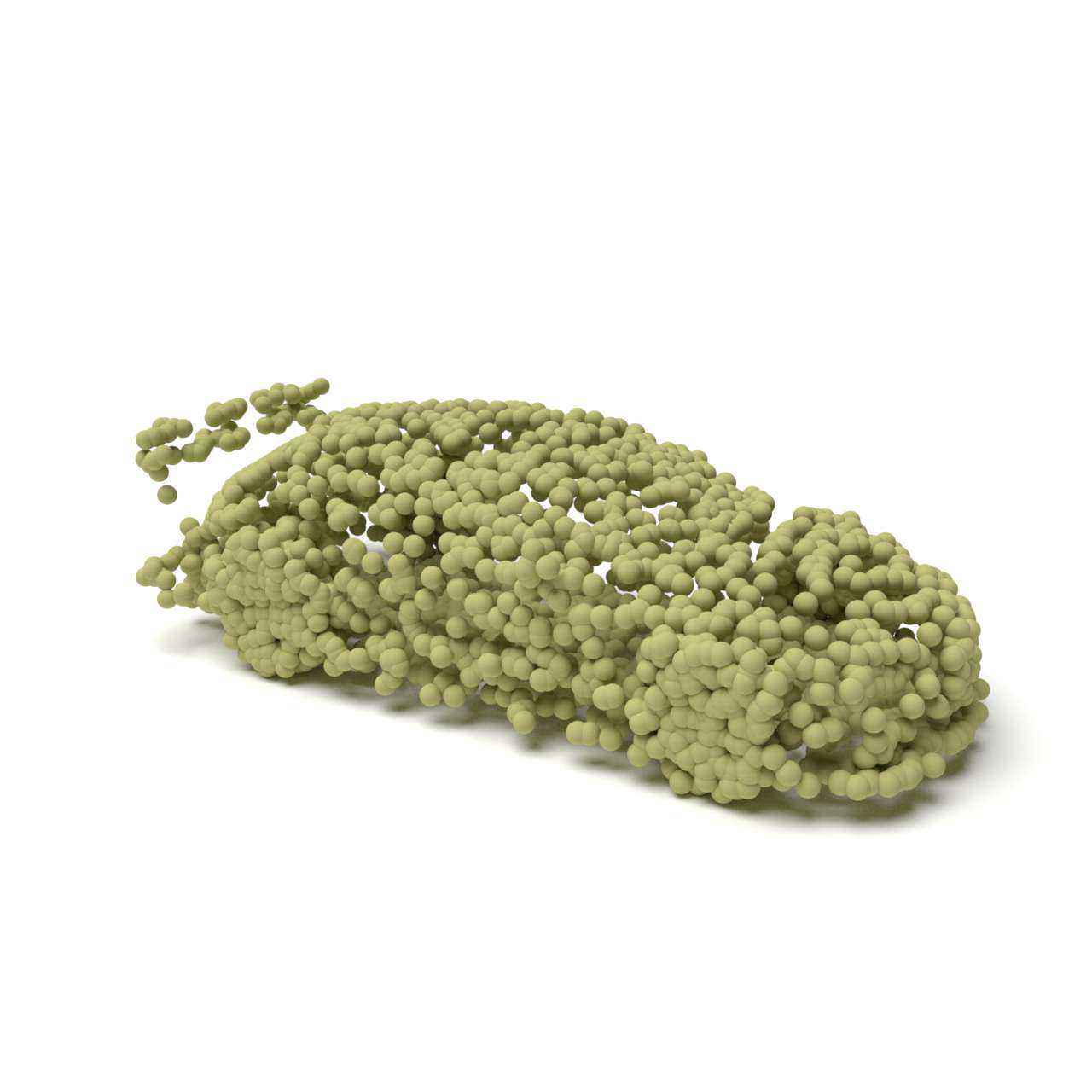} &
		\includegraphics[width=\sizea]{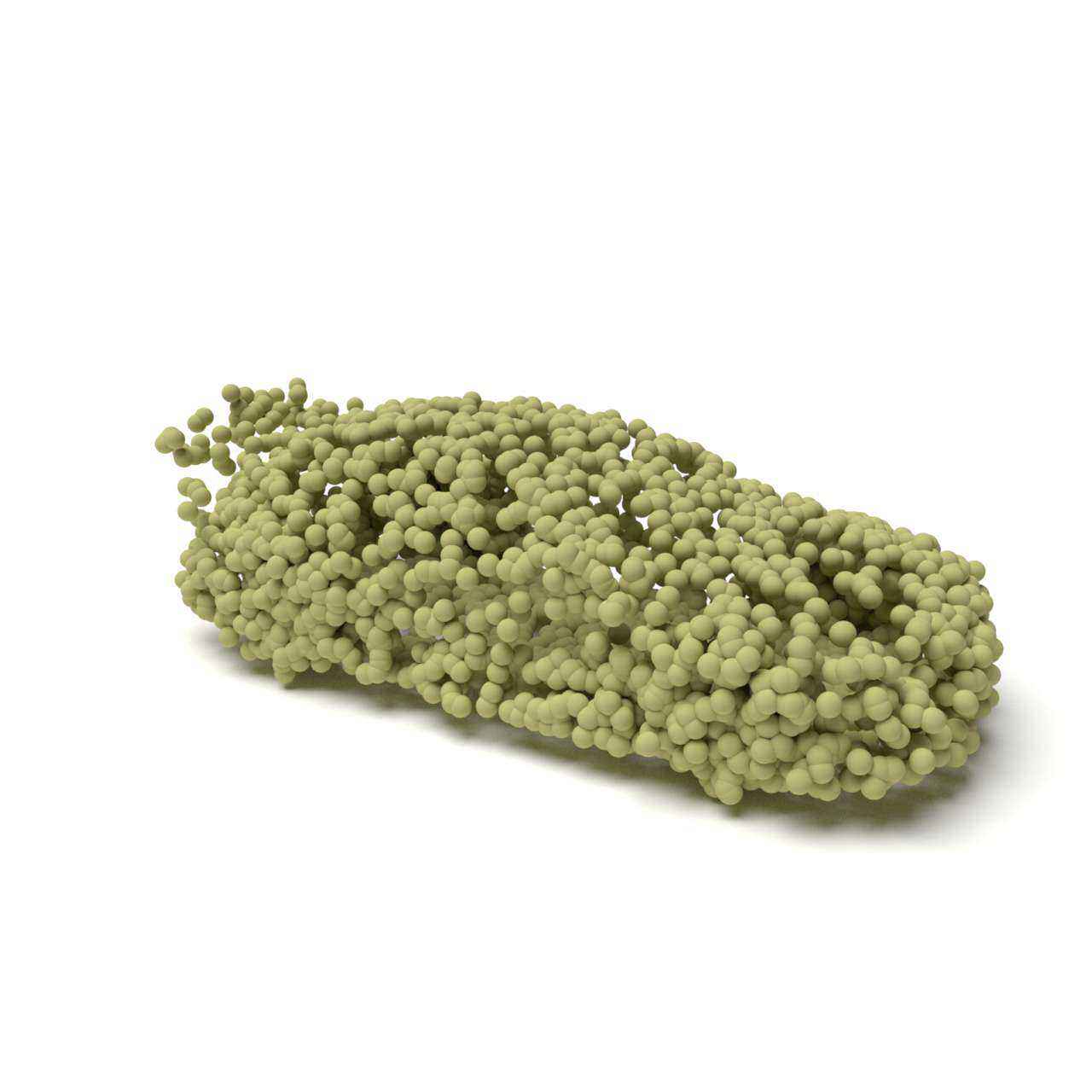} &
		\includegraphics[width=\sizea]{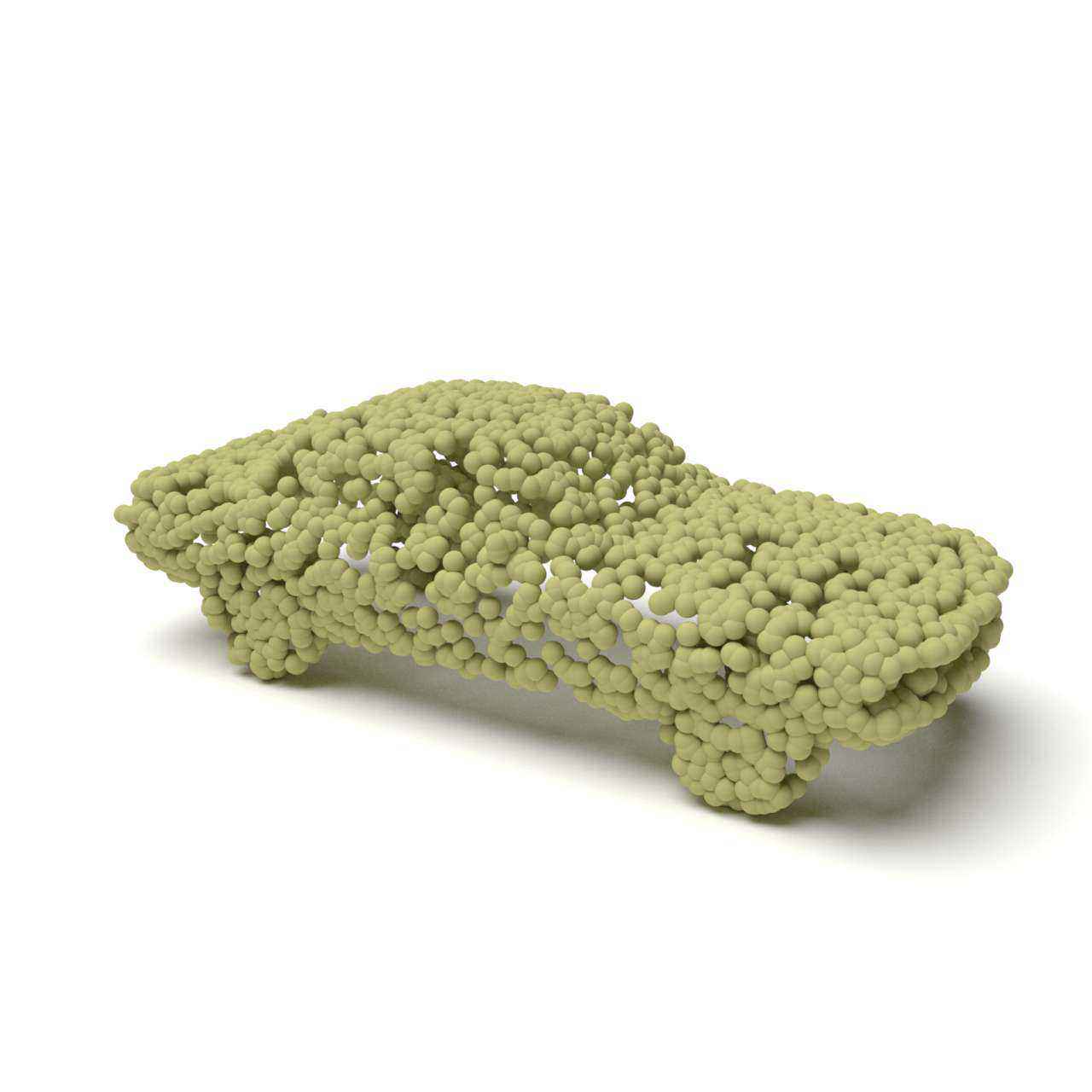} &
		\includegraphics[width=\sizea]{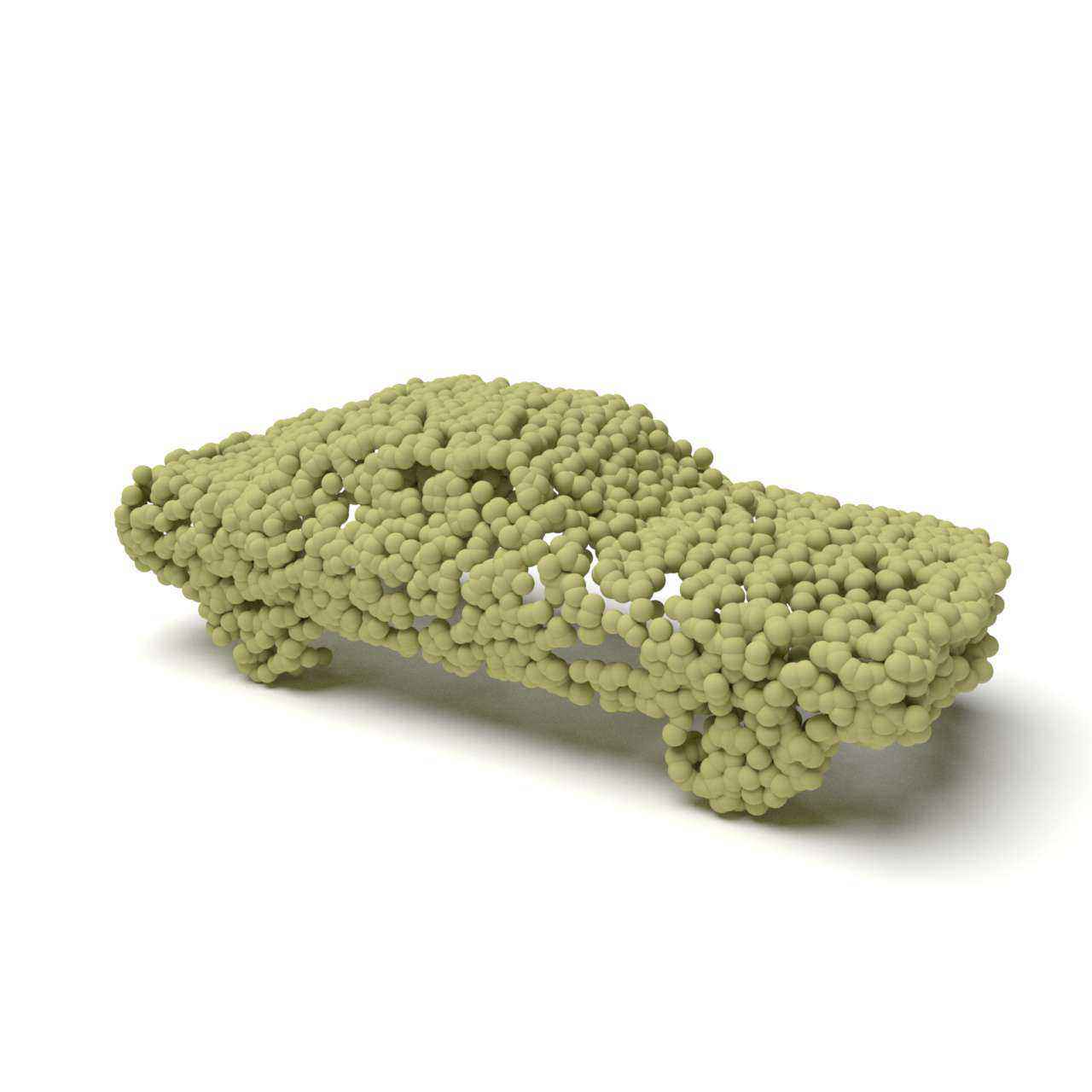}
	\end{tabular}	
	\caption{Examples of point clouds reconstructed from inputs. From top to bottom: airplane, chair, and car. On each side of the figure we show the input point cloud on the left and the reconstructed point cloud on the right.}
	\label{fig:auto-encoding}
\end{figure}

In Table~\ref{table:generation}, we report the performance of different models, as well as their number of parameters in total (full) or in the generative pathways (gen).
We first note that all the previous metrics~(JSD, MMD, and COV) sometimes assign a better score to point clouds generated by models than those from the training set~(marked in \textcolor{mygray}{gray}). The 1-NNA metric does not seem to have this problem and always gives a better score to shapes from the training set. Our model outperforms all baselines across all three categories according to 1-NNA and also obtains the best score in most cases as evaluated by other metrics. Besides, our model has the fewest parameters among compared models. In  \ifthenelse{\boolean{arxiv}}{Section~\ref{sec:more-comparision} of the appendix}{the supplementary materials}, we perform additional ablation studies to show the effectiveness of different components of our model. Figure~\ref{fig:generation} shows some examples of novel point clouds generated by our model. Figure~\ref{fig:auto-encoding} shows examples of point clouds reconstructed from given inputs.

\begin{table}[]
	\centering
	\caption{Unsupervised feature learning. Models are first trained on ShapeNet to learn shape representations, which are then evaluated on ModelNet40~(MN40) and ModelNet10~(MN10) by comparing the accuracy of off-the-shelf SVMs trained using the learned representations.}
	\label{table:classification}
	\begin{threeparttable}
		\begin{tabular}{cccc}
			\toprule
			Method                & MN40 (\%) & MN10 (\%) \\
			\midrule
			SPH~\cite{SPH}                   & 68.2      & 79.8      \\
			LFD~\cite{LFD}                   & 75.5      & 79.9      \\
			T-L Network~\cite{TLNet}           & 74.4      & -         \\
			VConv-DAE~\cite{VConvDAEDV}             & 75.5      & 80.5      \\
			3D-GAN~\cite{3dgan}                & 83.3      & 91.0        \\
			l-GAN~(EMD)~\cite{achlioptas_L3DP}           & 84.0               & \textbf{95.4}      \\
			l-GAN~(CD)~\cite{achlioptas_L3DP}            & 84.5               & \textbf{95.4}      \\
			PointGrow~\cite{pointgrow}             & 85.7      & -         \\
			MRTNet-VAE~\cite{MultiresolutionTN}    & 86.4      & -         \\
			FoldingNet~\cite{foldingnet}            & \textbf{88.4}      & 94.4      \\
			\midrule
			l-GAN~(CD)~\cite{achlioptas_L3DP}~\tnote{\dag}            & \textbf{87.0}      & 92.8      \\
			l-GAN~(EMD)~\cite{achlioptas_L3DP}~\tnote{\dag}           & 86.7               & 92.2      \\
			PointFlow~(ours)                  & 86.8      & \textbf{93.7}      \\
			\bottomrule
		\end{tabular}
		\begin{tablenotes}
			\item[\dag] We run the official code of l-GAN on our pre-processed dataset using the same encoder architecture as our model.
		\end{tablenotes}
	\end{threeparttable}
\end{table}

\subsection{Auto-encoding}
We further quantitatively compare the reconstruction ability of our flow-based auto-encoder with l-GAN~\cite{achlioptas_L3DP} and AtlasNet~\cite{atlasnet}. Following the setting of AtlasNet, the state-of-the-art in this task, we train our auto-encoder on all shapes in the ShapeNet dataset. The auto-encoder is trained with the reconstruction likelihood objective $\mathcal{L}_{\text{recon}}$ only. At test time, we sample $4096$ points per shape and split them into an input set and a reference set, each consisting of $2048$ points. We then compute the distance (CD or EMD) between the reconstructed input set and the reference set~\footnote{We use a separate reference set because we expect the auto-encoder to learn the point distribution. Exactly reproducing the input points is acceptable behavior, but should not be given a higher score than randomly sampling points from the underlying point distribution.}. Although our model is not directly trained with EMD, it obtains the best EMD score, even higher than l-GAN trained with EMD and AtlasNet which has more than $40$ times more parameters.

\begin{table}[]
	\centering
	\caption{Auto-encoding performance evaluated by CD and EMD. AtlasNet is trained with CD and l-GAN is trained on CD or EMD. Our method is not trained on CD or EMD. CD scores are multiplied by $10^4$; EMD scores are multiplied by $10^2$.}
	\label{table:auto-encoding}
	\begin{tabular}{@{}lccc@{}}
		\toprule
		Model                               & \# Parameters (M) & CD            & EMD          \\ \midrule
		l-GAN (CD)  ~\cite{achlioptas_L3DP} & 1.77          & 7.12          & 7.95          \\
		l-GAN (EMD) ~\cite{achlioptas_L3DP} & 1.77          & 8.85          & 5.26          \\
		AtlasNet~\cite{atlasnet}            & 44.9          & \textbf{5.13} & 5.97          \\
		PointFlow~(ours)                                & \textbf{1.30}          & 7.54          & \textbf{5.18} \\ \bottomrule
	\end{tabular}
\end{table}

\subsection{Unsupervised representation learning}

We finally evaluate the representation learning ability of our auto-encoders. 
Specifically, we extract the latent representations of our auto-encoder trained in the full ShapeNet dataset and train a linear SVM classifier on top of it on ModelNet10 or ModelNet40~\cite{modelnet}.
Only for this task, we normalize each individual point cloud to have zero-mean per axis and unit-variance globally, following prior works~\cite{deepsets,achlioptas_L3DP}.
We also apply random rotations along the gravity axis when training the auto-encoder.

A problem with this task is that different authors have been using different encoder architectures with a different number of parameters, making it hard to perform an apples-to-apples comparison. In addition, different authors may use different pre-processing protocols~(as also noted by Yang~\etal~\cite{foldingnet}), which could also affect the numbers. 

In Table~\ref{table:classification}, we still show the numbers reported by previous papers, but also include a comparison with l-GAN~\cite{achlioptas_L3DP} trained using the same encoder architecture and the exact same data as our model. On ModelNet10, the accuracy of our model is $1.5\%$ and $0.9\%$ higher than l-GAN~(EMD) and l-GAN~(CD), respectively. On ModelNet40, the performance of the three models is very close.

\section{Conclusion and future works}
In this paper, we propose PointFlow, a generative model for point clouds consisting of two levels of continuous normalizing flows trained with variational inference.
Future work includes applications to other tasks such as point cloud reconstruction from a single image.

\clearpage
\section{Acknowledgment}
This work was supported in part by a research gift from Magic Leap.
Xun Huang was supported by NVIDIA Graduate Fellowship.

{\small
    \bibliographystyle{ieee_fullname}
	\bibliography{ref}

\begin{thebibliography}{10}\itemsep=-1pt

\bibitem{achlioptas_L3DP}
Panos Achlioptas, Olga Diamanti, Ioannis Mitliagkas, and Leonidas Guibas.
\newblock Learning representations and generative models for 3d point clouds.
\newblock In {\em ICML}, 2018.

\bibitem{WGAN}
Martin Arjovsky, Soumith Chintala, and L{\'e}on Bottou.
\newblock Wasserstein generative adversarial networks.
\newblock In {\em ICML}, 2017.

\bibitem{shapenet}
Angel~X. Chang, Thomas Funkhouser, Leonidas Guibas, Pat Hanrahan, Qixing Huang,
  Zimo Li, Silvio Savarese, Manolis Savva, Shuran Song, Hao Su, Jianxiong Xiao,
  Li Yi, and Fisher Yu.
\newblock {ShapeNet: An Information-Rich 3D Model Repository}.
\newblock Technical Report arXiv:1512.03012 [cs.GR], Stanford University ---
  Princeton University --- Toyota Technological Institute at Chicago, 2015.

\bibitem{LFD}
Ding-Yun Chen, Xiao-Pei Tian, Edward Yu-Te Shen, and Ming Ouhyoung.
\newblock On visual similarity based 3d model retrieval.
\newblock {\em Comput. Graph. Forum}, 22:223--232, 2003.

\bibitem{neuralODE}
Tian~Qi Chen, Yulia Rubanova, Jesse Bettencourt, and David~K Duvenaud.
\newblock Neural ordinary differential equations.
\newblock In {\em NeurIPS}, 2018.

\bibitem{VLAE}
Xi Chen, Diederik~P Kingma, Tim Salimans, Yan Duan, Prafulla Dhariwal, John
  Schulman, Ilya Sutskever, and Pieter Abbeel.
\newblock Variational lossy autoencoder.
\newblock In {\em ICLR}, 2016.

\bibitem{danihelka2017comparison}
Ivo Danihelka, Balaji Lakshminarayanan, Benigno Uria, Daan Wierstra, and Peter
  Dayan.
\newblock Comparison of maximum likelihood and gan-based training of real nvps.
\newblock {\em arXiv preprint arXiv:1705.05263}, 2017.

\bibitem{Nice}
Laurent Dinh, David Krueger, and Yoshua Bengio.
\newblock Nice: Non-linear independent components estimation.
\newblock {\em CoRR}, abs/1410.8516, 2014.

\bibitem{RealNVP}
Laurent Dinh, Jascha Sohl-Dickstein, and Samy Bengio.
\newblock Density estimation using real nvp.
\newblock In {\em ICLR}, 2017.

\bibitem{learning2sample}
Oren Dovrat, Itai Lang, and Shai Avidan.
\newblock Learning to sample.
\newblock {\em arXiv preprint arXiv:1812.01659}, 2018.

\bibitem{nerual_stats}
Harrison~A Edwards and Amos~J. Storkey.
\newblock Towards a neural statistician.
\newblock In {\em ICLR}, 2017.

\bibitem{pointsetgen}
Haoqiang Fan, Hao Su, and Leonidas~J Guibas.
\newblock A point set generation network for 3d object reconstruction from a
  single image.
\newblock In {\em CVPR}, 2017.

\bibitem{MultiresolutionTN}
Matheus Gadelha, Rui Wang, and Subhransu Maji.
\newblock Multiresolution tree networks for 3d point cloud processing.
\newblock In {\em ECCV}, 2018.

\bibitem{TLNet}
Rohit Girdhar, David~F. Fouhey, Mikel Rodriguez, and Abhinav Gupta.
\newblock Learning a predictable and generative vector representation for
  objects.
\newblock In {\em ECCV}, 2016.

\bibitem{GAN}
Ian Goodfellow, Jean Pouget-Abadie, Mehdi Mirza, Bing Xu, David Warde-Farley,
  Sherjil Ozair, Aaron Courville, and Yoshua Bengio.
\newblock Generative adversarial nets.
\newblock In {\em NeurIPS}, 2014.

\bibitem{ffjord}
Will Grathwohl, Ricky T.~Q. Chen, Jesse Bettencourt, Ilya Sutskever, and David
  Duvenaud.
\newblock Ffjord: Free-form continuous dynamics for scalable reversible
  generative models.
\newblock In {\em ICLR}, 2019.

\bibitem{atlasnet}
Thibault Groueix, Matthew Fisher, Vladimir~G. Kim, Bryan Russell, and Mathieu
  Aubry.
\newblock {AtlasNet: A Papier-M\^ach\'e Approach to Learning 3D Surface
  Generation}.
\newblock In {\em CVPR}, 2018.

\bibitem{FlowGAN}
Aditya Grover, Manik Dhar, and Stefano Ermon.
\newblock Flow-gan: Combining maximum likelihood and adversarial learning in
  generative models.
\newblock In {\em AAAI}, 2018.

\bibitem{iWGAN}
Ishaan Gulrajani, Faruk Ahmed, Martin Arjovsky, Vincent Dumoulin, and Aaron~C
  Courville.
\newblock Improved training of wasserstein gans.
\newblock In {\em NeurIPS}, 2017.

\bibitem{NAF}
Chin-Wei Huang, David Krueger, Alexandre Lacoste, and Aaron~C. Courville.
\newblock Neural autoregressive flows.
\newblock In {\em ICML}, 2018.

\bibitem{bn}
Sergey Ioffe and Christian Szegedy.
\newblock Batch normalization: Accelerating deep network training by reducing
  internal covariate shift.
\newblock {\em arXiv preprint arXiv:1502.03167}, 2015.

\bibitem{GALGA}
Li Jiang, Shaoshuai Shi, Xiaojuan Qi, and Jiaya Jia.
\newblock Gal: Geometric adversarial loss for single-view 3d-object
  reconstruction.
\newblock In {\em ECCV}, 2018.

\bibitem{StyleGAN}
Tero Karras, Samuli Laine, and Timo Aila.
\newblock A style-based generator architecture for generative adversarial
  networks.
\newblock In {\em CVPR}, 2019.

\bibitem{SPH}
Michael~M. Kazhdan, Thomas~A. Funkhouser, and Szymon Rusinkiewicz.
\newblock Rotation invariant spherical harmonic representation of 3d shape
  descriptors.
\newblock In {\em Symposium on Geometry Processing}, 2003.

\bibitem{GLOW}
Diederik~P. Kingma and Prafulla Dhariwal.
\newblock Glow: Generative flow with invertible 1x1 convolutions.
\newblock In {\em NeurIPS}, 2018.

\bibitem{IAF}
Diederik~P. Kingma, Tim Salimans, and Max Welling.
\newblock Improving variational inference with inverse autoregressive flow.
\newblock In {\em NeurIPS}, 2016.

\bibitem{VAE}
Diederik~P Kingma and Max Welling.
\newblock Auto-encoding variational bayes.
\newblock In {\em ICLR}, 2014.

\bibitem{VideoFlow}
Manoj Kumar, Mohammad Babaeizadeh, Dumitru Erhan, Chelsea Finn, Sergey Levine,
  Laurent Dinh, and Durk Kingma.
\newblock Videoflow: A flow-based generative model for video.
\newblock {\em arXiv preprint arXiv:1903.01434}, 2019.

\bibitem{DeformNetFD}
Andrey Kurenkov, Jingwei Ji, Animesh Garg, Viraj Mehta, JunYoung Gwak,
  Christopher~B. Choy, and Silvio Savarese.
\newblock Deformnet: Free-form deformation network for 3d shape reconstruction
  from a single image.
\newblock In {\em WACV}, 2018.

\bibitem{PC-GAN}
Chun-Liang Li, Manzil Zaheer, Yang Zhang, Barnabas Poczos, and Ruslan
  Salakhutdinov.
\newblock Point cloud gan.
\newblock {\em arXiv preprint arXiv:1810.05795}, 2018.

\bibitem{EfficientDP}
Kejie Li, Trung Pham, Huangying Zhan, and Ian~D. Reid.
\newblock Efficient dense point cloud object reconstruction using deformation
  vector fields.
\newblock In {\em ECCV}, 2018.

\bibitem{tst}
David Lopez-Paz and Maxime Oquab.
\newblock Revisiting classifier two-sample tests.
\newblock In {\em ICLR}, 2017.

\bibitem{maaten2008visualizing}
Laurens van~der Maaten and Geoffrey Hinton.
\newblock Visualizing data using t-sne.
\newblock {\em Journal of machine learning research}, 9(Nov):2579--2605, 2008.

\bibitem{AAE}
Alireza Makhzani, Jonathon Shlens, Navdeep Jaitly, Ian Goodfellow, and Brendan
  Frey.
\newblock Adversarial autoencoders.
\newblock {\em arXiv preprint arXiv:1511.05644}, 2015.

\bibitem{PixelRNN}
Aaron van~den Oord, Nal Kalchbrenner, and Koray Kavukcuoglu.
\newblock Pixel recurrent neural networks.
\newblock In {\em ICML}, 2016.

\bibitem{MAF}
George Papamakarios, Theo Pavlakou, and Iain Murray.
\newblock Masked autoregressive flow for density estimation.
\newblock In {\em NeurIPS}, 2017.

\bibitem{WaveGlow}
Ryan Prenger, Rafael Valle, and Bryan Catanzaro.
\newblock Waveglow: A flow-based generative network for speech synthesis.
\newblock {\em CoRR}, abs/1811.00002, 2018.

\bibitem{pointnet}
Charles~R Qi, Hao Su, Kaichun Mo, and Leonidas~J Guibas.
\newblock Pointnet: Deep learning on point sets for 3d classification and
  segmentation.
\newblock In {\em CVPR}, 2017.

\bibitem{pointnet++}
Charles~Ruizhongtai Qi, Li Yi, Hao Su, and Leonidas~J Guibas.
\newblock Pointnet++: Deep hierarchical feature learning on point sets in a
  metric space.
\newblock In {\em NeurIPS}, 2017.

\bibitem{NormalizingFlow}
Danilo~Jimenez Rezende and Shakir Mohamed.
\newblock Variational inference with normalizing flows.
\newblock In {\em ICML}, 2015.

\bibitem{stochastic_bp}
Danilo~Jimenez Rezende, Shakir Mohamed, and Daan Wierstra.
\newblock Stochastic backpropagation and approximate inference in deep
  generative models.
\newblock In {\em ICML}, 2014.

\bibitem{VConvDAEDV}
Abhishek Sharma, Oliver Grau, and Mario Fritz.
\newblock Vconv-dae: Deep volumetric shape learning without object labels.
\newblock In {\em ECCV Workshops}, 2016.

\bibitem{shoef2019pointwise}
Matan Shoef, Sharon Fogel, and Daniel Cohen-Or.
\newblock Pointwise: An unsupervised point-wise feature learning network.
\newblock {\em arXiv preprint arXiv:1901.04544}, 2019.

\bibitem{pointgrow}
Yongbin Sun, Yue Wang, Ziwei Liu, Joshua~E Siegel, and Sanjay~E Sarma.
\newblock Pointgrow: Autoregressively learned point cloud generation with
  self-attention.
\newblock {\em arXiv preprint arXiv:1810.05591}, 2018.

\bibitem{usenko2015reconstructing}
Vladyslav Usenko, Jakob Engel, J{\"o}rg St{\"u}ckler, and Daniel Cremers.
\newblock Reconstructing street-scenes in real-time from a driving car.
\newblock In {\em 3DV}, 2015.

\bibitem{SylvesterNF}
Rianne van~den Berg, Leonard Hasenclever, Jakub~M. Tomczak, and Max Welling.
\newblock Sylvester normalizing flows for variational inference.
\newblock In {\em UAI}, 2018.

\bibitem{pixelCNN}
Aaron Van~den Oord, Nal Kalchbrenner, Lasse Espeholt, Oriol Vinyals, Alex
  Graves, et~al.
\newblock Conditional image generation with pixelcnn decoders.
\newblock In {\em NeurIPS}, 2016.

\bibitem{3dgan}
Jiajun Wu, Chengkai Zhang, Tianfan Xue, William~T Freeman, and Joshua~B
  Tenenbaum.
\newblock {Learning a Probabilistic Latent Space of Object Shapes via 3D
  Generative-Adversarial Modeling}.
\newblock In {\em NeurIPS}, 2016.

\bibitem{modelnet}
Zhirong Wu, Shuran Song, Aditya Khosla, Fisher Yu, Linguang Zhang, Xiaoou Tang,
  and Jianxiong Xiao.
\newblock 3d shapenets: A deep representation for volumetric shapes.
\newblock In {\em CVPR}, 2015.

\bibitem{1nnacc}
Qiantong Xu, Gao Huang, Yang Yuan, Chuan Guo, Yu Sun, Felix Wu, and Kilian
  Weinberger.
\newblock An empirical study on evaluation metrics of generative adversarial
  networks.
\newblock {\em arXiv preprint arXiv:1806.07755}, 2018.

\bibitem{foldingnet}
Yaoqing Yang, Chen Feng, Yiru Shen, and Dong Tian.
\newblock Foldingnet: Point cloud auto-encoder via deep grid deformation.
\newblock In {\em CVPR}, 2018.

\bibitem{patch-based-upsampling}
Wang Yifan, Shihao Wu, Hui Huang, Daniel Cohen-Or, and Olga Sorkine-Hornung.
\newblock Patch-based progressive 3d point set upsampling.
\newblock {\em arXiv preprint arXiv:1811.11286}, 2018.

\bibitem{EC-Net}
Lequan Yu, Xianzhi Li, Chi-Wing Fu, Daniel Cohen-Or, and Pheng-Ann Heng.
\newblock Ec-net: an edge-aware point set consolidation network.
\newblock In {\em ECCV}, 2018.

\bibitem{PU-net}
Lequan Yu, Xianzhi Li, Chi-Wing Fu, Daniel Cohen-Or, and Pheng-Ann Heng.
\newblock Pu-net: Point cloud upsampling network.
\newblock In {\em CVPR}, 2018.

\bibitem{deepsets}
Manzil Zaheer, Satwik Kottur, Siamak Ravanbakhsh, Barnabas Poczos, Ruslan~R
  Salakhutdinov, and Alexander~J Smola.
\newblock Deep sets.
\newblock In {\em NeurIPS}, 2017.

\bibitem{3D-AAE}
Maciej Zamorski, Maciej Zieba, Rafa{\l} Nowak, Wojciech Stokowiec, and Tomasz
  Trzci{\'n}ski.
\newblock Adversarial autoencoders for generating 3d point clouds.
\newblock {\em arXiv preprint arXiv:1811.07605}, 2018.

\end{thebibliography}
}

\clearpage

\newcolumntype{Y}{>{\centering\arraybackslash}X}

\begin{appendix}
\section{Overview}\label{sec:sup-overview}
In the appendix, we first describe the detailed hyper-parameters and model architectures for our experiments in Section~\ref{sec:hyper-param}.
We then compare our model with additional baselines to understand the effect of different model components in Section~\ref{sec:more-comparision}.
Limitations and typical failure cases are discussed in Section~\ref{sec:limitation}.
Finally, additional visualizations of latent space t-SNE, interpolations and flow transformations are presented in Section~\ref{sec:tsne}, Section~\ref{sec:interpolation}, and Section~\ref{sec:more-visualization} respectively.

\section{Training details}\label{sec:hyper-param}

In this section, we provide details about our network architectures and training hyper-parameters. 
We will release the code to reproduce our experiments.
Please refer to algorithm~\ref{alg:pointflow-train} for the detailed training procedure.

\textbf{Encoder.}
The architecture of our encoder follows that of Achlioptas \etal~\cite{achlioptas_L3DP}.
Specifically, we first use 1D Convolution with filter size $128$, $128$, $256$, and $512$ to process each point independently and then use max pooling to create a $512$-dimension feature as done in PointNet~\cite{pointnet}.
Such a feature is invariant to the permutation of points due to the max-pooling.
Finally, we apply a three-layer MLP with $256$ and $128$ hidden dimensions to convert the permutation invariant feature to a $D_z$-dimension one.
For the unsupervised representation learning experiment, we set $D_z=512$ following convention.
For all other experiments, $D_z$ is set to $128$.

\textbf{CNF prior.}
The CNF prior models the distribution $P_\psi(z)$. 
We follow FFJORD~\cite{ffjord}'s released code to use three \texttt{concatsquash} layers to model the dynamics $f_\psi$.
A \texttt{concatsquash} layer is defined as:
\begin{align}
    \operatorname{CS}(x, t) = (W_xx + b_x) \sigma(W_t t + b_t) + ( W_b t + b_b t),
\end{align}
where $W_x$, $b_x$, $W_t$, $b_t$, $W_b$, and $b_b$ are all trainable parameters and $\sigma(\cdot)$ is the sigmoid function.
$f_\psi$ uses three \texttt{concatsquash} layers with a hidden dimension $256$.
Tanh is used as the non-linearity between layers.

We use a Moving Batch Normalization layer to learn the scale of each dimension before and after the CNF, following FFJORD's released code~\cite{ffjord}.
Specifically, Moving Batch Normalization is defined as 
\begin{align}
    \operatorname{MBN}(x) = \frac{x - \mu}{\sigma} \gamma + \beta,
\end{align}
where $\gamma$ and $\beta$ are trainable parameters,
Different from batch normalization proposed by Ioffe and Szegedy~\cite{bn}, $\mu$ and $\sigma$ are running averages of the batch mean and standard deviation.
MovingBatchNorm is invertible : $ \operatorname{MBN}^{-1}(y) = \frac{y - \beta}{\gamma} \sigma + \mu$.
Its log determinant is given as:
\begin{align}
    \log{\det{\left|\frac{\partial \operatorname{MBN}(x)}{\partial x}\right|}} = \sum_i \log{|\gamma_i|} - \log{|\sigma_i|}.
\end{align}

\textbf{CNF decoder.}
The CNF decoder models the reconstruction likelihood $P_\theta(X|z)$.
We extend the \texttt{concatsquash} layer to condition on the latent vector $z$:
\begin{align}
    \operatorname{CCS}(x, z, t) &= (W_xx + b_x) \sigma(W_{tt} t + W_{tz} z + b_t) \nonumber \\
    &\quad  + ( W_{bt} t + W_{bz} z + b_b t),
\end{align}
where $W_x, W_{tt}, W_{tz}, W_{bt}, W_{bz}, b_t, b_b$ are all learnable parameters.
The CNF decoder uses four conditional \texttt{concatsquash} layers with a hidden dimension $512$ to model the dynamic $g_\theta$.
The non-linearity between layers is Tanh.
Similar to the CNF prior model, we also add a Moving Batch Normalization layer before and after the CNF.
In this case, all 3D points (from different shapes) from a batch are used to compute the batch statistics.

\textbf{Hyper-parameters.}
We use an Adam optimizer with an initial learning rate $0.002$, $\beta_{1}=0.9$, and $\beta_{2}=0.999$.
The learning rate decays linearly to $0$ starting at the 2000$^{th}$ epoch and ends at the $4000^{th}$ epoch.
We do not use any weight decay.
We also learn the integration time $t_1$ during training by back-propogation~\cite{neuralODE}.

\section{Additional comparisons}\label{sec:more-comparision}

\begin{table*}[t]
	\centering
	\caption{Ablation studies. $\uparrow$: the higher the better, $\downarrow$: the lower the better. The best scores are highlighted in bold. MMD-CD scores are multiplied by $10^3$; MMD-EMD scores are multiplied by $10^2$; JSDs are multiplied by $10^2$.}
	\label{table:generation-more}
	\begin{small}
		\begin{tabularx}{\textwidth}{ll*{11}{Y}}
			\toprule
			&  & \multicolumn{2}{c}{\# Parameters (M)} & \multirow{2}{*}{JSD~($\downarrow$)} & \multicolumn{2}{c}{MMD ($\downarrow$)} & \multicolumn{2}{c}{COV (\%, $\uparrow$)} & \multicolumn{2}{c}{1-NNA (\%, $\downarrow$)} \\ 
			\cmidrule(lr){3-4} \cmidrule(lr){6-7} \cmidrule(lr){8-9} \cmidrule(l){10-11} 
			Category                  & Model                  & Full              & Gen              &              & CD             & EMD      & CD     & EMD      & CD         & EMD  \\ \midrule
			\multirow{6}{*}{Airplane} & NS~\cite{nerual_stats} & 2.29              & 1.00             &\textbf{1.74} & 0.655          & 4.51     & 7.81   & 4.51     & 99.61      & 99.61        \\
			                          & VAECNF                 & \textbf{1.47}              & \textbf{0.92}    & 6.30         & 0.261          & 3.35     & 41.98  & 46.17    & 88.64      & 82.72   \\
			                          & WGAN-CNF               & 1.75              & 1.06             & 4.29         & 0.254          & \textbf{3.23}     & 42.47  & \textbf{48.40}     & 75.80      & 75.68            \\
			                          
			                          & PointFlow~(ours)       & 1.61     & 1.06             & 4.92         & \textbf{0.217} & 3.24     & \textbf{46.91}               & \textbf{48.40}              & \textbf{75.68}         & \textbf{75.06}        \\ 
			\cmidrule(l){2-11} 
			& Training set              & -                 & -                & 6.61                            & 0.226             & 3.08              & 42.72               & 49.14              & 70.62                  & 67.53                 \\ 
			\bottomrule
		\end{tabularx}
	\end{small}
\end{table*}

In this section, we compare our model to more baselines to show the effectiveness of the model design.
The first baseline is Neural Statistician (NS)~\cite{nerual_stats}, a state-of-the-art generative model for sets. We modify its official code for generating 2D spatial coordinates of MNIST digits to make it work with 3D point cloud coordinates. We use the same encoder architecture as our model, and use the VAE decoder provided by authors with the input dimension changed from $2$ to $3$. It differs from our model mainly in 1) using VAEs instead of CNFs to model the reconstruction likelihood, and 2) using a simple Gaussian prior instead of a flow-based one.
The second baseline is VAECNF, where we use the CNF to model the reconstruction likelihood but not prior.
Specifically, the VAECNF optimizes ELBO in the following form:
{\small\begin{align}
\mathcal{L}(X;\phi, \theta) &= \sum_{x\in X}\left(
\log{P(G_\theta^{-1}(x;z))} - \int_{t_0}^{t_1} \operatorname{Tr}\left(\frac{\partial g_\theta}{\partial y(t)}\right) dt \right) \nonumber \\
& +  D_{KL}(Q_\phi(z|X)||P(z))\,,
\end{align}}where $P(z)$ is a standard Gaussian $\mathcal{N}(0,I)$ and $D_{KL}$ is the KL-divergence.
As another baseline, we follow l-GAN~\cite{achlioptas_L3DP} to train a WGAN~\cite{iWGAN} in the latent space of our pretrained auto-encoder.
Both the discriminator and the generator are MLP with batch normalization between layers.
The generator has three layers with hidden dimensions 256.
The discriminator has three layers with hidden dimensions 512.

The results are presented in Table~\ref{table:generation-more}. Neural Statistician~\cite{nerual_stats} is able to learn the marginal point distribution but fails to learn the correct shape distribution, as it obtains the best marginal JSD but very poor scores according to metrics that measure similarities between shape distributions. Also, using a flexible prior parameterized by a CNF~(PointFlow) is better than using a simple Gaussian prior~(VAECNF) or a prior learned with a latent GAN~(WGAN-CNF) that requires two-stage training.

\begin{algorithm}[t]
  \caption{PointFlow training.}
  \label{alg:pointflow-train}
\begin{algorithmic}
\REQUIRE{
    Point cloud encoder $Q_\phi$; CNFs $G_\theta$ and $F_\psi$, whose dynamics are defined by $g_\theta$ and $f_\psi$, respectively; Integration time interval $[t_0, t_1]$; Learning rate $\alpha$; Total number of training iterations $T$; Data samples $X_t$.
}
\FOR{$t = 1, 2, \dots, T$ do} 
    \STATE {$\mu,\sigma \leftarrow Q_\phi(X_t)$} \COMMENT{$d$ is the dimension of $\mu$}
    \STATE {$\mathcal{L}_{ent} = \frac{d}{2}(1+\ln{(2\pi)}) + \sum_{i=1}^d \ln{\sigma_i}$}
    \STATE {$z \leftarrow \epsilon \odot \sigma + \mu $} \COMMENT{Reparameterization.}
    \STATE {$w \leftarrow F_\psi^{-1}(z)$}
    \STATE {$\mathcal{L}_{prior} = \log{\mathcal{N}(w; 0, I)} - \int_{t_0}^{t_1} \operatorname{Tr}\left(\frac{\partial f_\psi(w(t))}{\partial w(t)}\right)dt$}
    
    \STATE {$L \leftarrow 0$}
    \FOR {$x_i \in X_t$ do} 
        \STATE {$y_i \leftarrow G_\theta^{-1}(x_i; z)$}
        \STATE {$L_i \leftarrow \log{\mathcal{N}(y_i; 0, I)} - \int_{t_0}^{t_1} \operatorname{Tr}\left(\frac{\partial g_\theta(y_i(t))}{\partial y_i(t)}\right)dt$}
        \STATE {$L \leftarrow L + L_i$}    
    \ENDFOR 
    \STATE {$\mathcal{L}_{recon} = \frac{L}{|X_t|}$}
    \STATE {$\mathcal{L} = \mathcal{L}_{recon} + \mathcal{L}_{prior} + \mathcal{L}_{ent}$}
    \STATE {$\phi, \psi, \theta \leftarrow Adam(\mathcal{L}, \phi, \psi, \theta)$}
\ENDFOR
\STATE 
\textbf{return} $Q_\phi$, $G_\theta$, $F_\psi$
\end{algorithmic}
\end{algorithm}

\section{Limitation and failure cases}\label{sec:limitation}

\begin{figure}[]
	\centering	
	\newcommand{\sizea}{0.235\linewidth}
	\setlength{\arrayrulewidth}{.5pt}%
	\setlength{\tabcolsep}{1pt}
	\renewcommand{\arraystretch}{0}
	\begin{tabular}{cc;{2pt/2pt}cc}
		\includegraphics[width=\sizea]{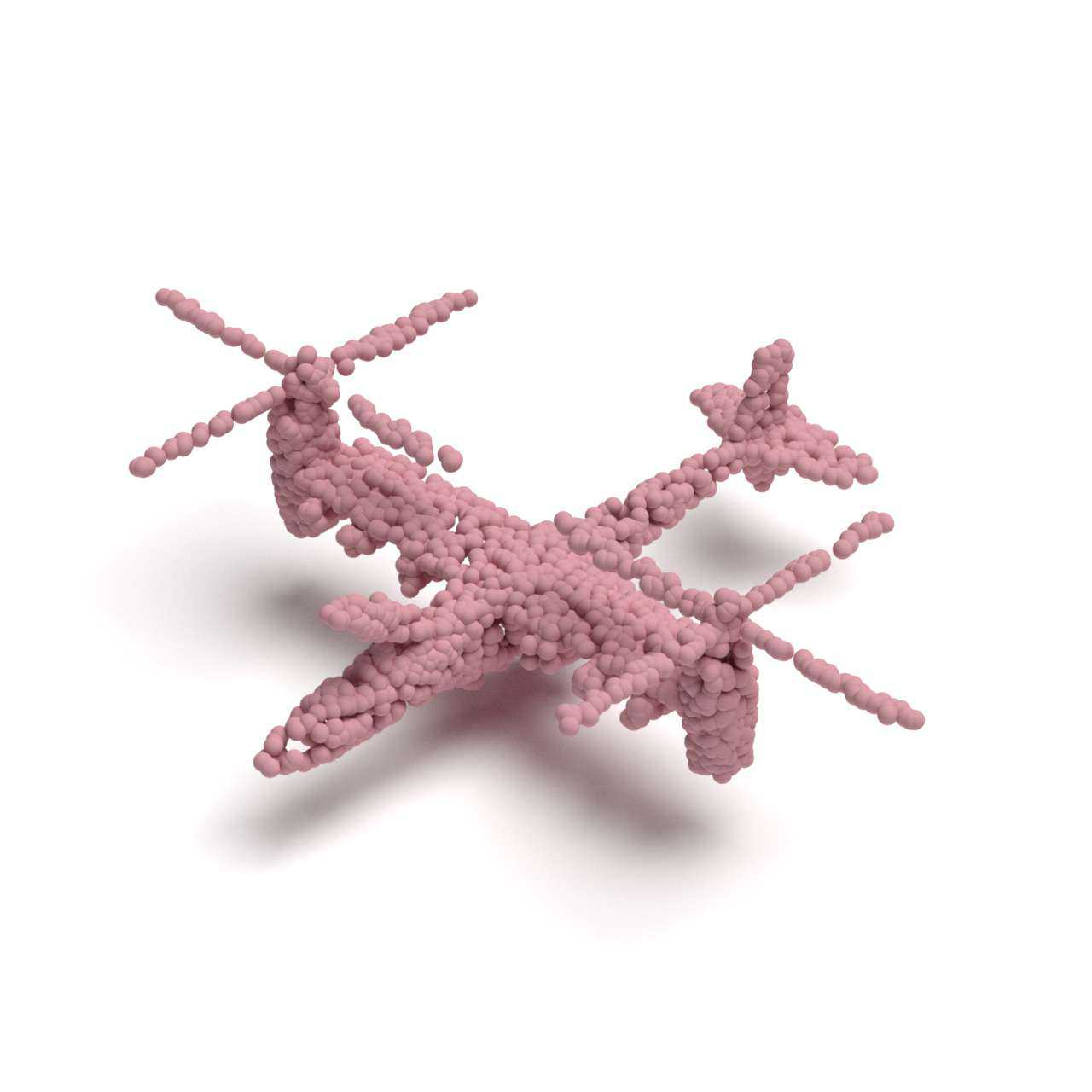} &
		\includegraphics[width=\sizea]{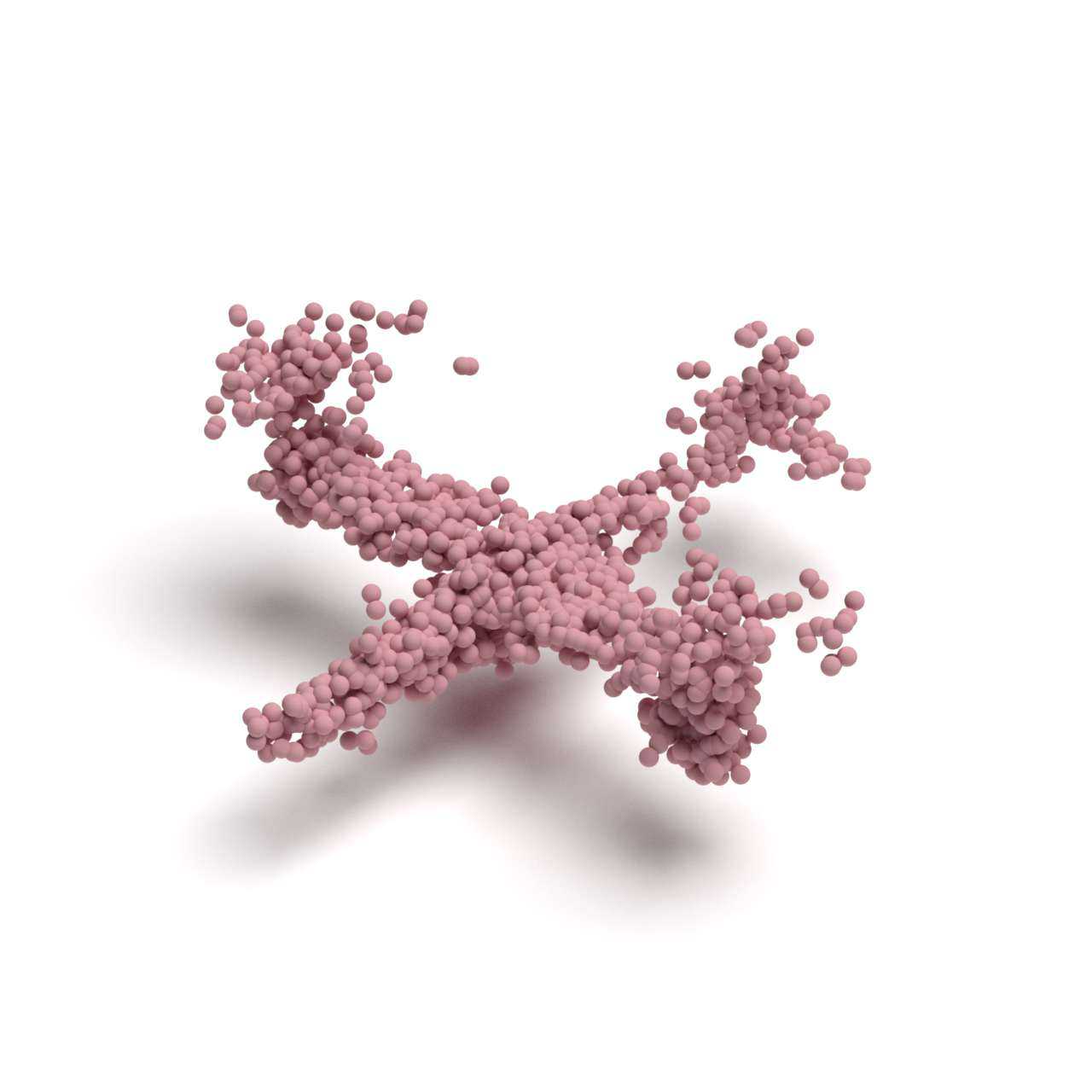} &
		\includegraphics[width=\sizea]{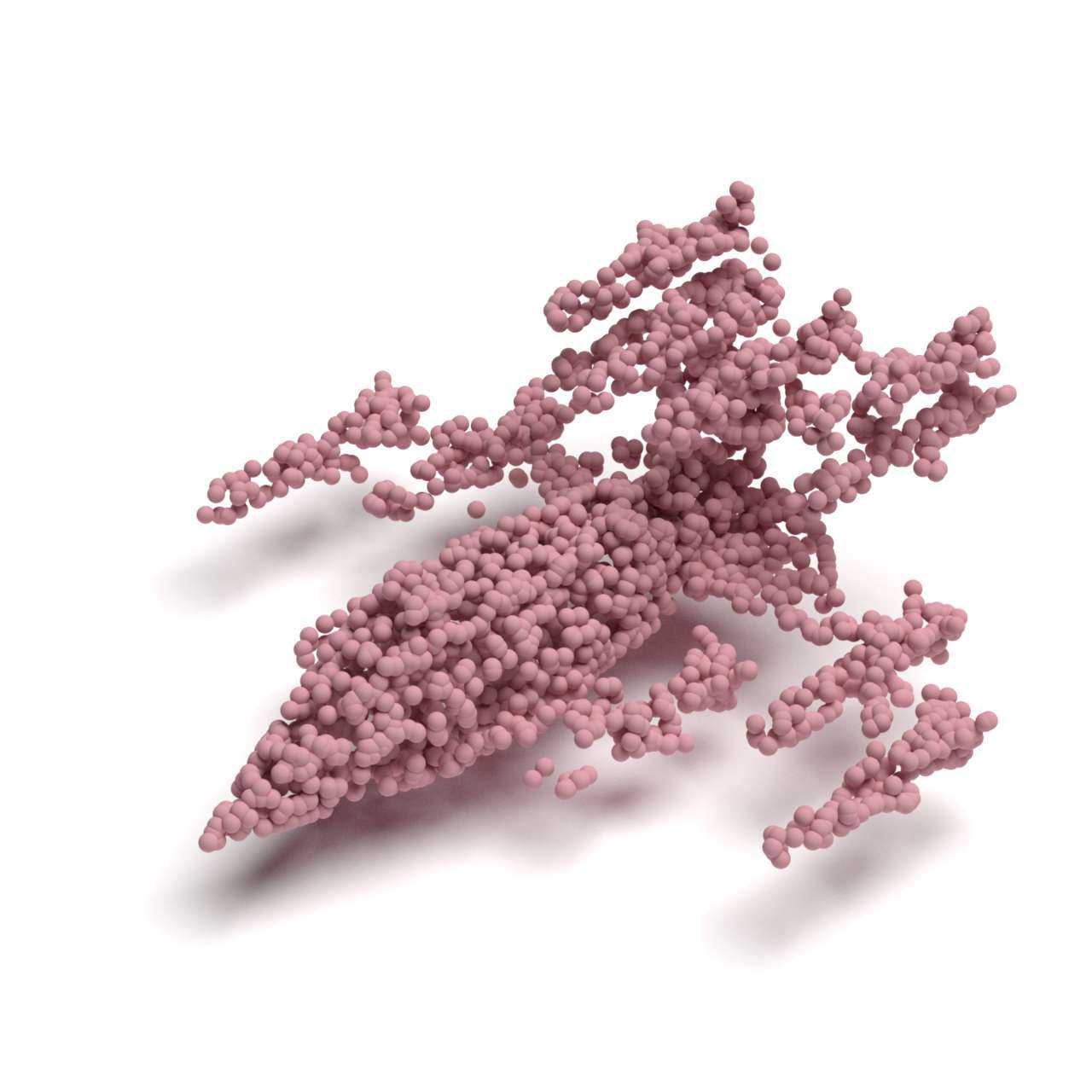} &
		\includegraphics[width=\sizea]{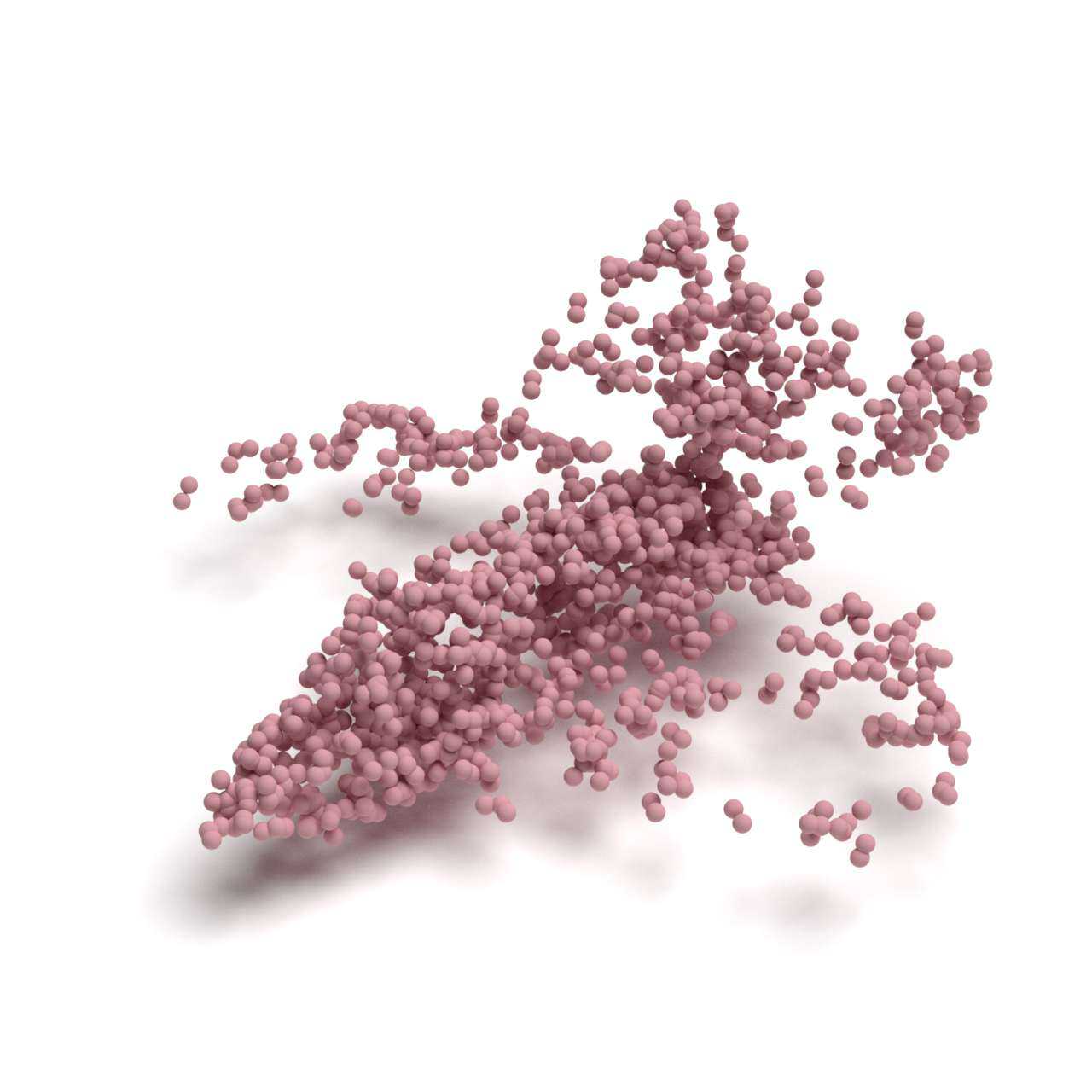} \\
		\includegraphics[width=\sizea]{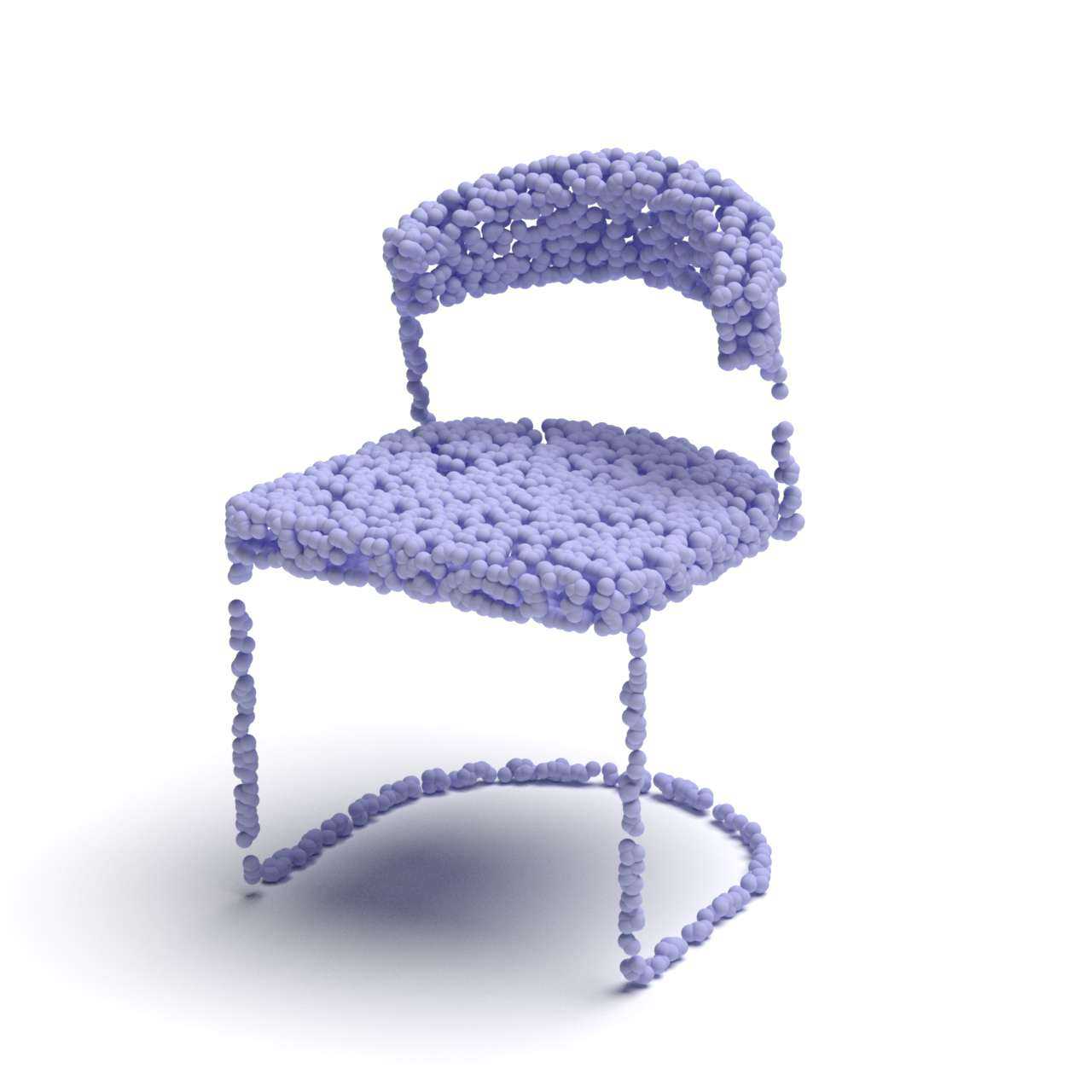} &
		\includegraphics[width=\sizea]{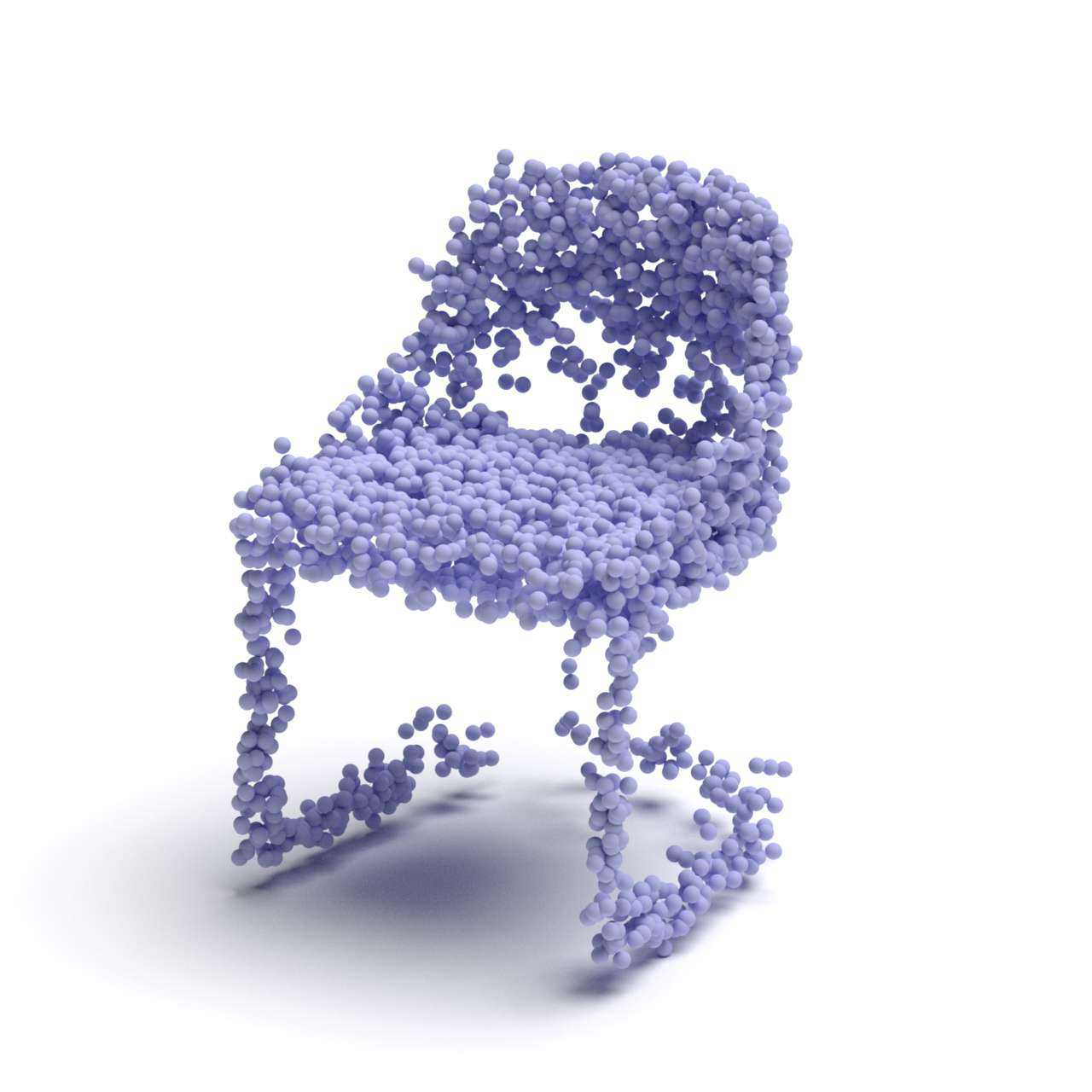} &
		\includegraphics[width=\sizea]{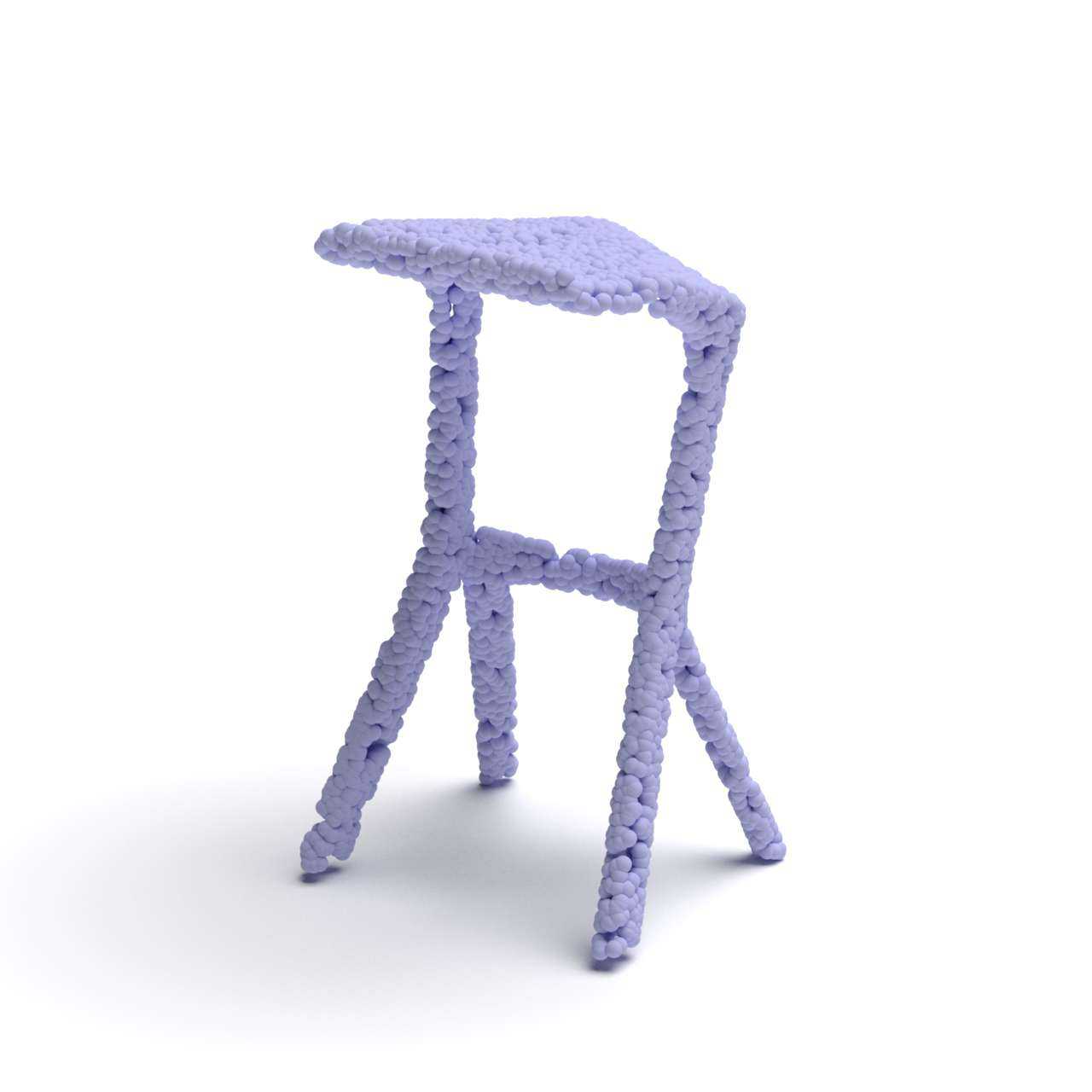} &
		\includegraphics[width=\sizea]{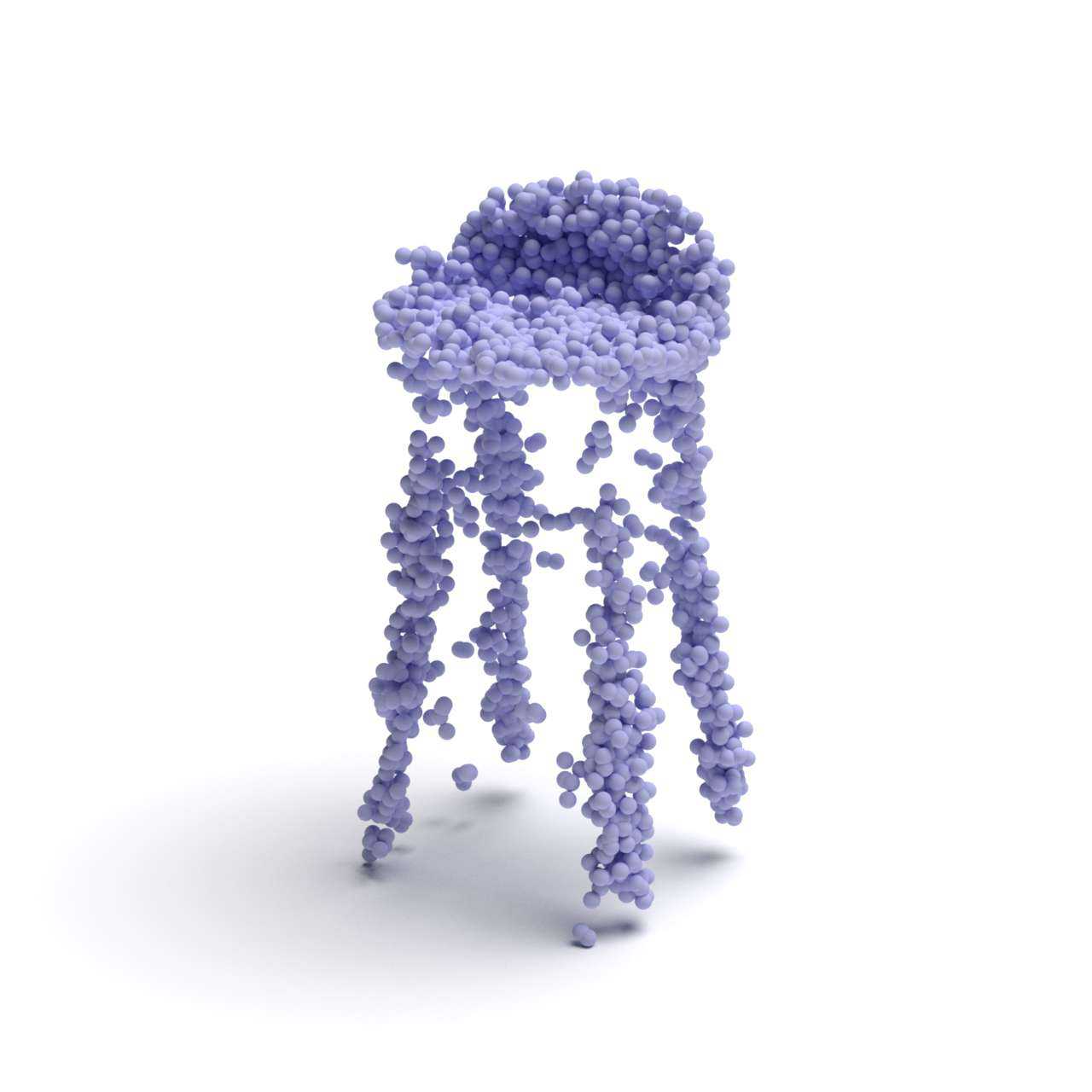} \\
		\includegraphics[width=\sizea]{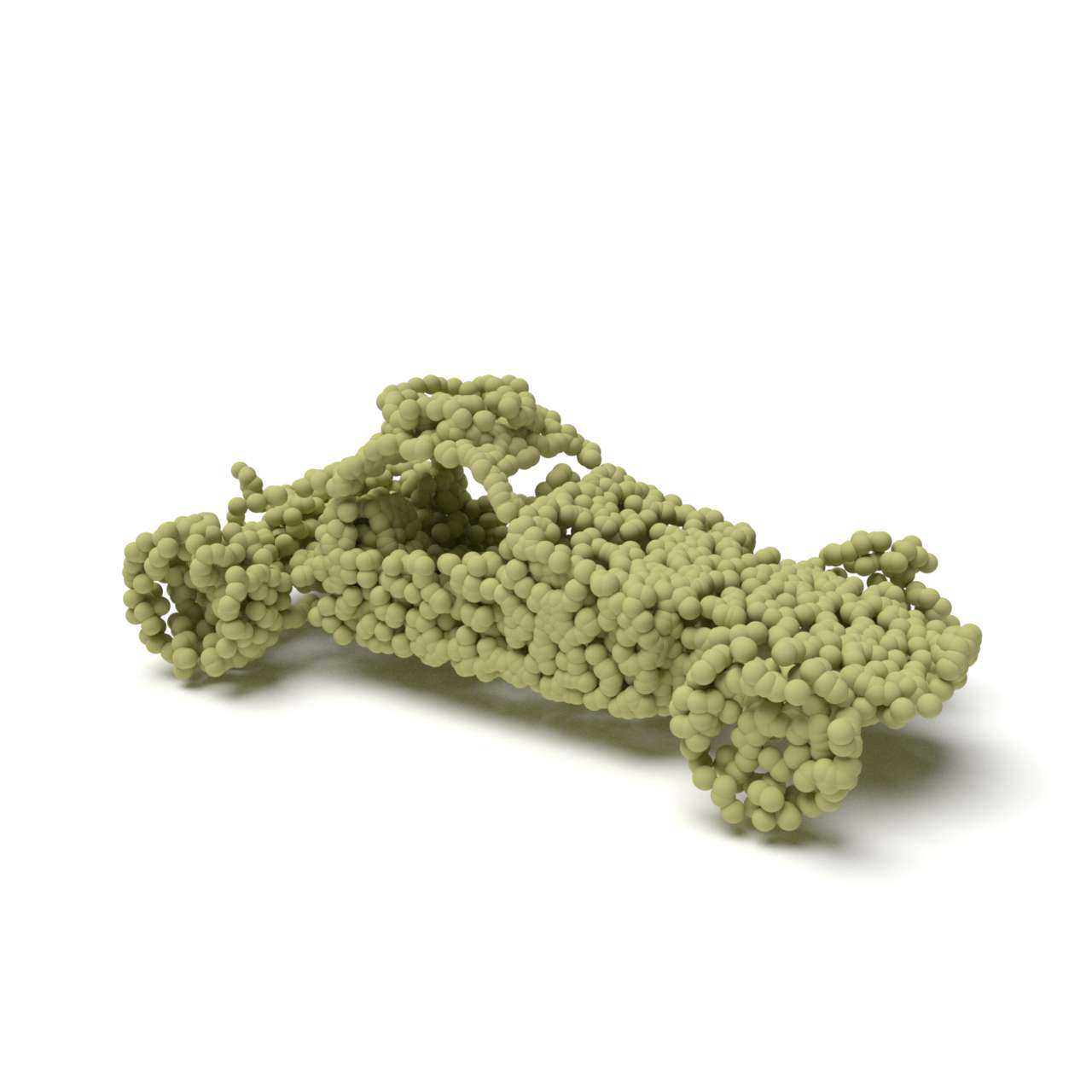} &
		\includegraphics[width=\sizea]{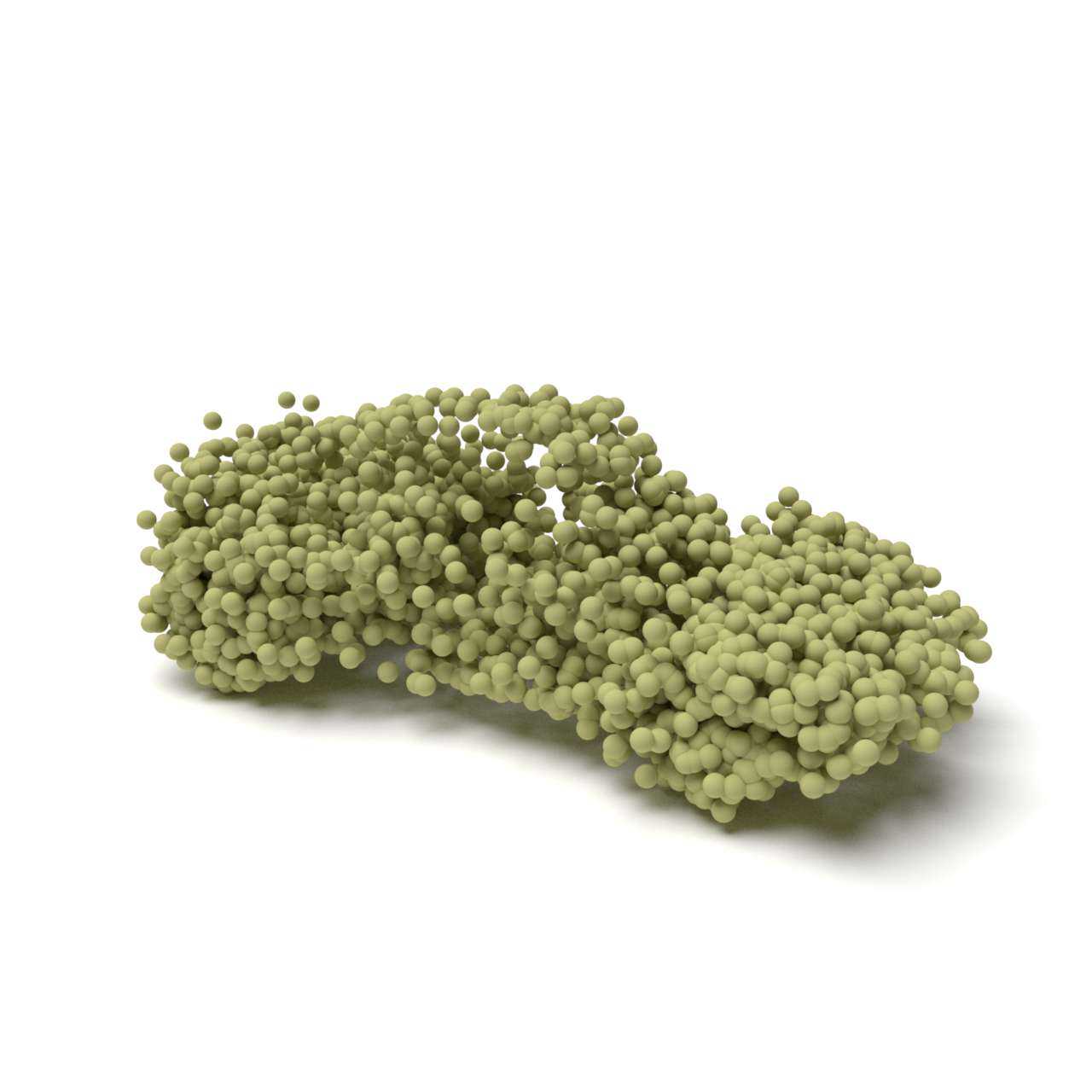} &
		\includegraphics[width=\sizea]{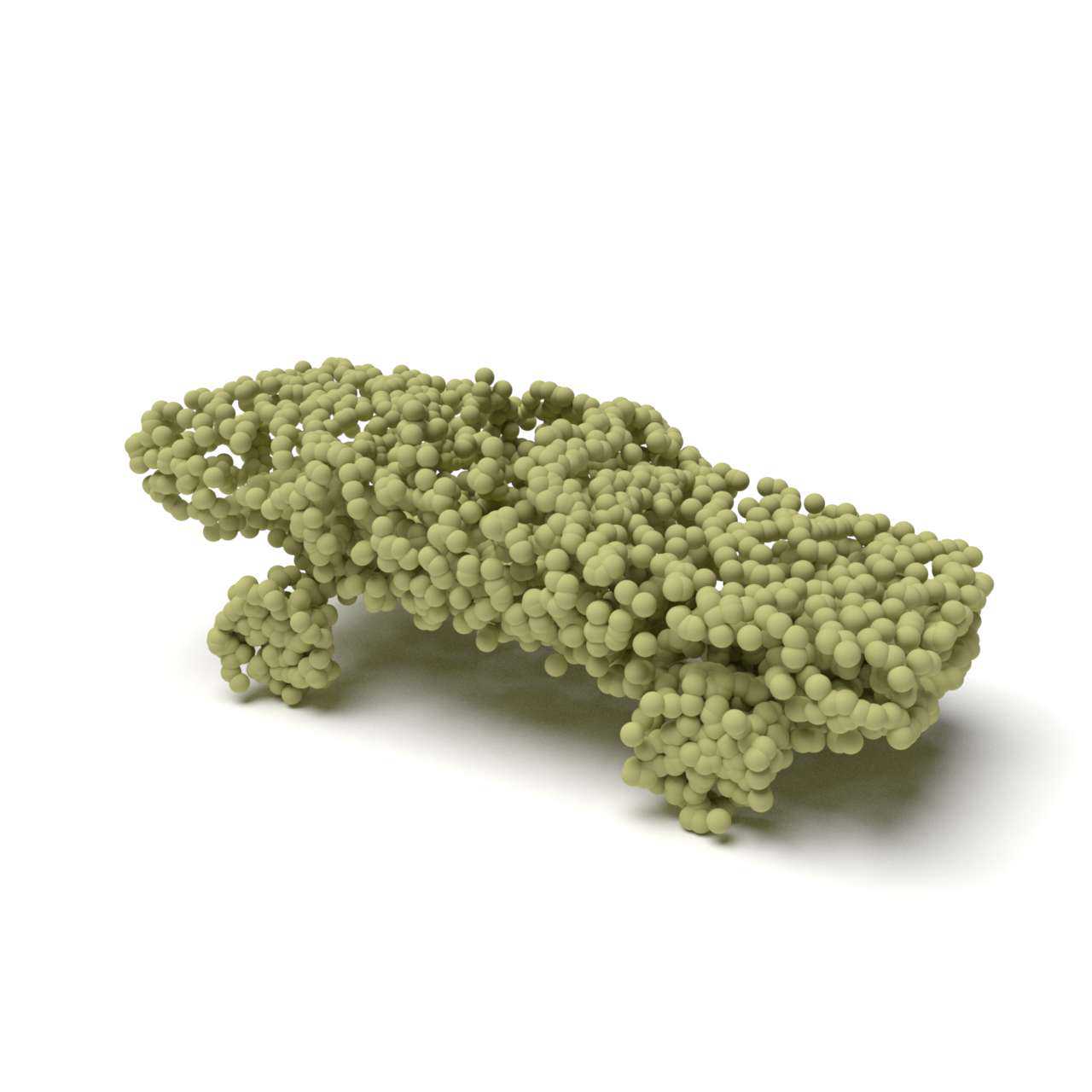} &
		\includegraphics[width=\sizea]{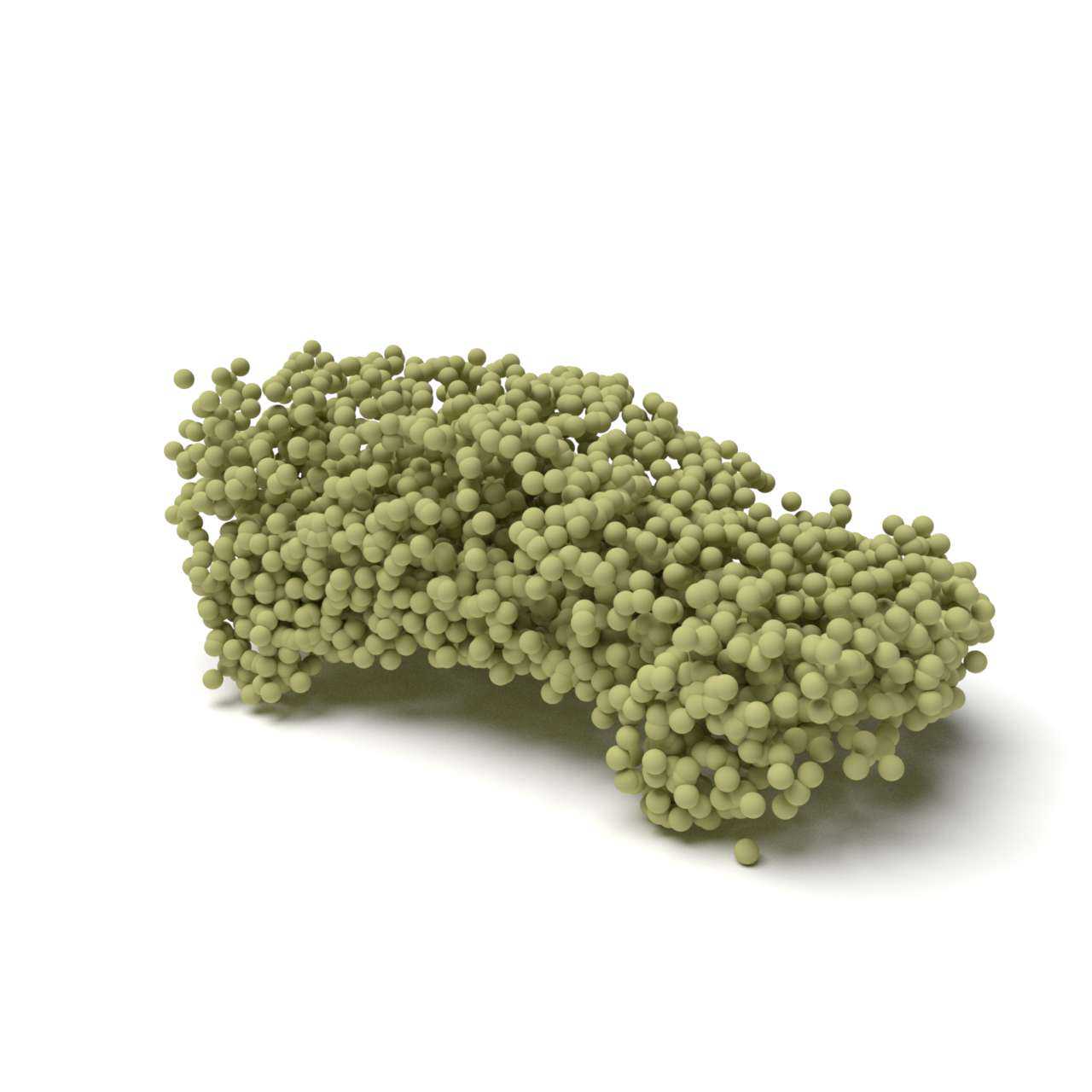}
	\end{tabular}	
	\caption{Difficult cases for our model. Rare shapes or shapes that contain many thin structures are usually hard to reconstruct in high quality.}
	\label{fig:failure-case}

\end{figure}

In this section, we discuss the limitation of our model and present visualizations of difficult cases where our model fails.
As mentioned in FFJORD~\cite{ffjord}, each integration requires evaluating the neural networks modeling the dynamics multiple times.
The number of function evaluations tends to increase as the training proceeds since the dynamic becomes more complex and more function evaluations are needed to achieve the same numerical precision.
This issue limits our model size and makes the convergence slow.
Grathwohl \etal indicate that using regularization such as weight decay could alleviate such an issue, but we empirically find that using regularization tends to hurt performance.
Future advances in invertible models like CNF might help improve this issue.
Typical failure case appears when reconstructing or generating the rare shape or shapes with many thin structures as presented in Figure~\ref{fig:failure-case}.

\section{Latent space visualizations}\label{sec:tsne}
\begin{figure}[t]
	\centering	
	\includegraphics[width=0.49\linewidth]{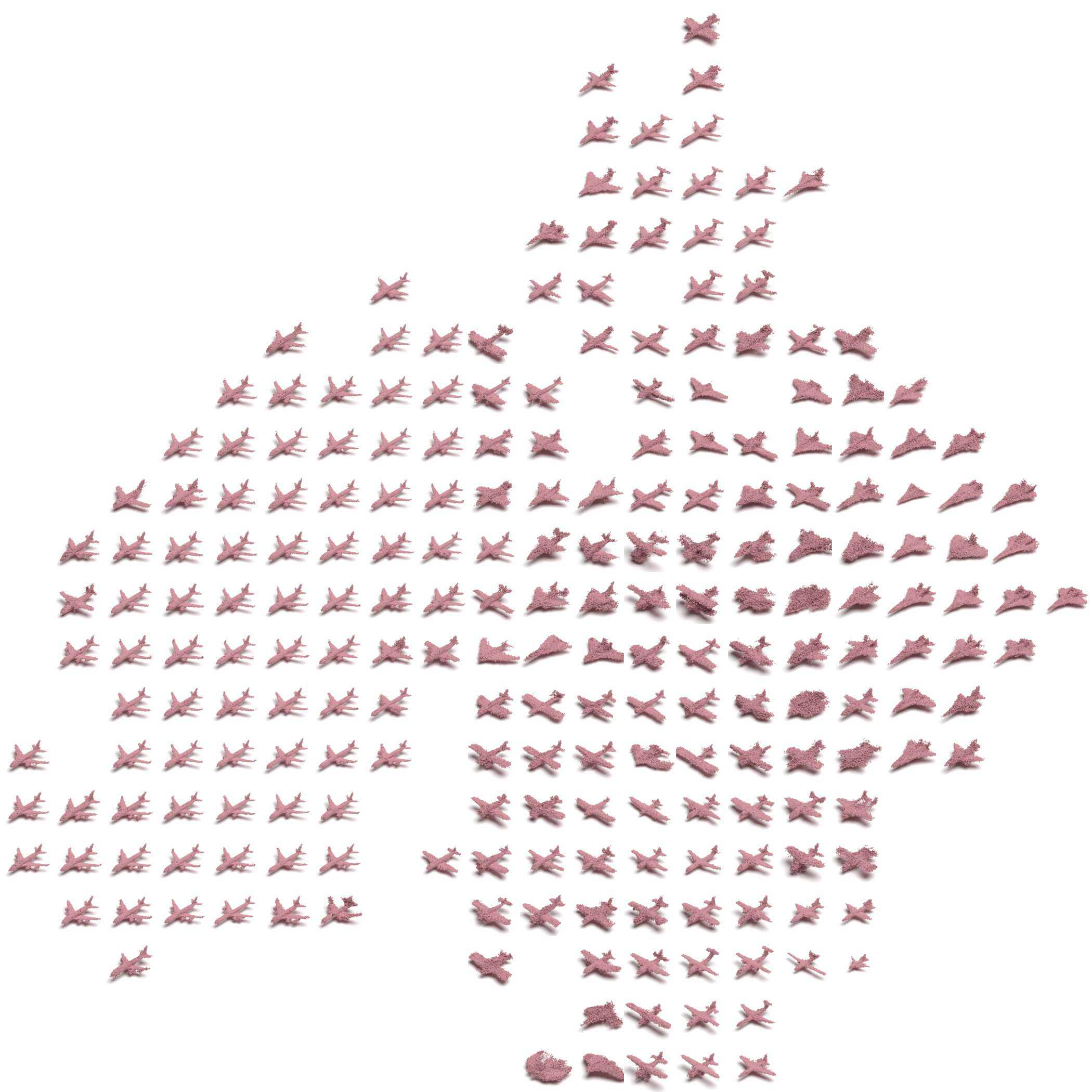}
	\includegraphics[width=0.49\linewidth]{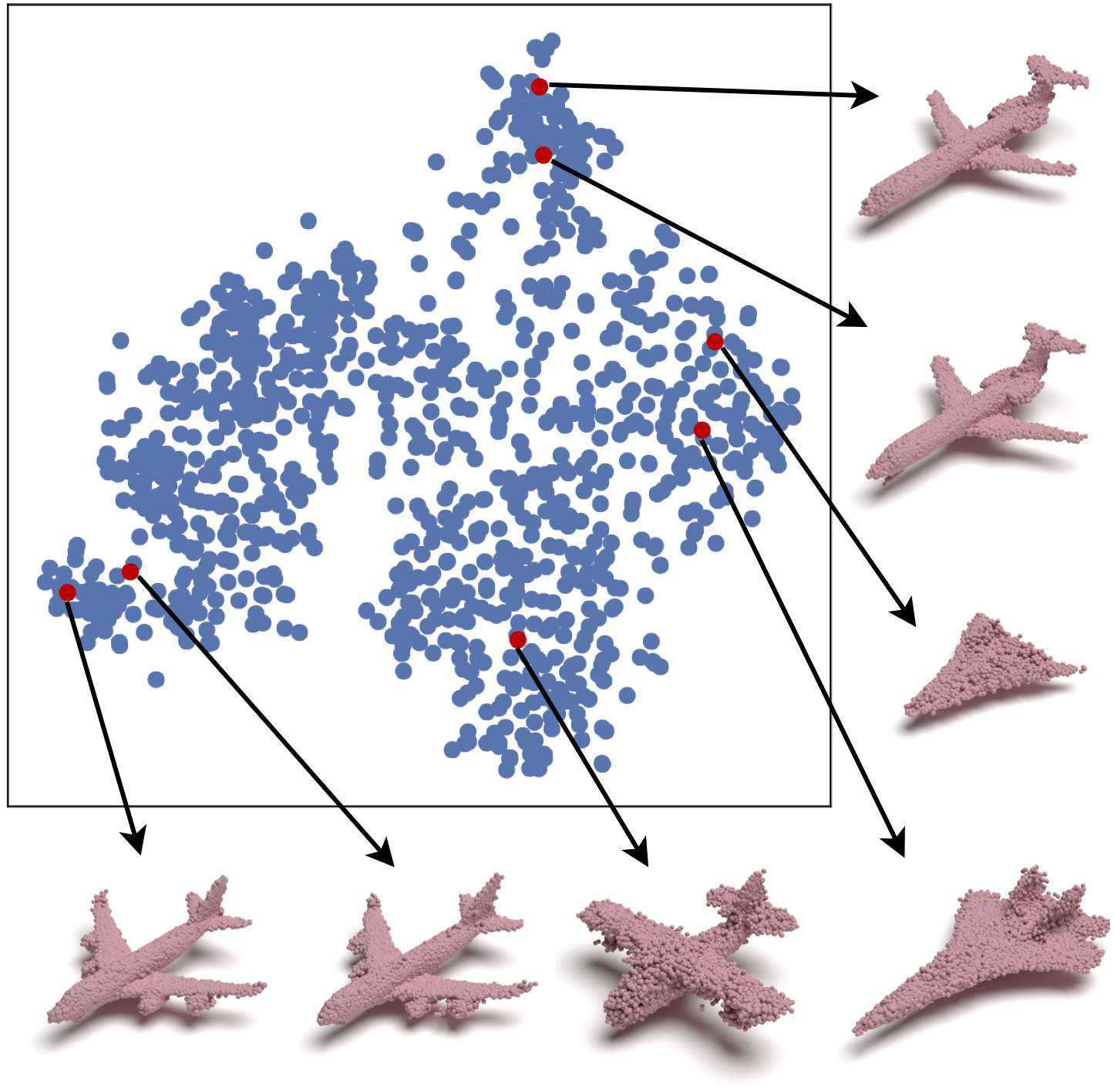}
    \caption{Visualization of latent space.}
	\label{fig:emb-vis}
\vspace{-0.1in}
\end{figure}

We provide visualization of the sampled latent vectors $z\in\mathbb{R}^{128}$ in Figure~\ref{fig:emb-vis}.
We sample $1000$ latent vectors and run t-SNE~\cite{maaten2008visualizing} to visualize these latent vectors in 2D. Shapes with similar styles are close in the latent space.

\section{Interpolation}\label{sec:interpolation}

\begin{figure*}
	\centering
	\newcommand{\sizea}{0.138\linewidth}
	\includegraphics[width=\sizea]{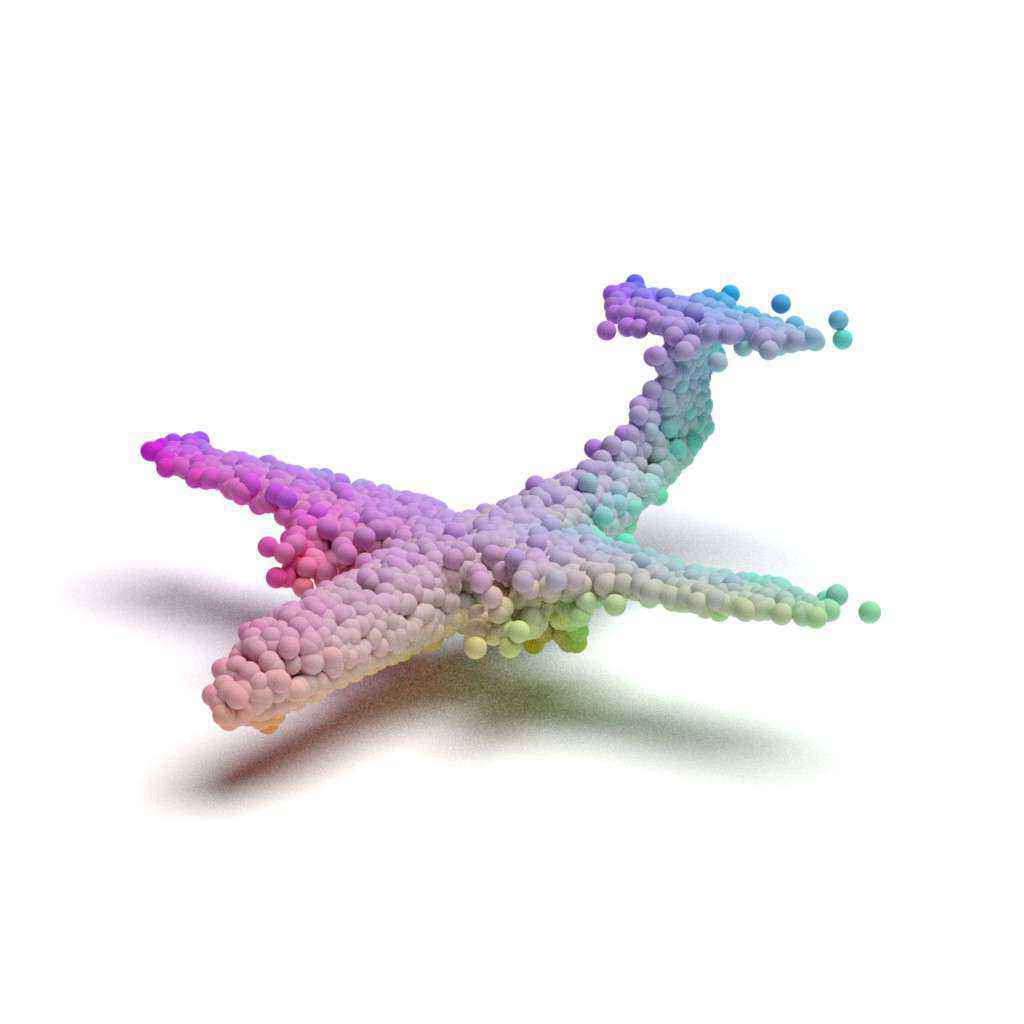}
	\includegraphics[width=\sizea]{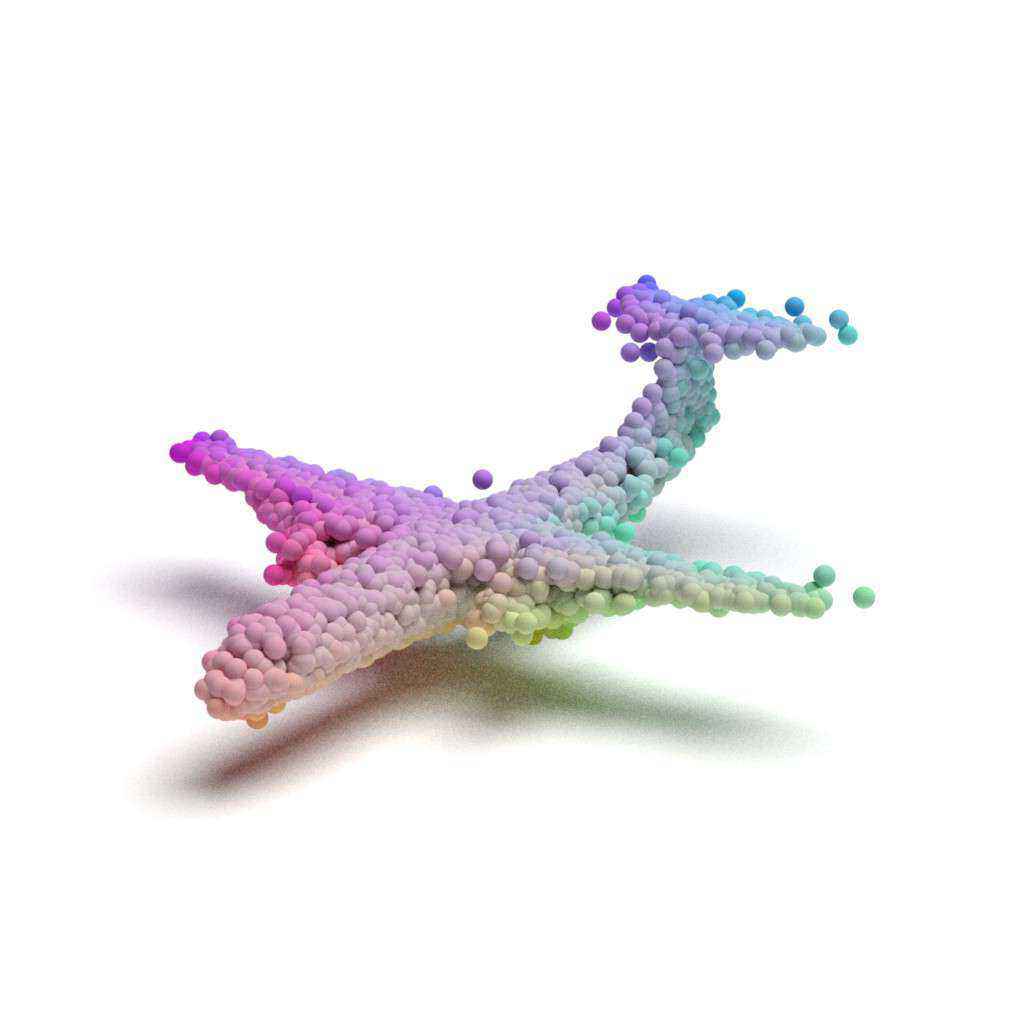}
	\includegraphics[width=\sizea]{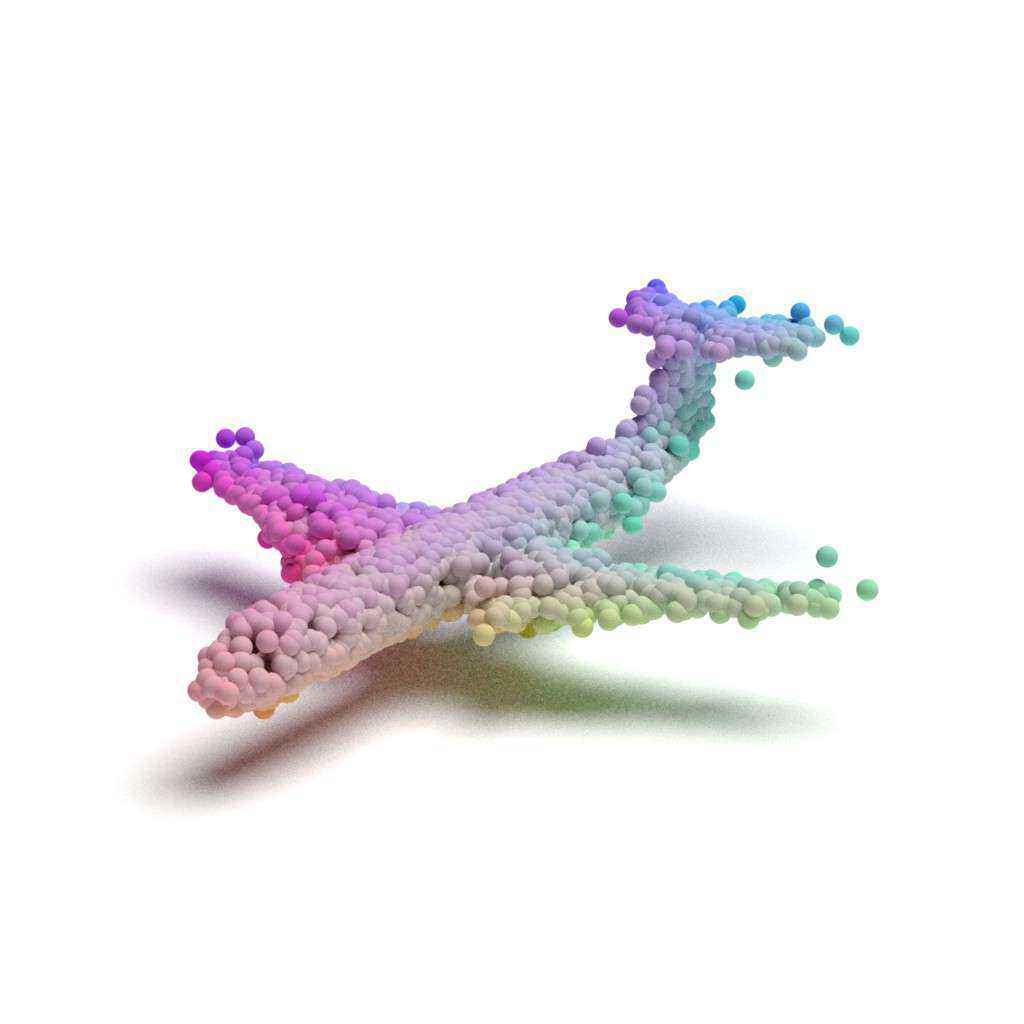}
	\includegraphics[width=\sizea]{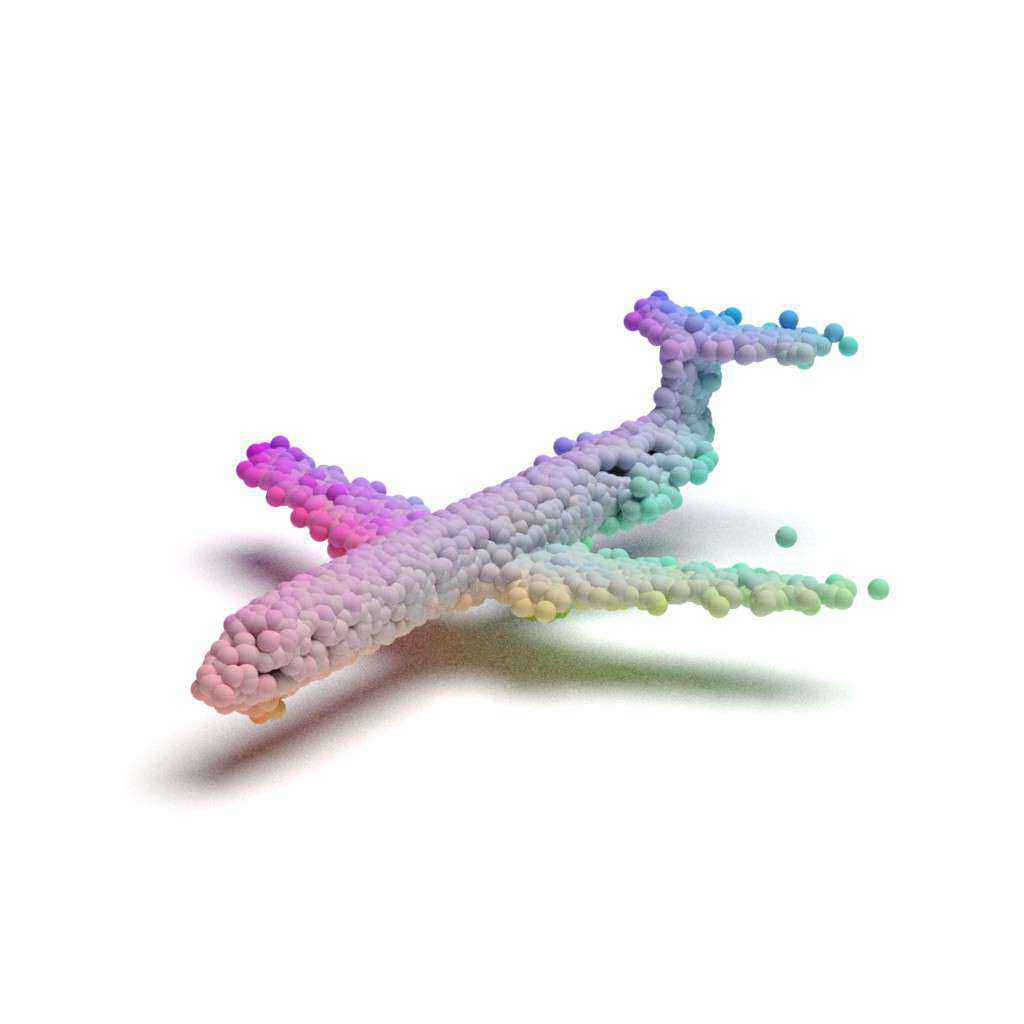}
	\includegraphics[width=\sizea]{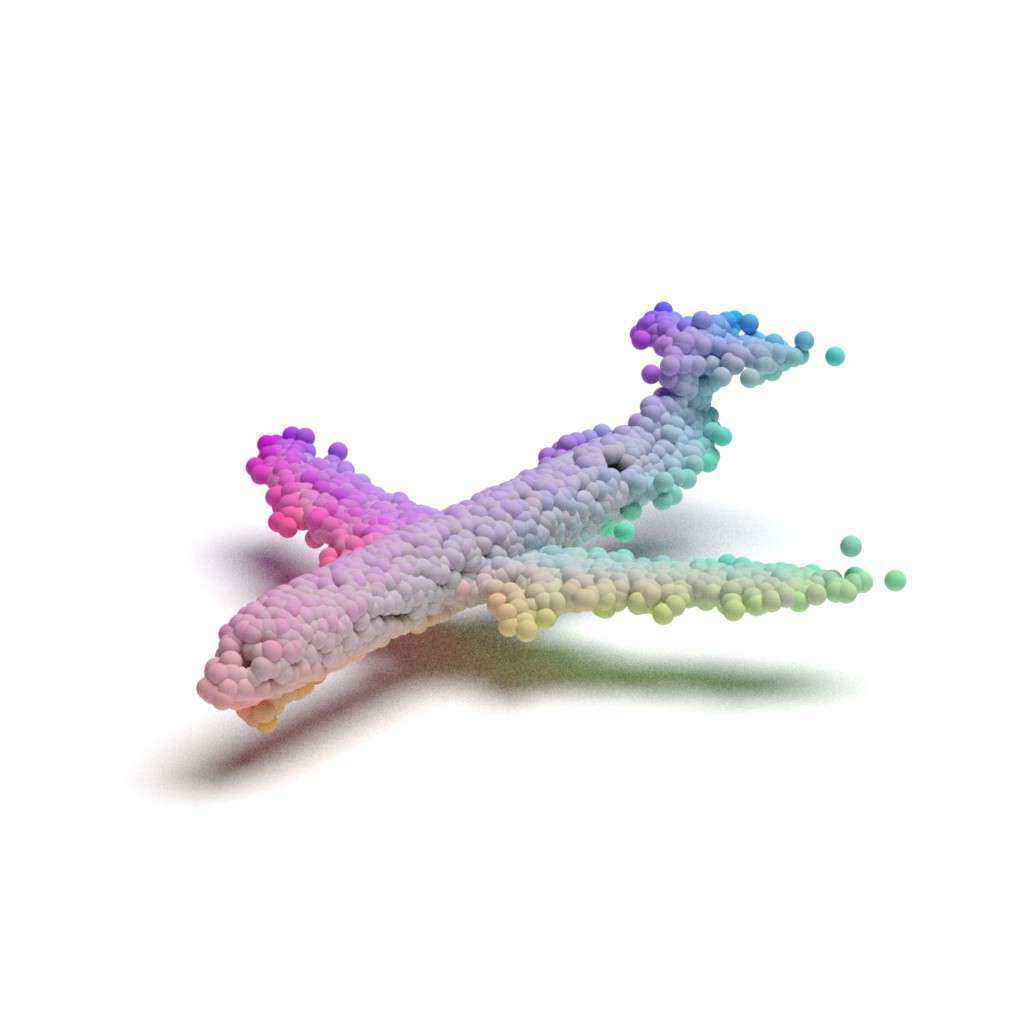}
	\includegraphics[width=\sizea]{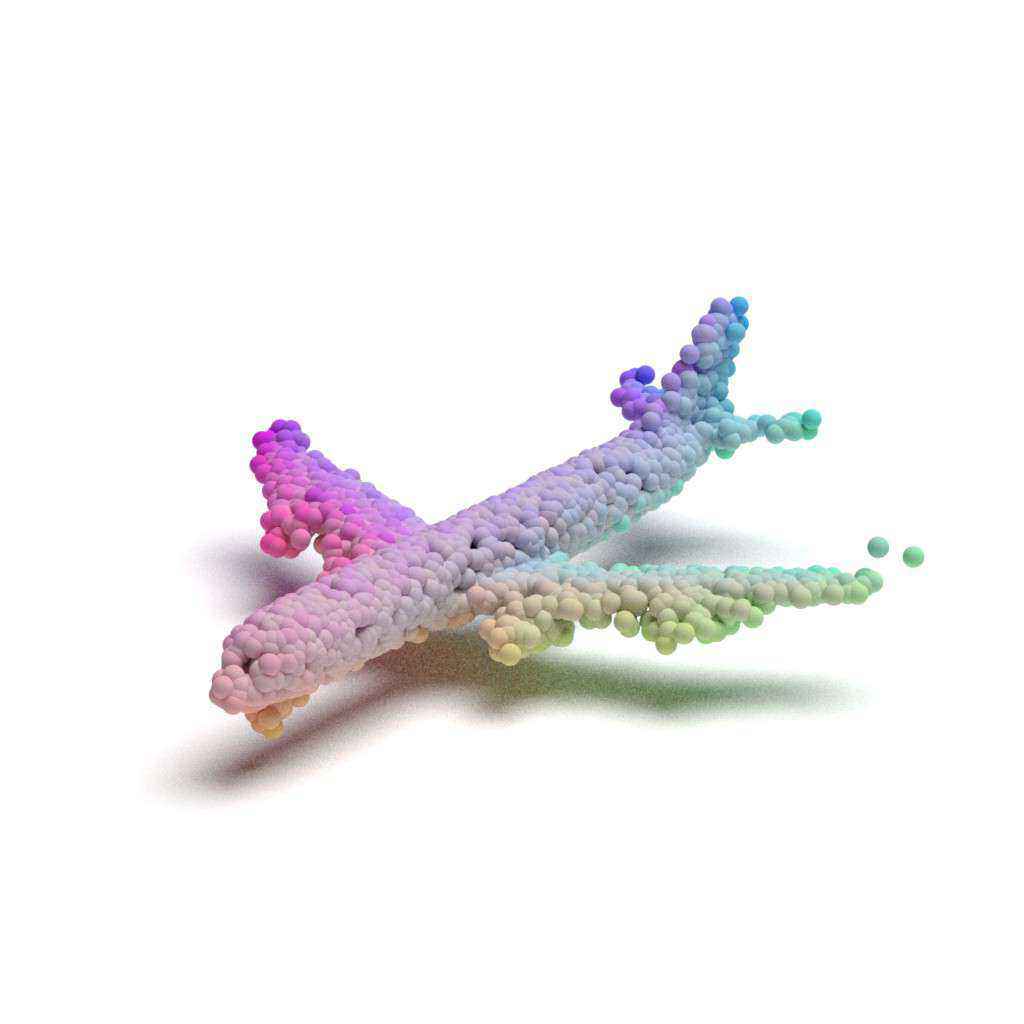}
	\includegraphics[width=\sizea]{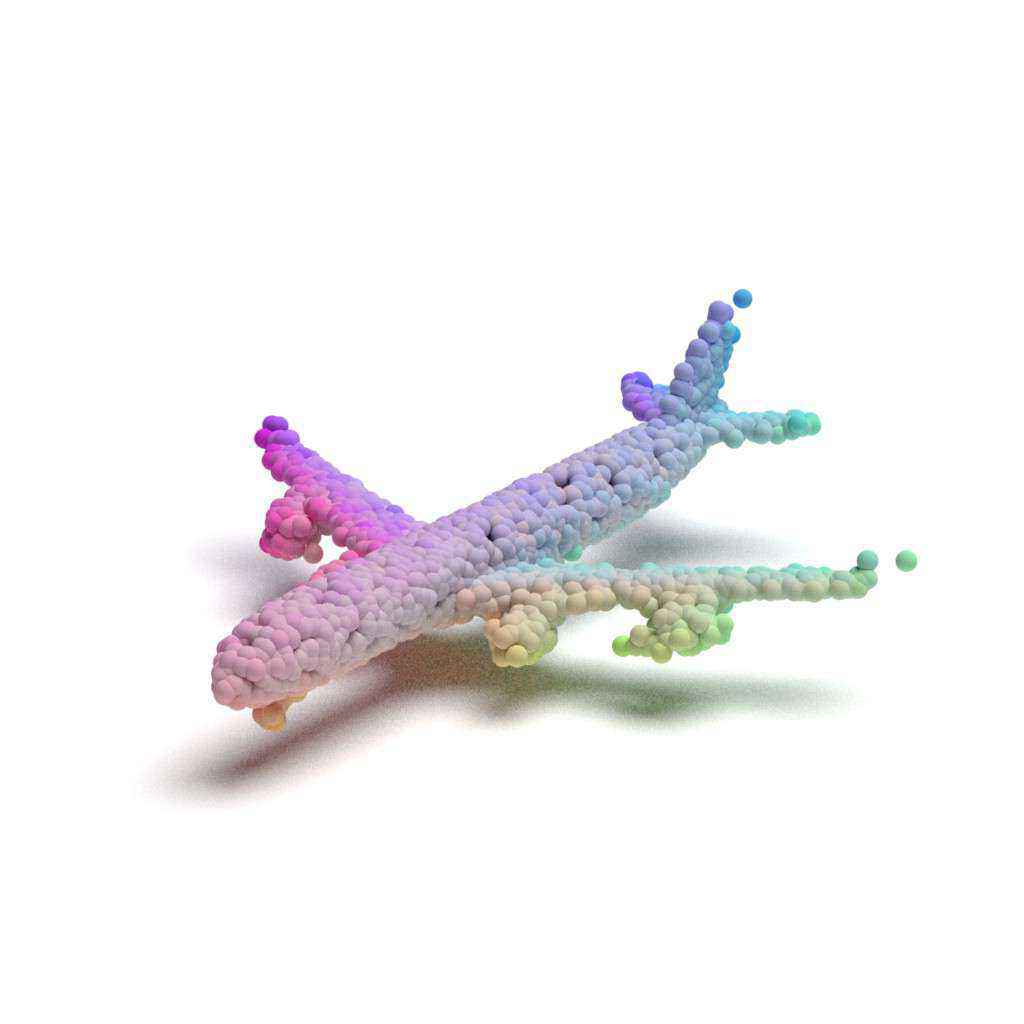}\\
	\includegraphics[width=\sizea]{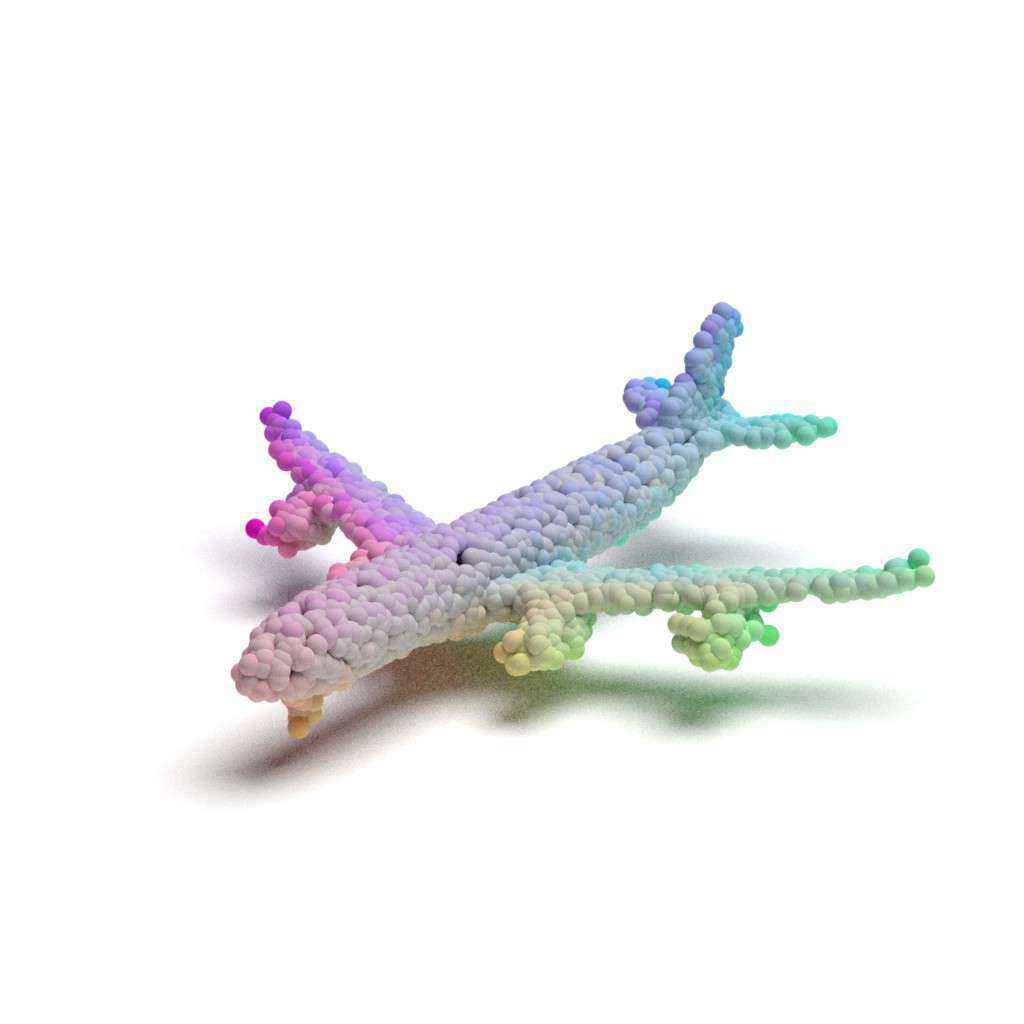}
	\includegraphics[width=\sizea]{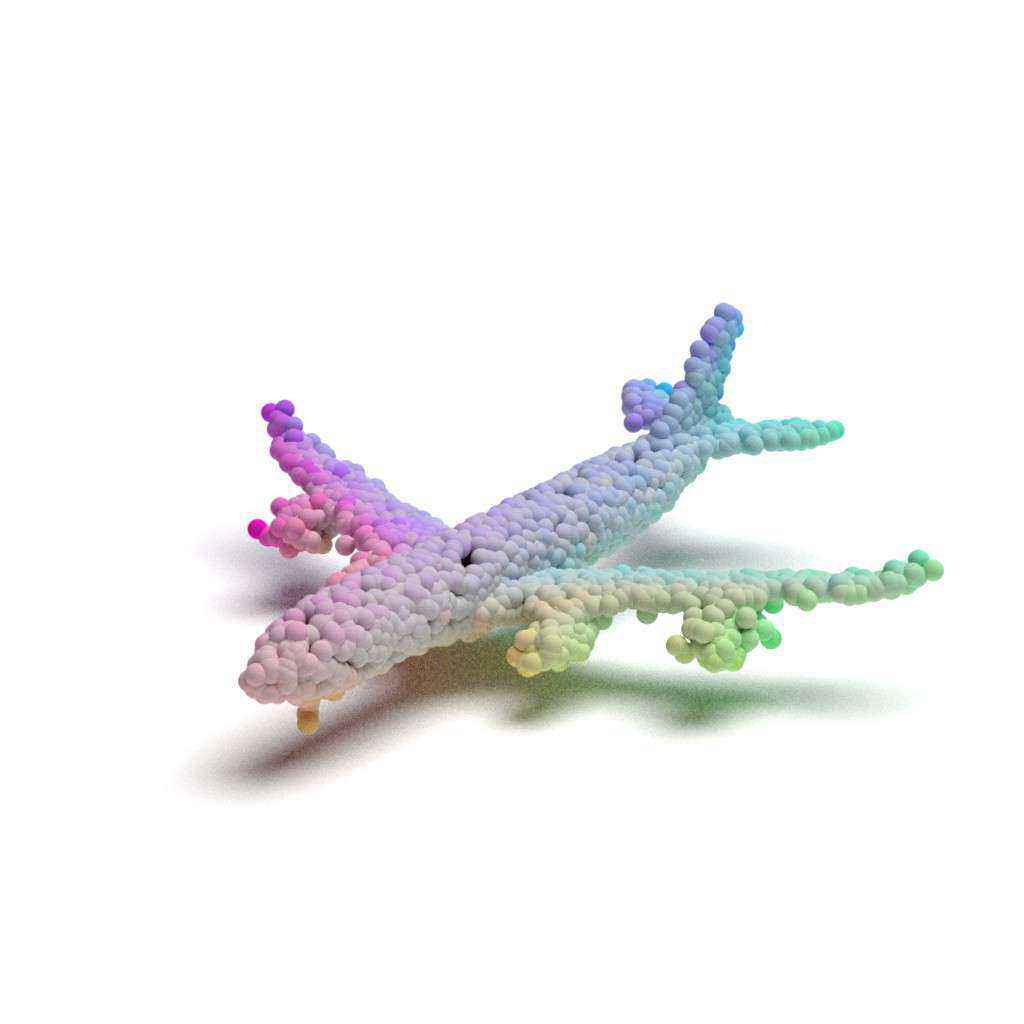}
	\includegraphics[width=\sizea]{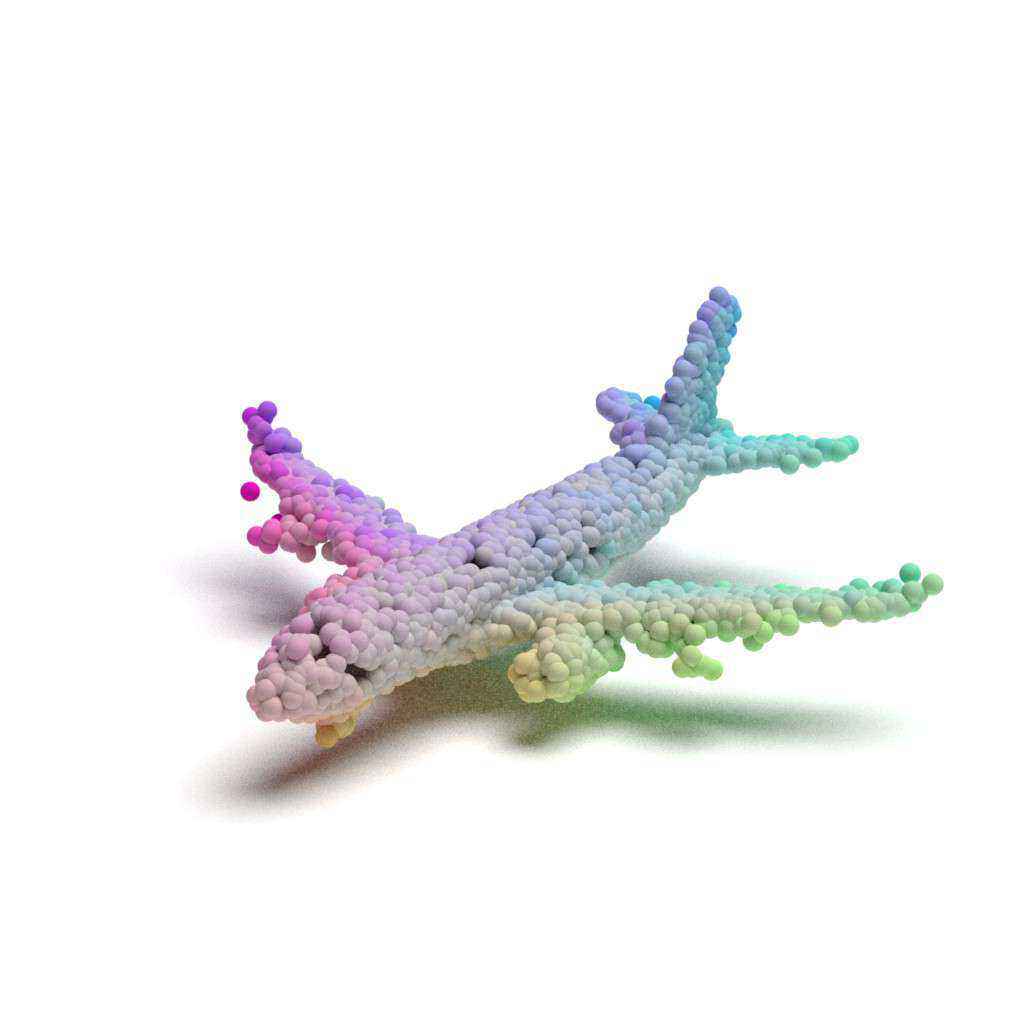}
	\includegraphics[width=\sizea]{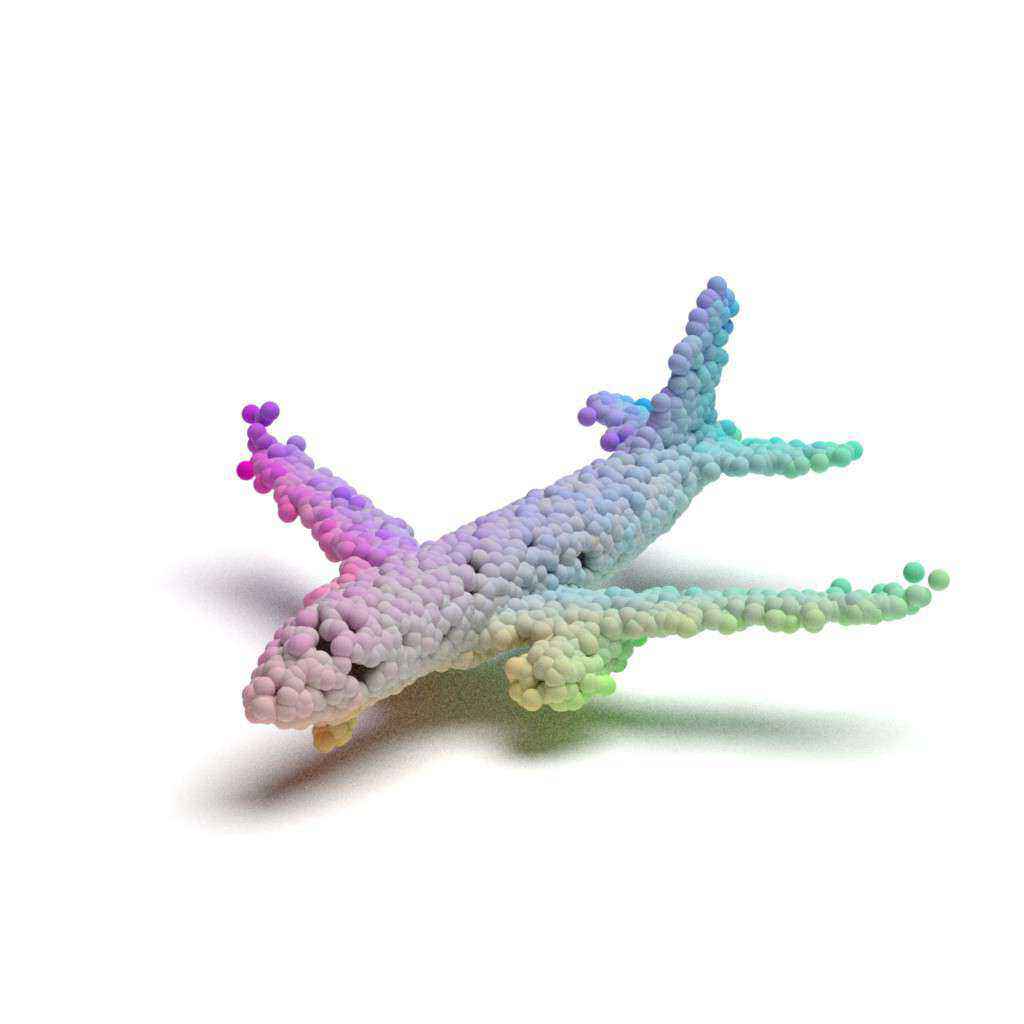}
	\includegraphics[width=\sizea]{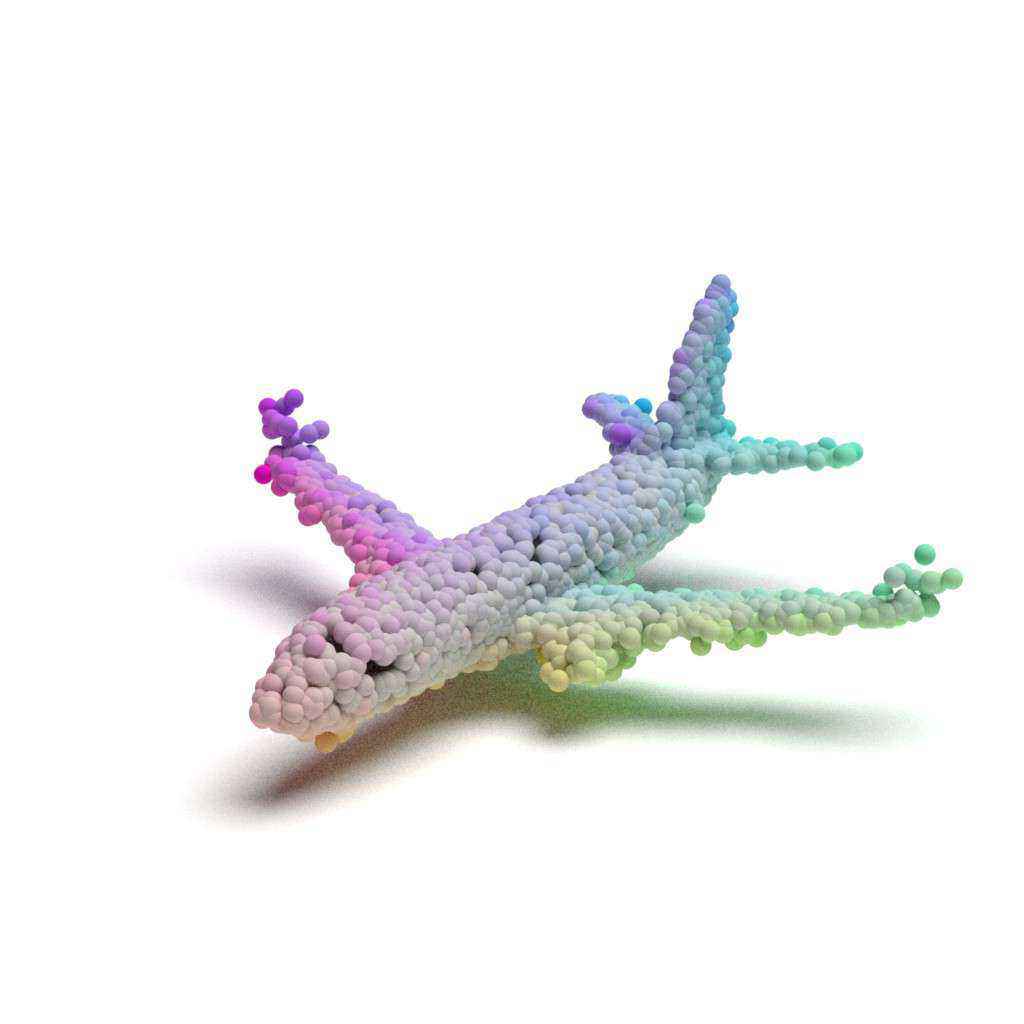}
	\includegraphics[width=\sizea]{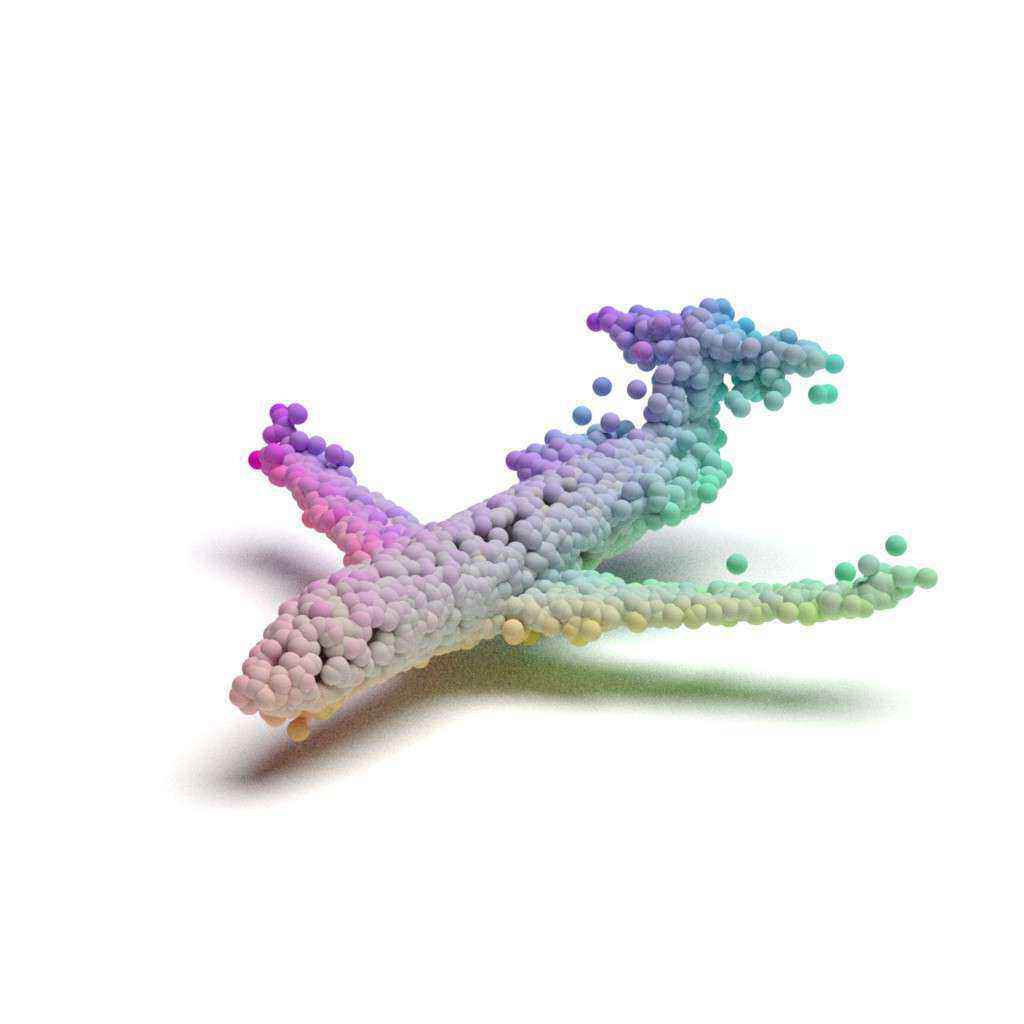}
	\includegraphics[width=\sizea]{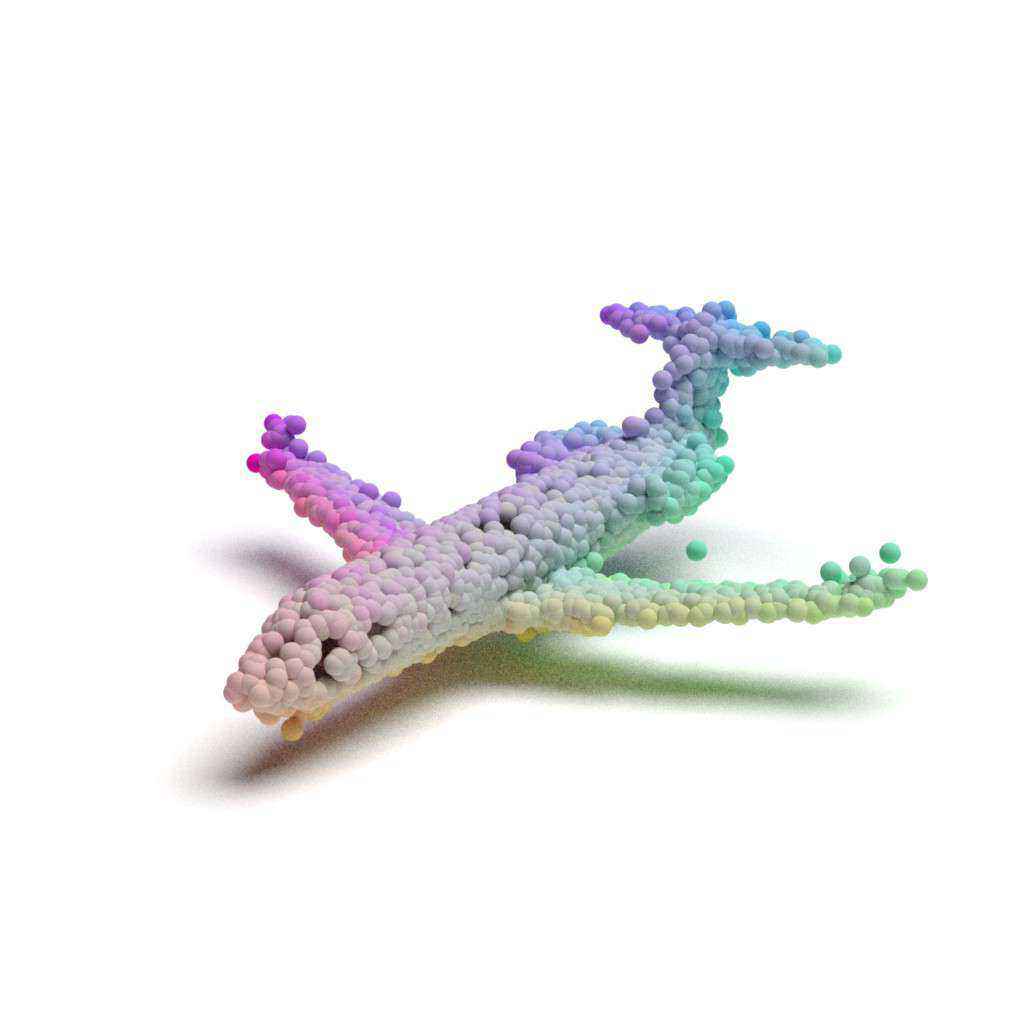}\\
	\includegraphics[width=\sizea]{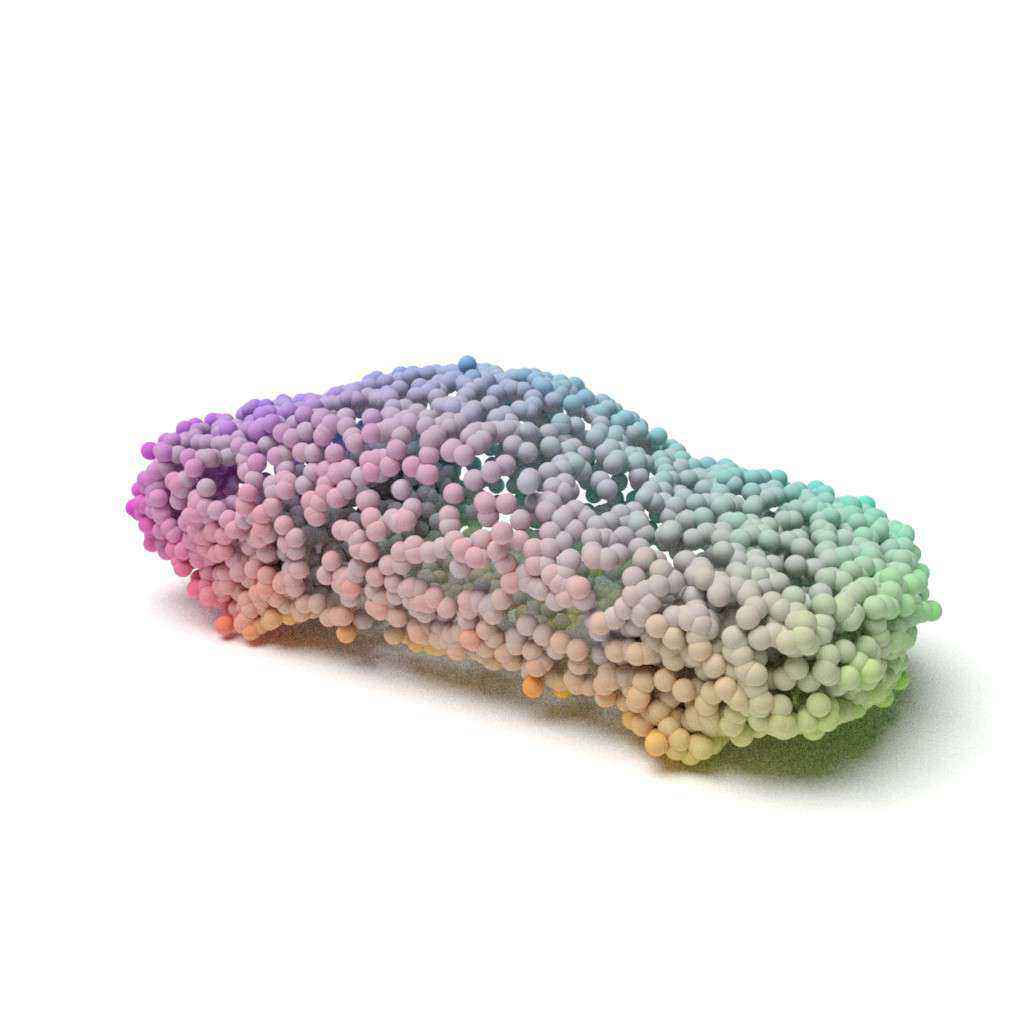}
	\includegraphics[width=\sizea]{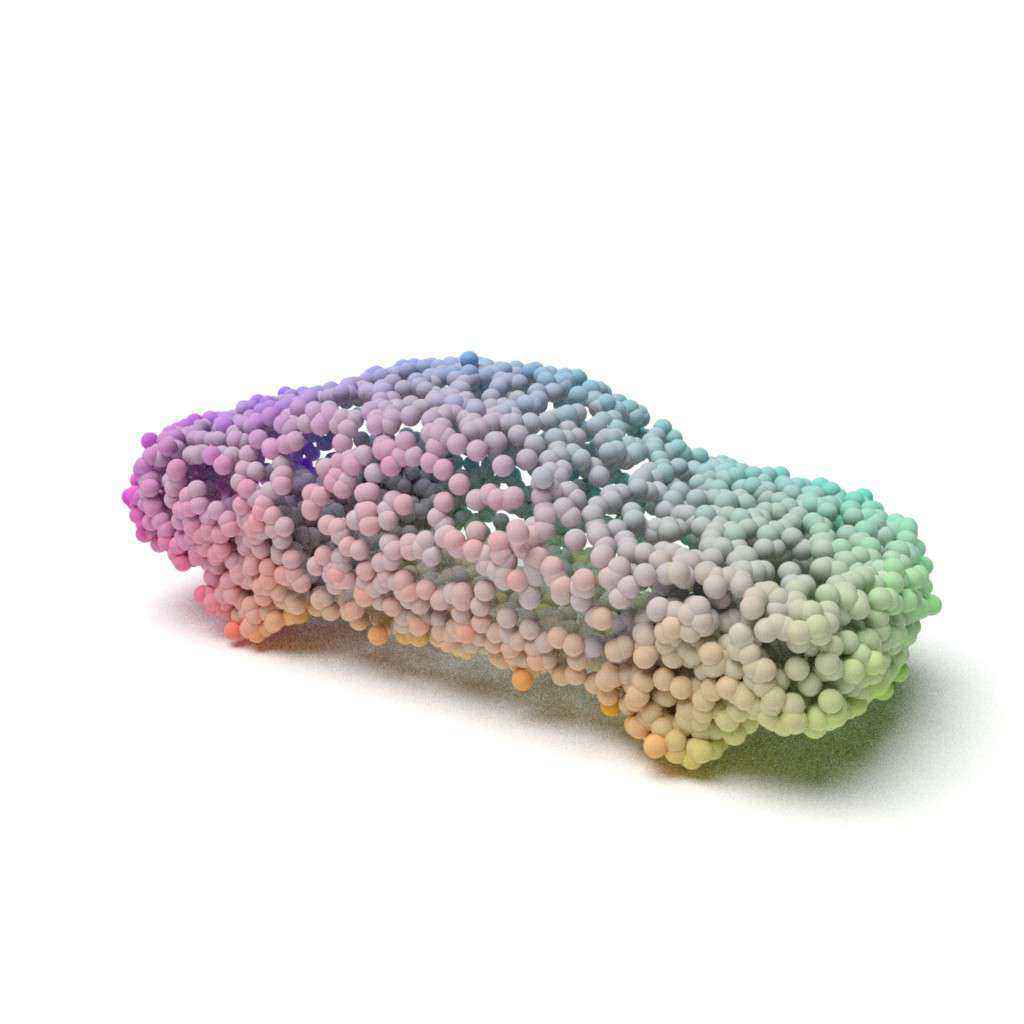}
	\includegraphics[width=\sizea]{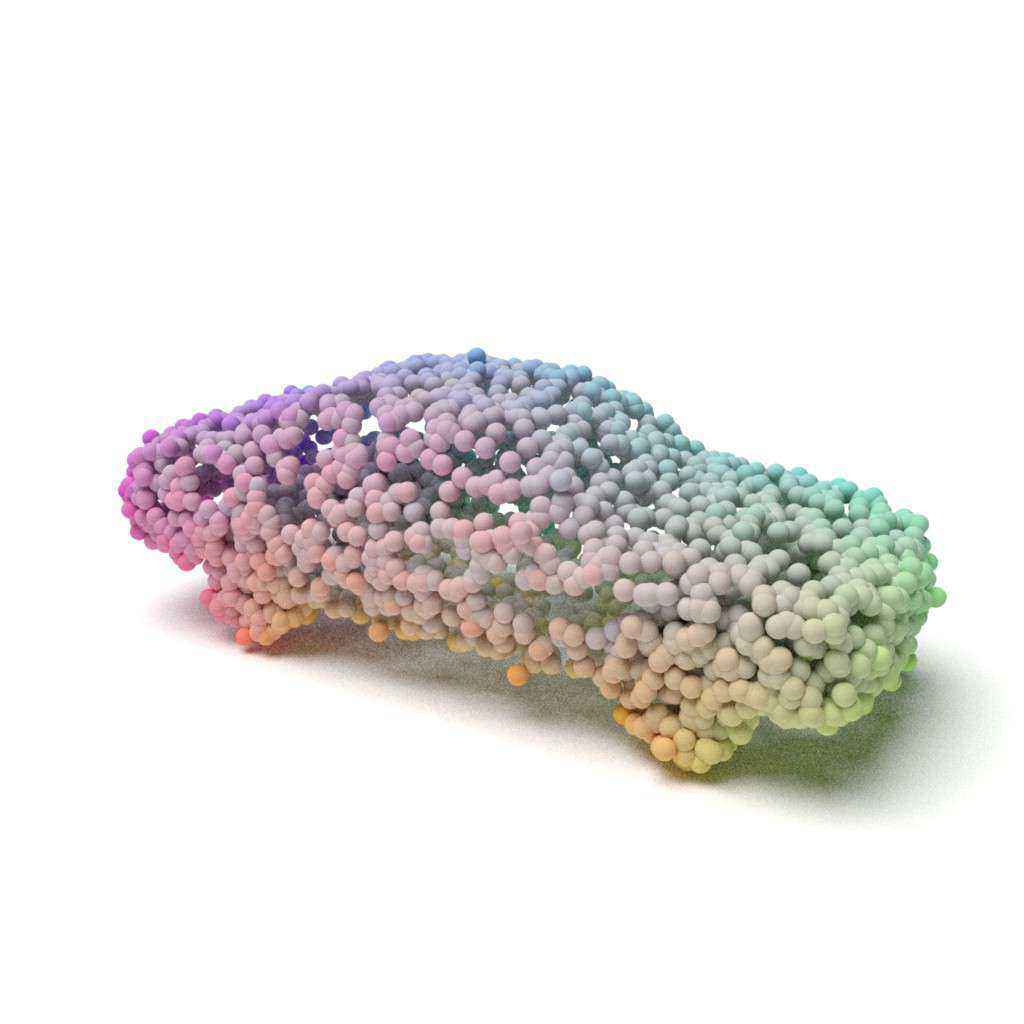}
	\includegraphics[width=\sizea]{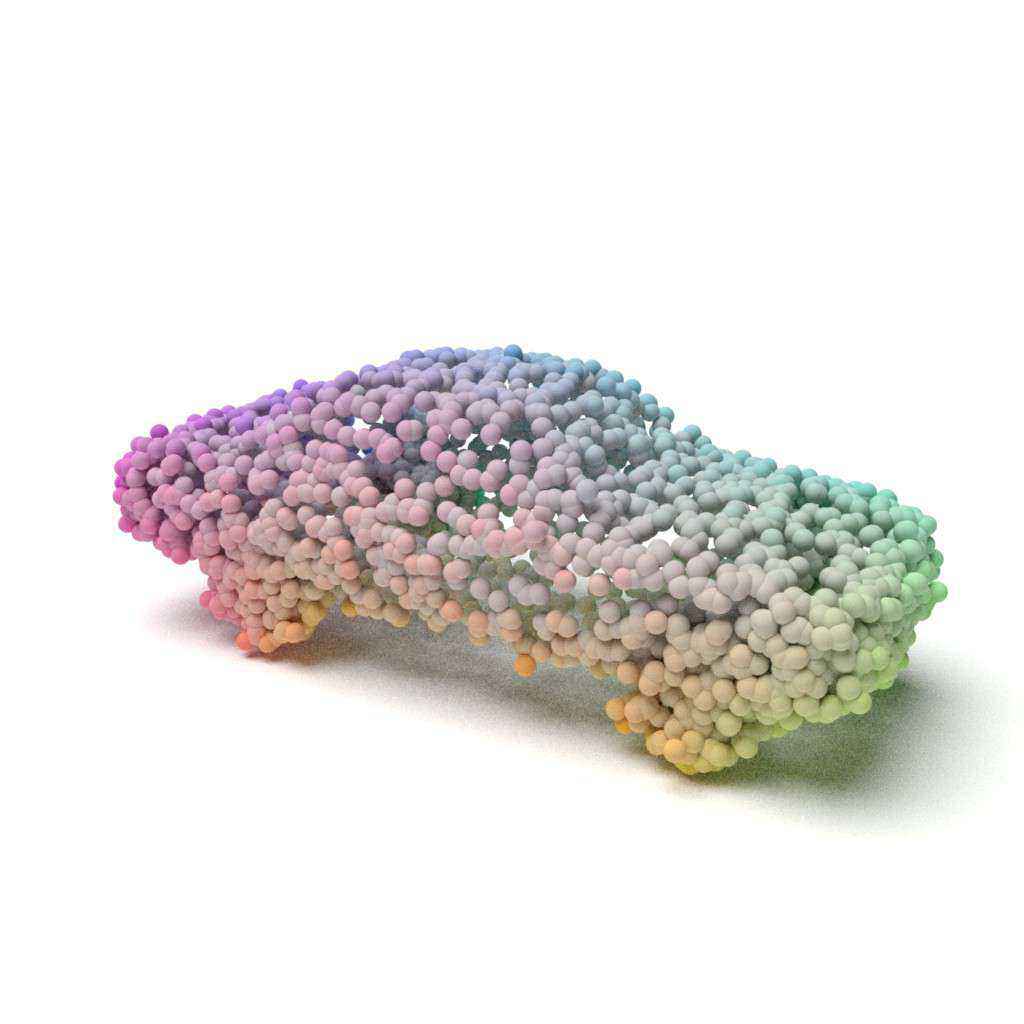}
	\includegraphics[width=\sizea]{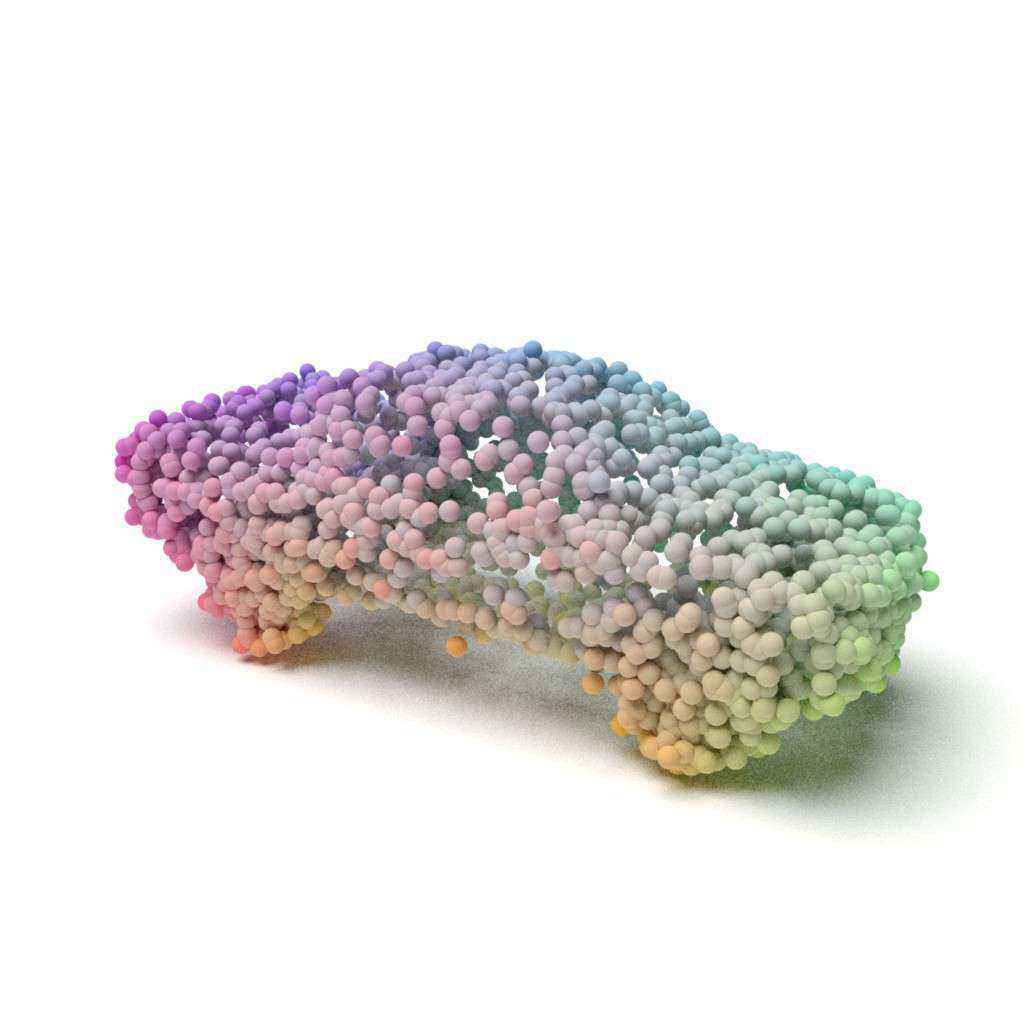}
	\includegraphics[width=\sizea]{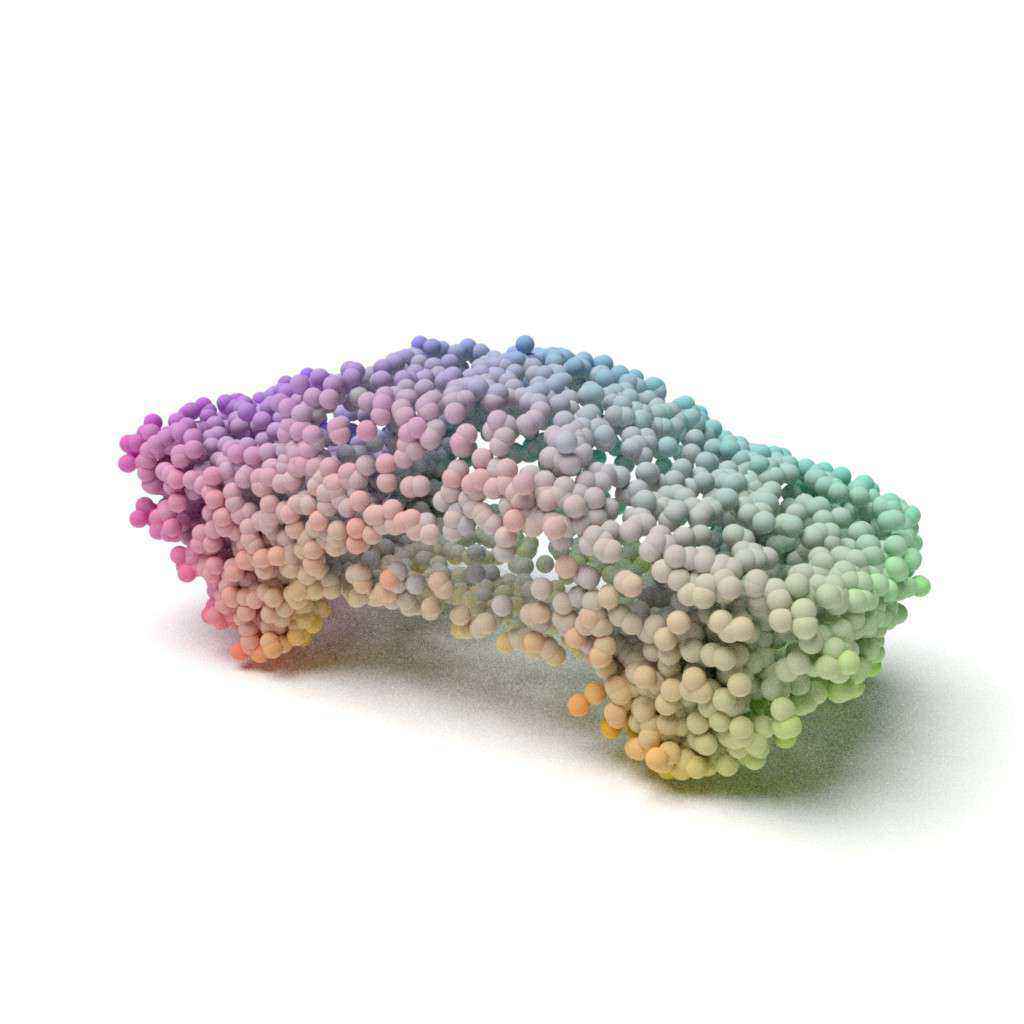}
	\includegraphics[width=\sizea]{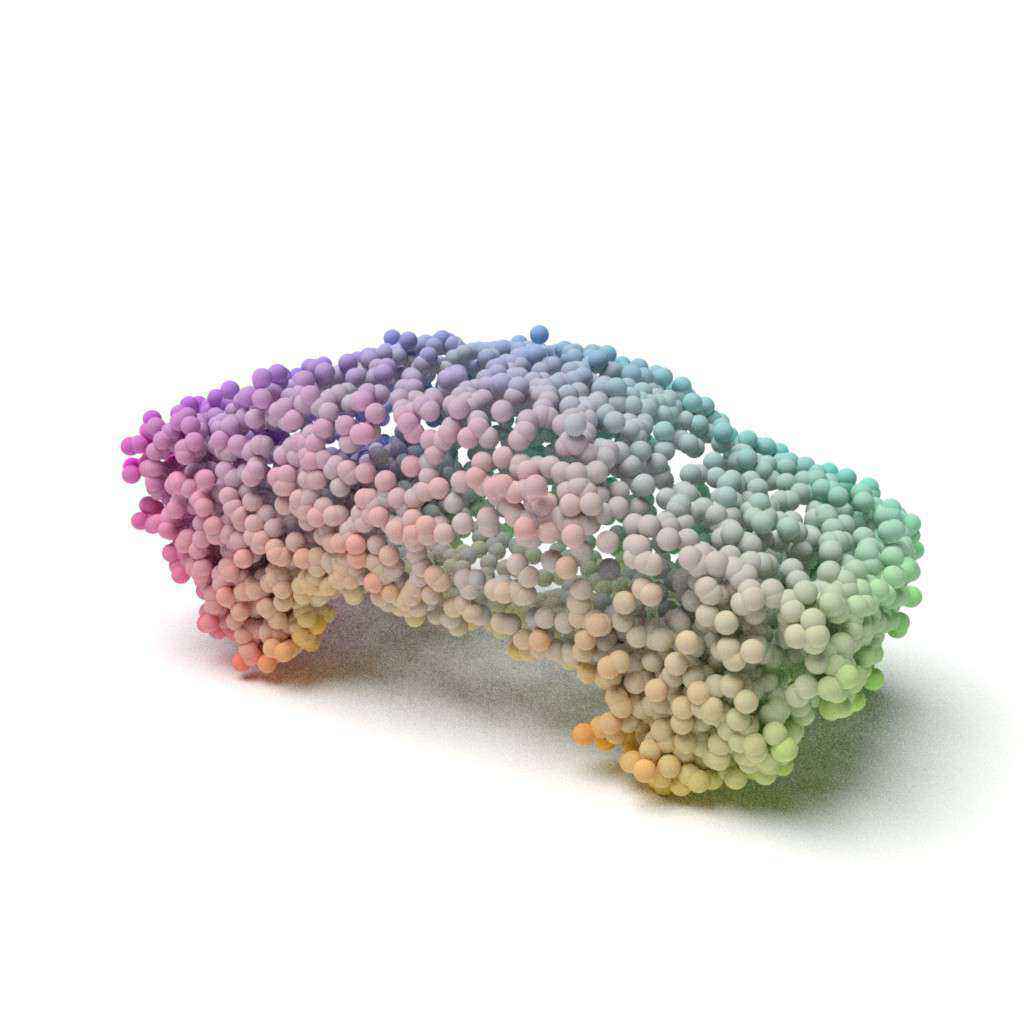}\\
	\includegraphics[width=\sizea]{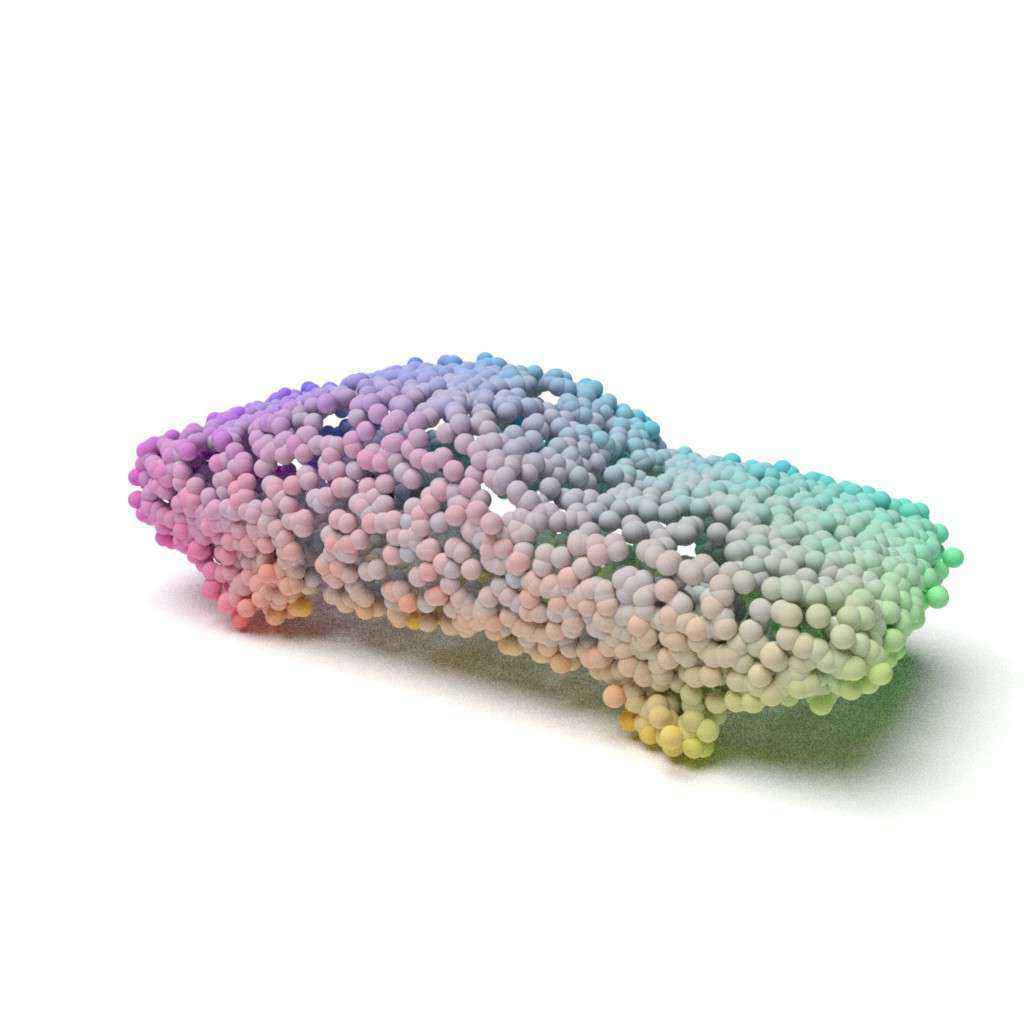}
	\includegraphics[width=\sizea]{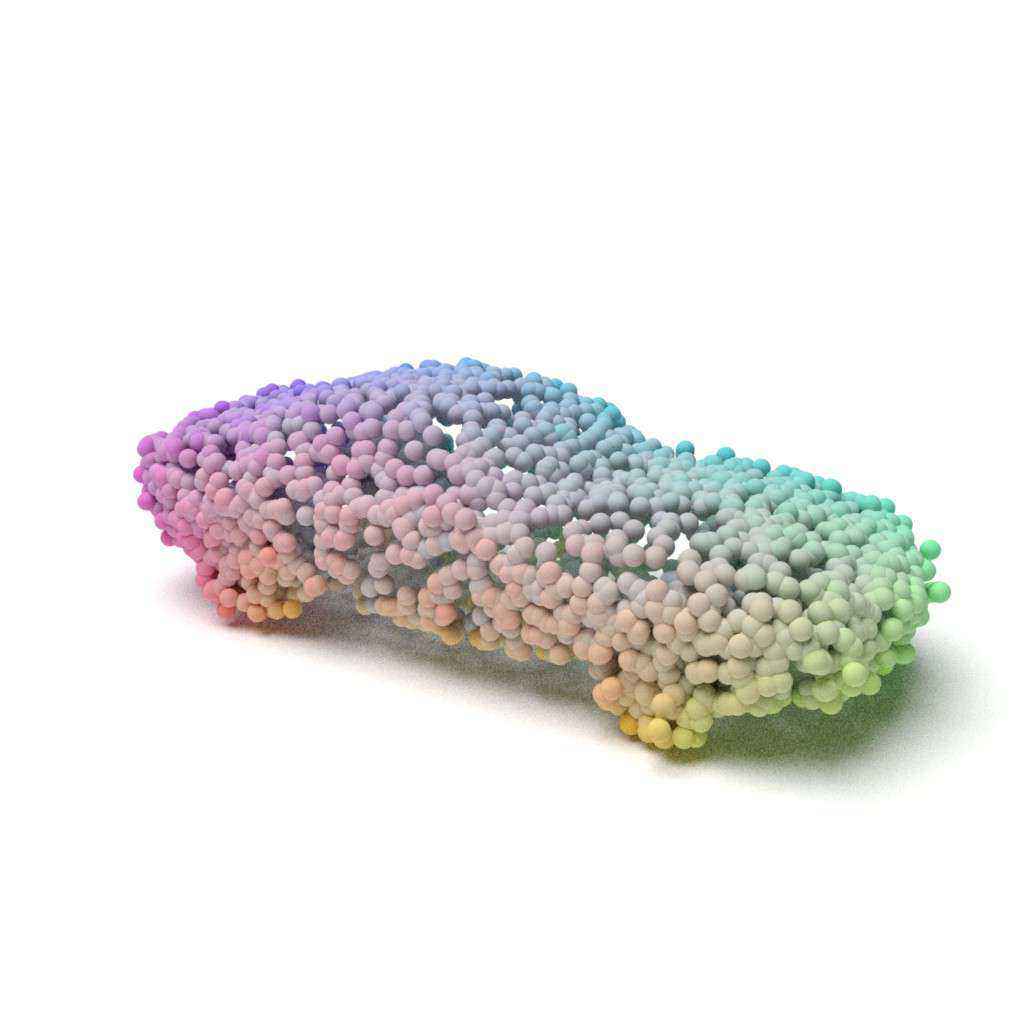}
	\includegraphics[width=\sizea]{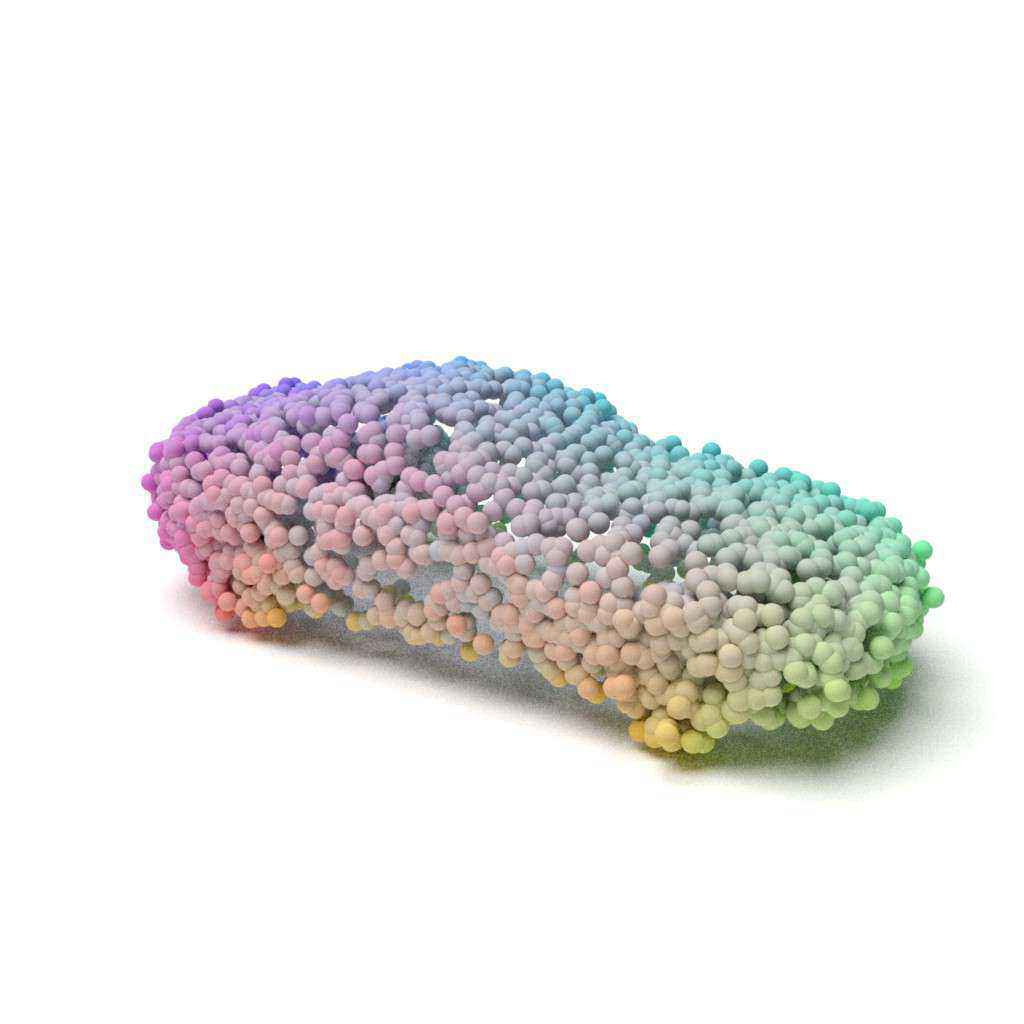}
	\includegraphics[width=\sizea]{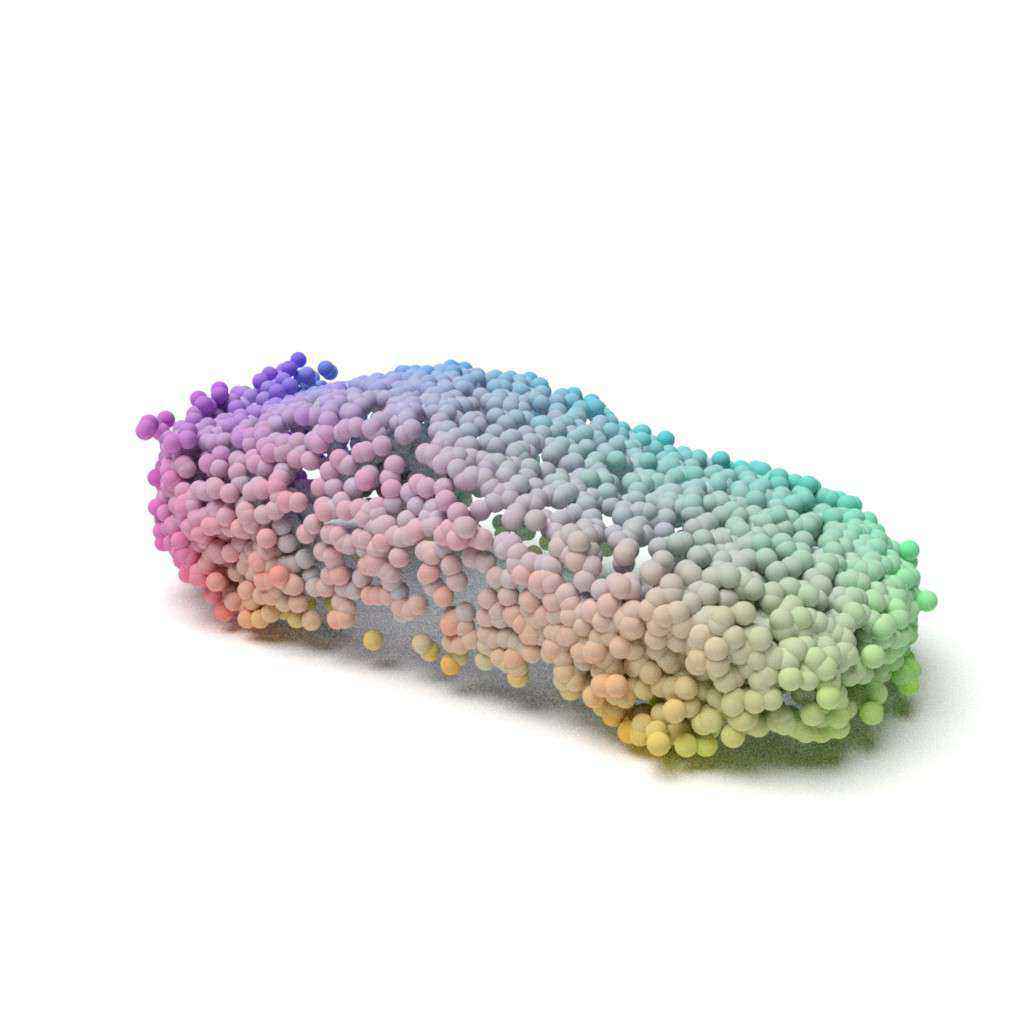}
	\includegraphics[width=\sizea]{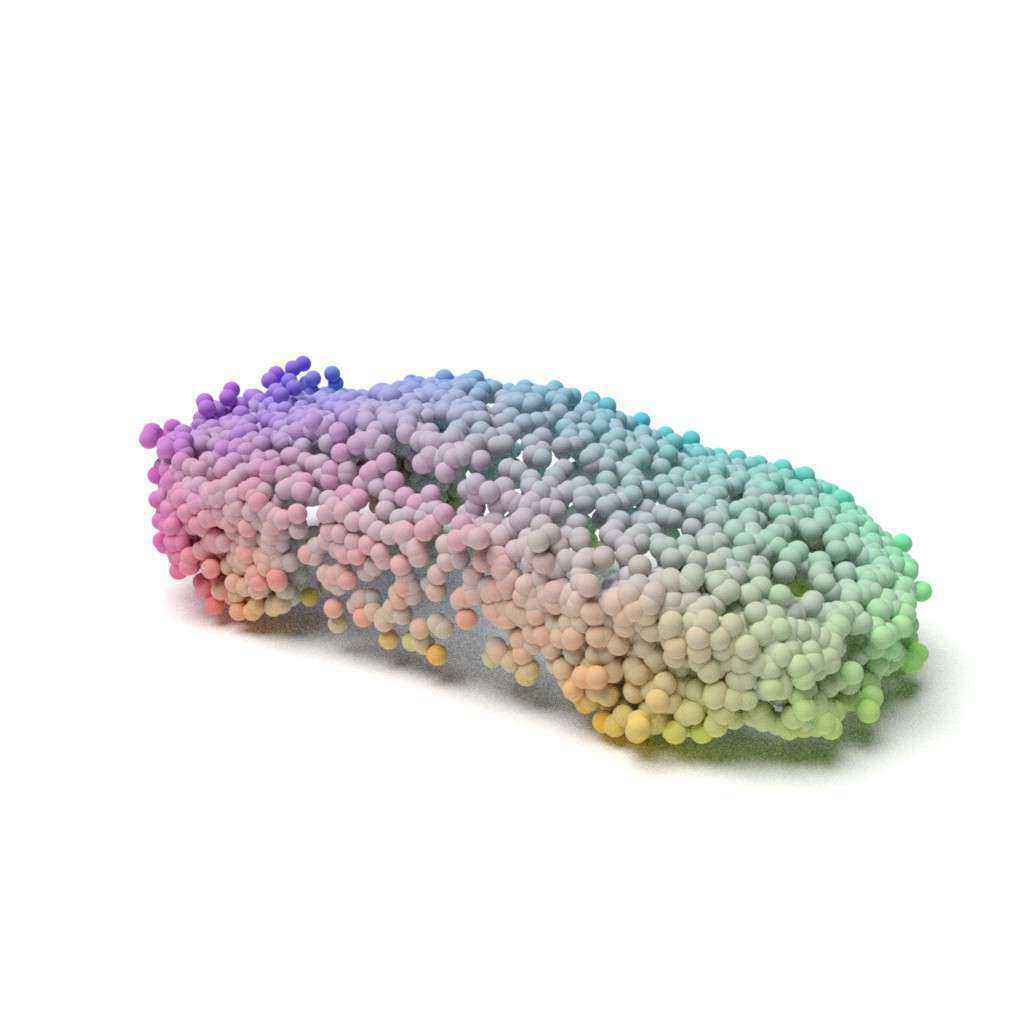}
	\includegraphics[width=\sizea]{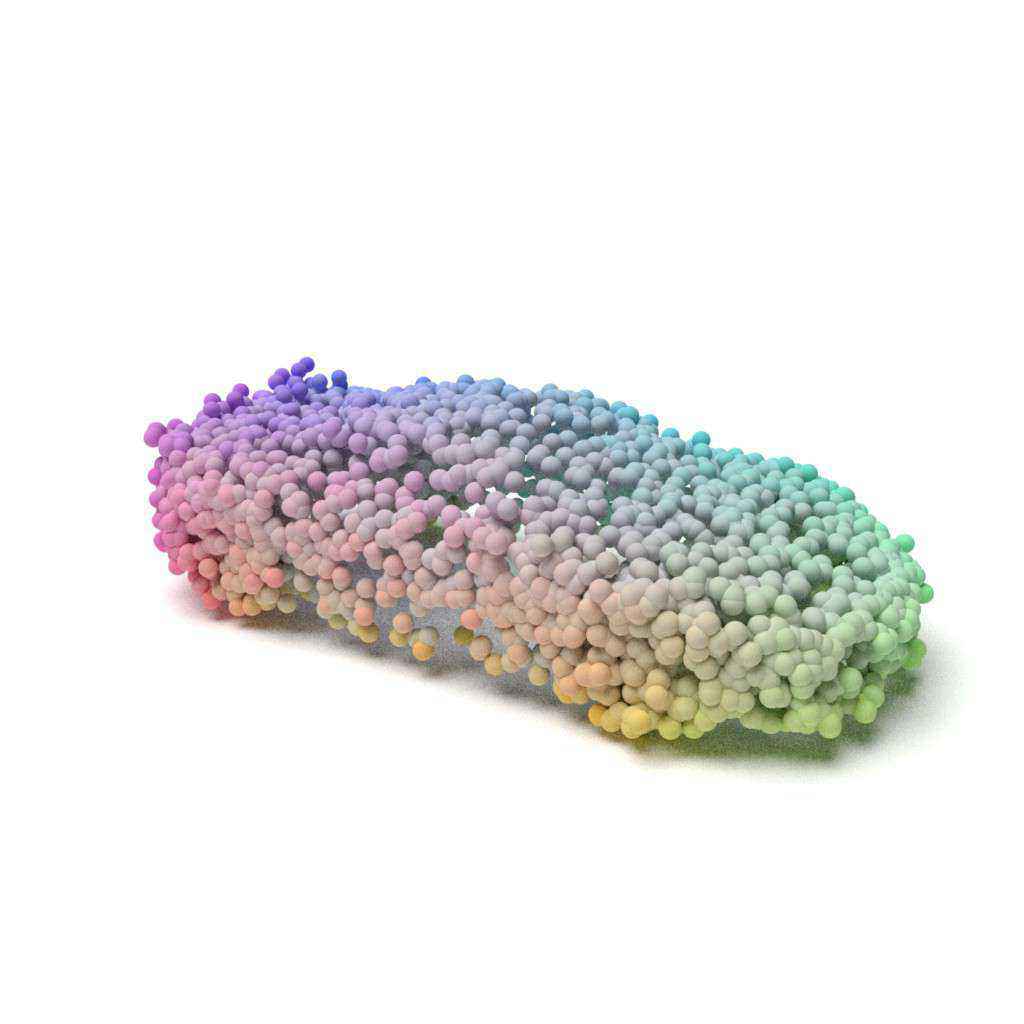}
	\includegraphics[width=\sizea]{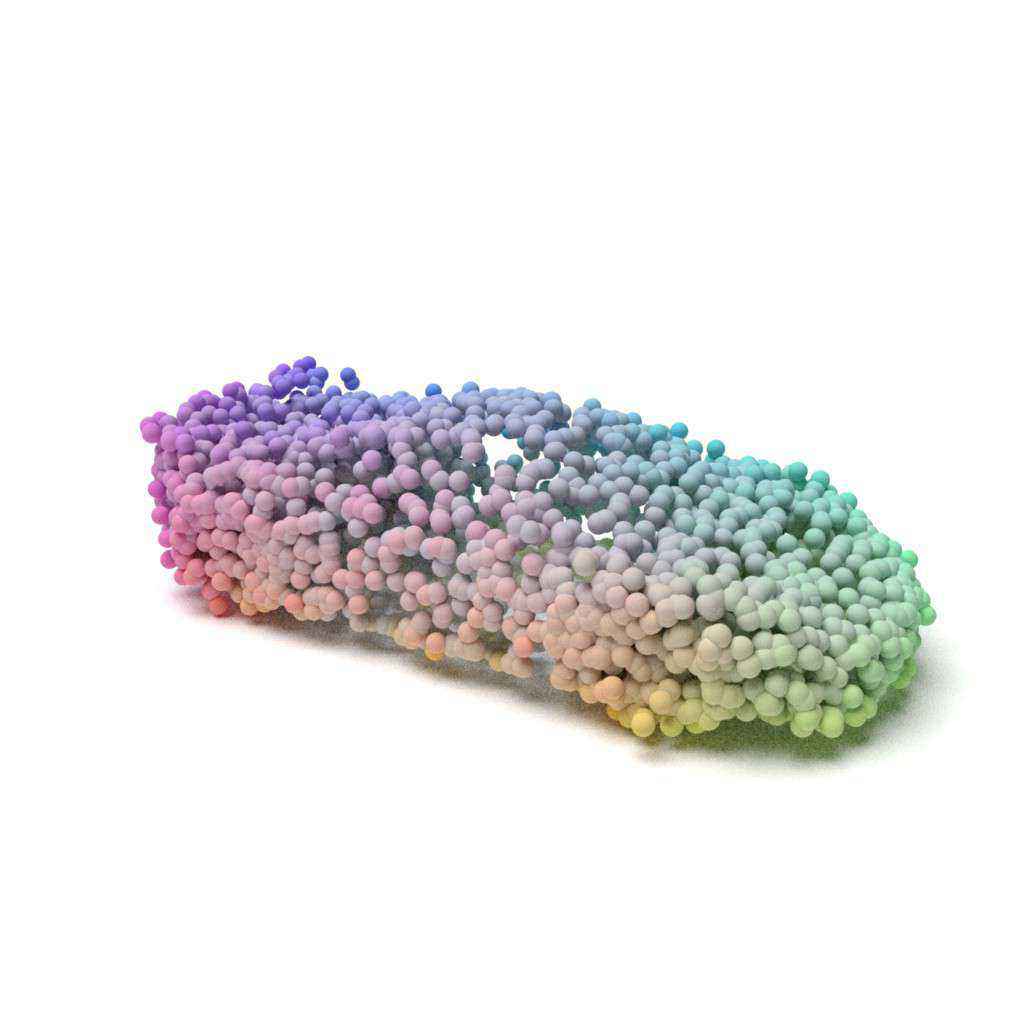}\\
	\includegraphics[width=\sizea]{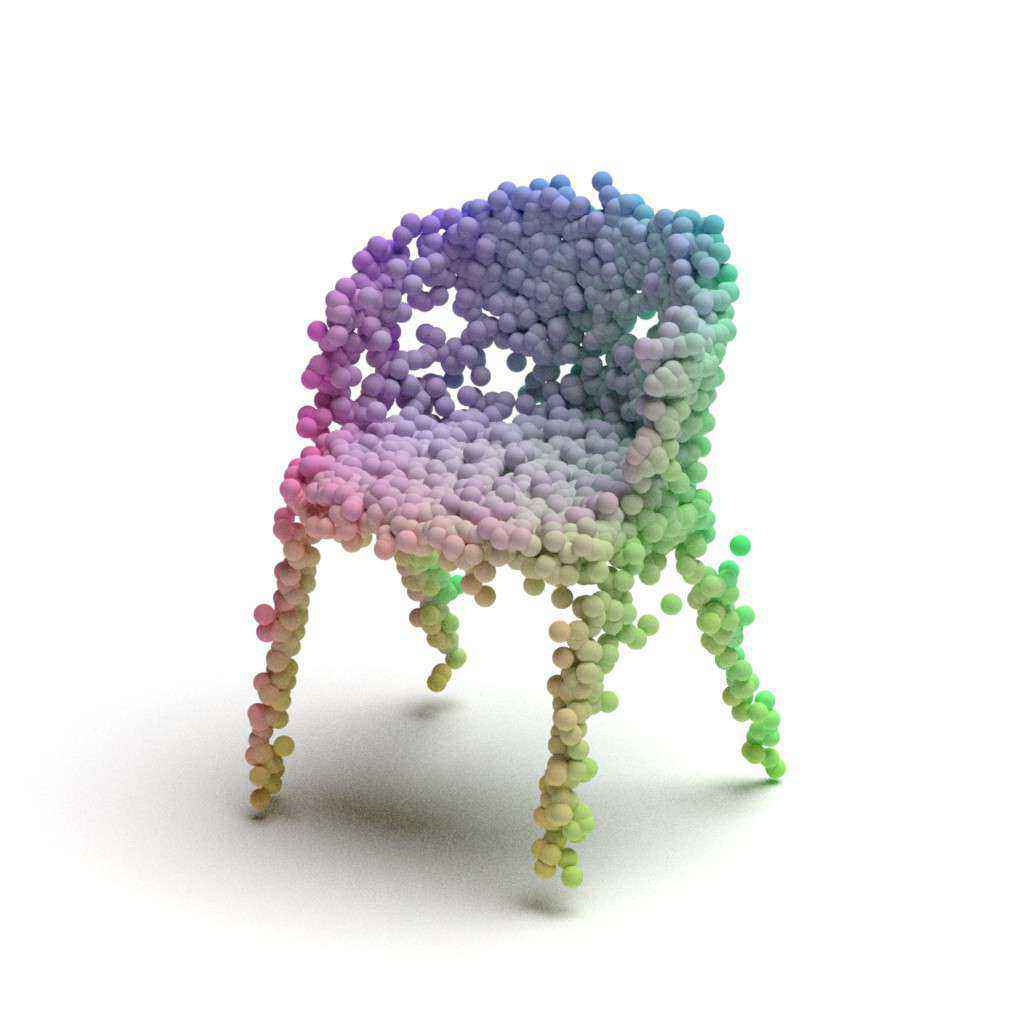}
	\includegraphics[width=\sizea]{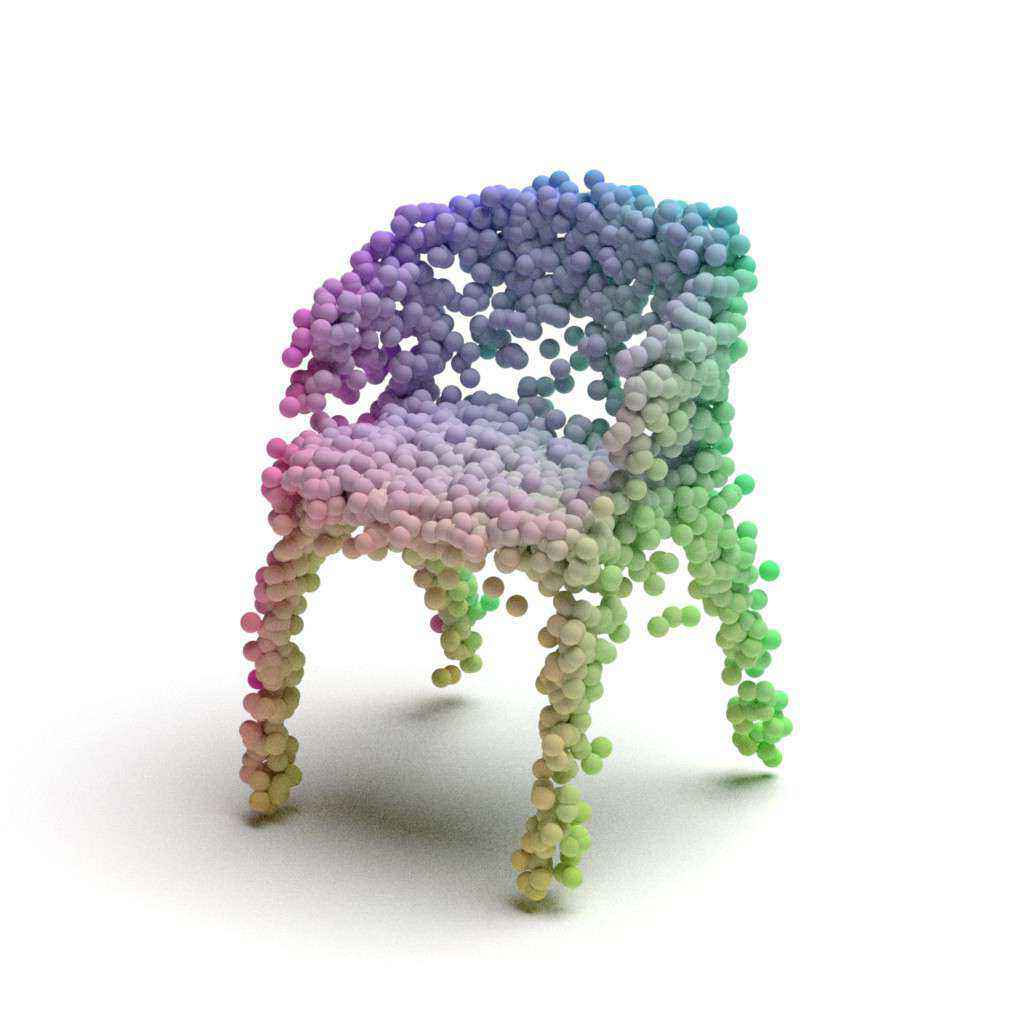}
	\includegraphics[width=\sizea]{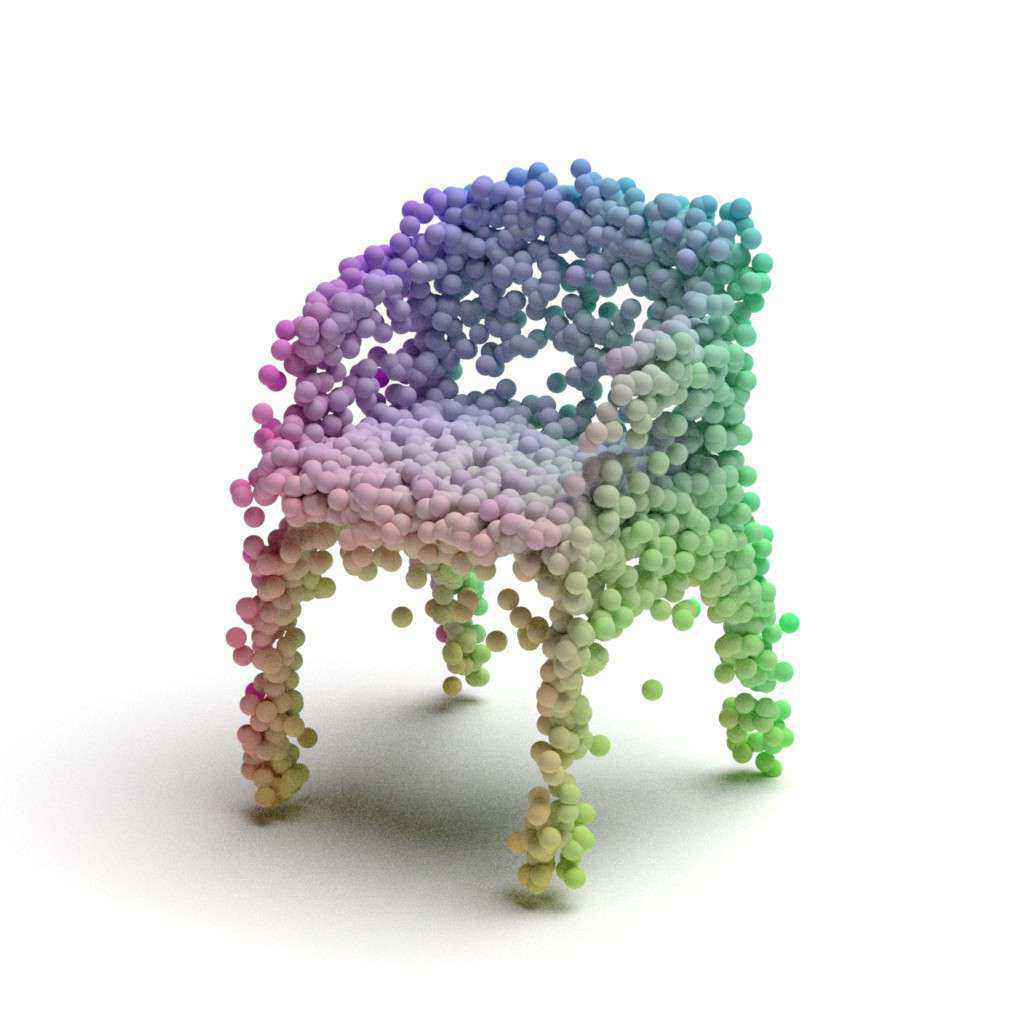}
	\includegraphics[width=\sizea]{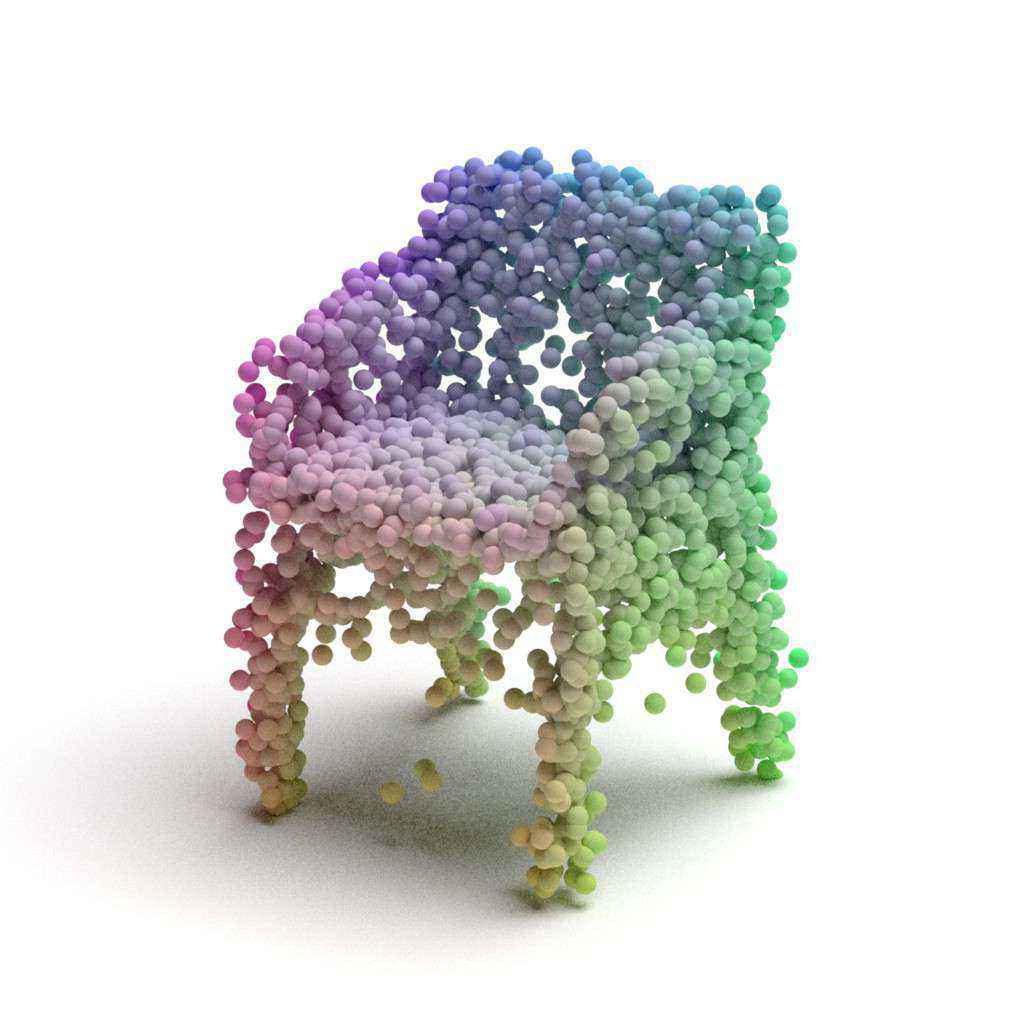}
	\includegraphics[width=\sizea]{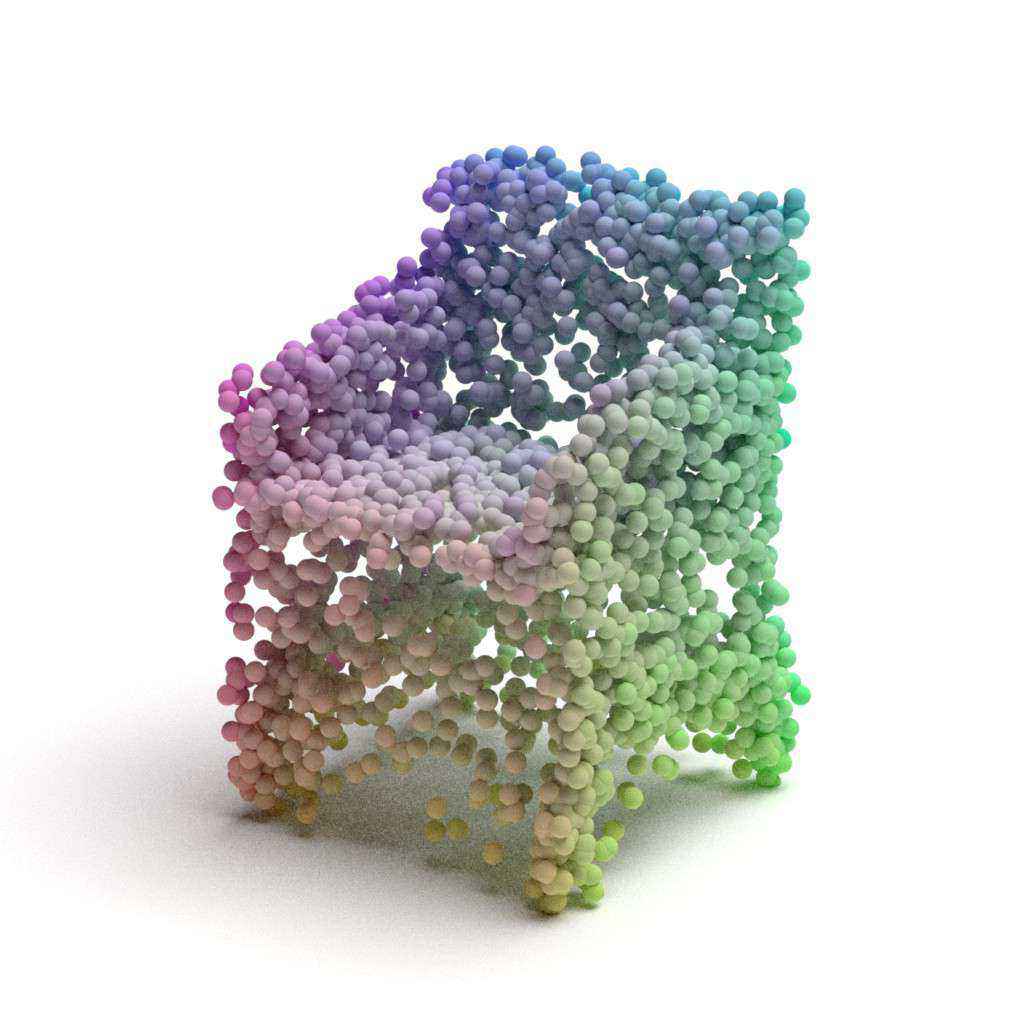}
	\includegraphics[width=\sizea]{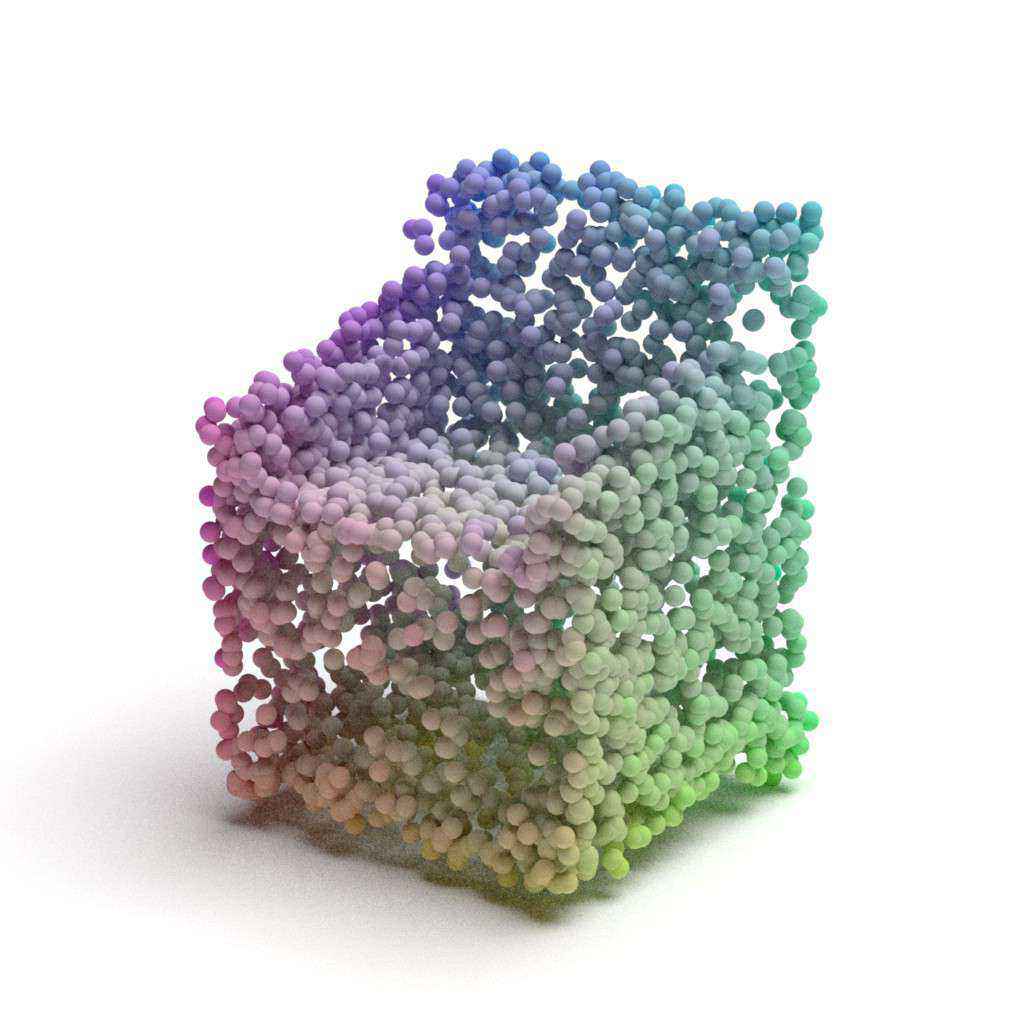}
	\includegraphics[width=\sizea]{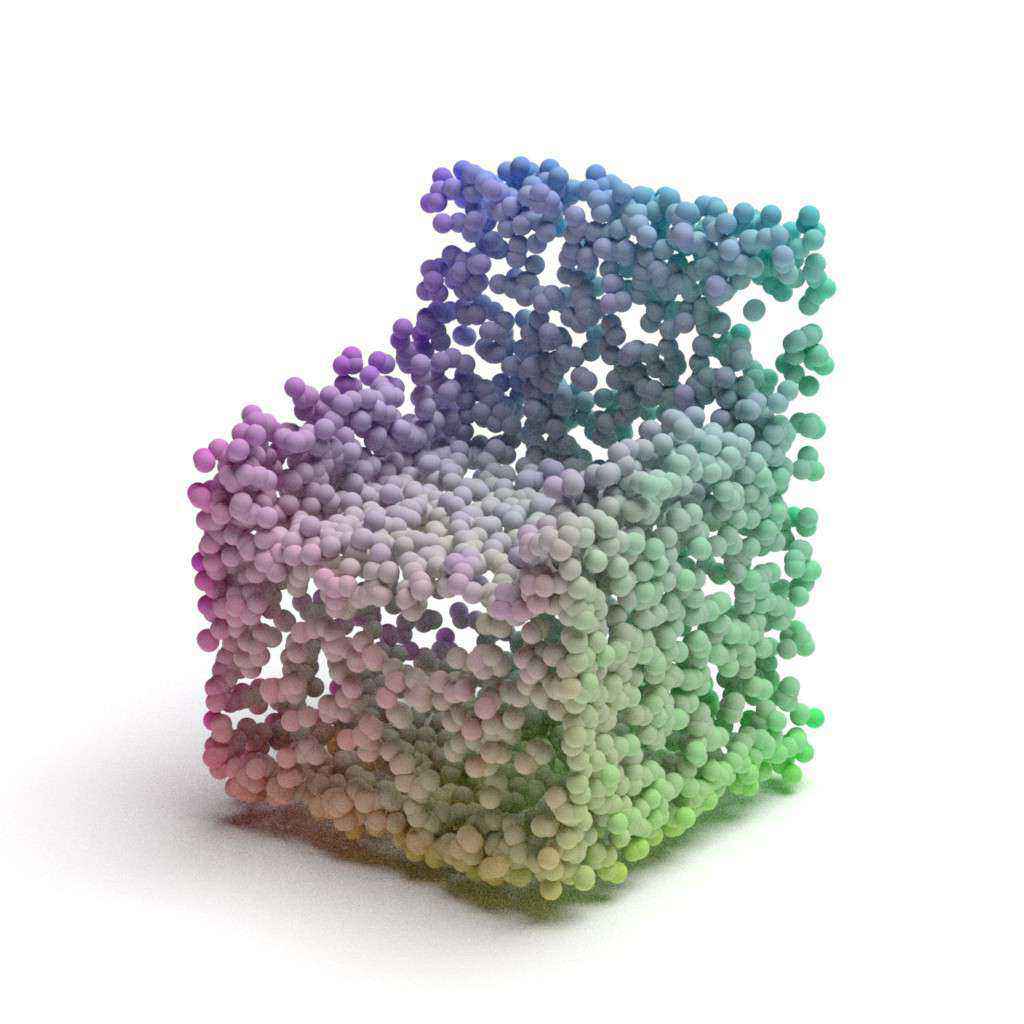}\\
	\includegraphics[width=\sizea]{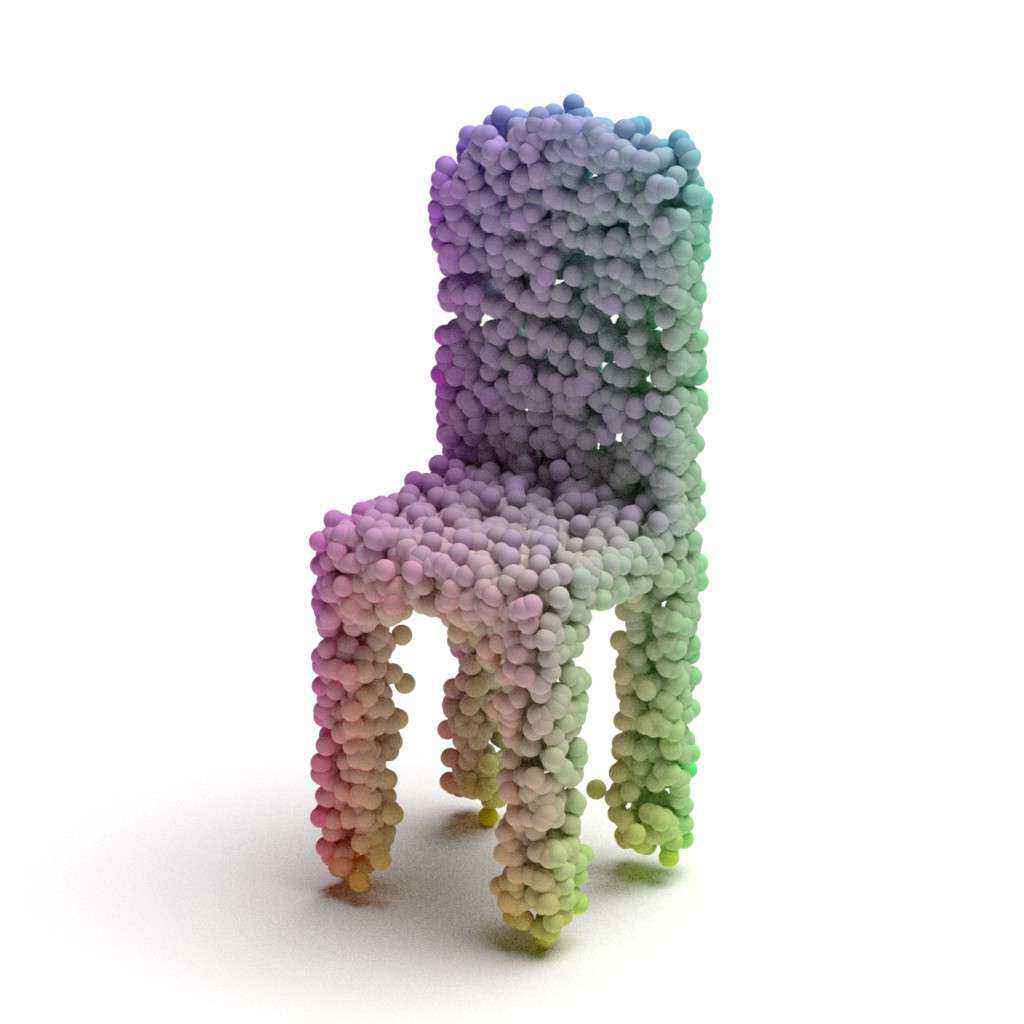}
	\includegraphics[width=\sizea]{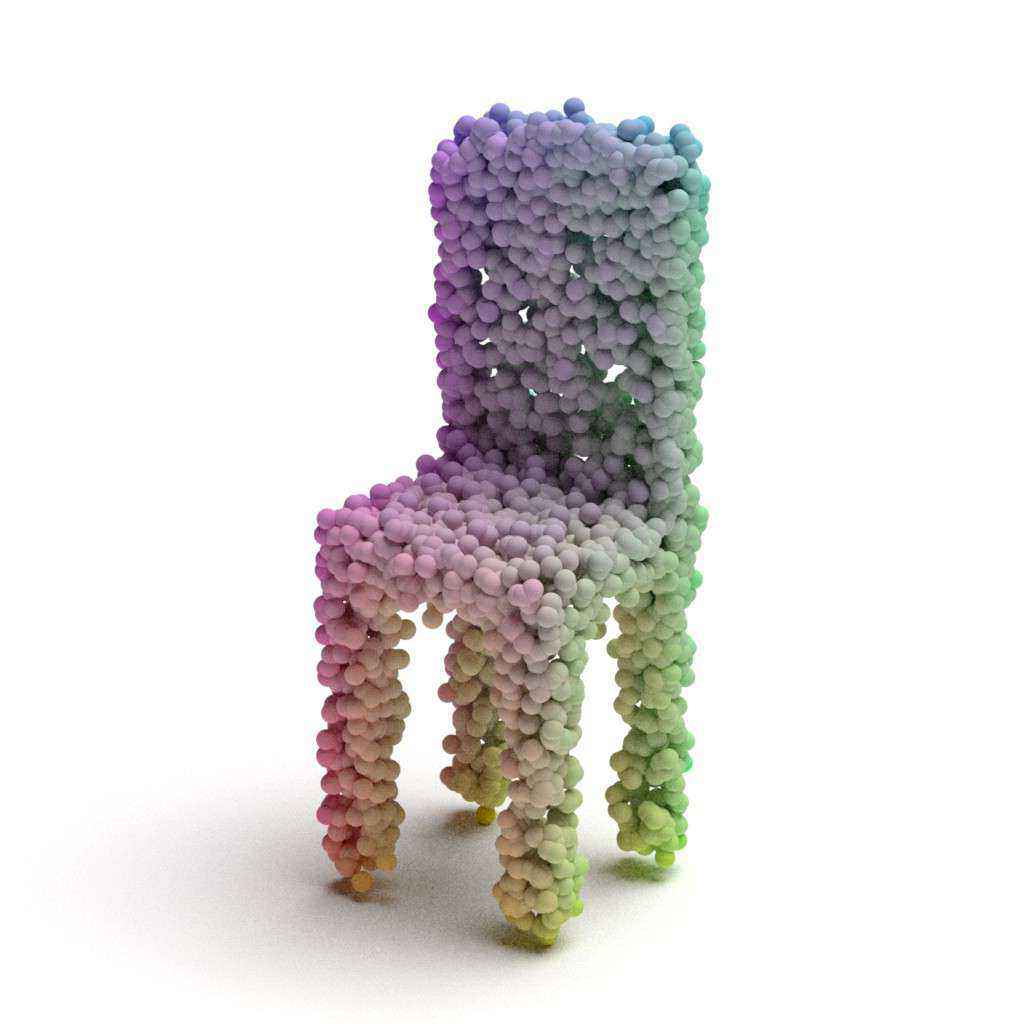}
	\includegraphics[width=\sizea]{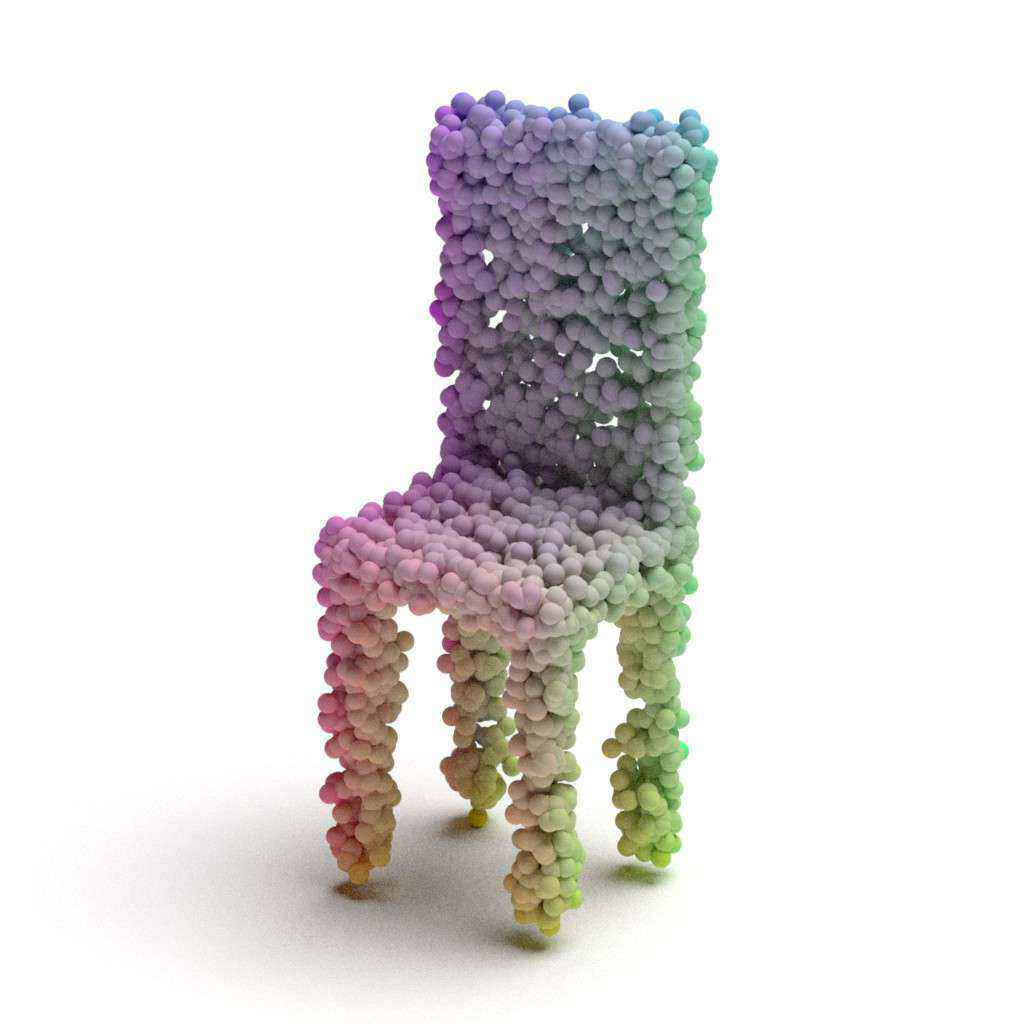}
	\includegraphics[width=\sizea]{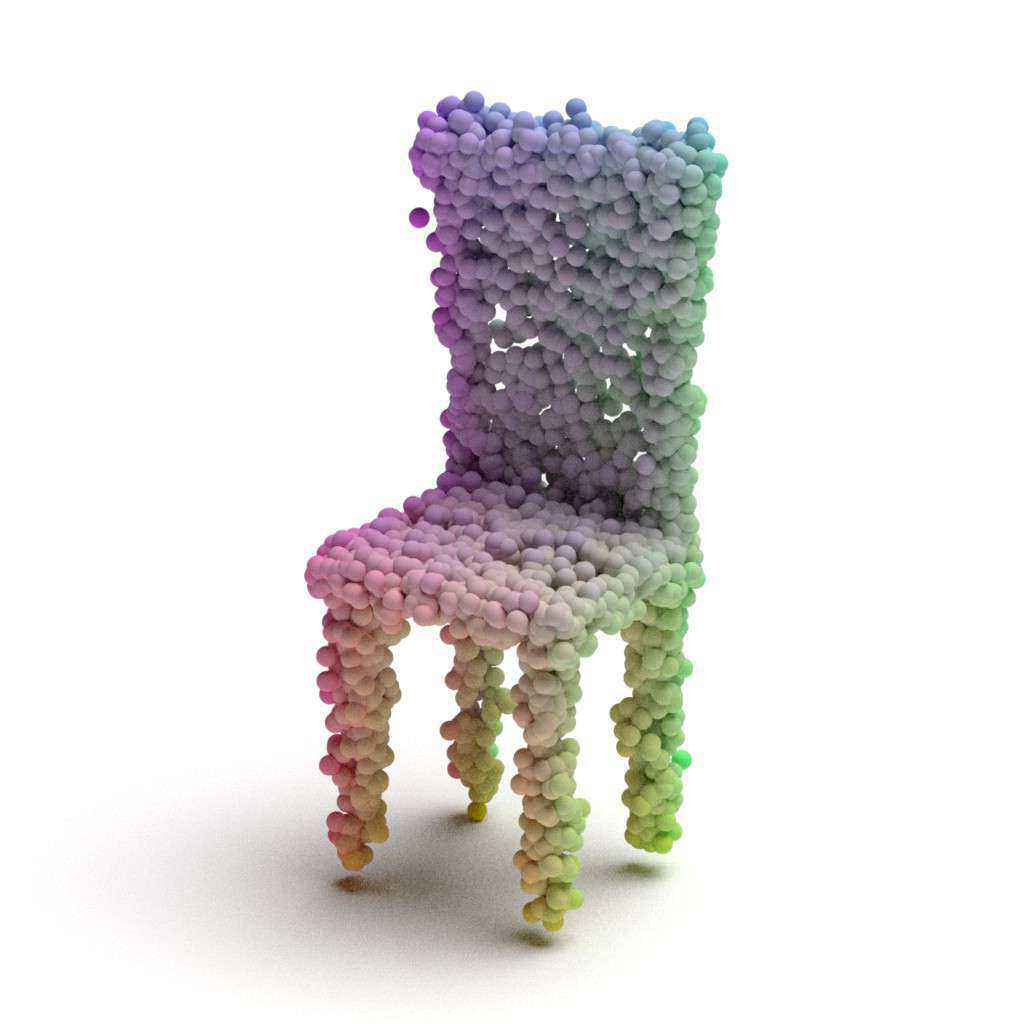}
	\includegraphics[width=\sizea]{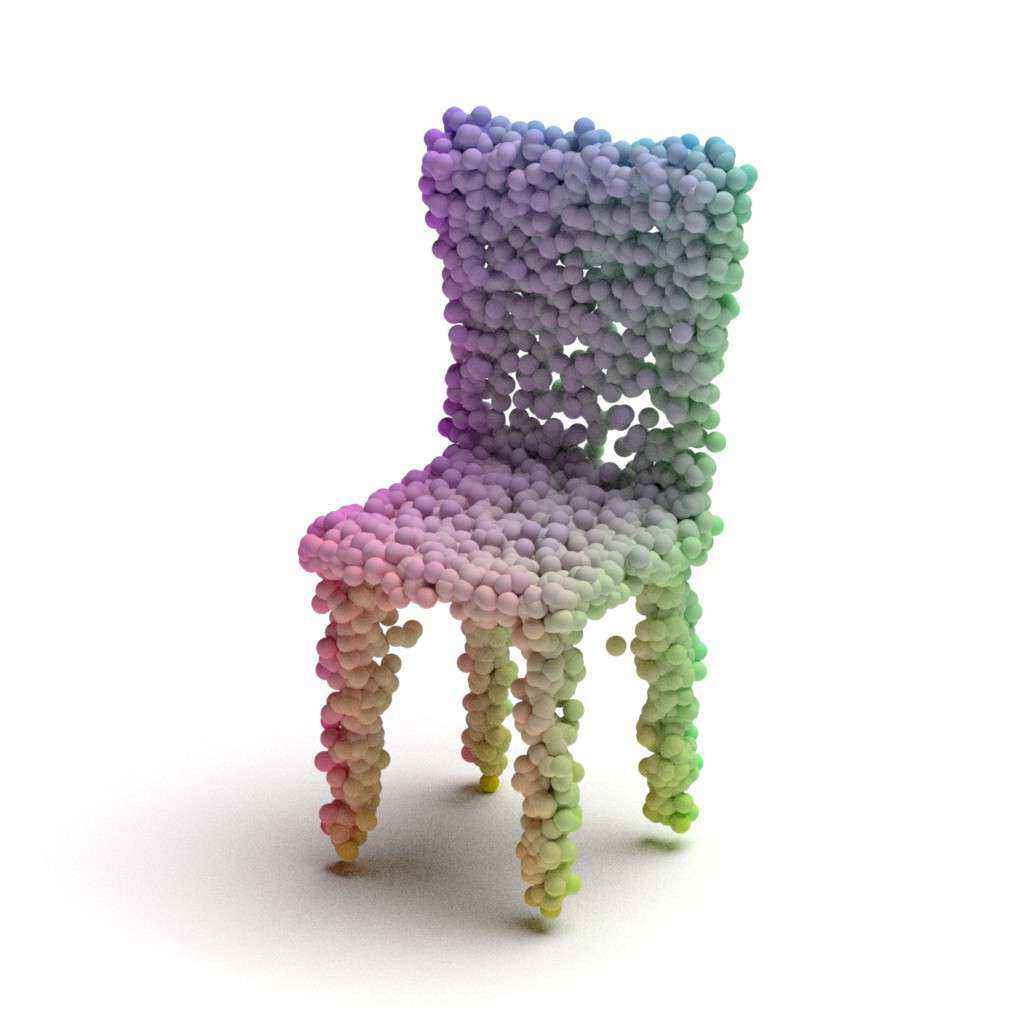}
	\includegraphics[width=\sizea]{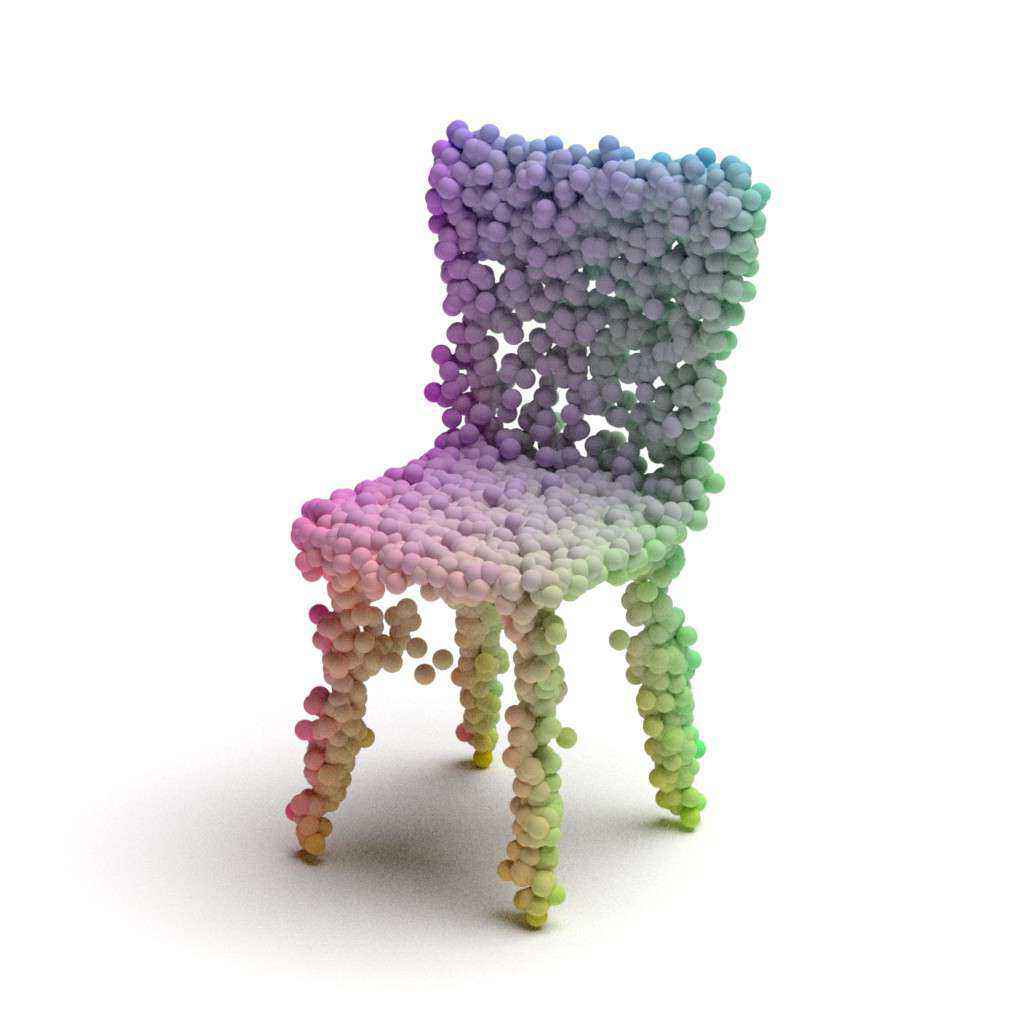}
	\includegraphics[width=\sizea]{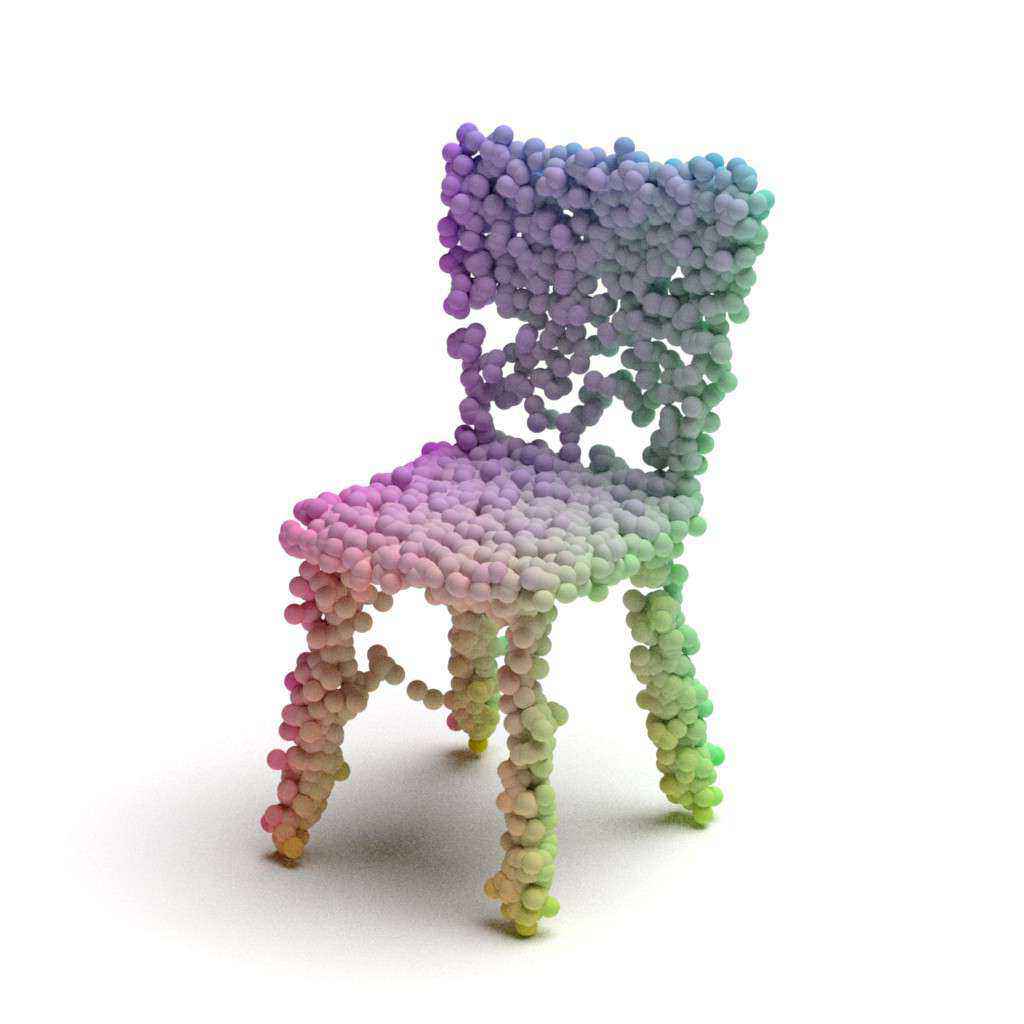}\\
	\captionof{figure}{Feature space interpolation. The left-most and the right-most shapes are sampled from scratch. The shapes in between are generated by interpolating the two shapes in the prior space.}
	\label{fig:geninterp}
	\vspace{0.2in}
\end{figure*}

In this section, we present interpolation between two different shapes using our model.
For two shapes $X_1$ and $X_2$, we first compute the mean of the posterior distribution using $Q_\theta(z|X)$.  
Let $\mu_1$ and $\mu_2$ be the means of the posterior distribution for $X_1$ and $X_2$ respectively.
We use $\mu_1$ and $\mu_2$ as the latent representation for these two shapes.
We then use the inverse prior flow $F_\psi^{-1}$ to transform $\mu_1$ and $\mu_2$ back to the prior space.
Let $w_1 = F_\psi^{-1}(\mu_1)$ and $w_2 = F_\psi^{-1}(\mu_2)$ be the corresponding vectors for $\mu_1$ and $\mu_2$ in the prior space.
We use spherical interpolation between $w_1$ and $w_2$ to retrieve a series of vectors $w_i$.
For each $w_i$, we use the CNF prior $F_\psi$ and the CNF decoder $G_\theta$ to generate the corresponding shape $X_i$.
Figure~\ref{fig:geninterp} contains examples of the interpolation.

\section{More flow transformation}\label{sec:more-visualization}

\begin{figure*}
	\centering
	\newcommand{\sizea}{0.138\linewidth}
	\includegraphics[width=\sizea]{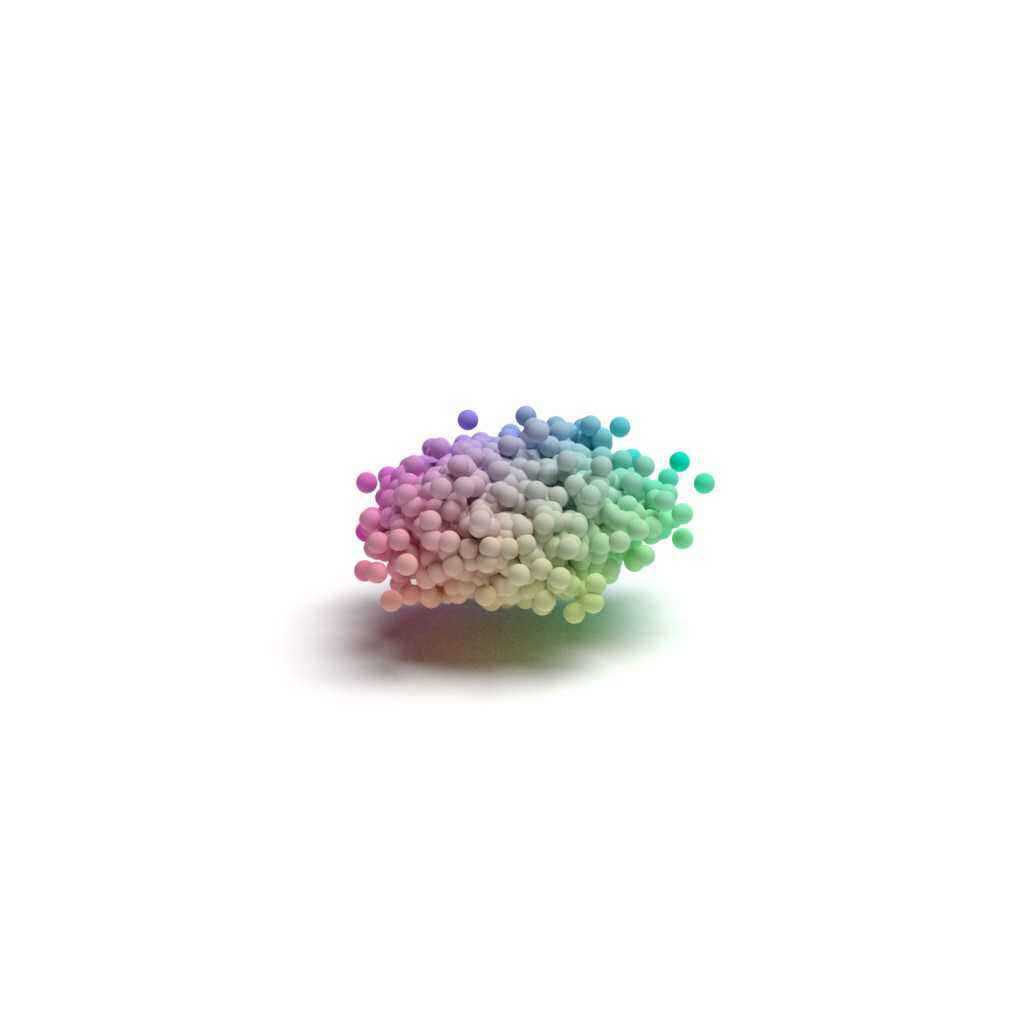}
	\includegraphics[width=\sizea]{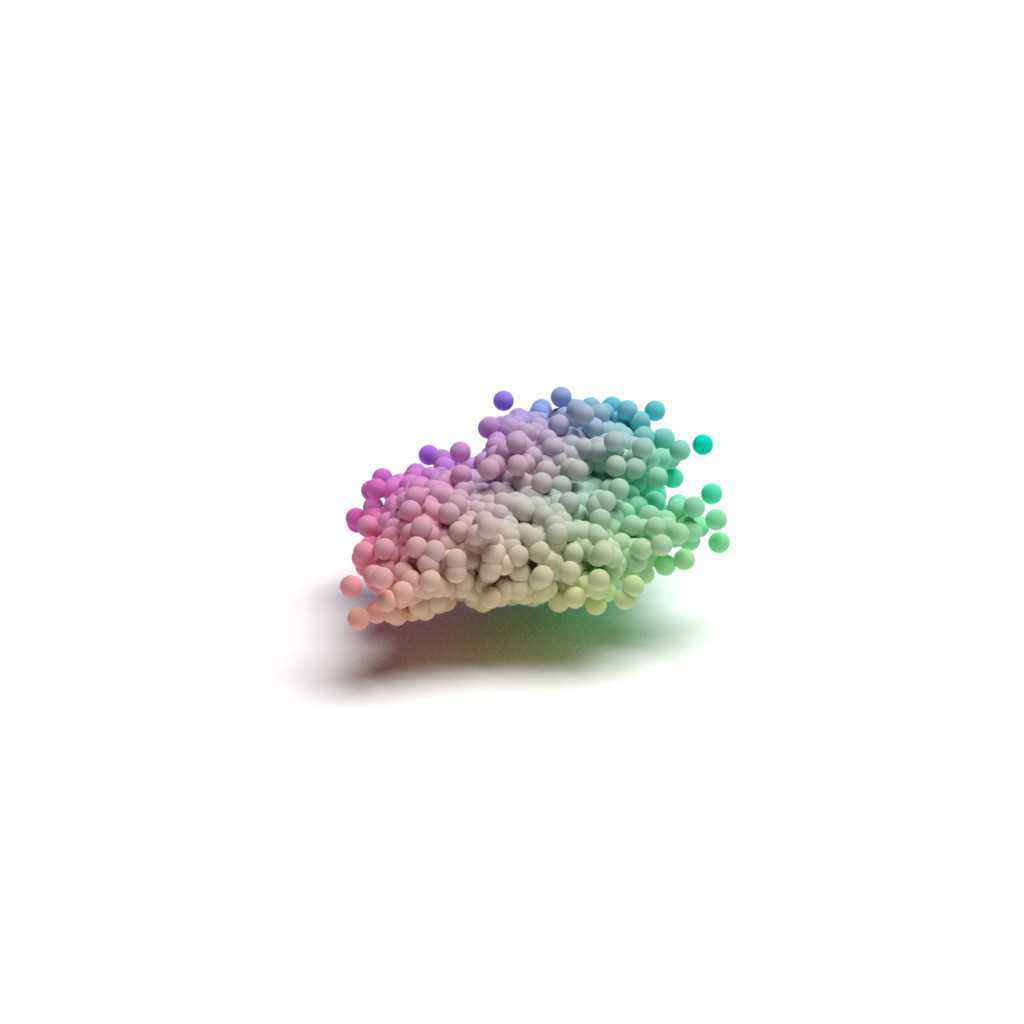}
	\includegraphics[width=\sizea]{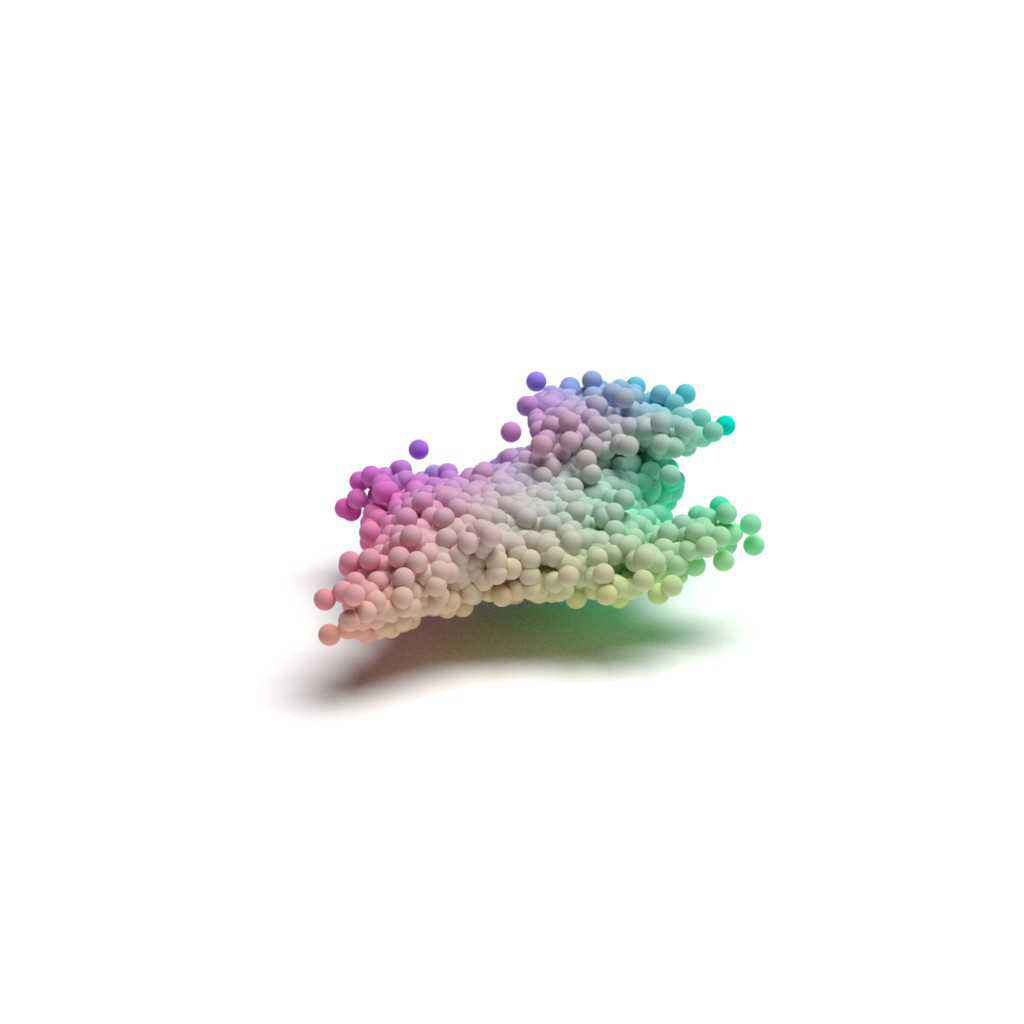}
	\includegraphics[width=\sizea]{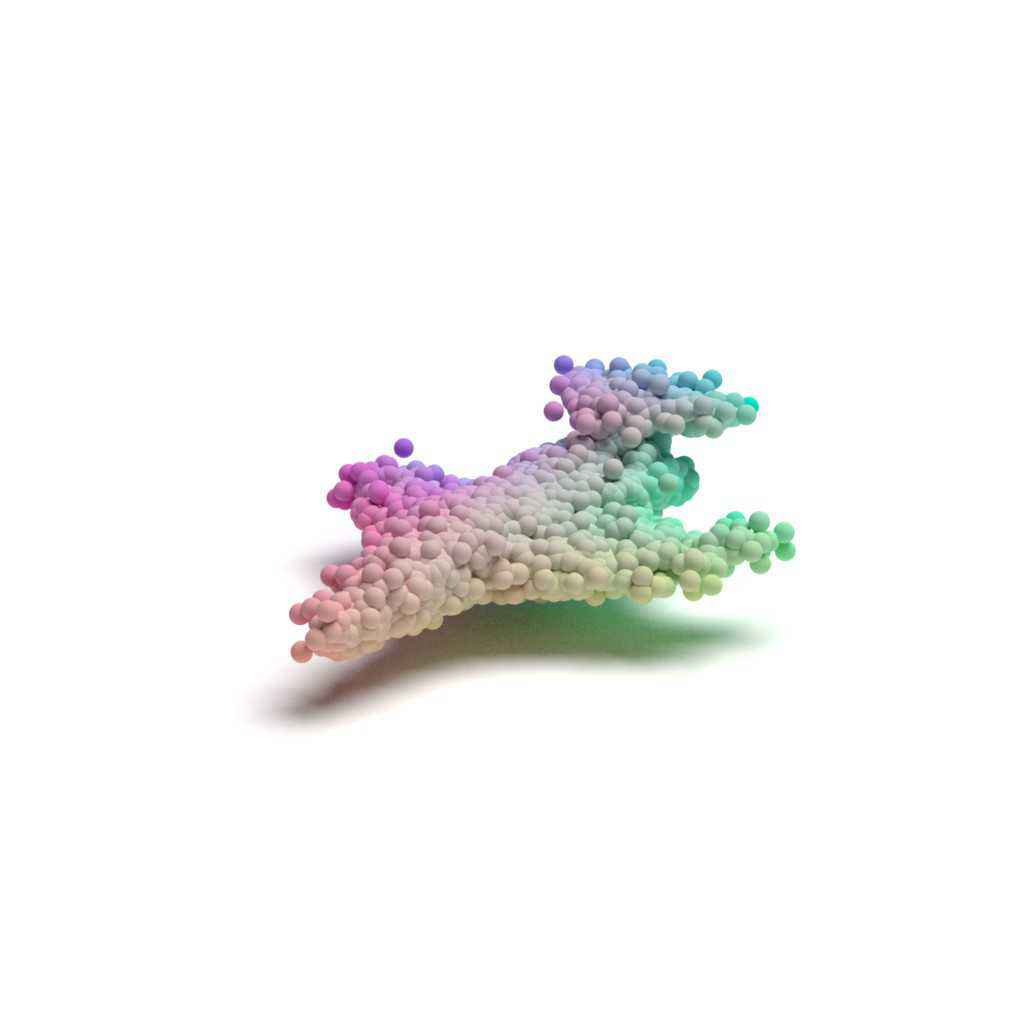}
	\includegraphics[width=\sizea]{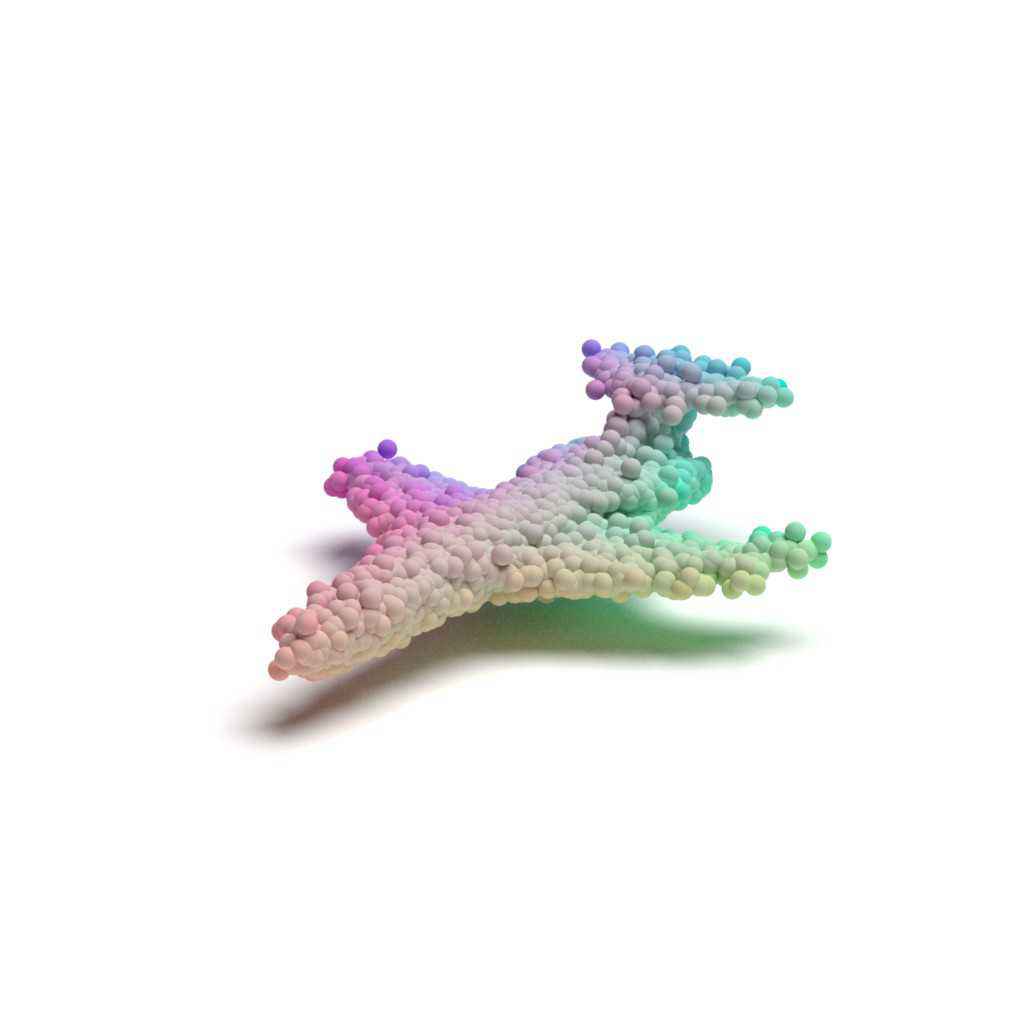}
	\includegraphics[width=\sizea]{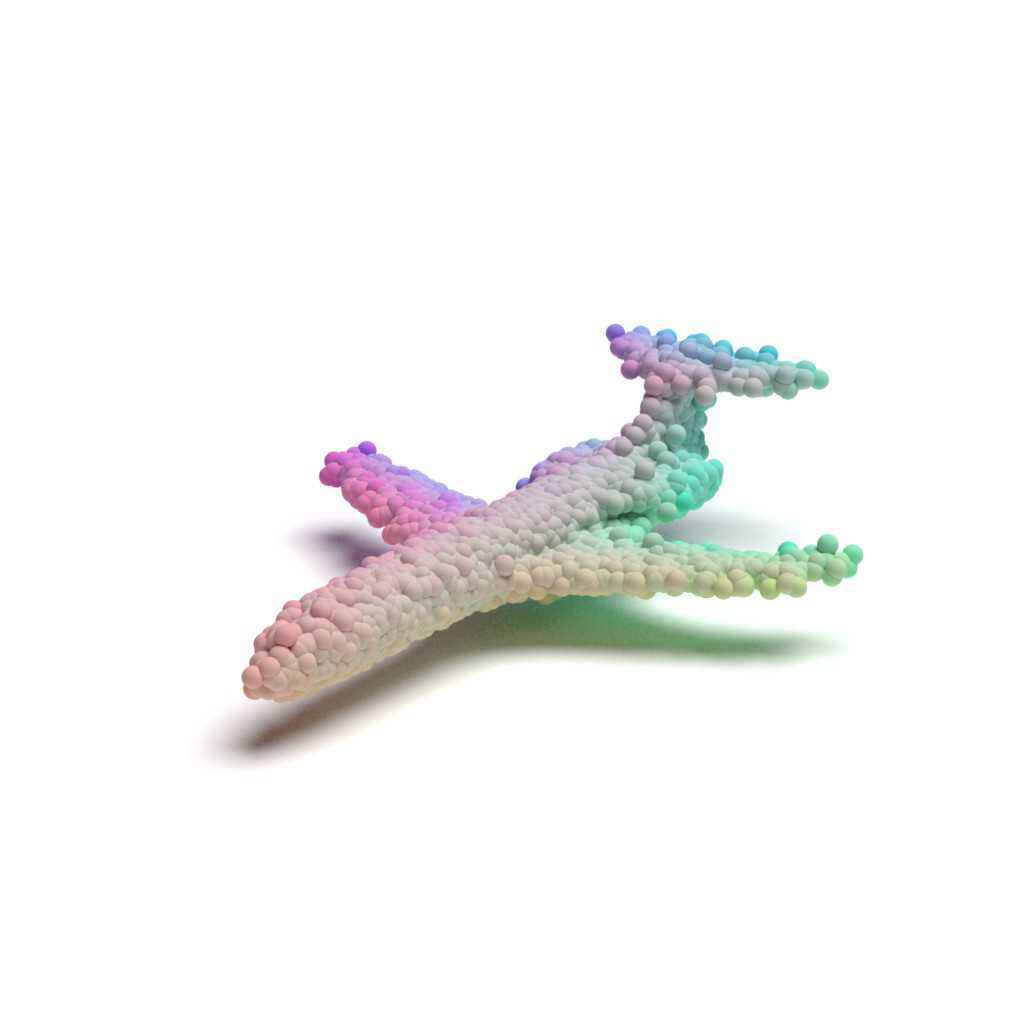}
	\includegraphics[width=\sizea]{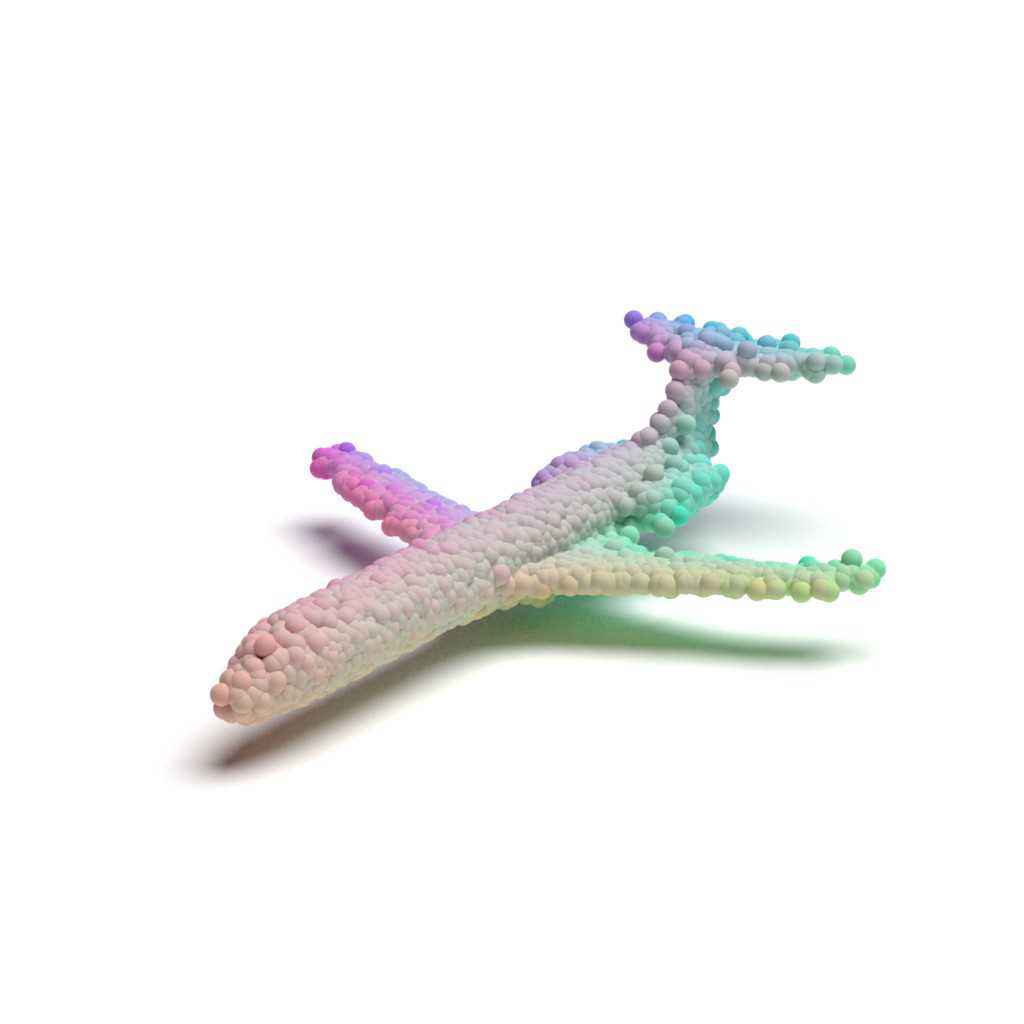}\\
	\includegraphics[width=\sizea]{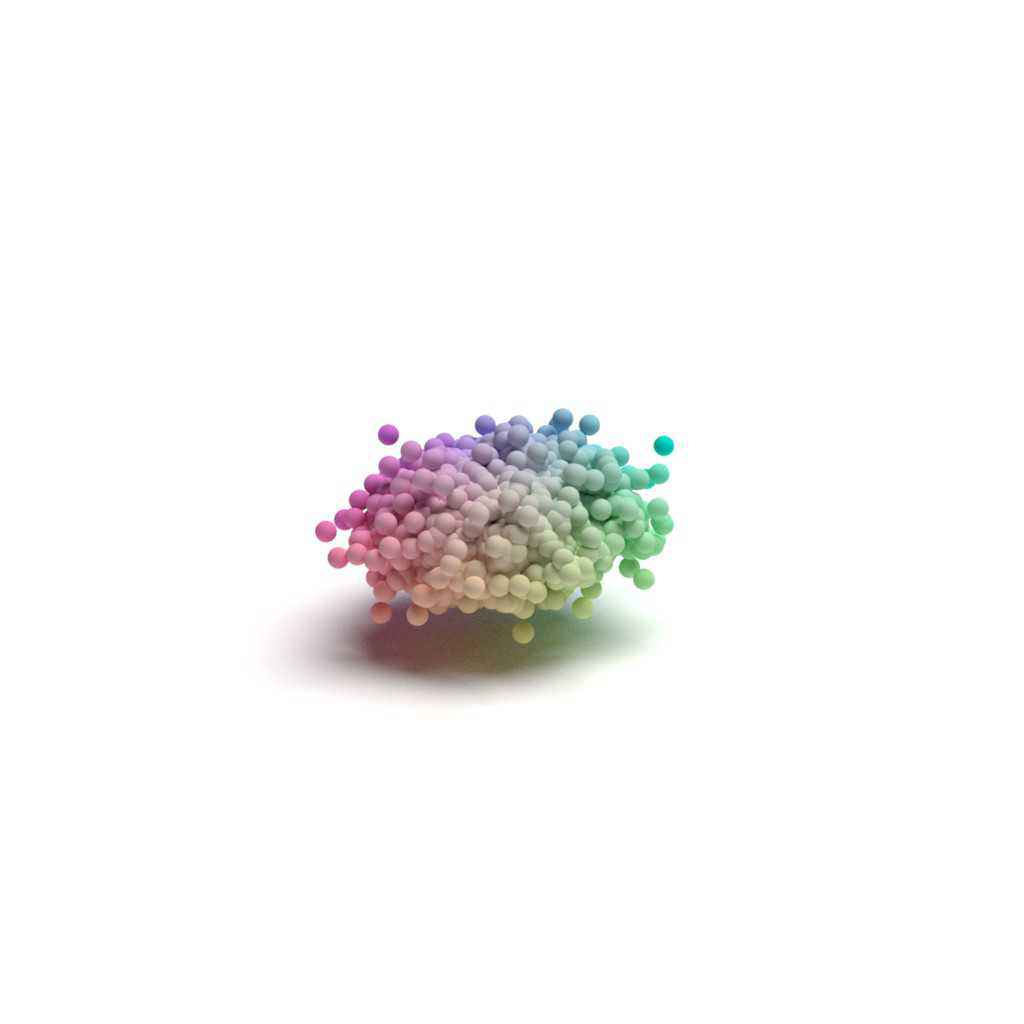}
	\includegraphics[width=\sizea]{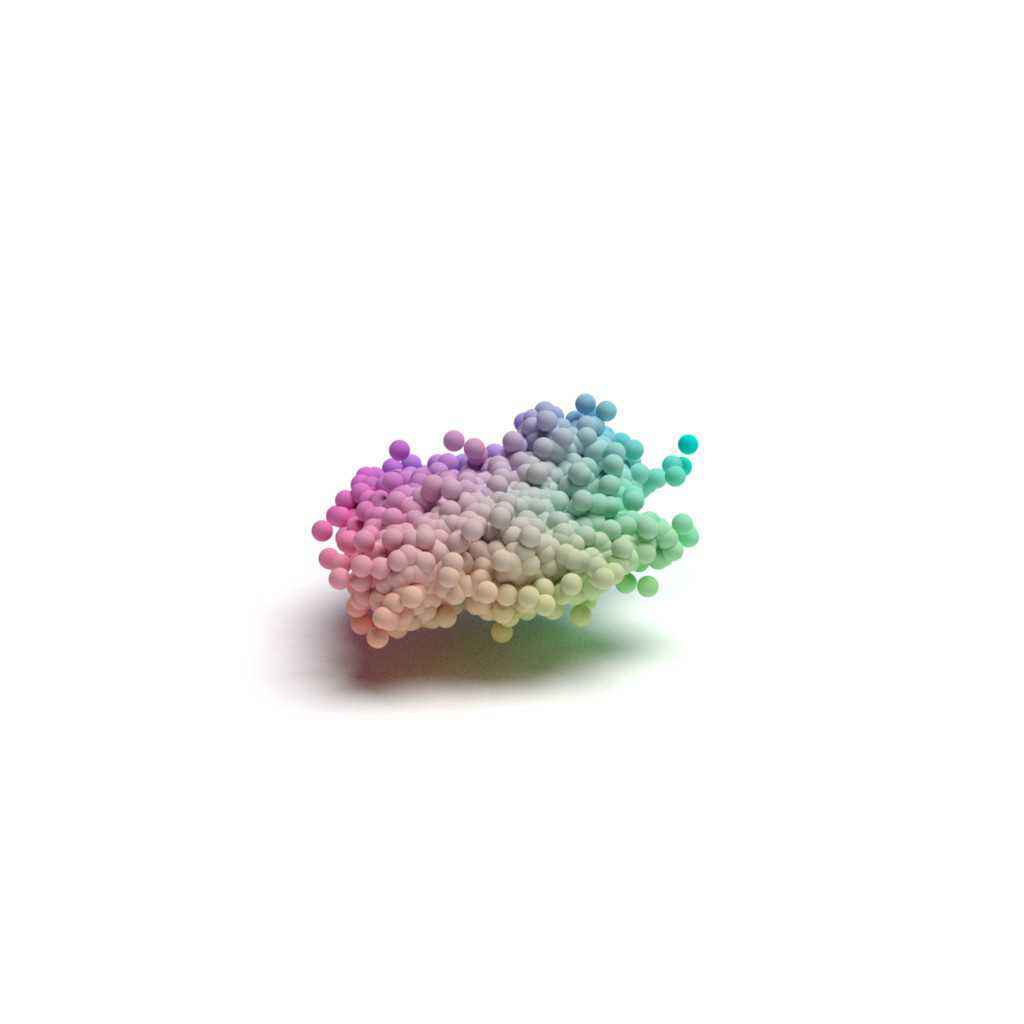}
	\includegraphics[width=\sizea]{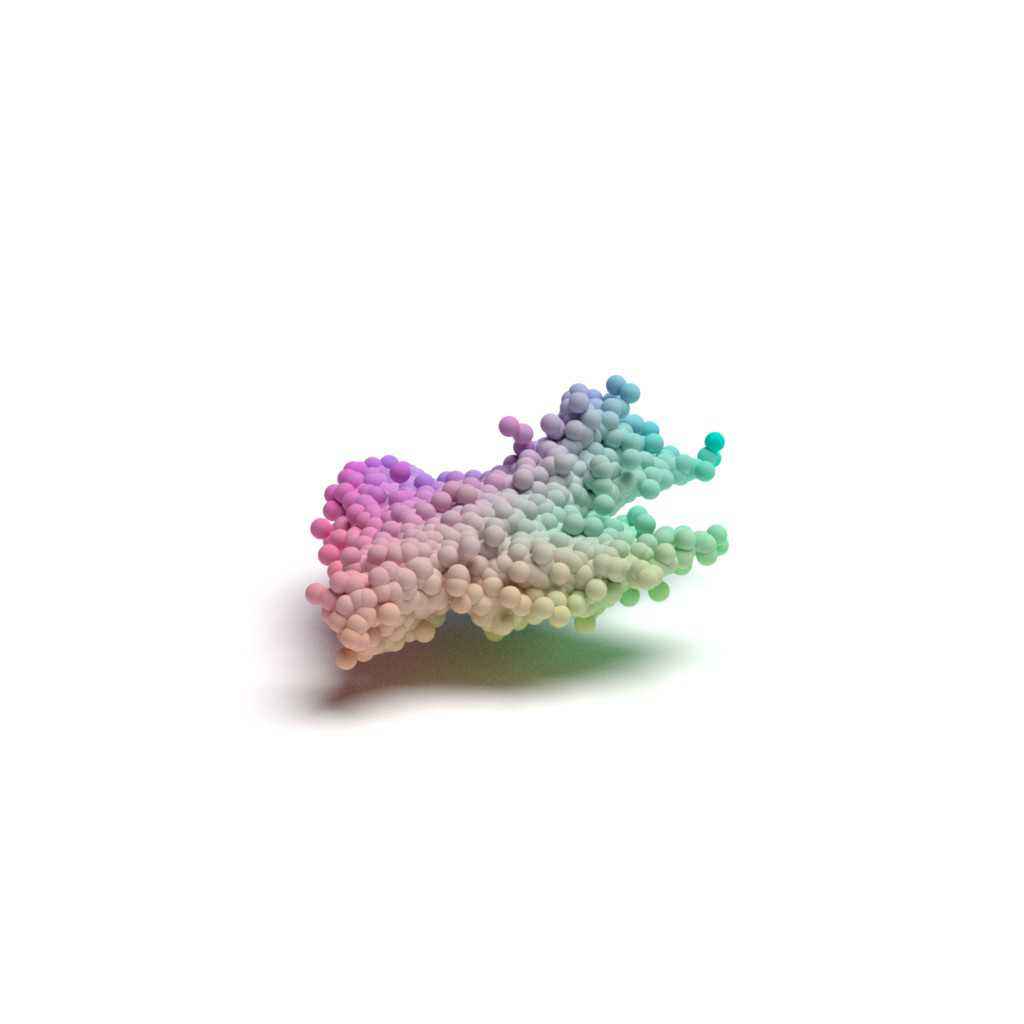}
	\includegraphics[width=\sizea]{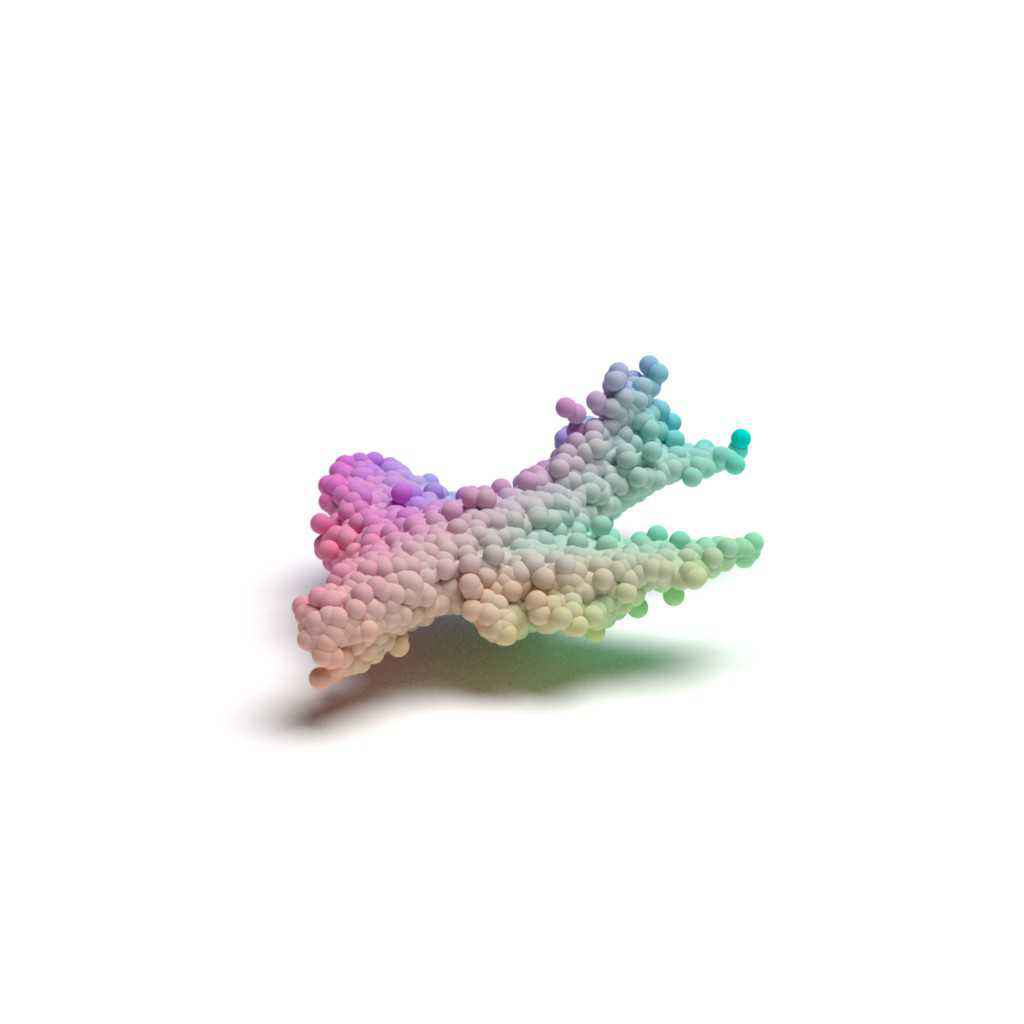}
	\includegraphics[width=\sizea]{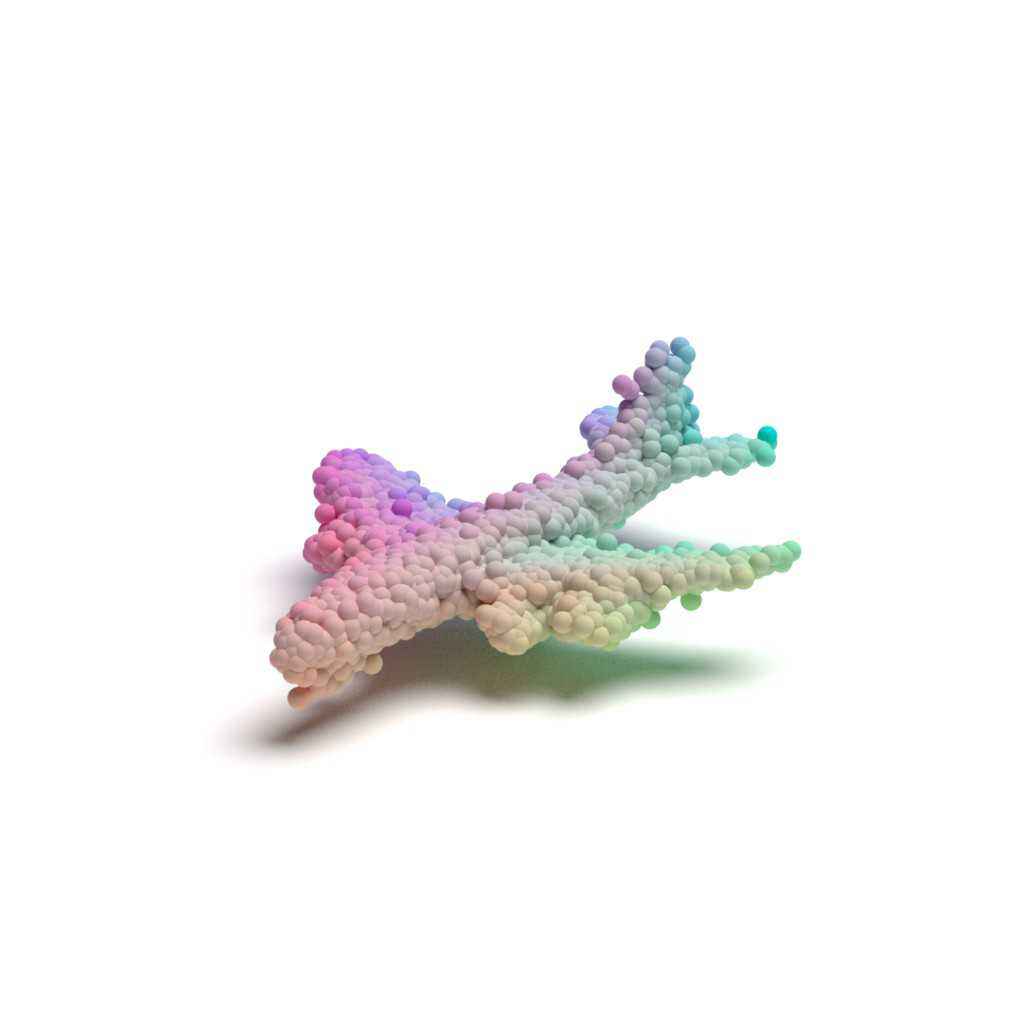}
	\includegraphics[width=\sizea]{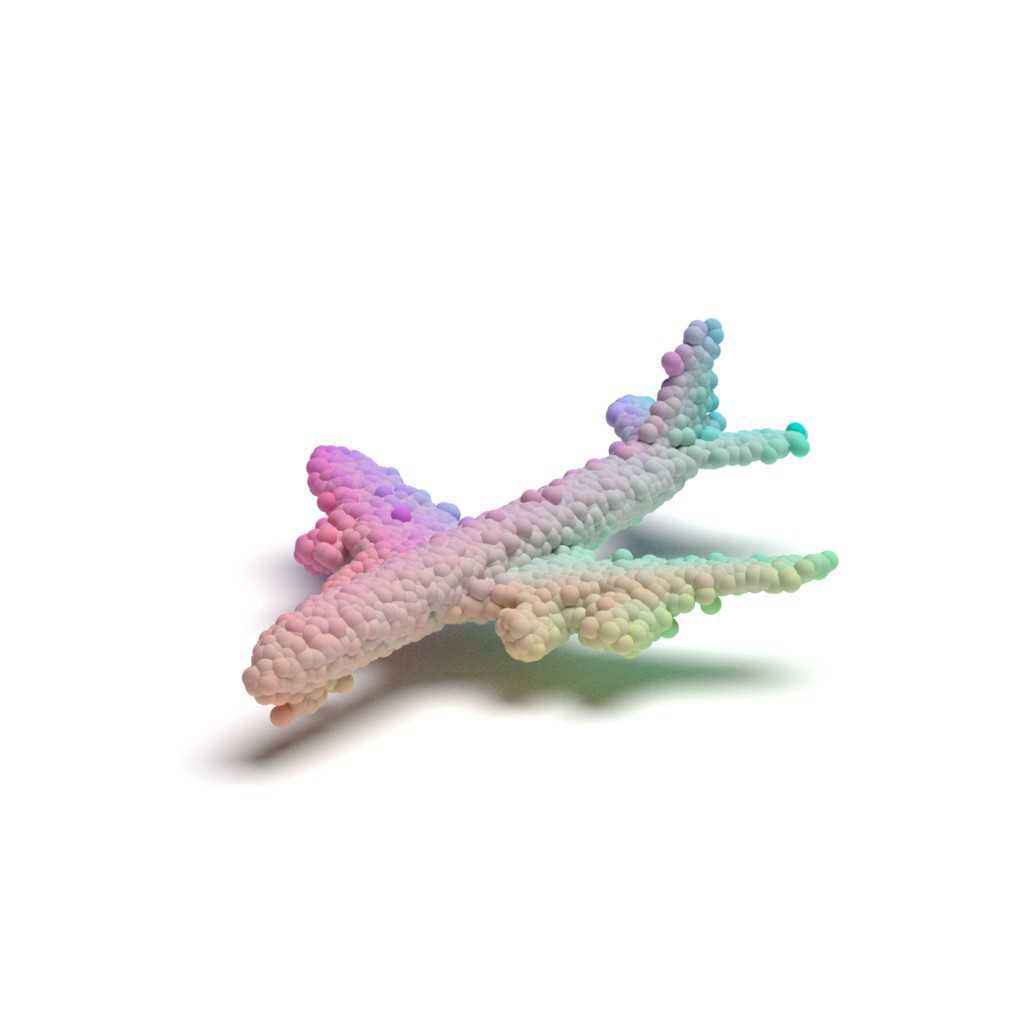}
	\includegraphics[width=\sizea]{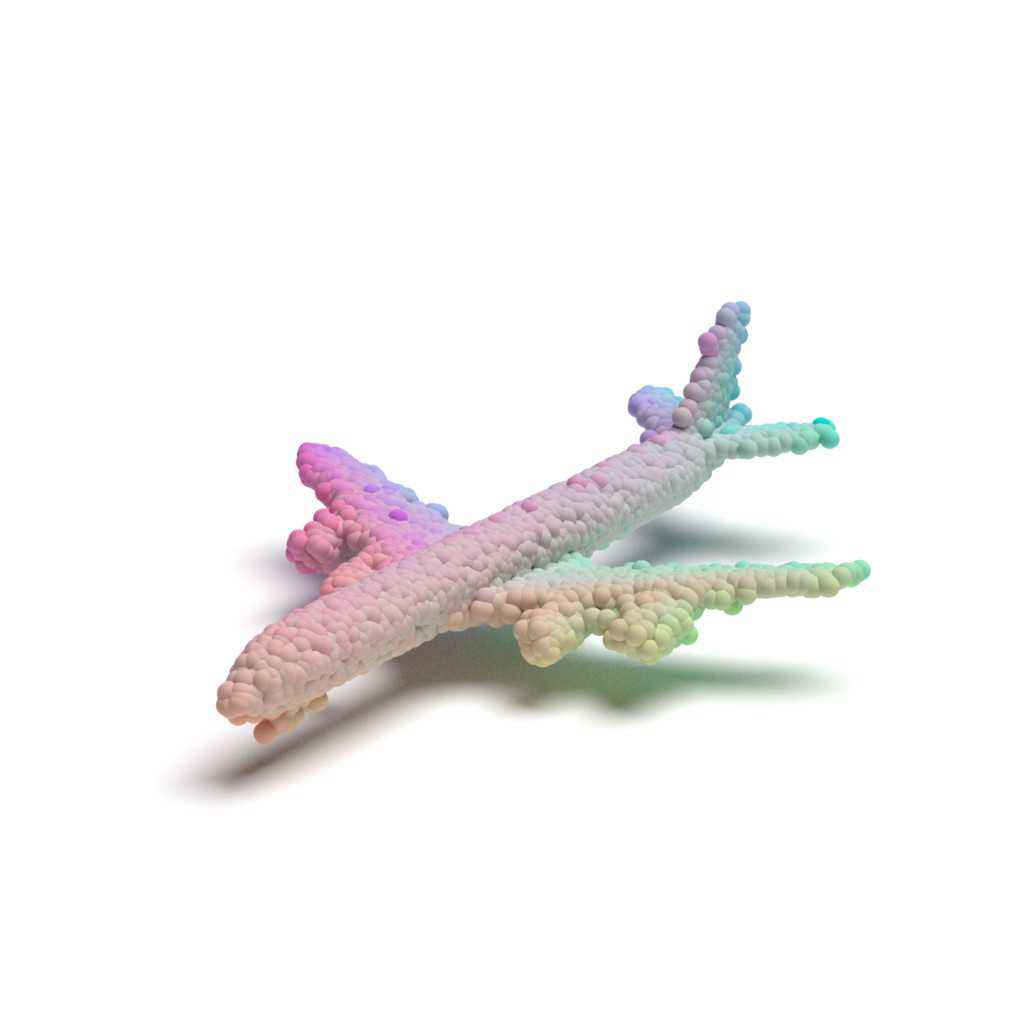}\\
	\includegraphics[width=\sizea]{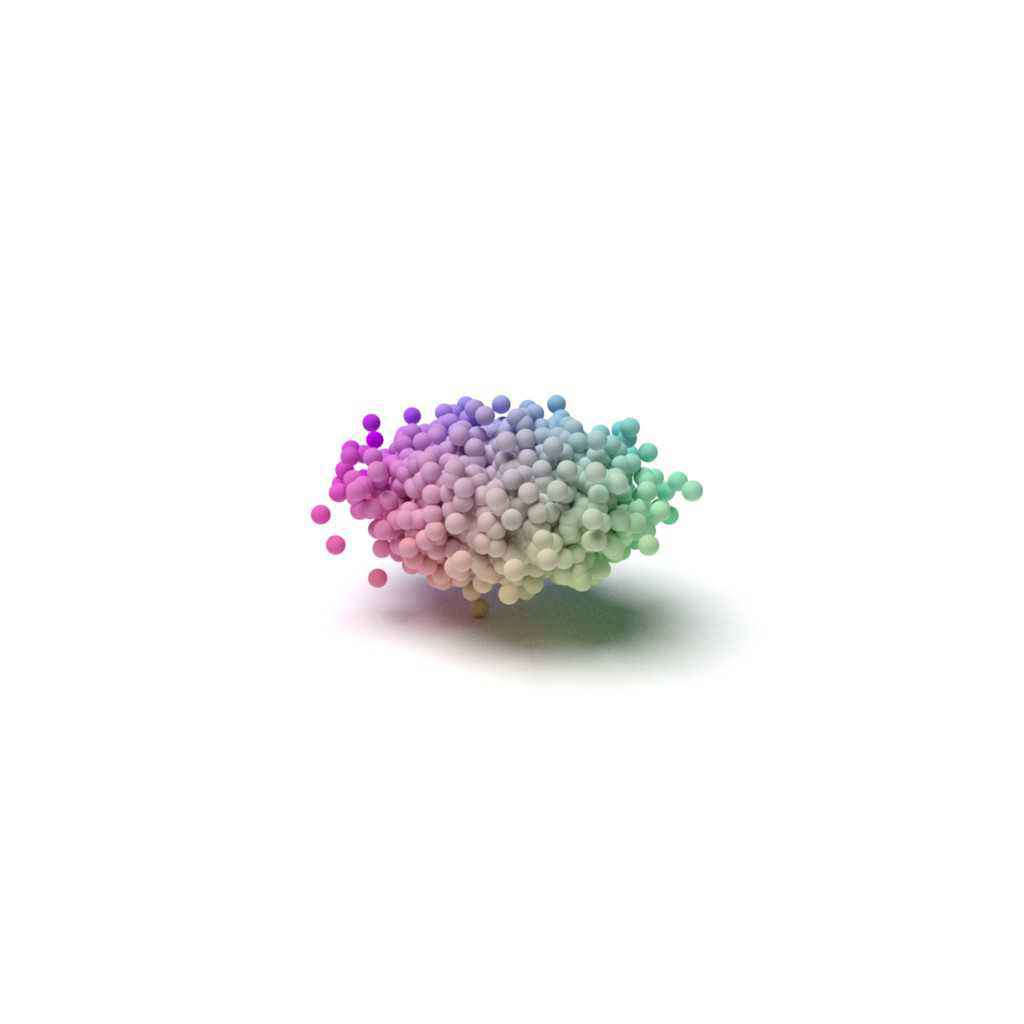}
	\includegraphics[width=\sizea]{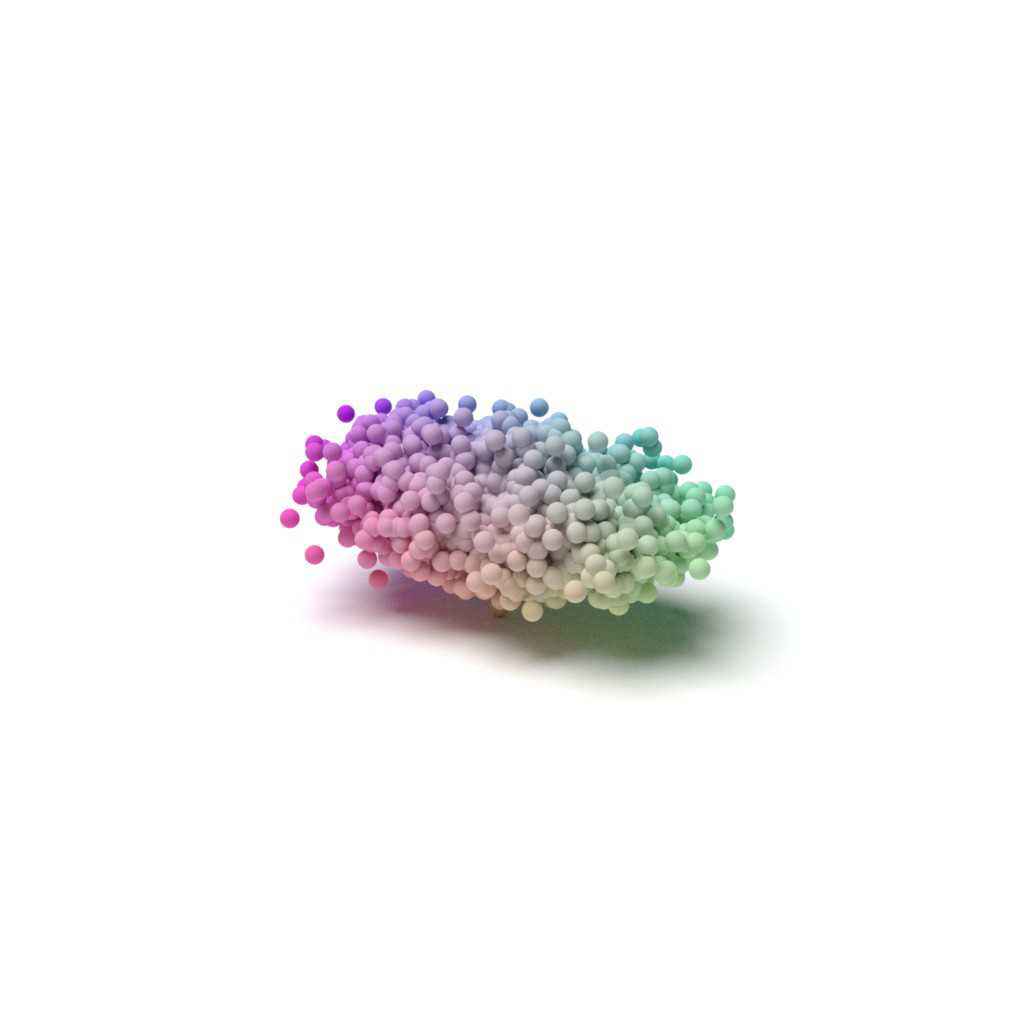}
	\includegraphics[width=\sizea]{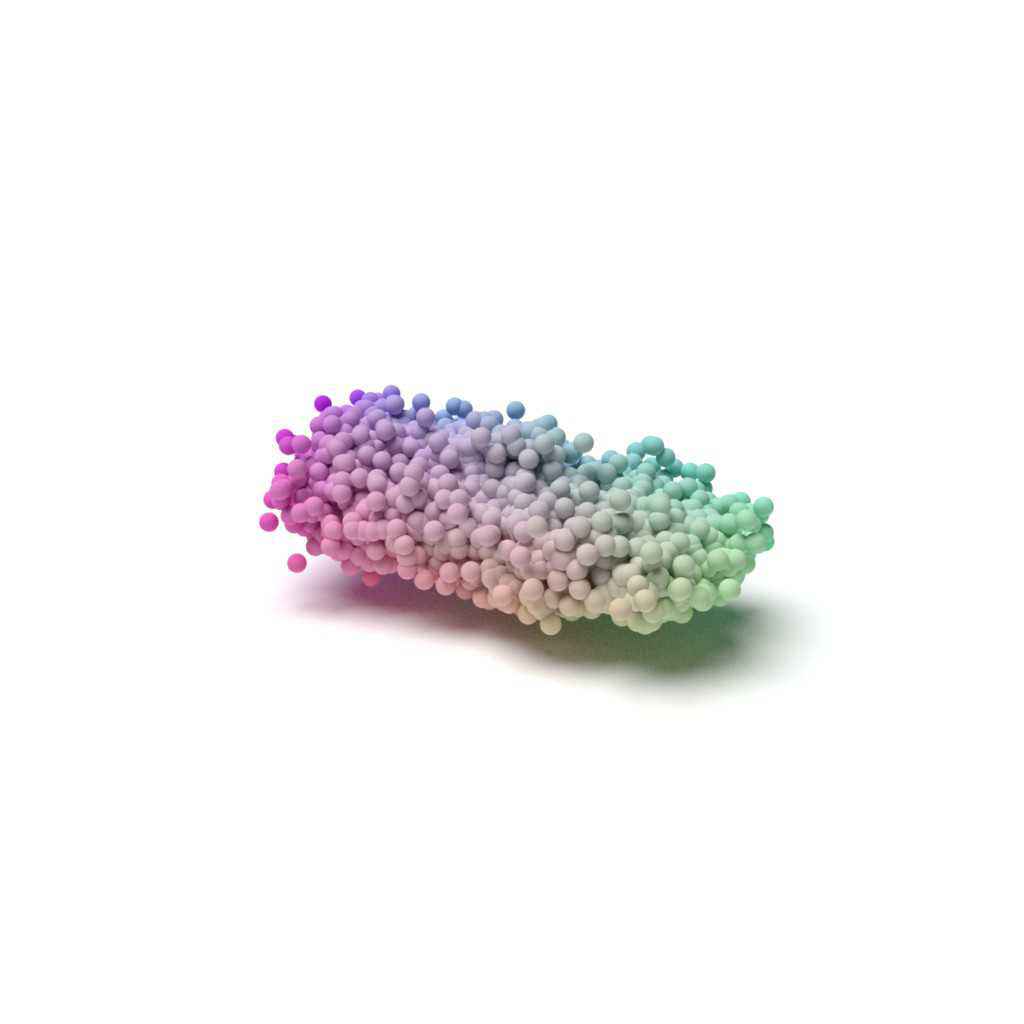}
	\includegraphics[width=\sizea]{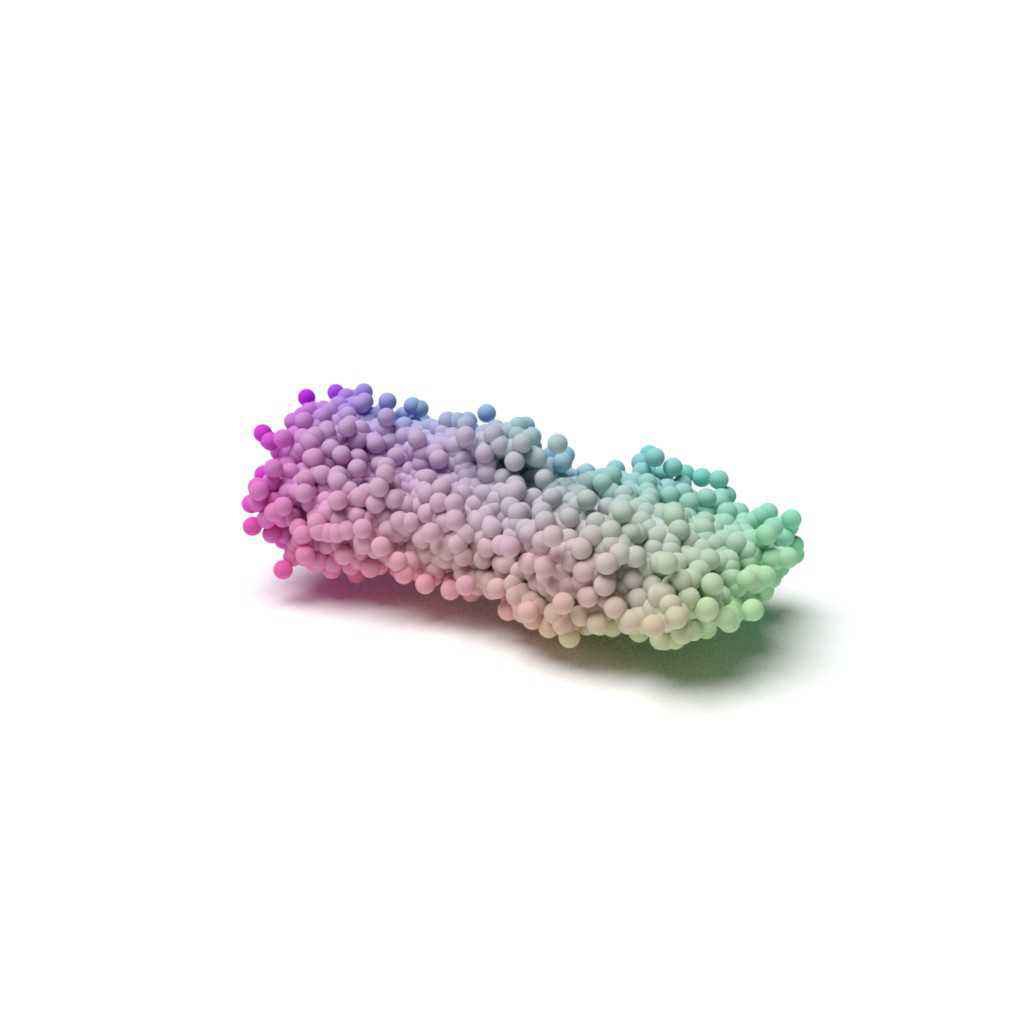}
	\includegraphics[width=\sizea]{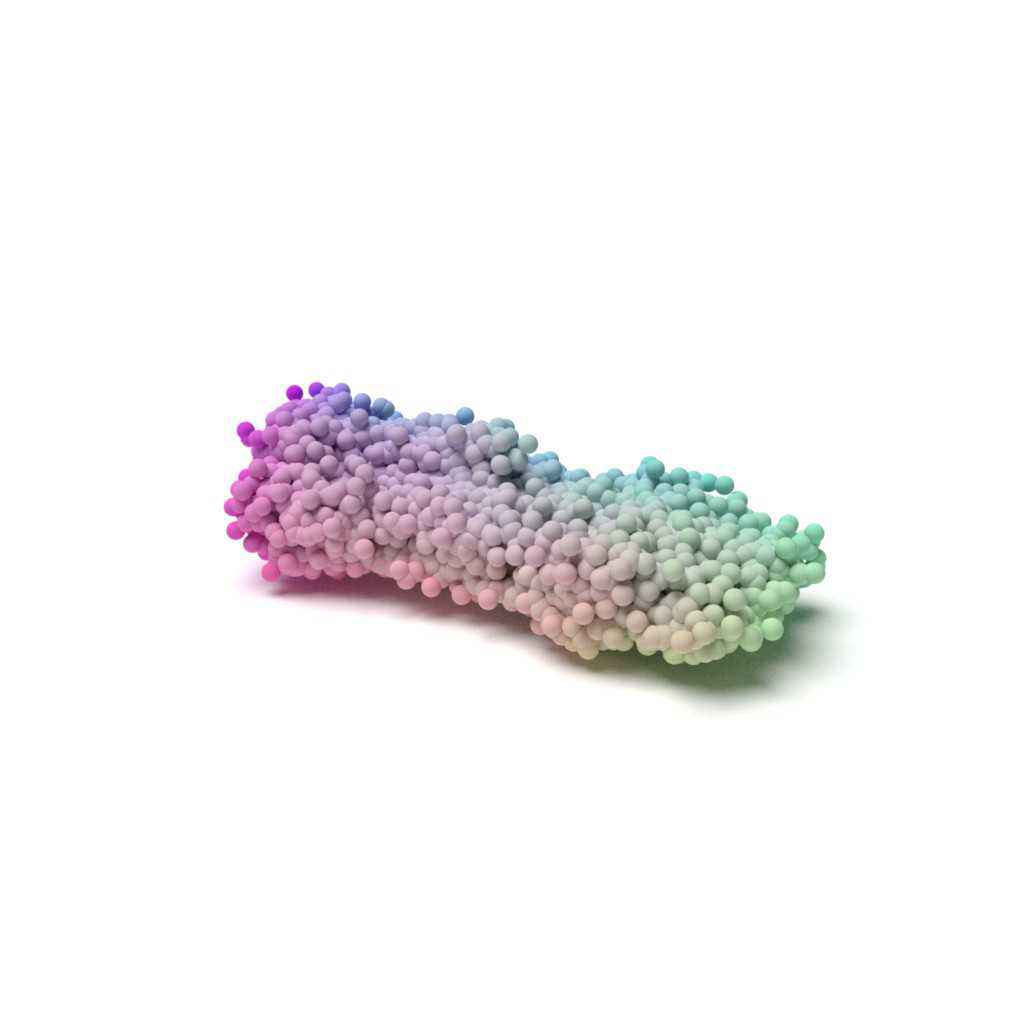}
	\includegraphics[width=\sizea]{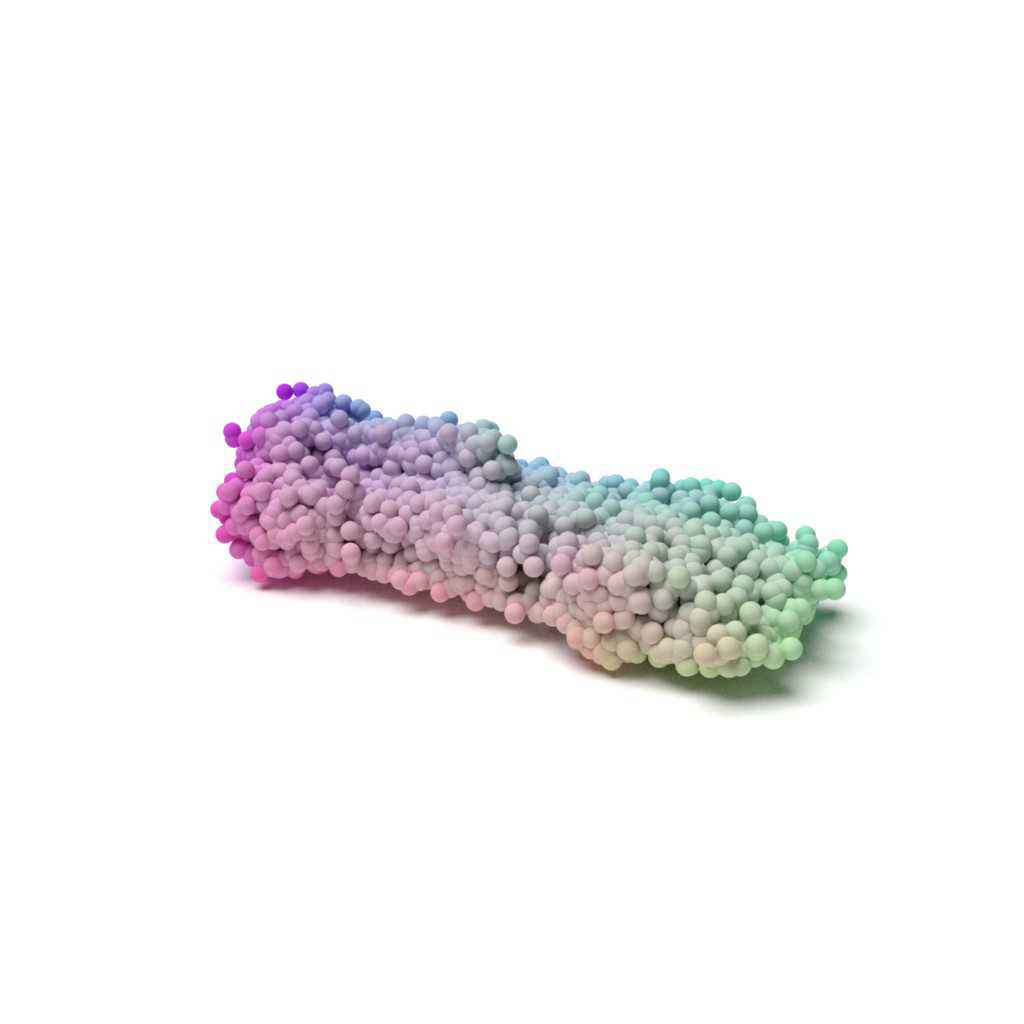}
	\includegraphics[width=\sizea]{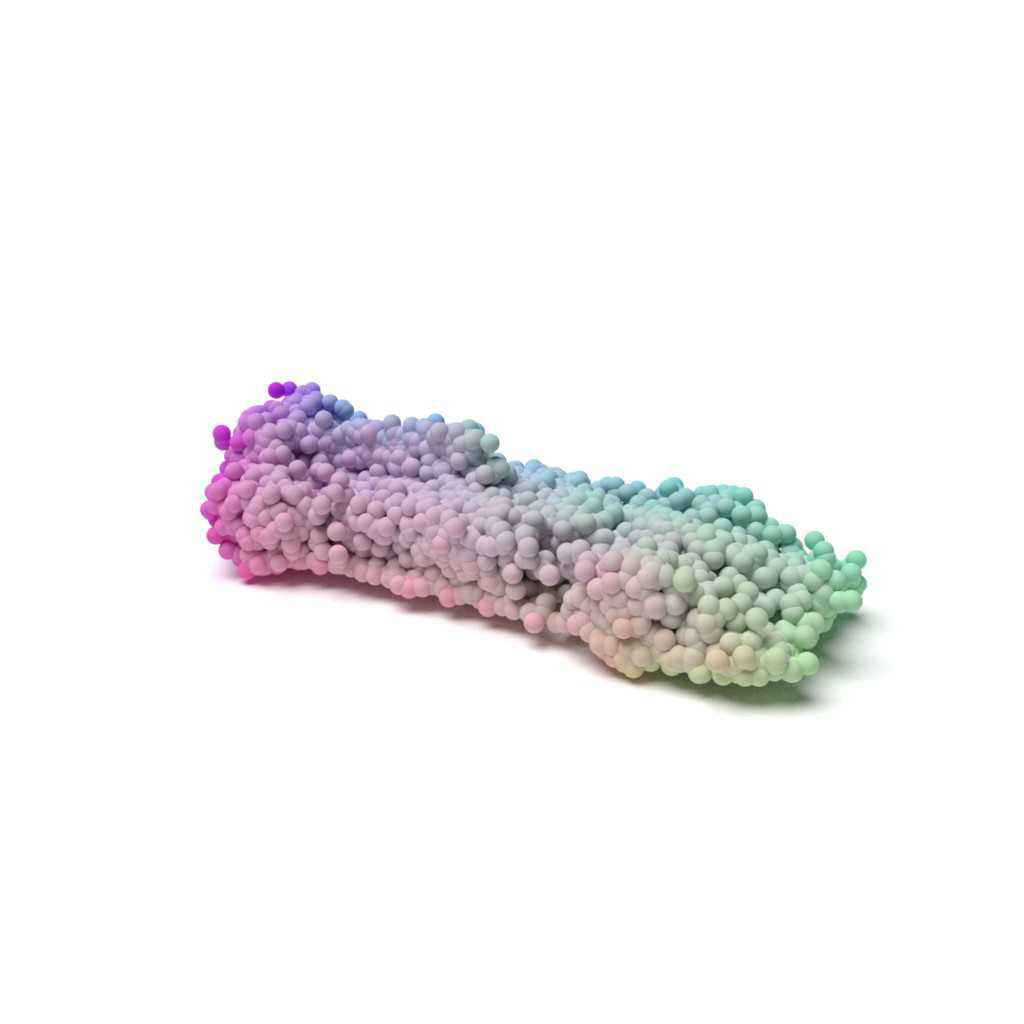}\\
	\includegraphics[width=\sizea]{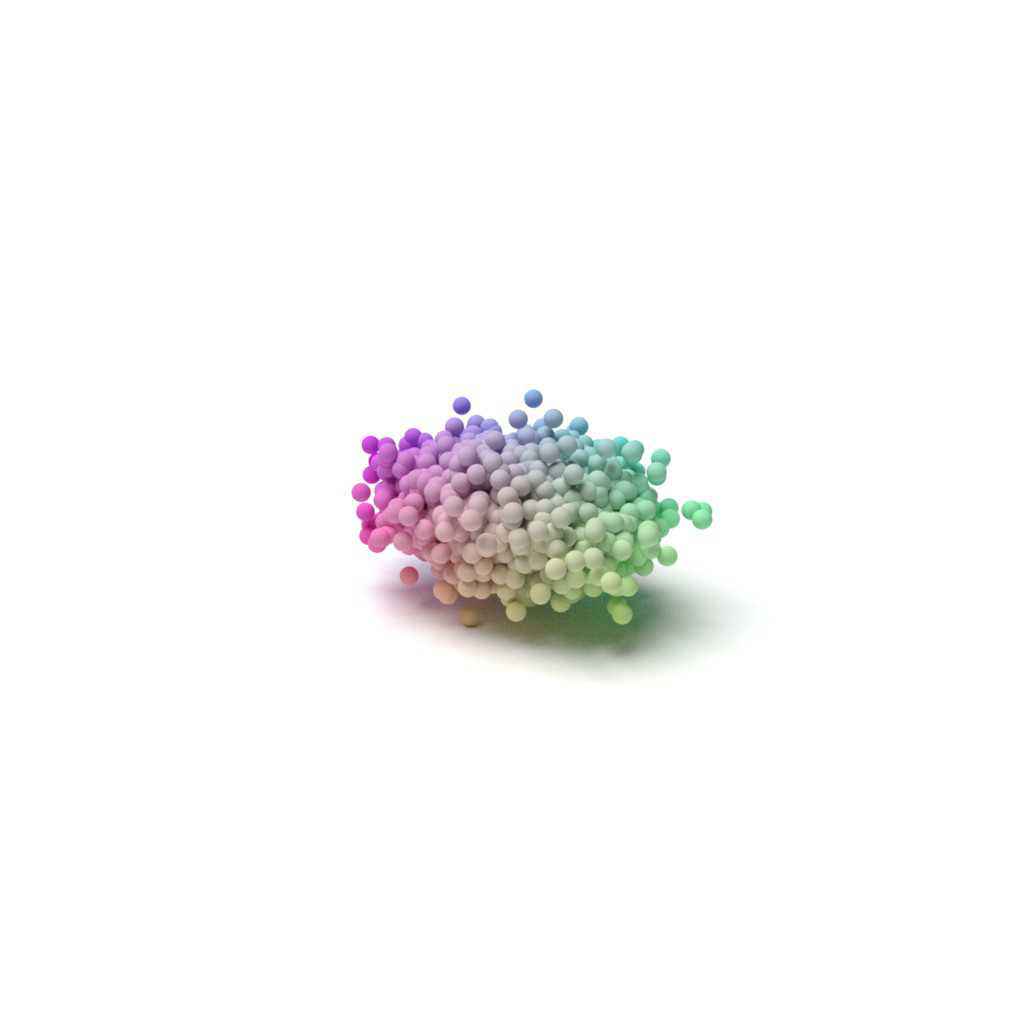}
	\includegraphics[width=\sizea]{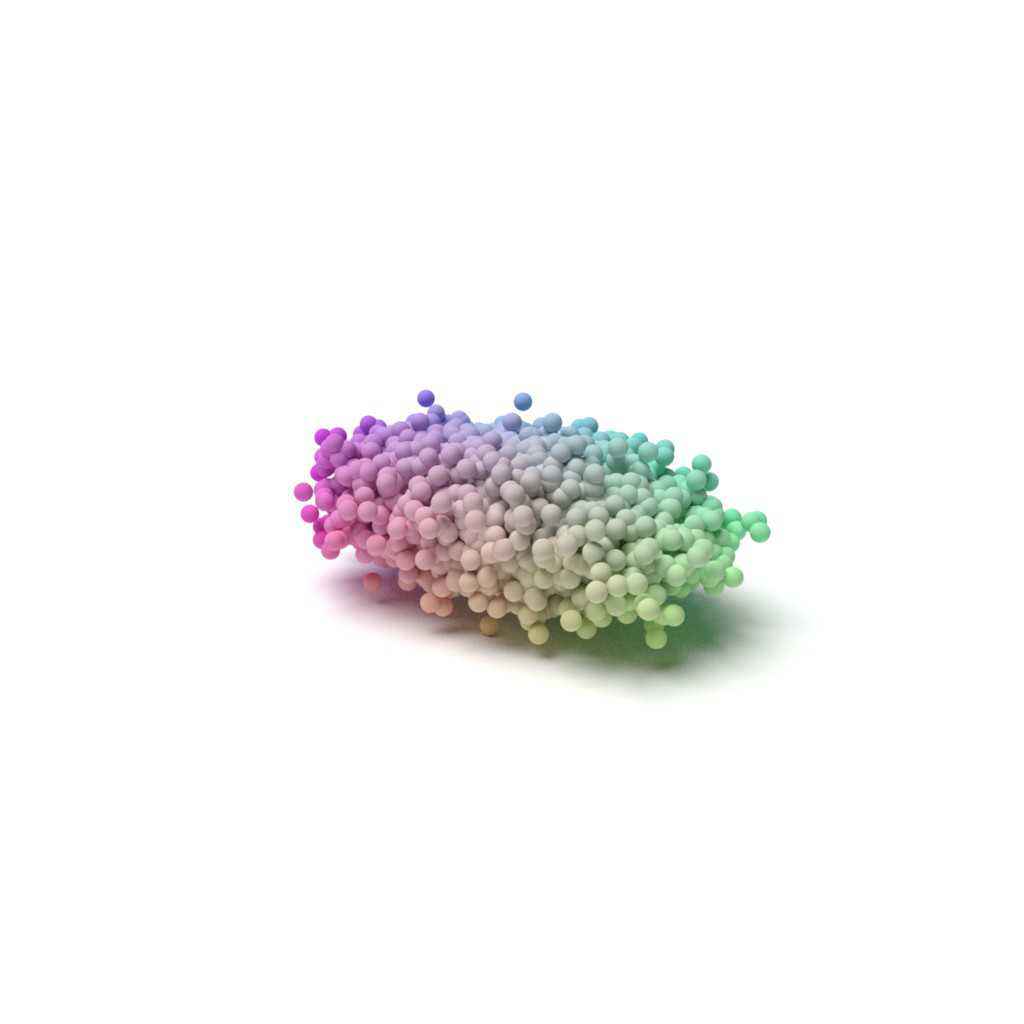}
	\includegraphics[width=\sizea]{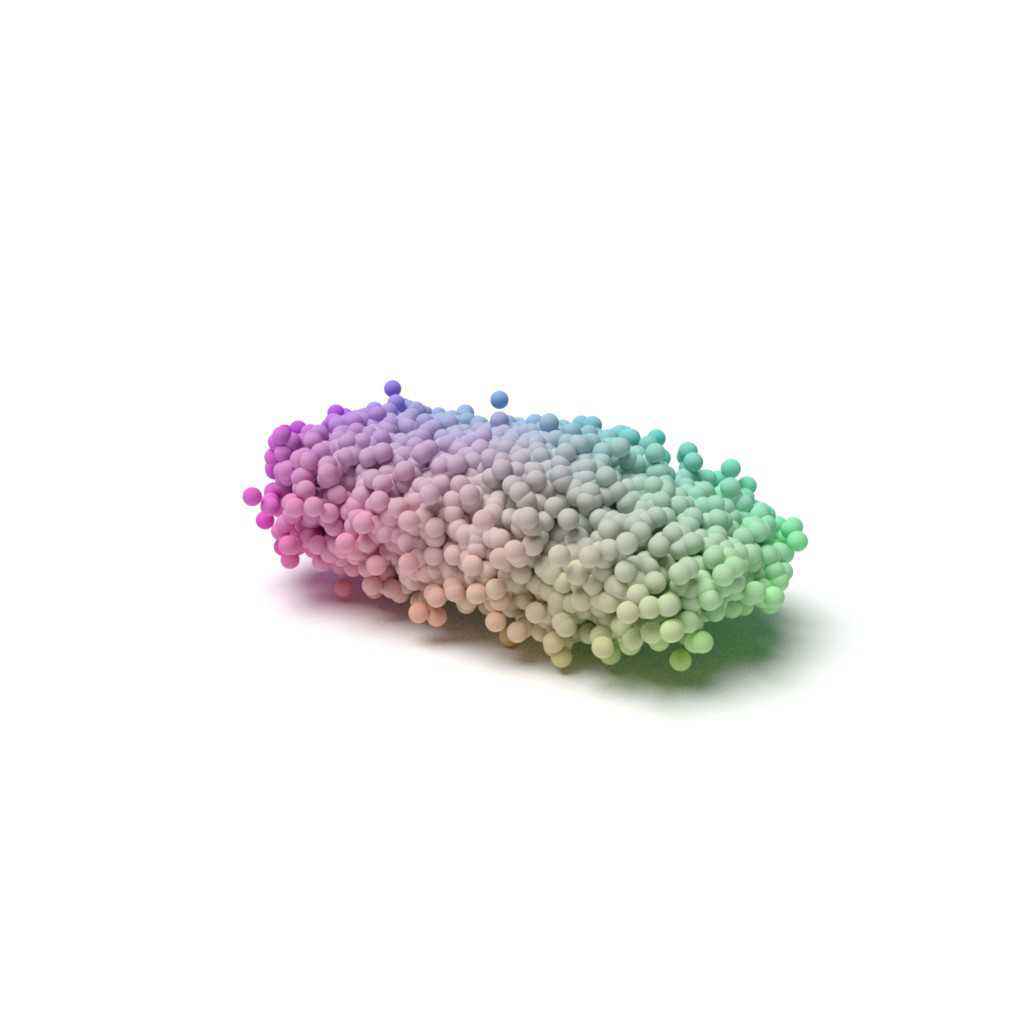}
	\includegraphics[width=\sizea]{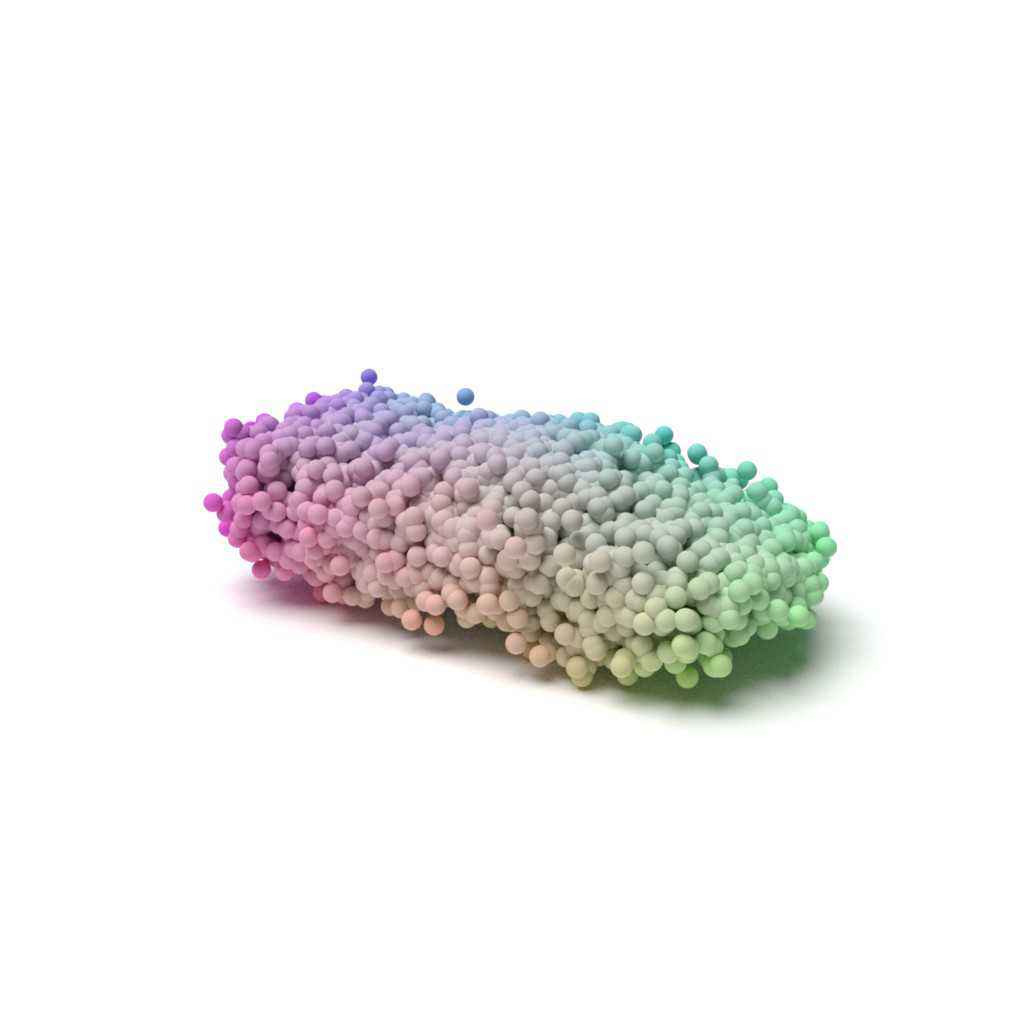}
	\includegraphics[width=\sizea]{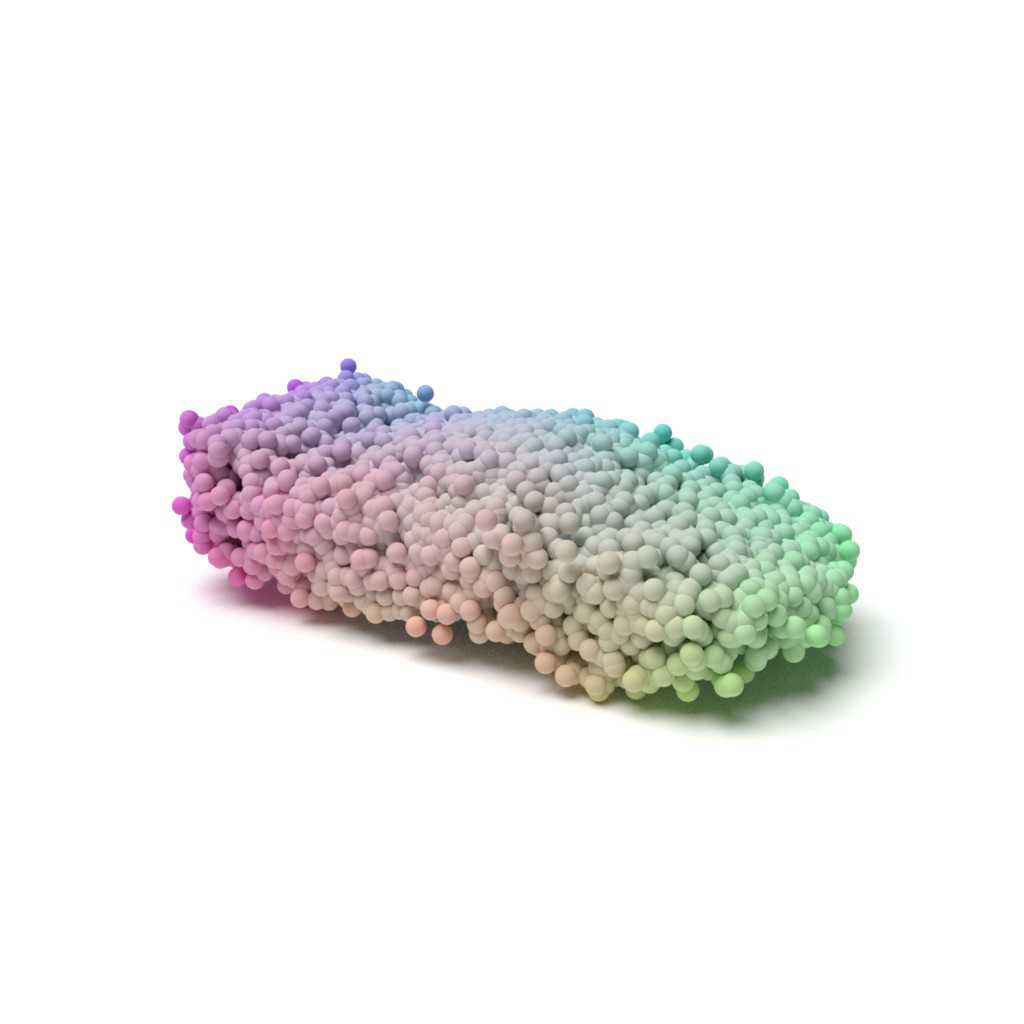}
	\includegraphics[width=\sizea]{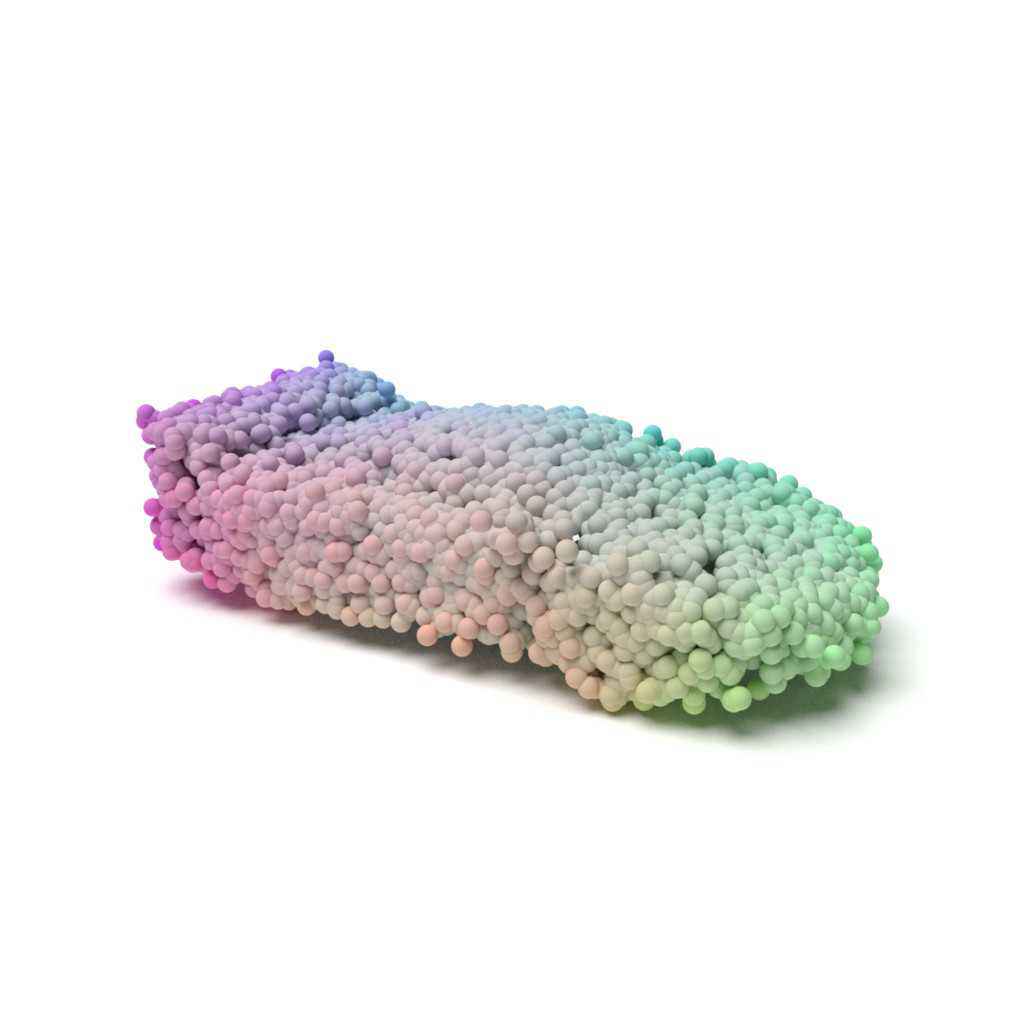}
	\includegraphics[width=\sizea]{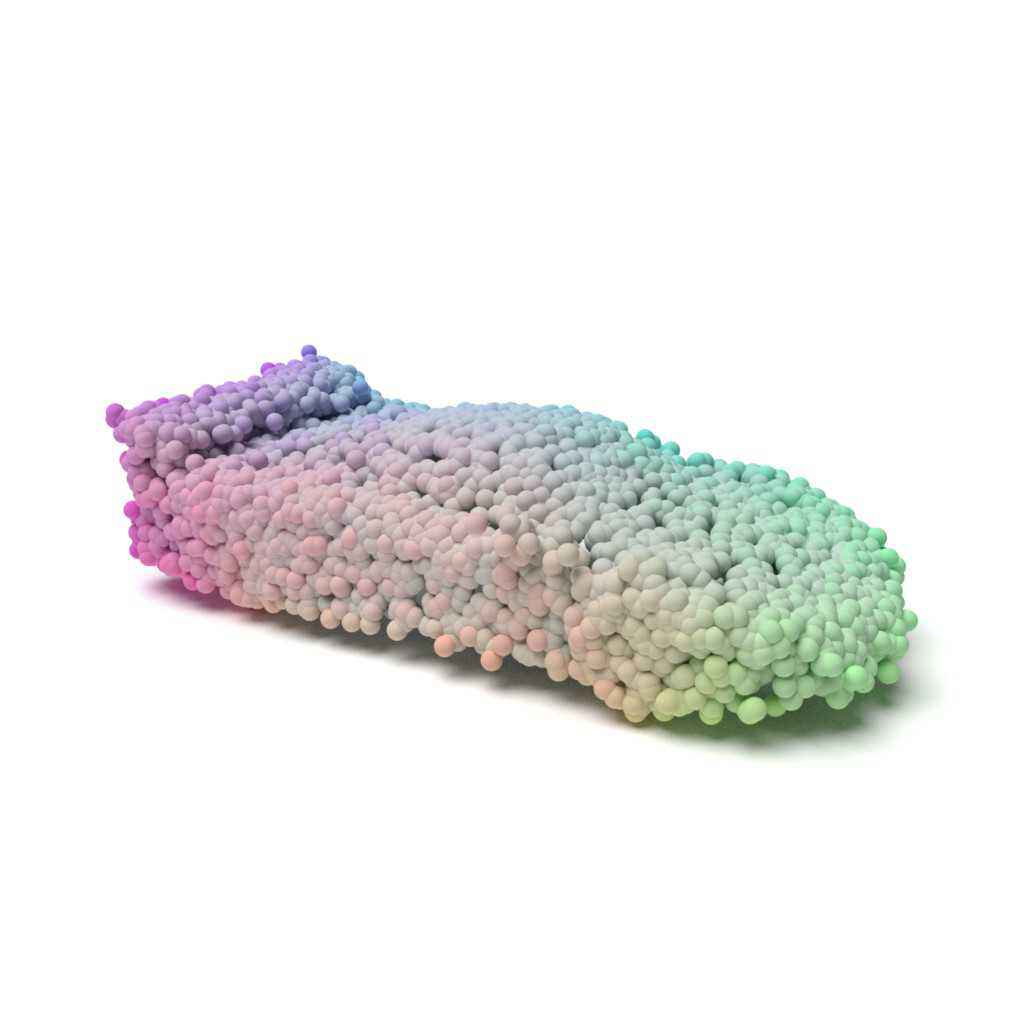}\\
	\includegraphics[width=\sizea]{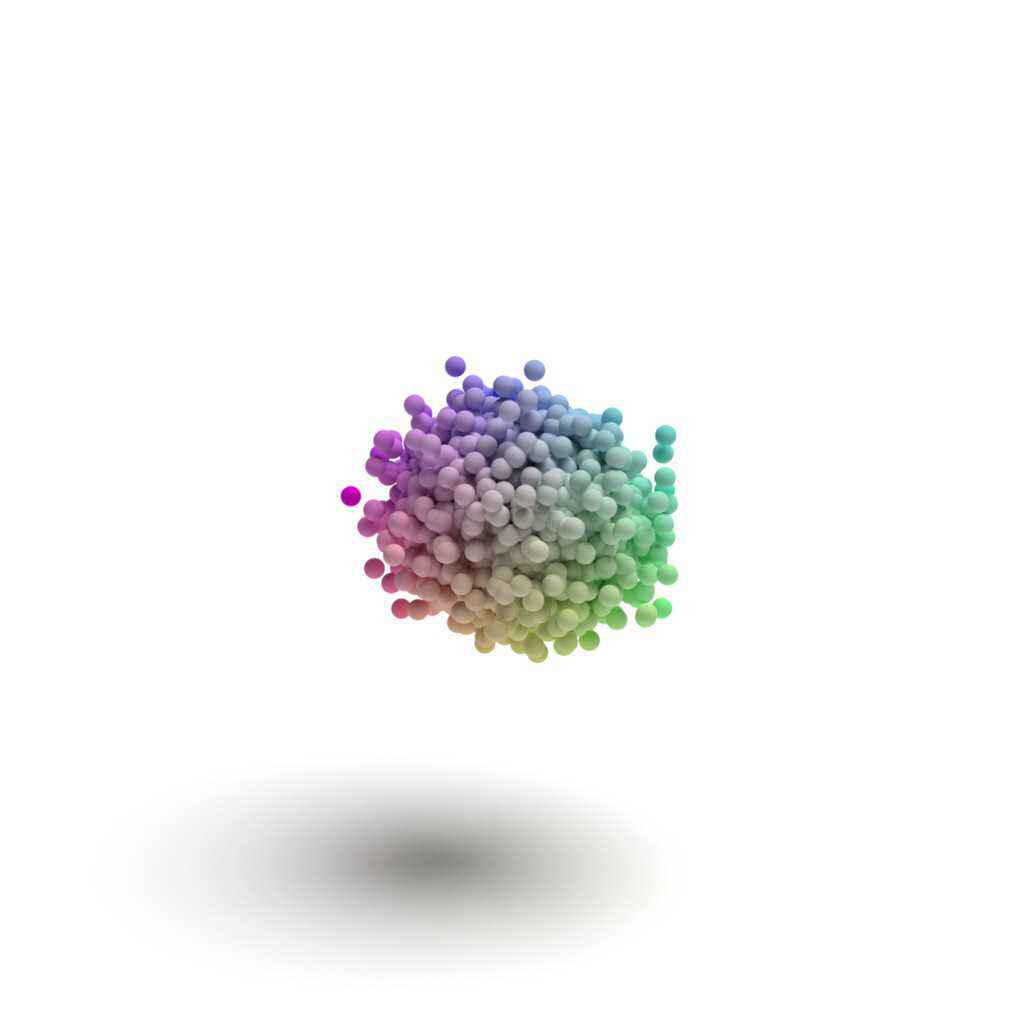}
	\includegraphics[width=\sizea]{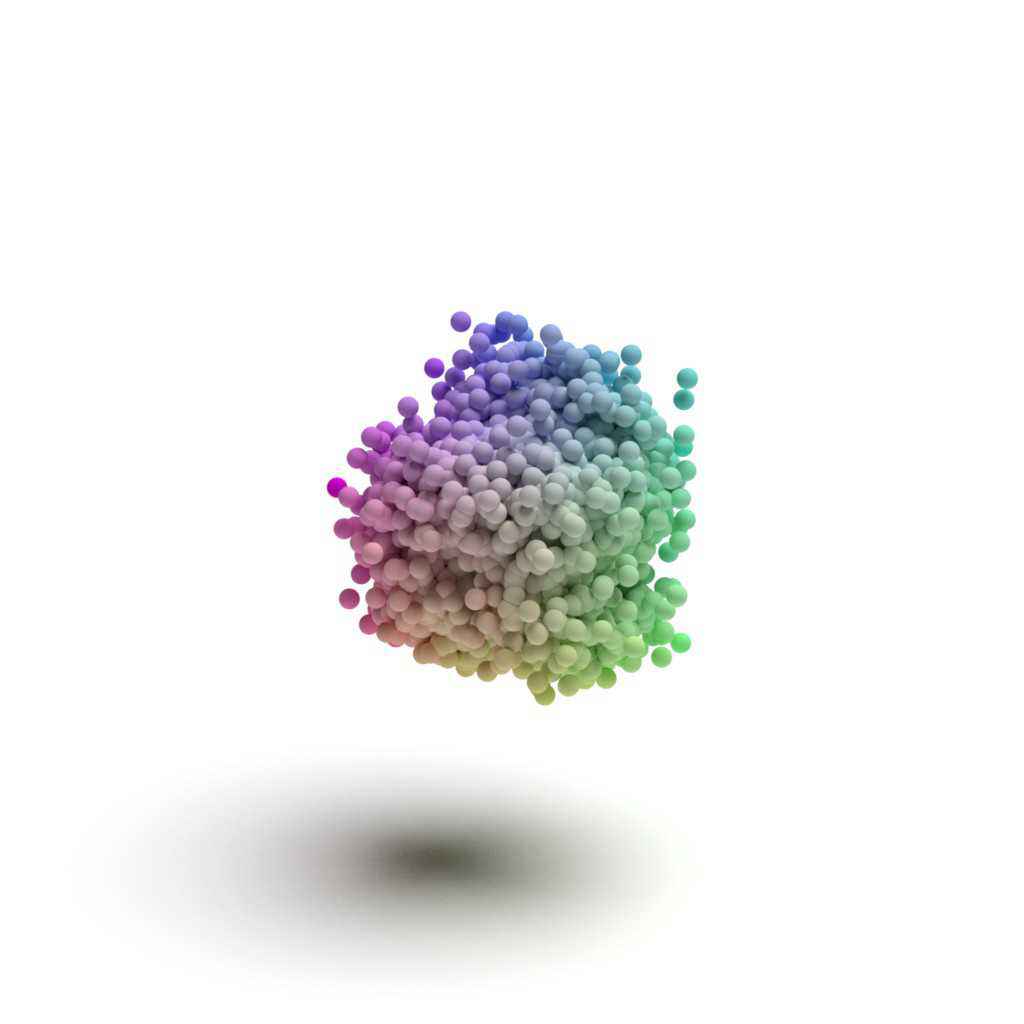}
	\includegraphics[width=\sizea]{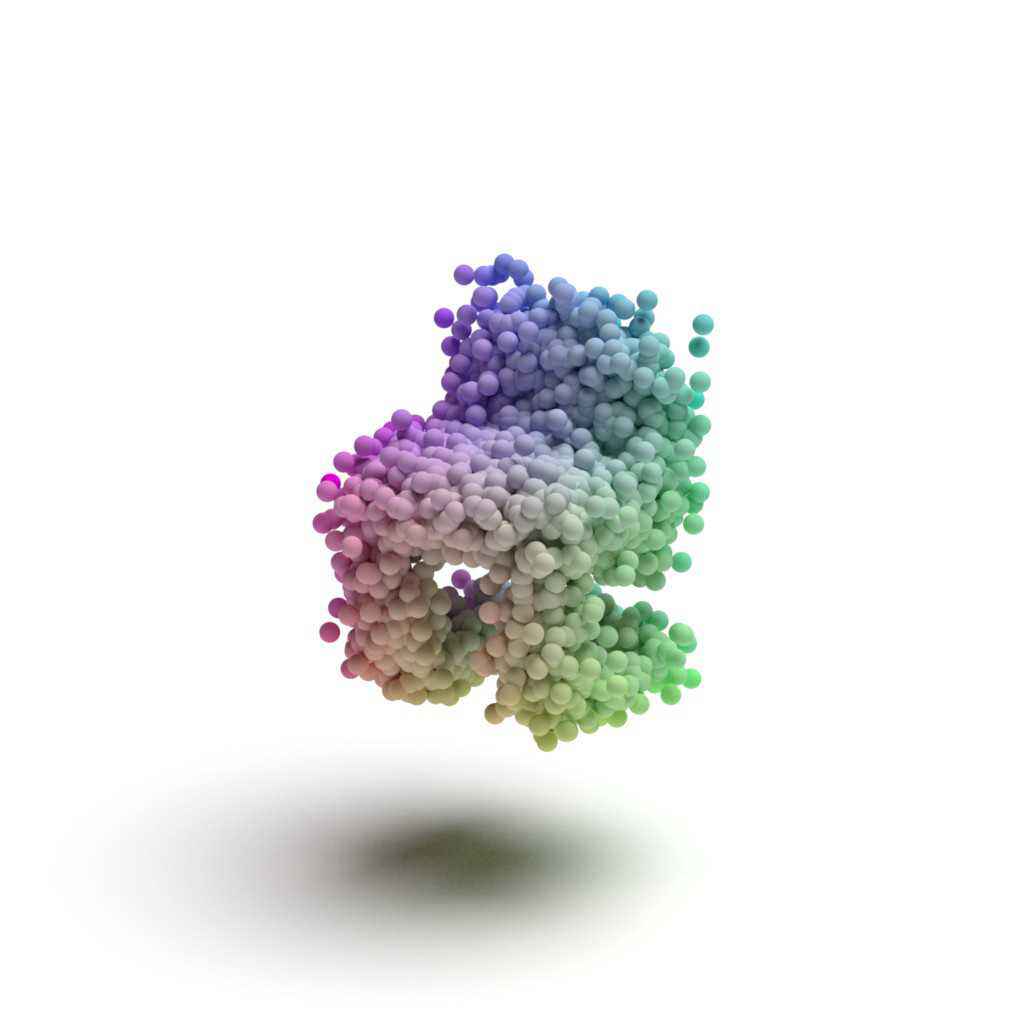}
	\includegraphics[width=\sizea]{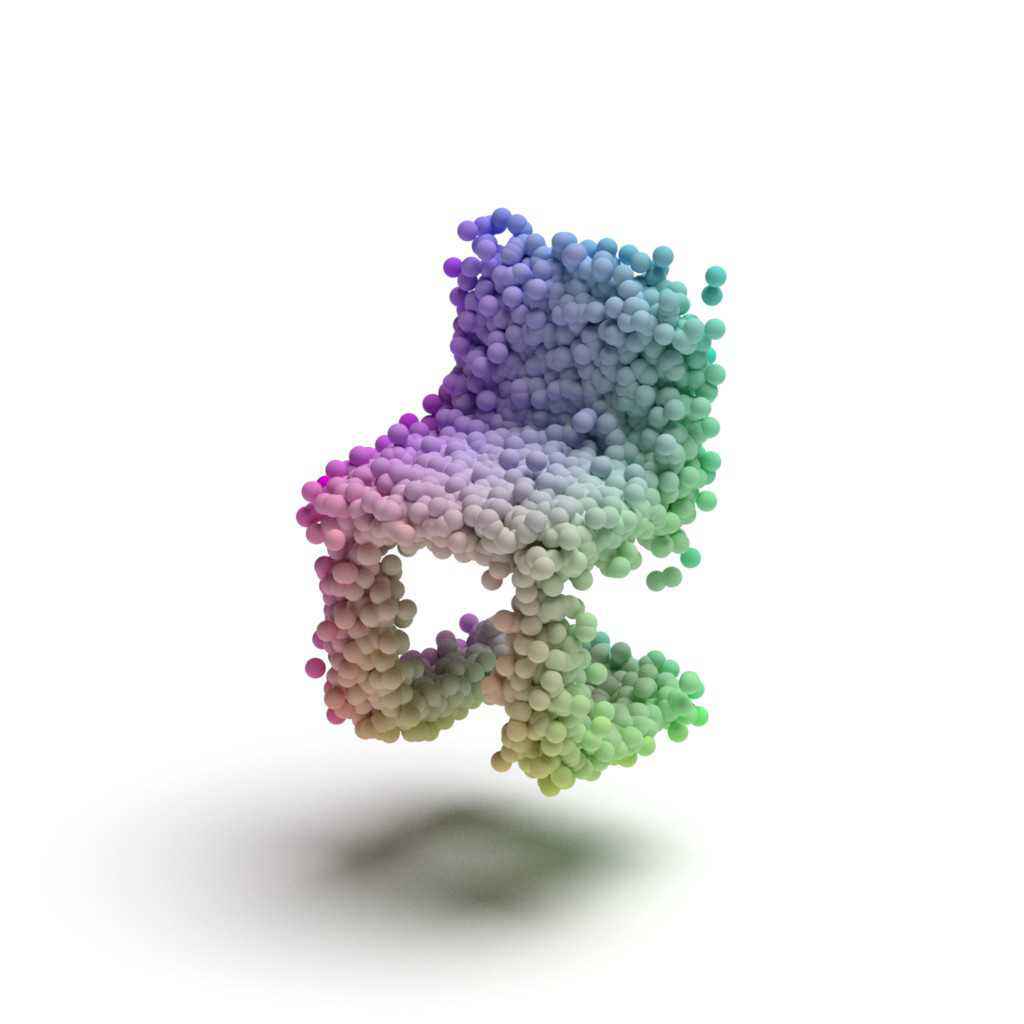}
	\includegraphics[width=\sizea]{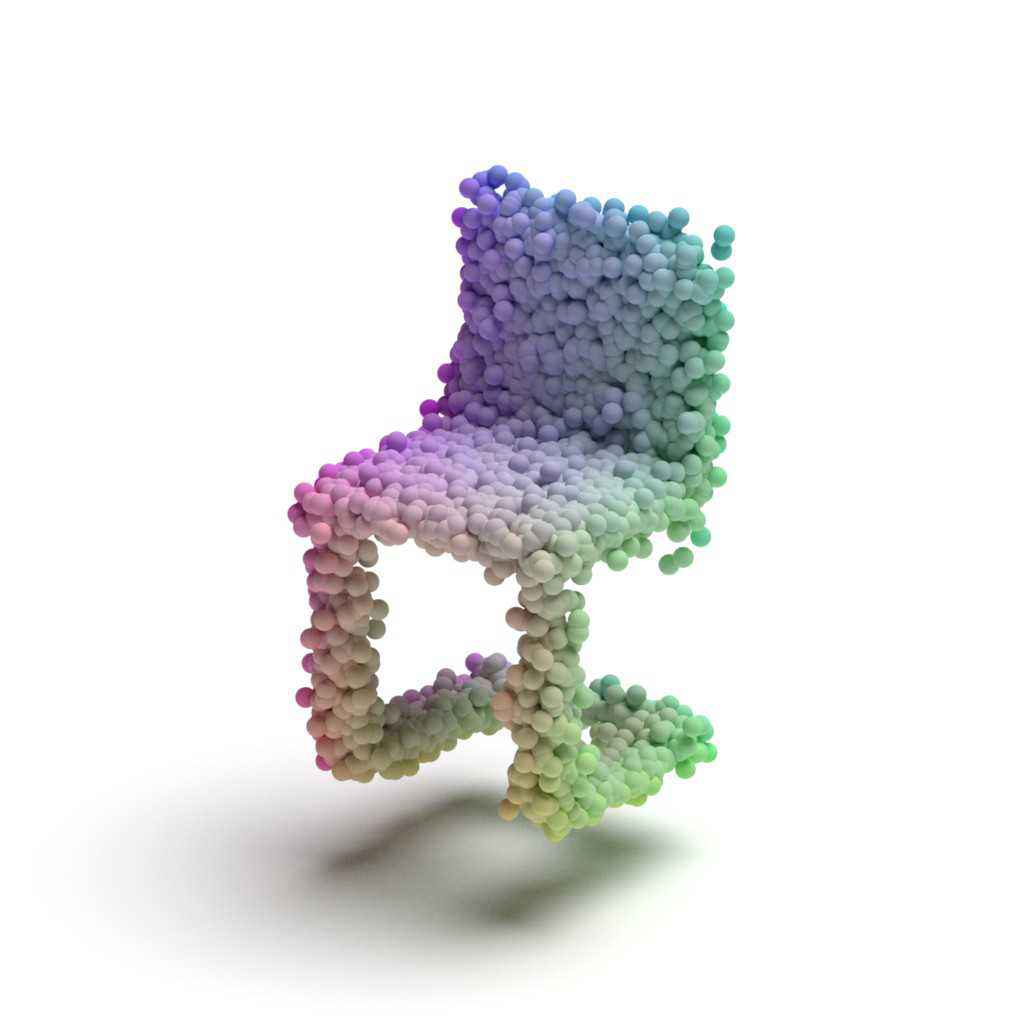}
	\includegraphics[width=\sizea]{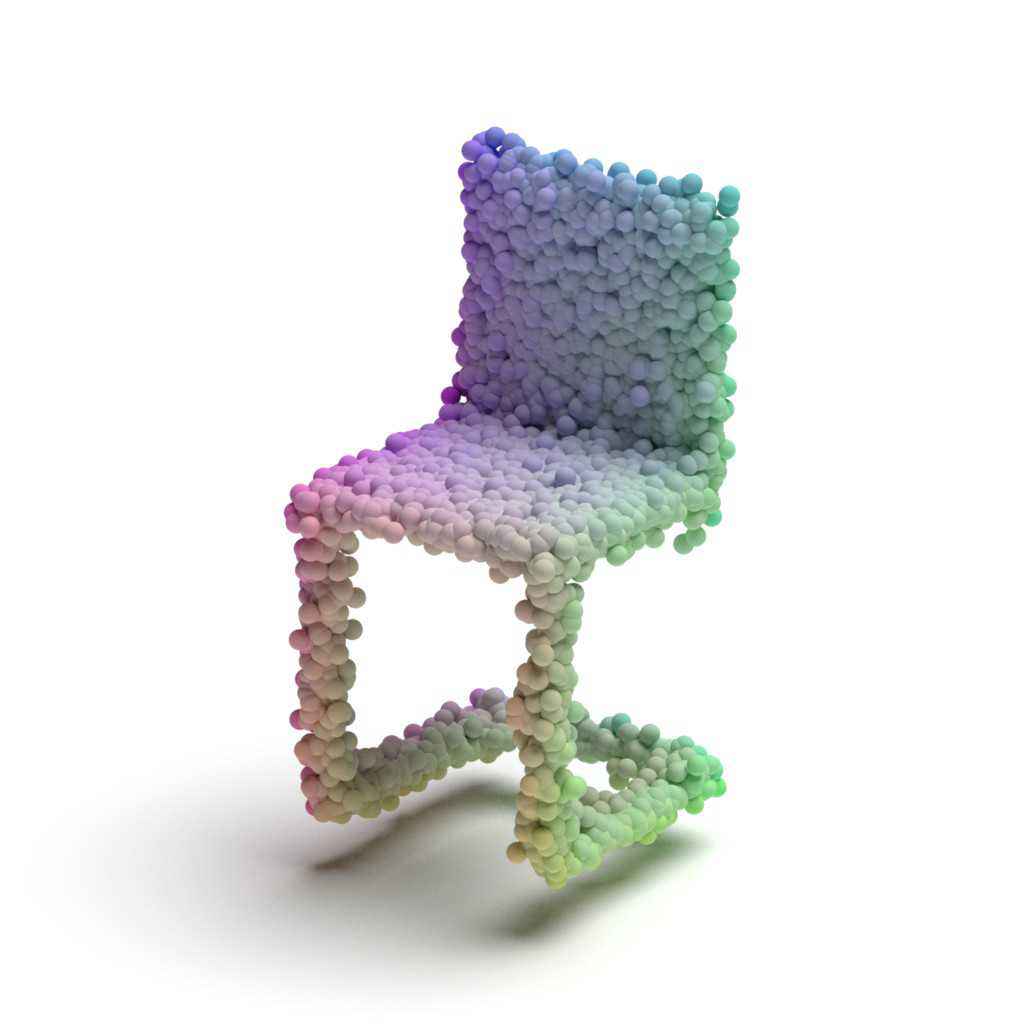}
	\includegraphics[width=\sizea]{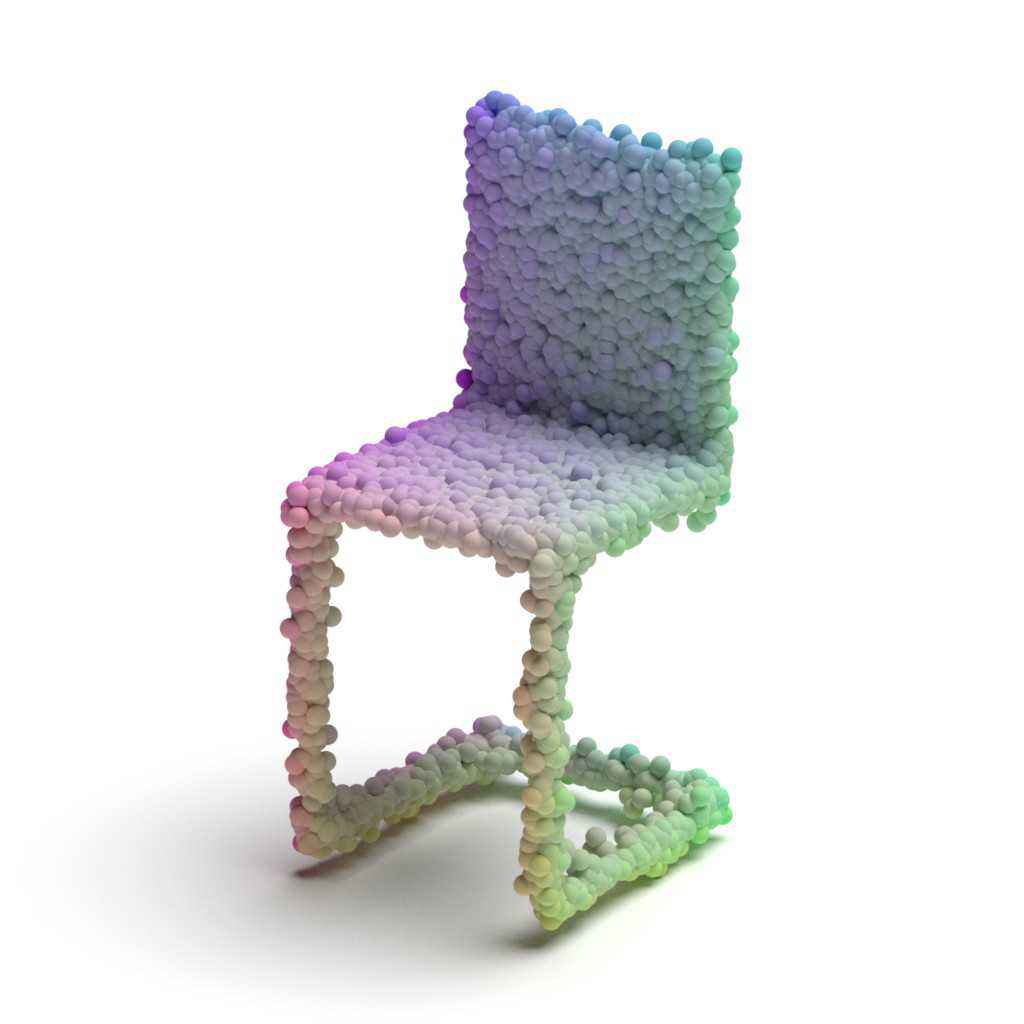}\\
	\includegraphics[width=\sizea]{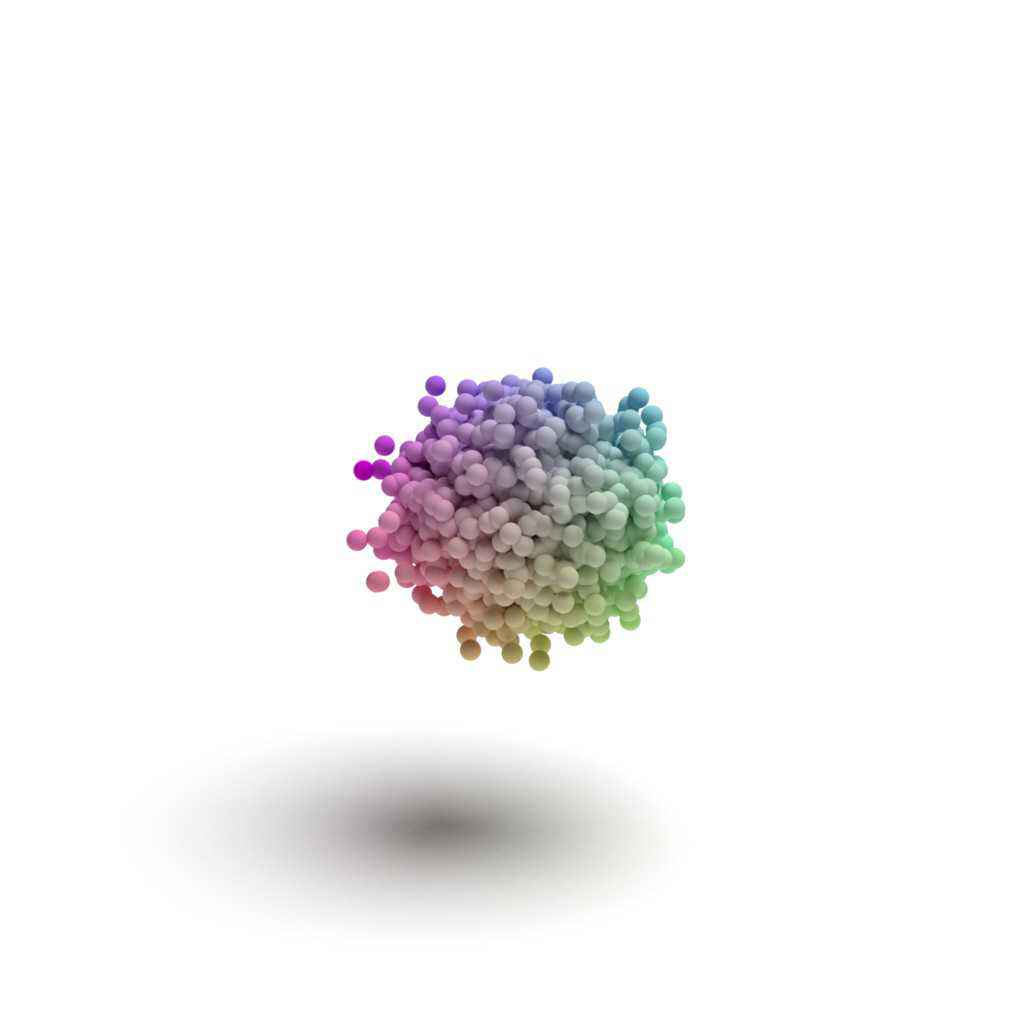}
	\includegraphics[width=\sizea]{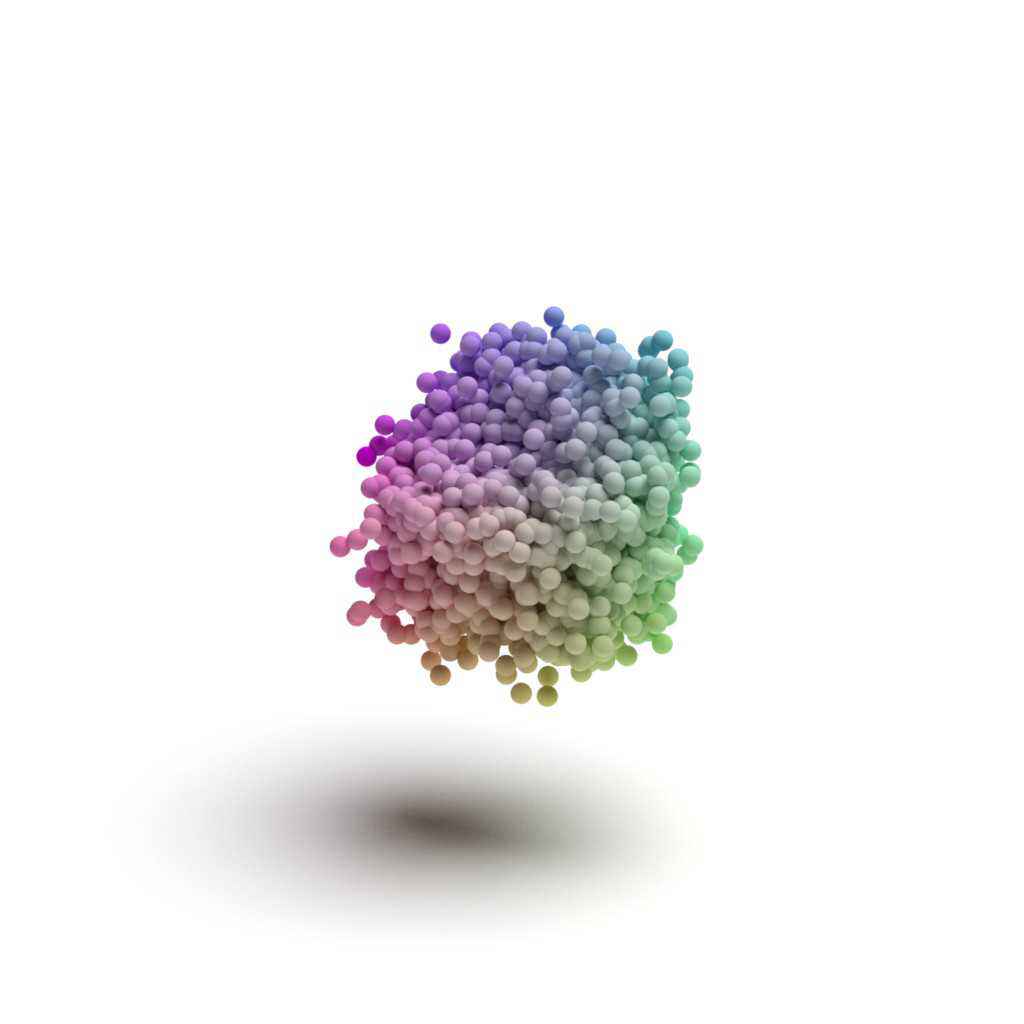}
	\includegraphics[width=\sizea]{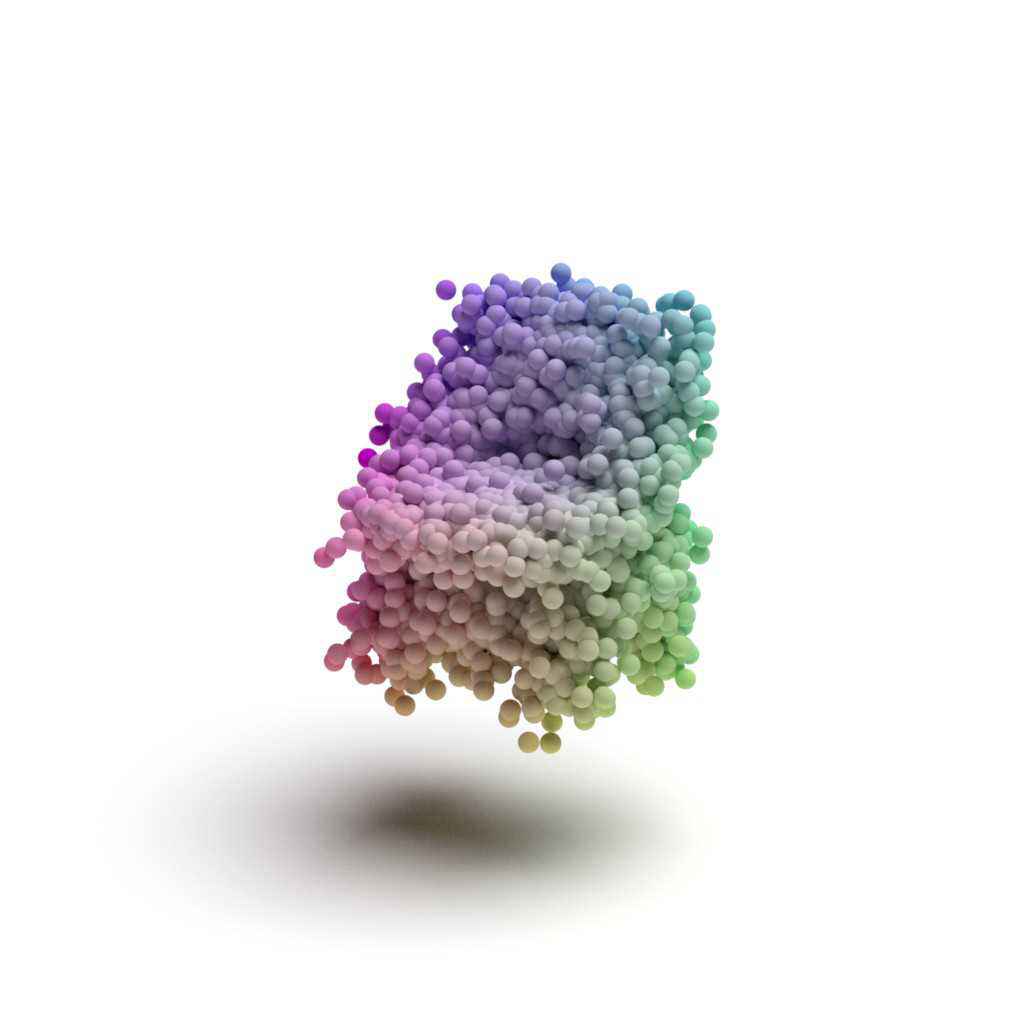}
	\includegraphics[width=\sizea]{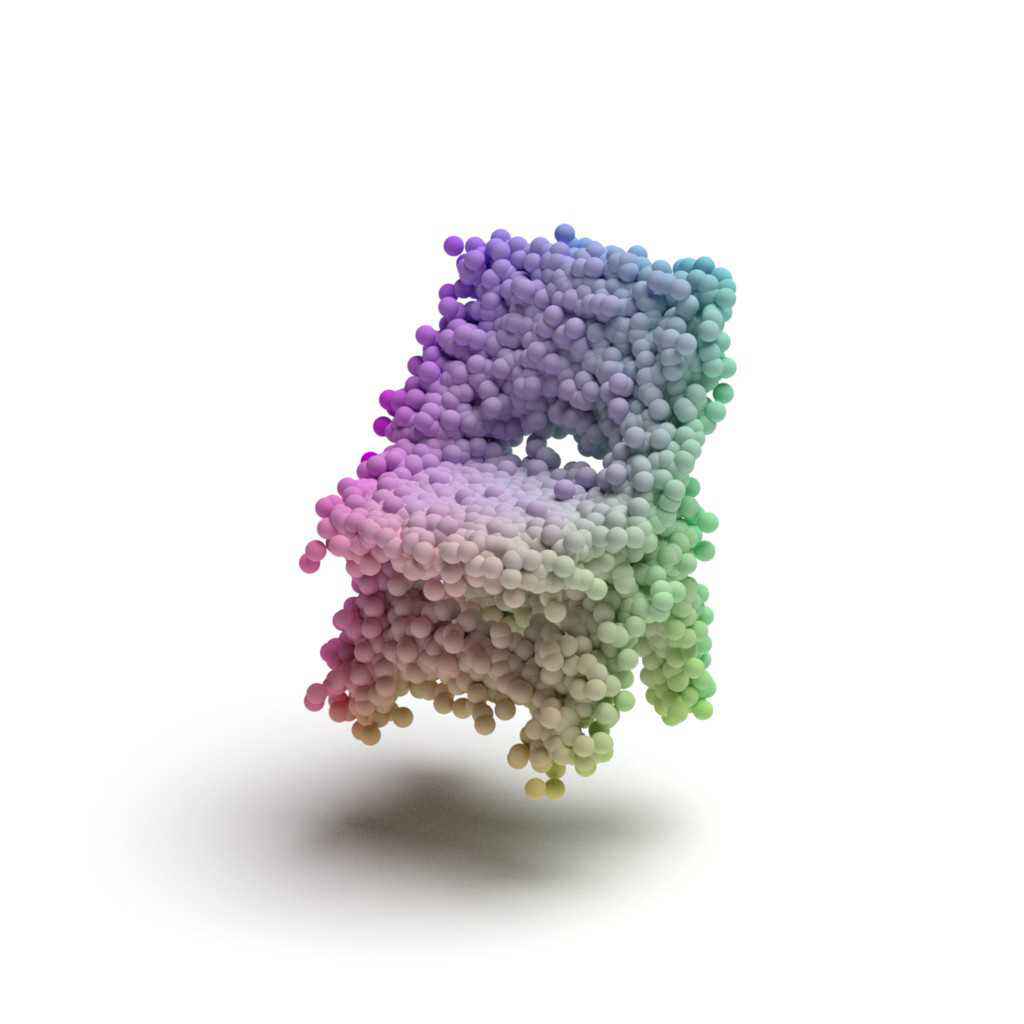}
	\includegraphics[width=\sizea]{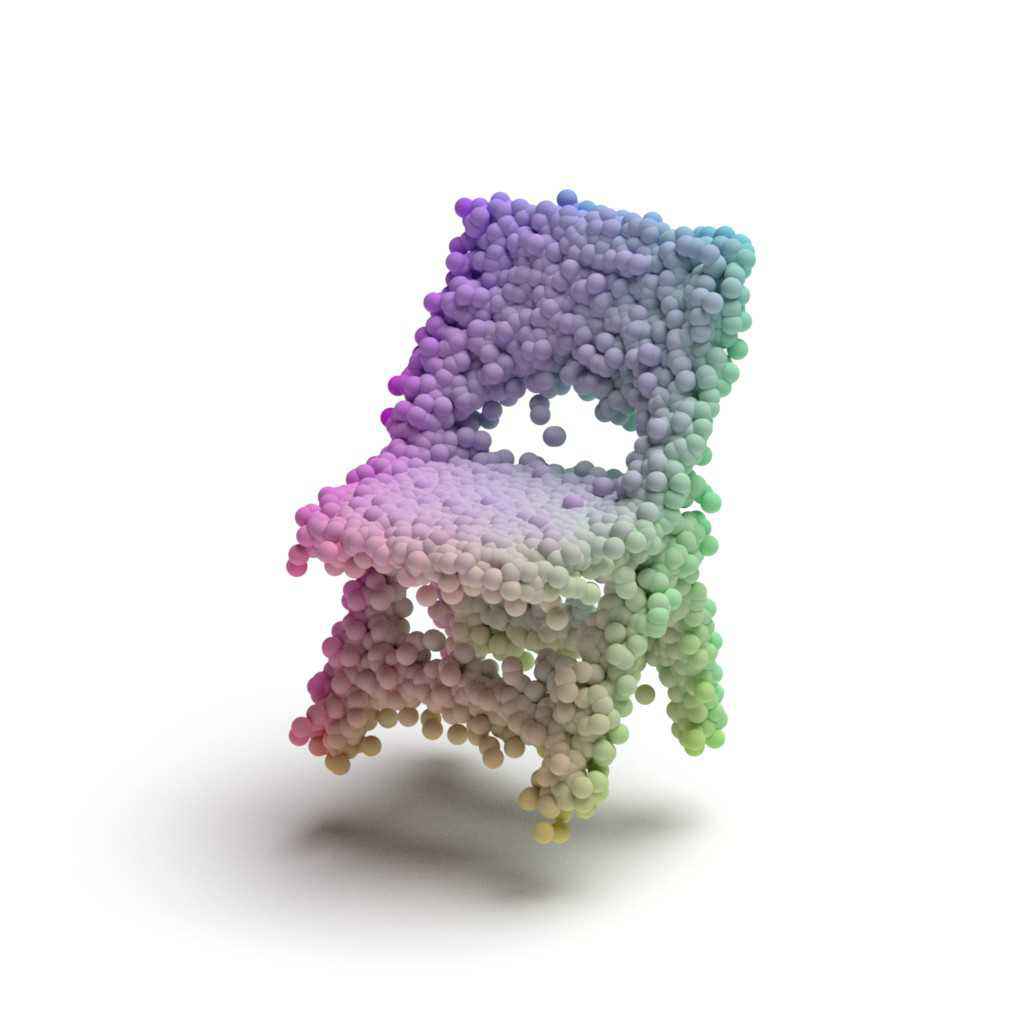}
	\includegraphics[width=\sizea]{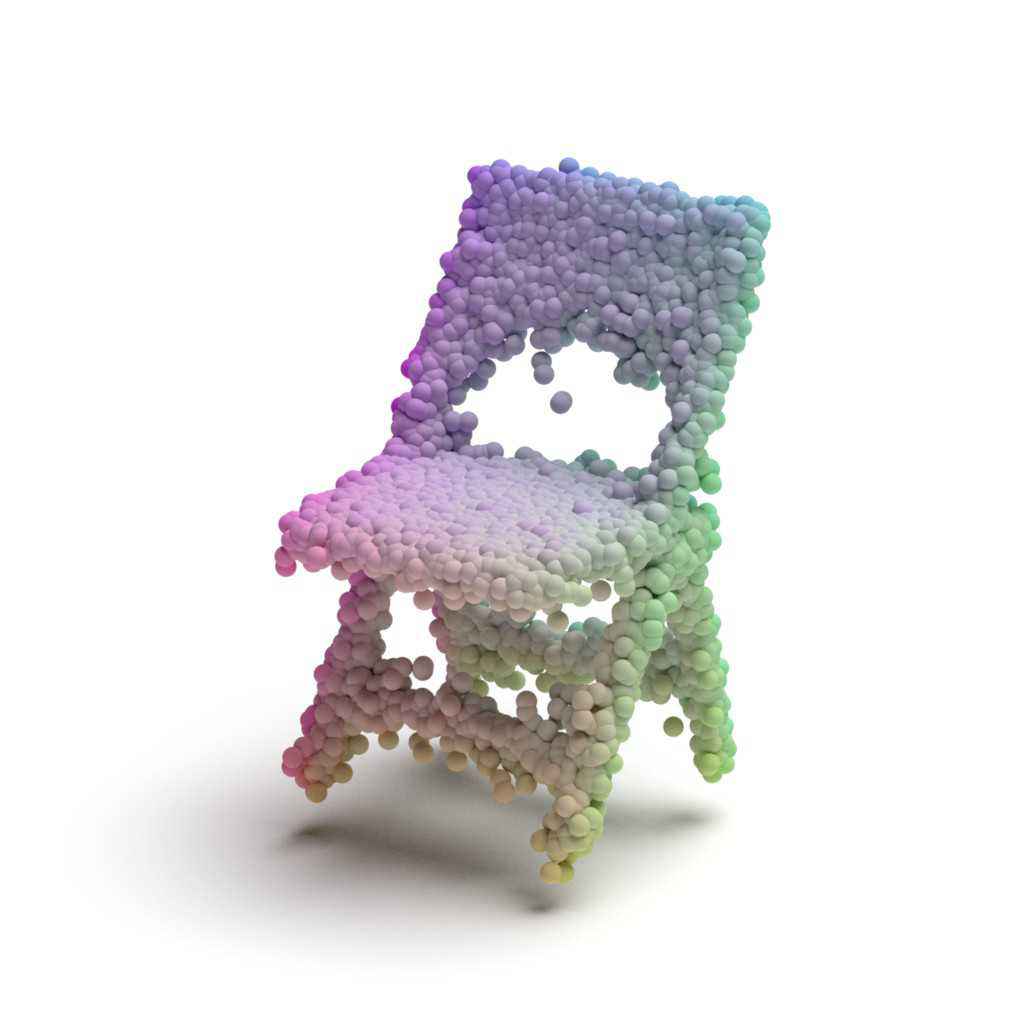}
	\includegraphics[width=\sizea]{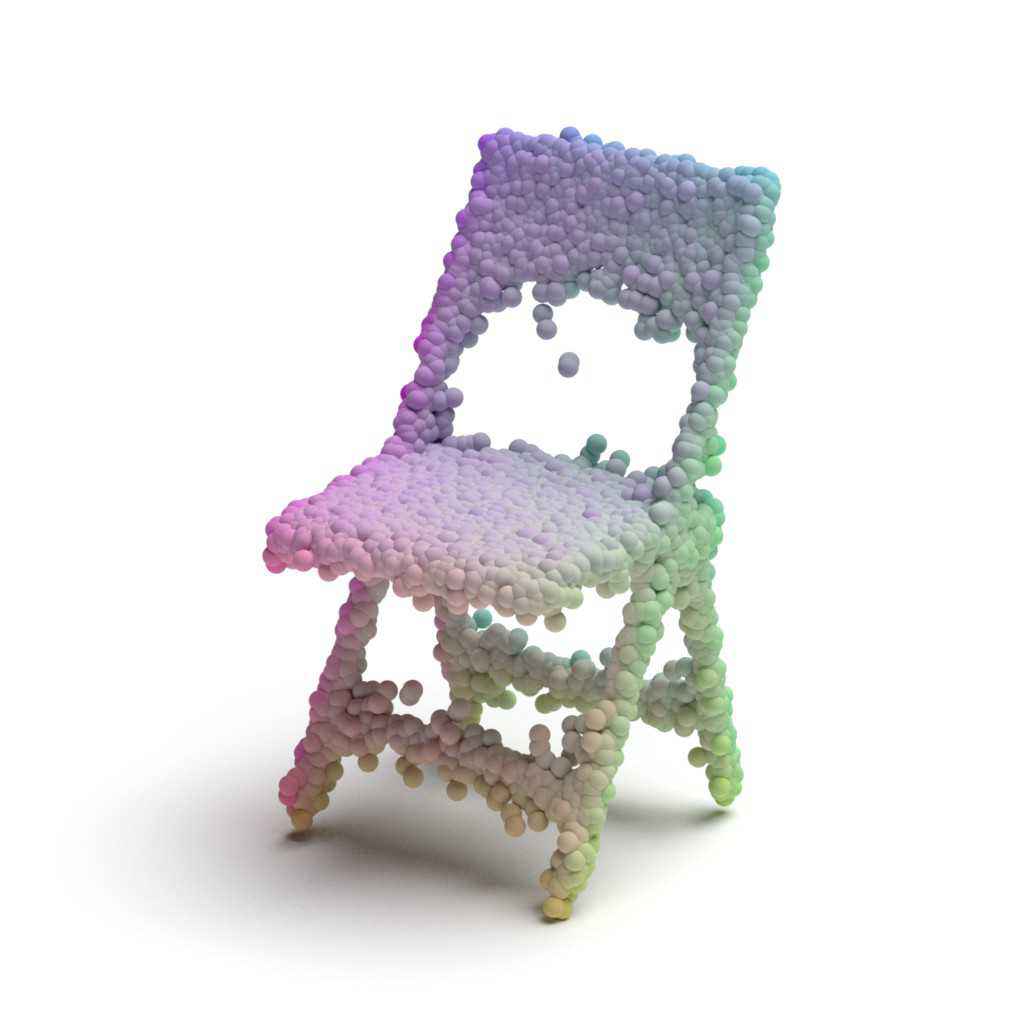}\\
	\captionof{figure}{Additional visualizations on the process of transforming prior to point cloud.}
	\label{fig:morevisflow}
	\vspace{0.2in}
\end{figure*}

Figure~\ref{fig:morevisflow} presents more examples of flow transformations from the Gaussian prior to different shapes.


\end{appendix}

\end{document}